%% file: acl_latex.tex
\documentclass[11pt]{article}
\usepackage{amsmath}
\usepackage[final]{acl}

\usepackage{times}
\usepackage{latexsym}

\usepackage[T1]{fontenc}

\usepackage[utf8]{inputenc}

\usepackage{microtype}

\usepackage{inconsolata}

\usepackage{graphicx}
\usepackage{xspace}
\usepackage{booktabs}
\usepackage{multirow}
\usepackage{nccmath}
\usepackage[most]{tcolorbox}
\usepackage{listings}
\usepackage{authblk}

\def\secref#1{\S\ref{sec:#1}}
\def\seclabel#1{\label{sec:#1}}

\title{Large Reasoning Models Are (Not Yet) Multilingual Latent Reasoners}

\author[]{\bf{Yihong Liu}$^{\text *}$}
\author[]{\bf{Raoyuan Zhao}$^{\text *}$}
\author[]{{\bf Hinrich Sch\"utze}$^\dag$}
\author[]{{\bf Michael A. Hedderich}$^\dag$}

\affil[]{Center for Information and Language Processing, LMU Munich \\Munich Center for Machine Learning (MCML)
 \protect\\ \texttt{\{yihong, rzhao, hedderich\}@cis.lmu.de}} 

\newcounter{notecounter}
\newcommand{\enotesoff}{\long\gdef\enote##1##2{}}
\newcommand{\enoteson}{\long\gdef\enote##1##2{{
\stepcounter{notecounter}
{\large\bf
\hspace{0cm}\arabic{notecounter} $<<<$ ##1: ##2
$>>>$\hspace{1cm}}}}}

\enoteson
\enotesoff

\begin{document}
\maketitle

\def\thefootnote{*}\footnotetext{Equal contribution.}\def\thefootnote{\arabic{footnote}}
\def\thefootnote{$\dag$}\footnotetext{Equal advising.}\def\thefootnote{\arabic{footnote}}

\begin{abstract}

Large reasoning models (LRMs) achieve strong performance on mathematical reasoning tasks, often attributed to their capability to generate explicit chain-of-thought (CoT) explanations.
However, recent work shows that LRMs often arrive at the correct answer before completing these textual reasoning steps, indicating the presence of \emph{latent reasoning} -- internal, non-verbal computation encoded in hidden states.
While this phenomenon has been explored in English, its multilingual behavior remains largely unknown.
In this paper, we conduct a systematic investigation of multilingual latent reasoning in LRMs across 11 languages.
Using a truncation-based strategy, we examine how the correct answer emerges as the model is given only partial reasoning traces, allowing us to measure stepwise latent prediction formation. 
Our results reveal clear evidence of multilingual latent reasoning, though unevenly: strong in resource-rich languages, weaker in low-resource ones, and broadly less observable on harder benchmarks.
To understand whether these differences reflect distinct internal mechanisms, we further perform representational analyses.
Despite surface-level disparities, we find that the internal evolution of predictions is highly consistent across languages and broadly aligns with English -- a pattern suggesting an English-centered latent reasoning pathway.\footnote{We make our code publicly available at: \url{https://github.com/cisnlp/multilingual-latent-reasoner}}

\end{abstract}

\section{Introduction}

Recent large reasoning models (LRMs) \citep{openai2024openaio1card,yang2025qwen3technicalreport,deepseekai2025deepseekr1incentivizingreasoningcapability} have rapidly advanced the state of the art on many challenging tasks, such as coding, mathematical reasoning, and logical reasoning \citep{li202512surveyreasoning}.
This is largely thought to be due to their capacity to generate explicit CoT explanations \citep{wei2022cot} that scaffold multi-step problem solving, especially through test-time scaling, where enough computation budget is given to allow the model to generate longer reasoning traces \citep{snell2024scalingllmtesttimecompute,muennighoff2025s1simpletesttimescaling}. 

Despite this reliance on explicit CoT explanations, emerging evidence shows that models often engage in \emph{latent reasoning} -- computing intermediate or final answers within hidden states.  
Such latent behavior has been observed in multi-hop factual knowledge recall \citep{yang-etal-2024-large-language-models,biran-etal-2024-hopping} and, in the context of CoT, in models that internally form solutions well before they articulate the answer in their reasoning \citep{lanham2023measuringfaithfulnesschainofthoughtreasoning,pfau2024letsthinkdotdot,mao2025earlystoppingchainofthoughtslarge}.  
This phenomenon aligns with recent findings that LLMs can ``think ahead'' by predicting future tokens directly from intermediate hidden states \citep{pal-etal-2023-future,wu2024languagemodelsplanahead,cai2024medusa}.
Together, these observations indicate that explicit CoT generation is not the sole mechanism through which LRMs solve problems and that reasoning may be occurring within the model's latent space.

However, existing studies of latent reasoning focus almost exclusively on English, leaving open how these latent reasoning processes behave across languages.  
At the explicit reasoning level, multilingual performance is already known to be uneven: 
models trained on English-centric corpora often struggle with underrepresented languages due to limited multilingual reasoning training data \citep{wang2025enhancing,huang2025englishcentrictrainingreinforcementlearning}, weaker language understanding ability \citep{yoon-etal-2024-langbridge,kang2025multilingualreasoninggapsemerge}, and lower-quality reasoning trace generation \citep{yong2025crosslingualreasoningtesttimescaling,zhao2025comprehensiveevaluationmultilingualchainofthought}.  
These findings raise a natural question: if explicit reasoning varies across languages, does latent reasoning exhibit similar disparities, or does it follow a language-independent mechanism?  
This motivates our two research questions: (\textbf{RQ1}) \emph{Do LRMs exhibit latent reasoning across languages, and how does the strength vary?} and (\textbf{RQ2}) \emph{Do languages follow different internal latent reasoning pathways, or do they share a common mechanism?}

To answer these questions, we conduct a systematic investigation of multilingual latent reasoning in LRMs using two mathematical reasoning benchmarks of different difficulty across 11 languages.  
To address \textbf{RQ1}, we quantify how strongly LRMs rely
on explicit reasoning traces by eliciting and evaluating
their stepwise early predictions
in truncated traces,
and we propose novel aggregate metrics capturing different dimensions of latent reasoning (cf.~\secref{surface_dynamics}).  
To address \textbf{RQ2}, we analyze the internal evolution of answer formation using the logit lens approach \citep{logit-lens}, examining when the correct answer becomes probable across layers in each language, and we compare hidden-state similarity trajectories across languages (cf.~\secref{latent_dynamics}).  
We further disentangle latent reasoning from potential memorization effects (cf.~\secref{memorization}).

Our key findings are as follows:
(\textbf{i}) Latent reasoning exists across languages.
However, resource-rich languages show strong early-emergent correctness, while low-resource ones display weaker latent reasoning signals.
(\textbf{ii}) Latent reasoning is less pronounced under increased task difficulty.
On harder benchmarks, early answer formation largely disappears across all languages and model sizes.
(\textbf{iii}) Internal latent reasoning dynamics are shared across languages.
Such dynamics converge to an English-centered pathway, especially for high-resource languages and correctly solved instances.
(\textbf{iv}) While models show partial memorization, latent reasoning remains evident for high-resource languages, and this capability scales with model size.

\section{Related Work}\seclabel{related_work}

\textbf{Multilingual Reasoning}
Multilingual reasoning remains challenging due to the strong language bias of most models \citep{ghosh-etal-2025-survey}.  
Since models often rely on English as a pivot, translate-then-solve strategies are frequently effective for under-resourced languages \citep{qin-etal-2023-cross,huang-etal-2023-languages,zhu-etal-2024-question}.  
Recent work shows that post-training on multilingual reasoning data can substantially improve crosslingual performance \citep{chen-etal-2024-breaking,huang2025englishcentrictrainingreinforcementlearning}.  
At inference time, model behavior is highly sensitive to the language used in the reasoning process \citep{wang-etal-2025-language-mixing,qi-etal-2025-models,yong2025crosslingualreasoningtesttimescaling,tam2025languagemattersmultilingualinput}.  
These performance gaps have been attributed to disparities in the quality of language-specific reasoning traces \citep{zhao2025comprehensiveevaluationmultilingualchainofthought} and to failures in basic understanding for low-resource inputs \citep{kang2025multilingualreasoninggapsemerge,bafna2025translationbarrierhypothesismultilingual}.  
However, existing studies focus almost exclusively on \emph{explicit} multilingual reasoning behavior.
Our work aims to investigate \emph{latent} reasoning across languages systematically.

\textbf{Implicit Latent Reasoning}
Unlike \emph{explicit} reasoning, where models produce step-by-step textual explanations, implicit or \emph{latent} reasoning refers to the internal computation that occurs in the model's hidden representations \citep{cheng2024compressedchainthoughtefficient,li202512surveyreasoning,chen2025reasoninglanguagecomprehensivesurvey}. 
Prior work shows that LLMs may pursue multiple latent reasoning paths in parallel, gradually increasing confidence in a particular solution as explicit reasoning unfolds \citep{Prystawski2023stepbystep,Dutta2024step,qian2025demystifyingreasoningdynamicsmutual}.  
Yet even when the model has internally formed the correct answer, it may continue generating unnecessary reasoning steps, referred to as ``overthinking’’  \citep{chen2025think23overthinkingo1like,sui2025overthink}.
Motivated by these observations, there have been new approaches aiming to train models to reason directly in latent space without producing full textual traces \citep{deng2024explicitcotimplicitcot,hao2025traininglargelanguagemodels,lin-etal-2025-implicit,Saunshi2025reasoning,xu-etal-2025-softcot}.
However, this line of work focuses almost exclusively on English, leaving open whether latent reasoning behaviors emerge across languages.  
Our work addresses this gap by systematically evaluating and comparing latent reasoning dynamics in a multilingual setting.

\section{Experimental Setup}

\subsection{Models}

We use three distilled variants of \texttt{DeepSeek-R1} \citep{deepseekai2025deepseekr1incentivizingreasoningcapability}, \texttt{DeepSeek-R1-Distill-Qwen-\{7B, 14B, 32B\}}, whose backbone models are based on the \texttt{Qwen2.5} family \citep{qwen2025qwen25technicalreport}.  
These models are selected because they exhibit strong
reasoning performance while providing multiple sizes,
enabling us to analyze how multilingual latent reasoning varies with model capacity.

\subsection{Datasets and Languages}

\textbf{MGSM} 
Multilingual Grade School Math dataset \citep{mgsm} contains 250 grade-school math problems sourced from GSM8K \citep{cobbe2021trainingverifierssolvemath}, originally written in English and manually translated into 10 additional languages:  French (FR), German (DE), Chinese (ZH), Japanese (JA), Russian (RU), Spanish (ES), Swahili (SW), Bengali (BN), Telugu (TE), and Thai (TH).
Since the underlying mathematical problems are identical across languages, it is well-suited for studying multilingual (latent) reasoning dynamics.

\textbf{Multilingual AIME}  
The Multilingual American Invitational Mathematics Examination (AIME) datasets are translated versions of AIME2024 and AIME2025 introduced by \citet{qi-etal-2025-models}, covering the same 11 languages as MGSM.  
These datasets contain substantially more challenging, competition-level math problems, enabling us to examine how increased problem difficulty influences multilingual latent reasoning dynamics.

We categorize the languages considered in this study into \textbf{high-resource} (EN, ES, DE, FR, RU, ZH), \textbf{mid-resource} (BN, JA, TH), and \textbf{low-resource} (SW, TE) groups.
This categorization is based on the relative availability of large-scale training resources and the degree of language coverage in contemporary multilingual LLMs \citep{joshi-etal-2020-state,blasi-etal-2022-systematic,xu2025survey}

\subsection{Language Control}

LLMs may generate explicit reasoning traces in a language different from that of the prompt \citep{wang-etal-2025-language-mixing,qi-etal-2025-models}, which is undesirable for crosslingual analysis of latent reasoning.
To ensure that explicit reasoning is produced in the same language as the input, we employ a \emph{prompt-hacking strategy} \citep{qi-etal-2025-models,zhao2025comprehensiveevaluationmultilingualchainofthought} that inserts a language-specific prefix immediately after the \texttt{<think>} token, reliably steering the reasoning trace to the target language (see \secref{language_control} for details).

\section{Latent Reasoning Identification}\seclabel{surface_dynamics}

To address \textbf{RQ1}: \emph{Do LRMs exhibit latent reasoning across languages, and how does the strength vary?}, we analyze the model's early predictions under reasoning-trace truncation. 
This protocol connects the \emph{explicit} reasoning process with the model's \emph{internal} answer construction: if the model already knows the answer early in the trace, it should often answer correctly even when only a small portion of the reasoning is visible. 
This method is similar to concurrent work on early stopping and stepwise answer prediction \citep{mao2025earlystoppingchainofthoughtslarge,wang2025thinkingcheatingdetectingimplicit,zhao2025comprehensiveevaluationmultilingualchainofthought,chen2026decouplingeffectchainofthoughtreasoning}, but we leverage such truncation to identify latent reasoning and complement it with novel metrics that quantify latent reasoning capability across languages.

\subsection{Truncating Reasoning Traces}\seclabel{truncation_method}

Let $x$ denote a math problem and let the model produce a full reasoning trace $c = (t_1, t_2, \dots, t_T)$ in the target language, followed by a final answer, where $t_i$ indicates the $i$-th reasoning step.\footnote{The reasoning trace is regarded as the tokens between special thinking markers, e.g., \texttt{<think>} and \texttt{</think>}.
We view each individual sentence as a reasoning step.
}
We then consider a set of truncation ratios
$
\mathcal{R} = \{ r_1, r_2, \dots, r_M \} \subset [0,1]
$,
where each $r \in \mathcal{R}$ specifies the fraction of the reasoning trace that is retained (e.g., 10\%).
For a ratio $r$, we define the truncation index
as $
m(r) = \bigl\lfloor r \cdot T \bigr\rfloor
$
and the truncated reasoning trace as
$
c_{\le r} = (t_1, t_2, \dots, t_{m(r)})
$.
We then ask the model to directly produce a numerical prediction based on the original math problem $x$ and the truncated reasoning trace $c_{\le r}$.\footnote{This is achieved by adding \texttt{</think>} right after the truncated reasoning trace and then appending a short prefix to elicit the numerical answer prediction (see \secref{language_control} for details).}

\subsection{Evaluation Metrics}
\seclabel{truncation_metrics}


We evaluate performance over a set of truncation ratios $r \in \mathcal{R}$ using the following metrics.\footnote{For MGSM, we consider every 10\%, i.e., $\mathcal{R}=\{0\%, 10\%, 20\%, \dots, 100\%\}$. For Multilingual AIME, we consider every 5\%, i.e., $\mathcal{R}=\{0\%, 5\%, 10\%, \dots, 100\%\}$. The choice is based on a preliminary analysis of the average number of steps across languages (see \secref{trace_statistics} for details).}

\begin{figure*}
    \centering
    \setlength{\abovecaptionskip}{-0.03cm}
    \setlength{\belowcaptionskip}{-0.5cm}
    \includegraphics[width=0.4\columnwidth]{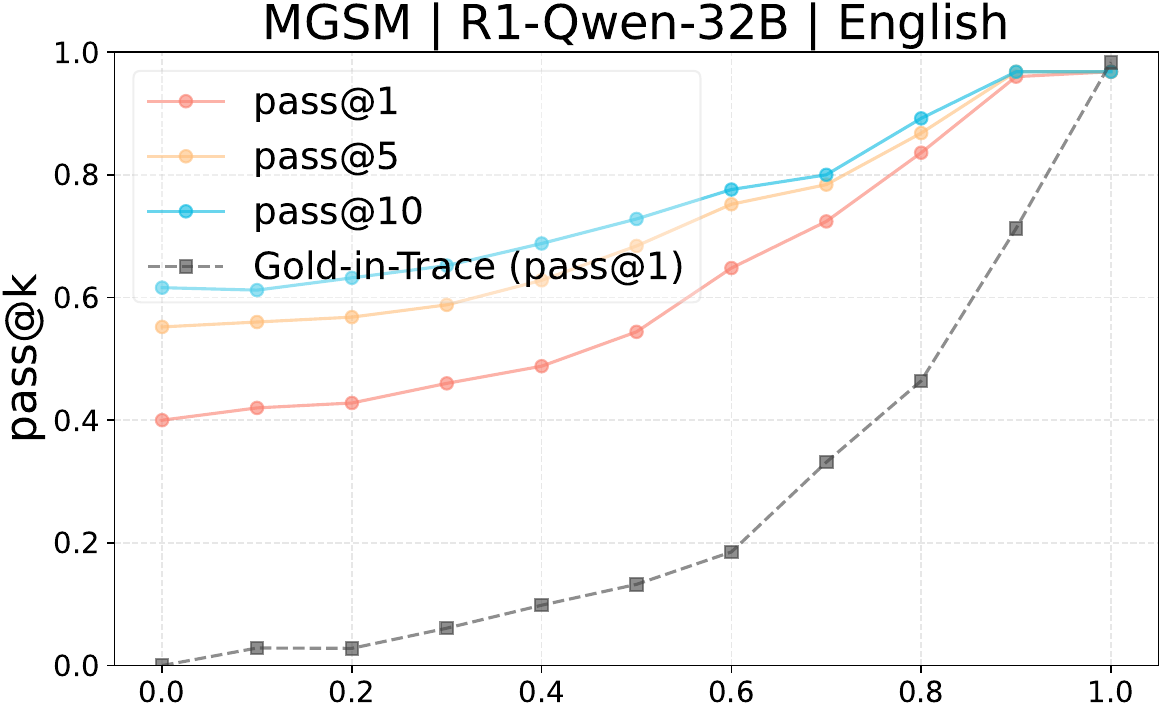}
    \includegraphics[width=0.4\columnwidth]{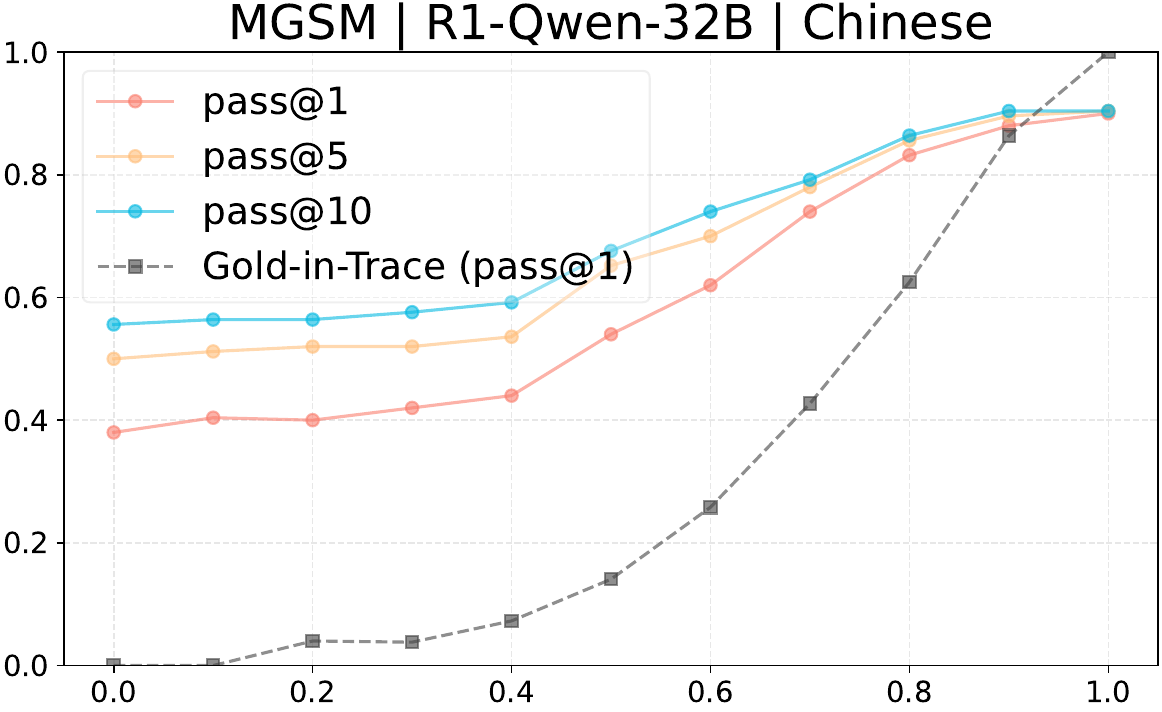}
    \includegraphics[width=0.4\columnwidth]{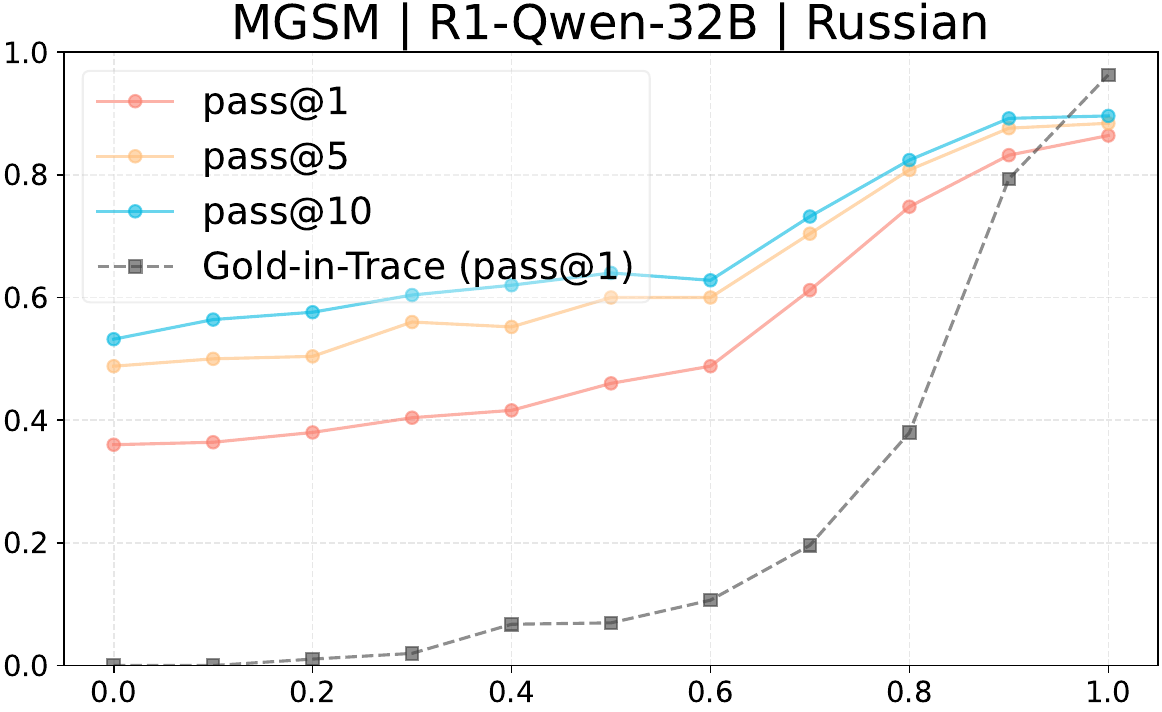}
    \includegraphics[width=0.4\columnwidth]{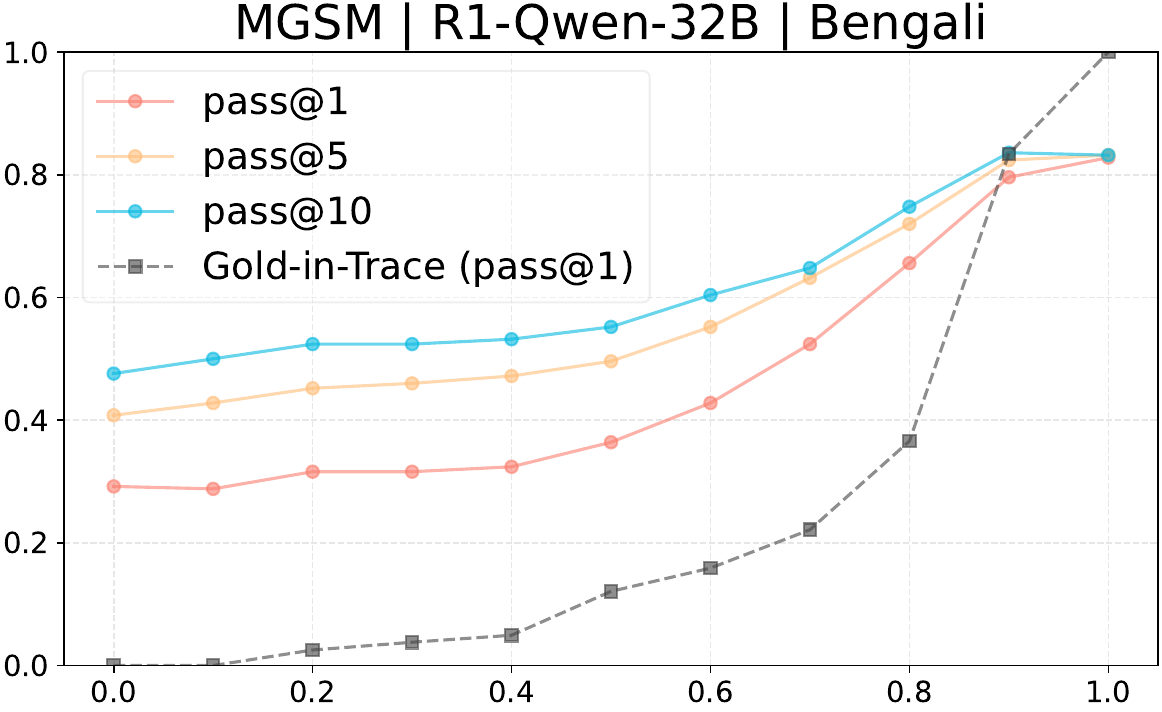}
    \includegraphics[width=0.4\columnwidth]{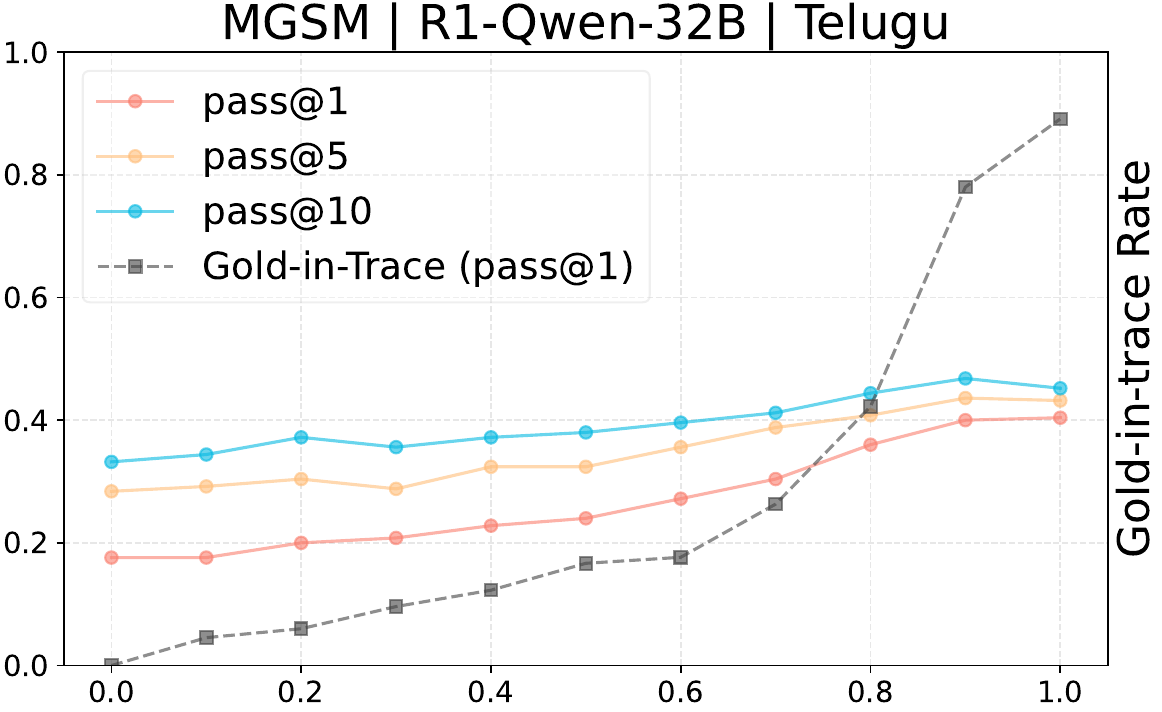}
    \includegraphics[width=0.4\columnwidth]{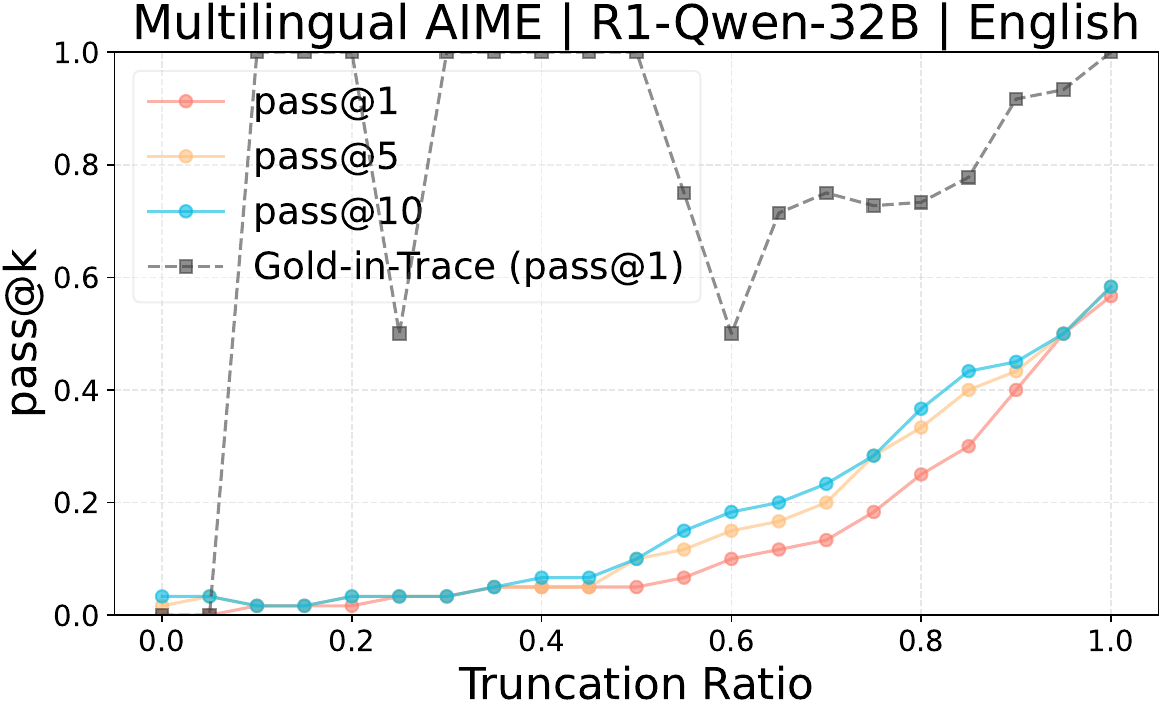}
    \includegraphics[width=0.4\columnwidth]{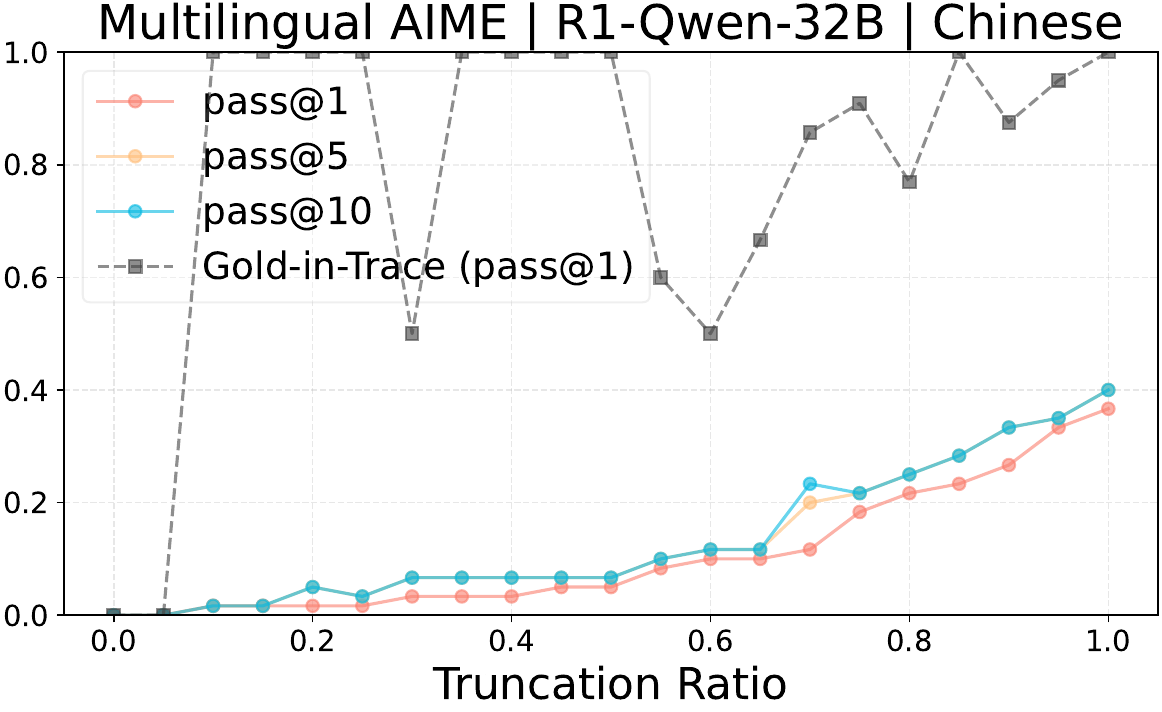}
    \includegraphics[width=0.4\columnwidth]{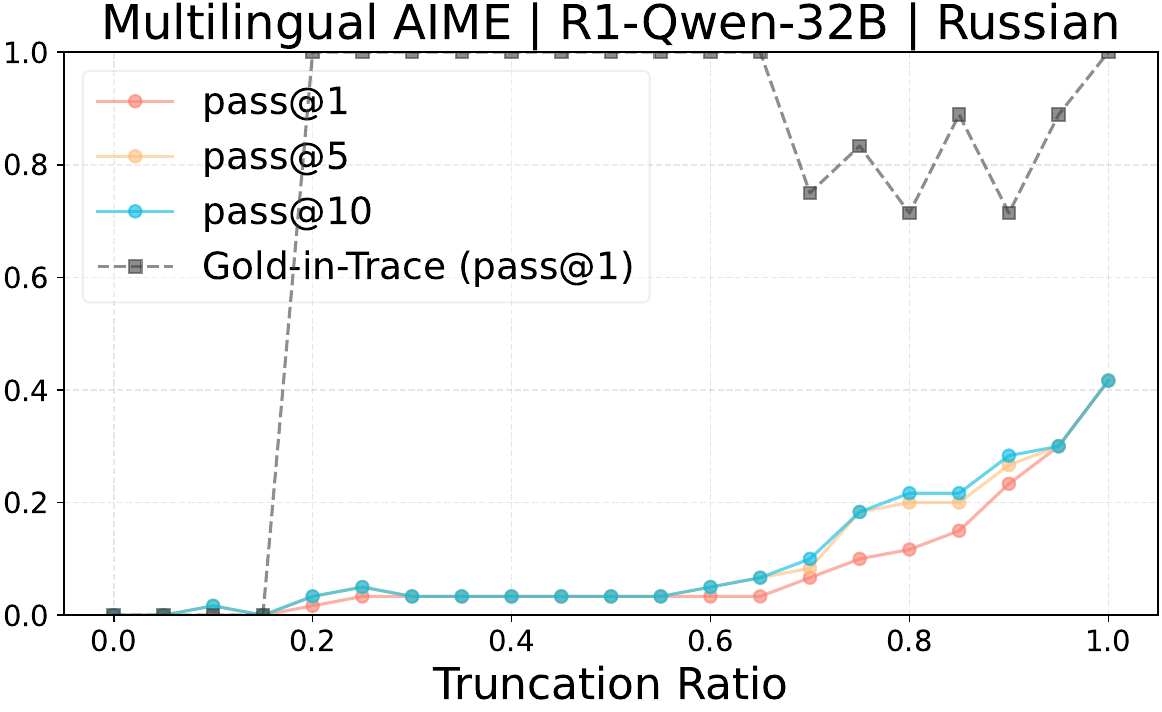}
    \includegraphics[width=0.4\columnwidth]{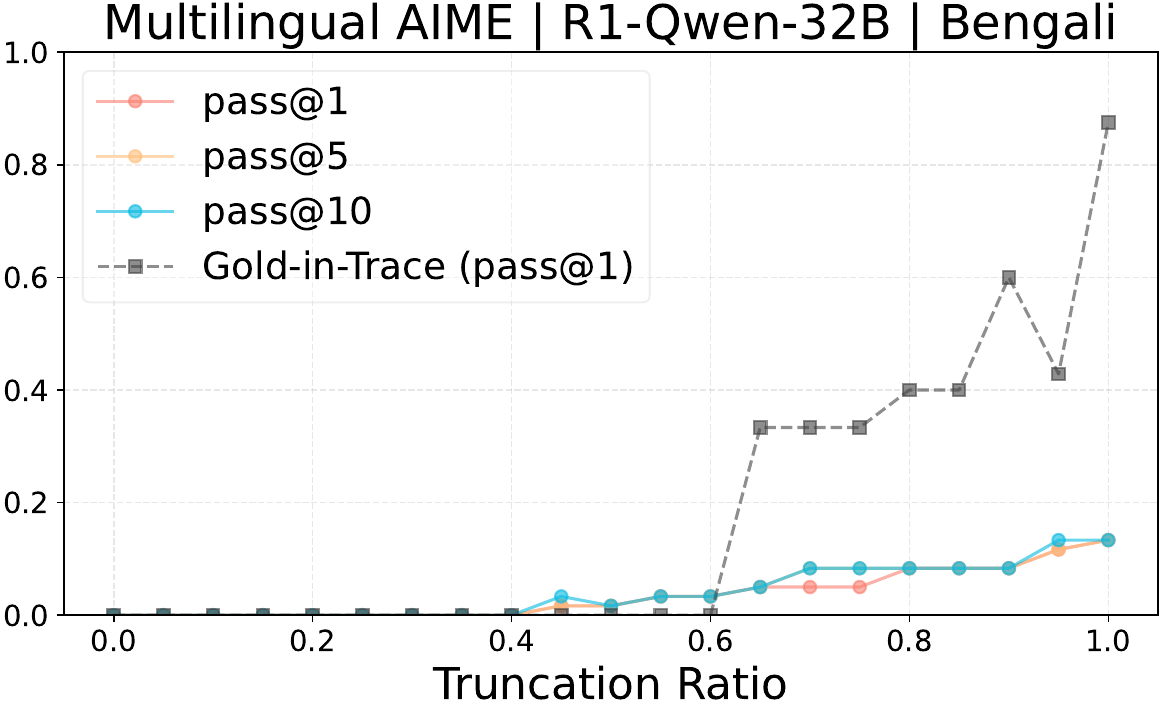}
    \includegraphics[width=0.4\columnwidth]{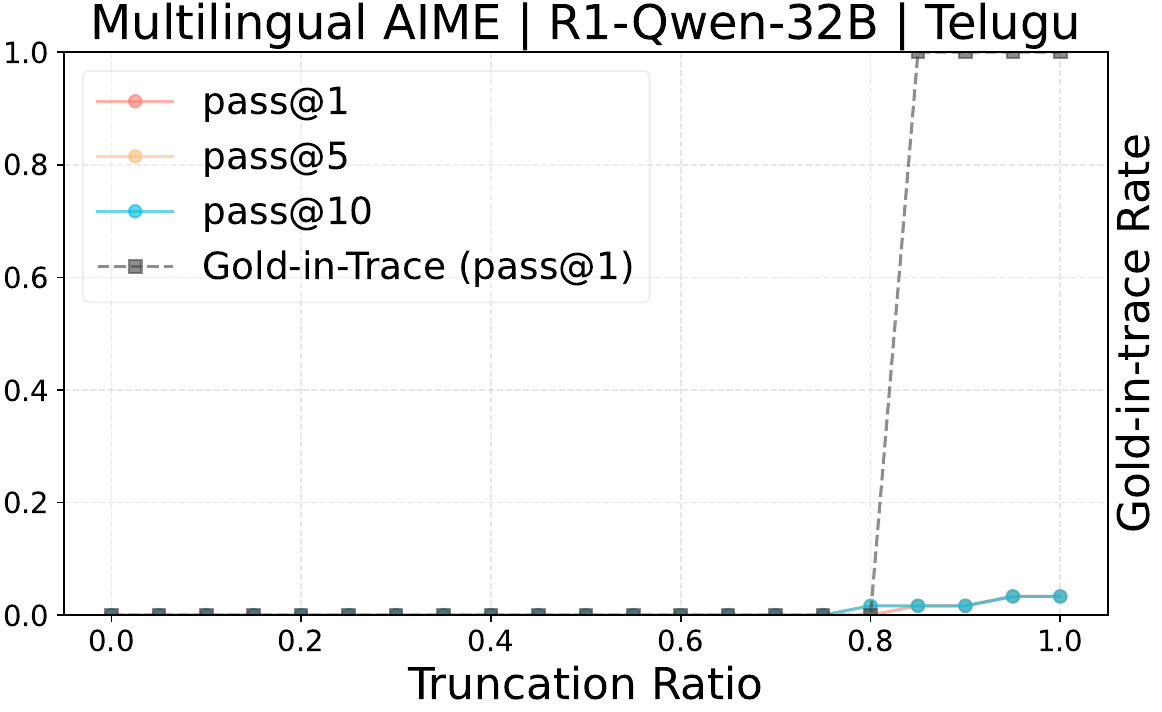} 
    \vspace{0.4em}
    \caption{Pass@$k$ accuracy ($k=1,5,10$) and gold-in-trace rate under reasoning-trace truncation for R1-Qwen-32B. 
    High accuracy with a low gold-in-trace rate indicates latent reasoning.
    The model shows strong evidence of latent reasoning in high-resource languages (e.g., English) on MGSM, but it is less detectable on 
    Multilingual AIME.}
    \label{fig:truncation_32b}
\end{figure*}

\enote{hs}{figure 1: aime results for english, chinese,
  russian seem puzzling: why are the ``pass'' results so bad
  even though the answer is already present in the trace?}

\enote{yl}{the gold-in-trace rate only considers the questions which are correctly answered. in figure 1, it is gold-in-trace (pass@1), so the number would read as ``out of the correctly answered questions under pass@1, how many of them already contain the gold answer so far in the reasoning trace''.
i bold the important text describing the definition in the gold-in-trace rate paragraph.
}

\begin{figure}
    \centering
    \setlength{\belowcaptionskip}{-0.5cm}
    \includegraphics[width=0.32\columnwidth]{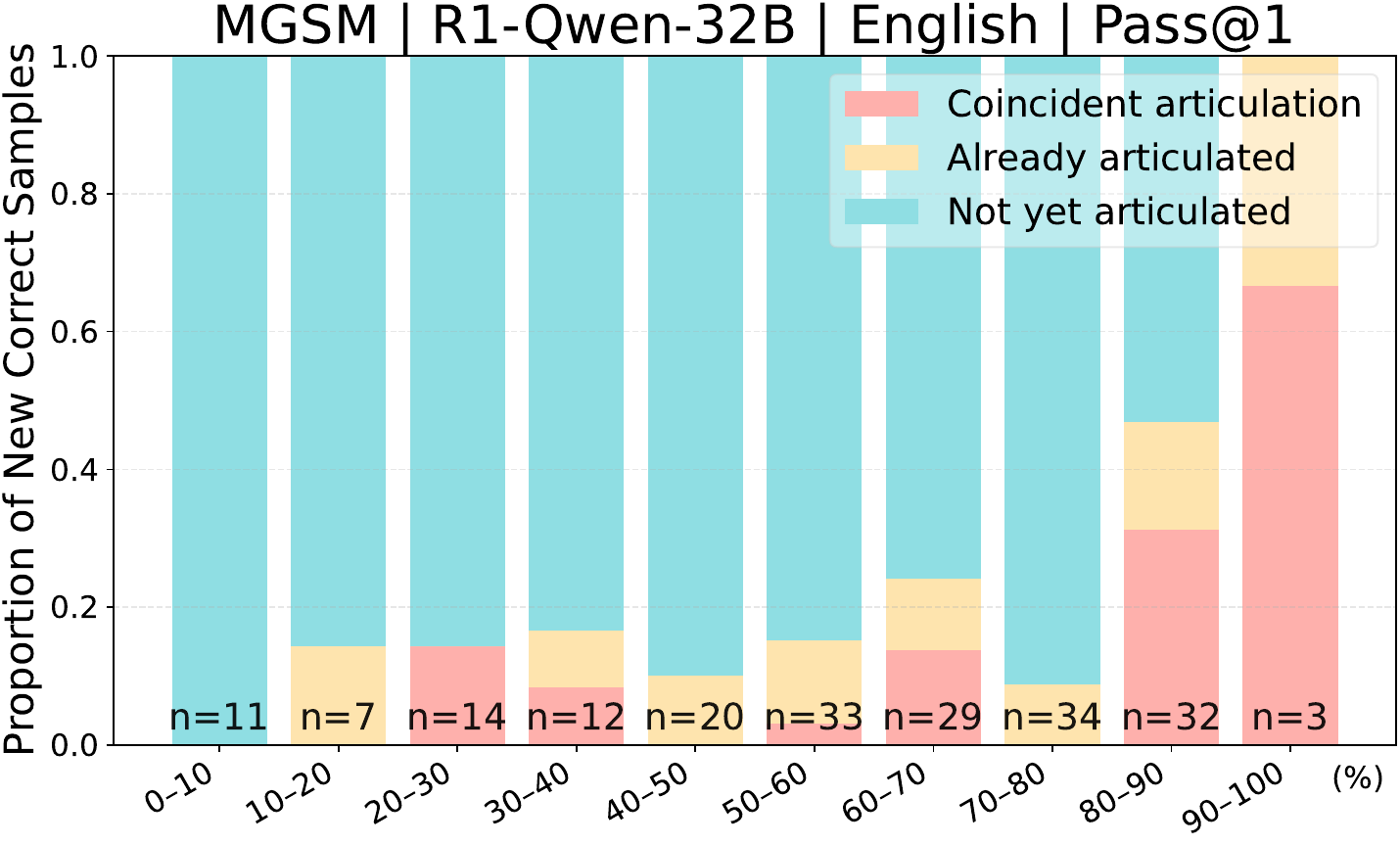}
    \includegraphics[width=0.32\columnwidth]{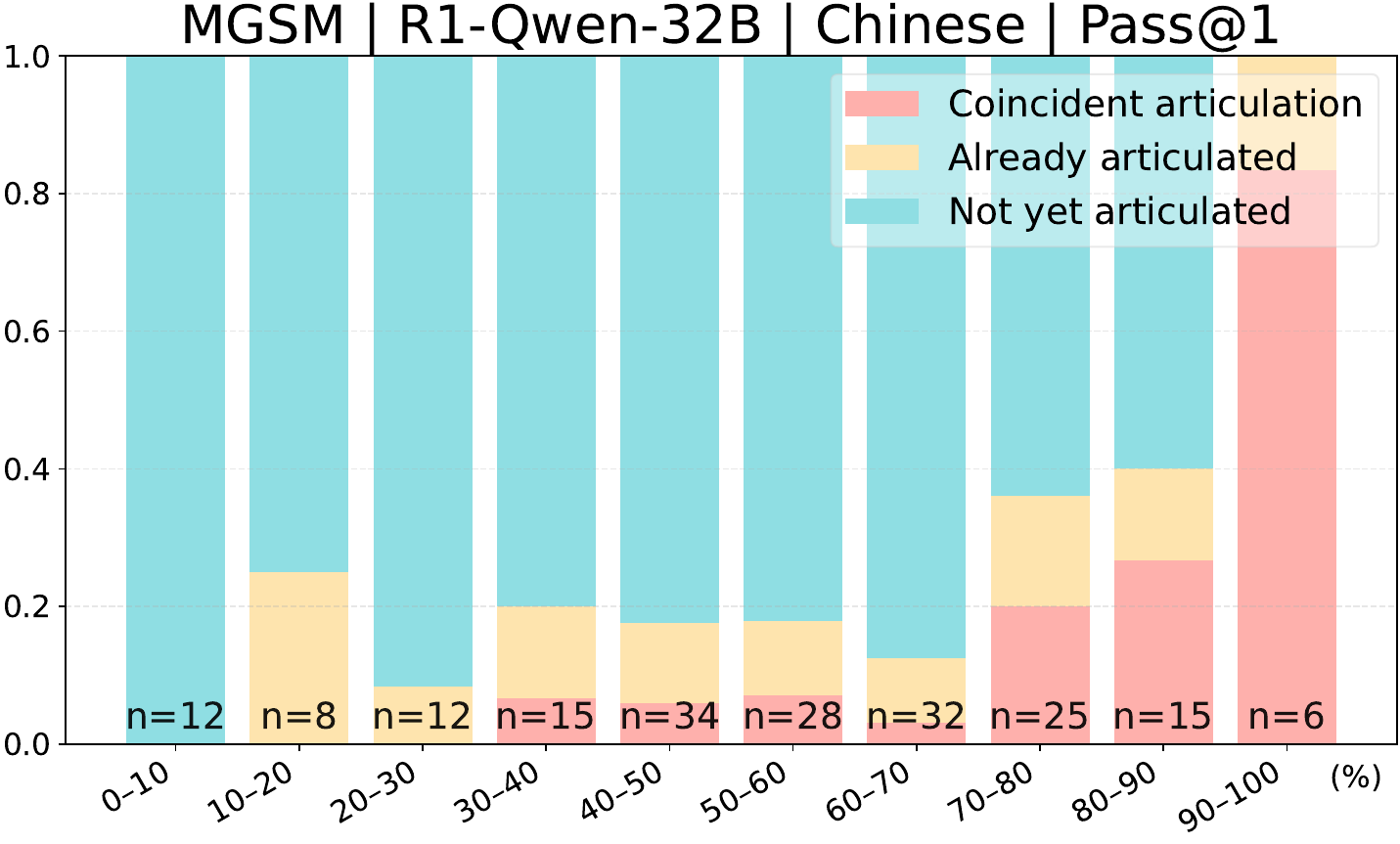}
    \includegraphics[width=0.32\columnwidth]{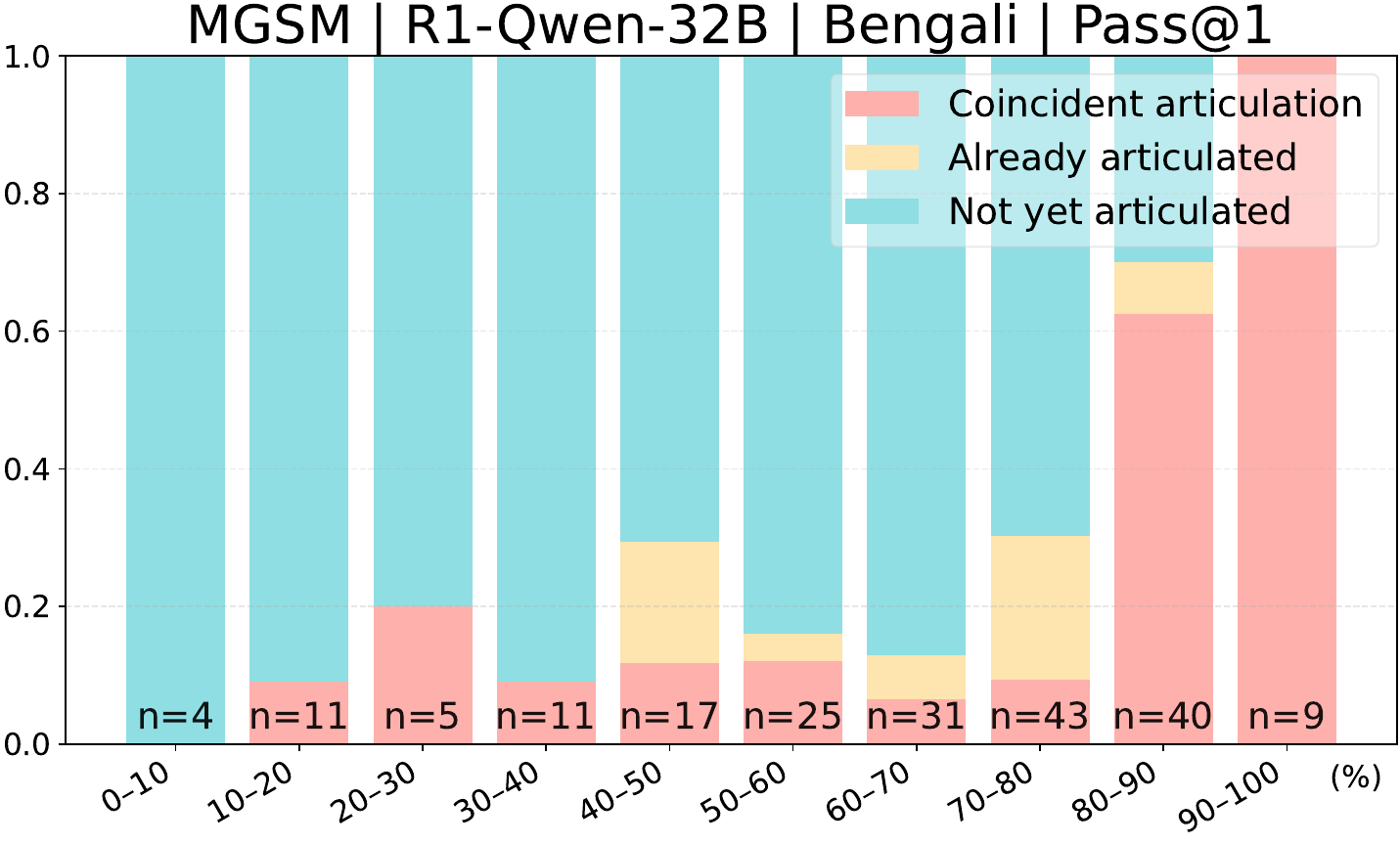}
    \includegraphics[width=0.32\columnwidth]{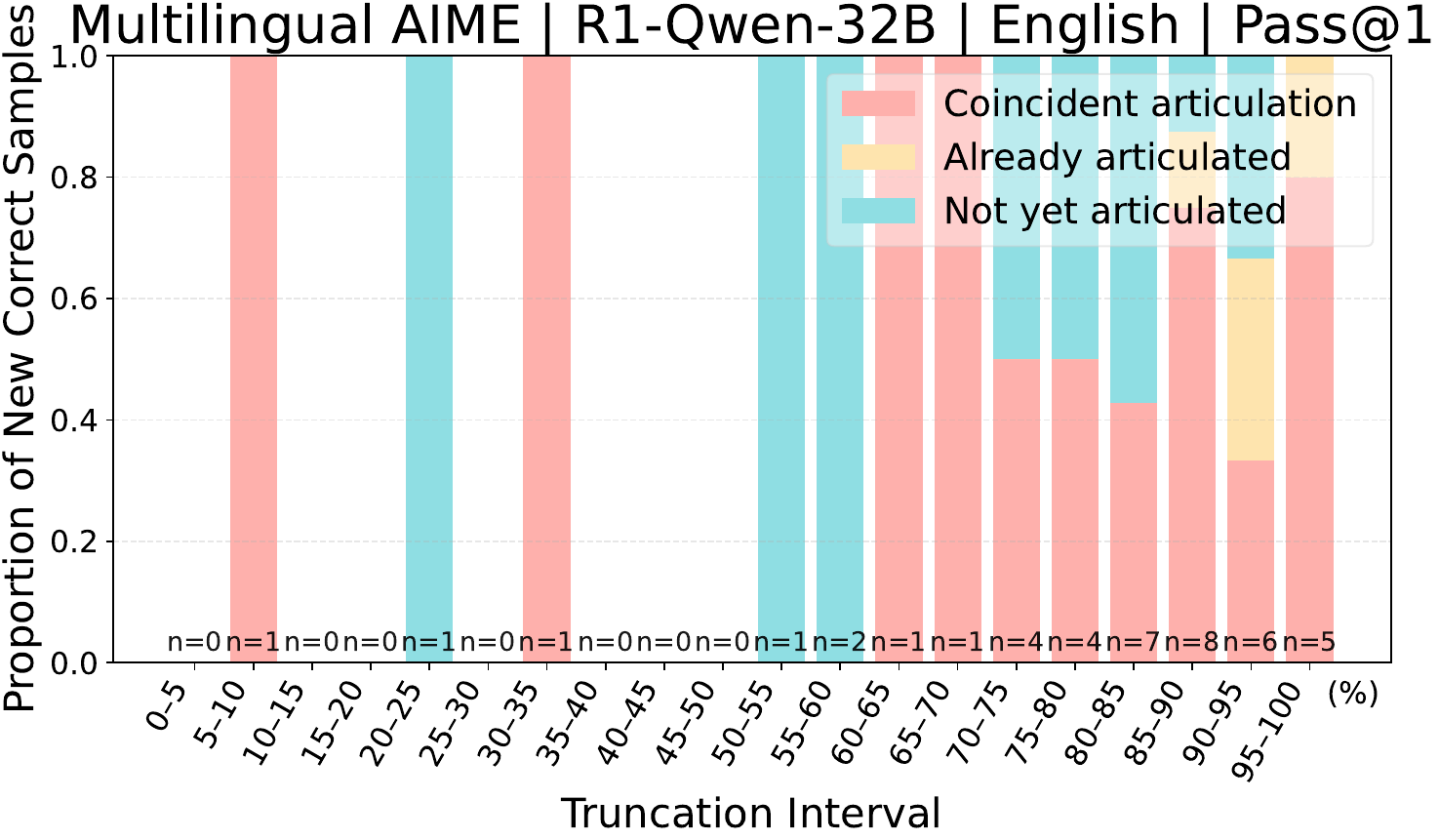}
    \includegraphics[width=0.32\columnwidth]{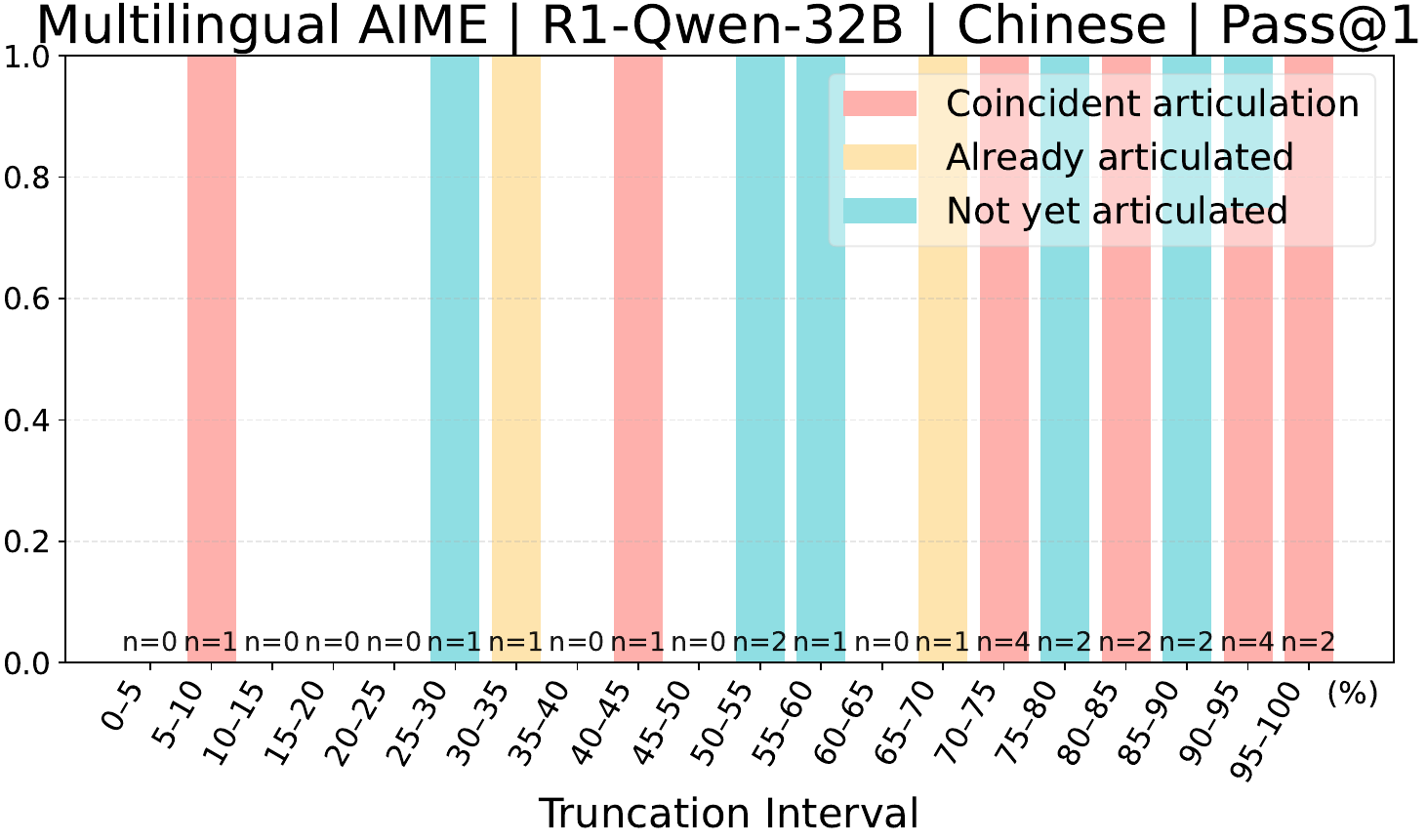}
    \includegraphics[width=0.32\columnwidth]{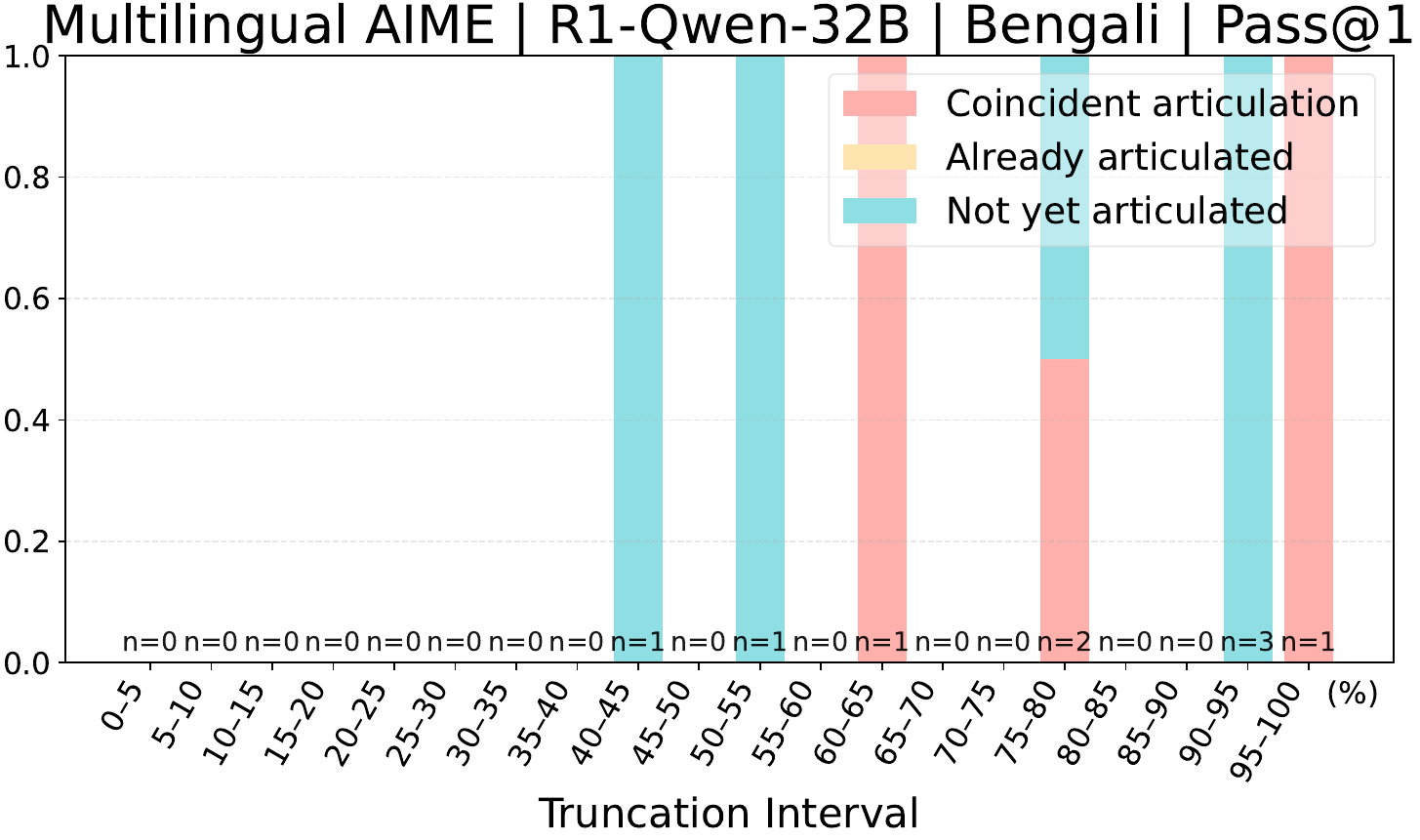}
    \caption{
    Causal decomposition of newly correct predictions across truncation intervals.
    Each bar partitions gains into three cases: (\textbf{i}) the gold answer is first articulated in the newly added reasoning steps, (\textbf{ii}) it was already articulated in earlier steps, or (\textbf{iii}) it has not yet appeared in the visible truncated trace.
    On MGSM, performance improvements at early and intermediate truncation ratios are dominated by case (iii), indicating that many gains arise from latent reasoning.
    }
    \label{fig:interval}
\end{figure}

\textbf{Truncated Pass@$k$.}
This metric estimates the probability that at least one correct answer appears among the top-$\boldsymbol{k}$ attempts for a given problem \citep{Kulal2019passk,chen2021evaluatinglargelanguagemodels}.
Let $a_k(r)$ denote the pass@$k$ accuracy at truncation ratio $r$, i.e.,
{\setlength{\abovedisplayskip}{0.1cm}%
 \setlength{\belowdisplayskip}{0.1cm}%
\[
a_k(r) = \frac{1}{N} \sum_{i=1}^{N} \mathbf{1}\!\left[\,\exists j \le k : \hat{y}^{(i)}_j(r) = y^{(i)\star}\,\right]
\]}\noexpand
where $N$ is the number of problems, $\hat{y}^{(i)}_j(r)$ is the $j$-th sampled prediction for problem $i$ based on the truncated reasoning trace $c^{(i)}_{\le r}$, and $y^{(i)\star}$ is the gold answer.
This metric measures performance under partial reasoning: if $a_k(r)$ is high even for small $r$, the model may not rely on fully explicit reasoning traces and instead perform latent reasoning.

\enote{hs}{I don't understand the definition of
  $a_k(r)$. For a given problem $i$, I would think that
  $c^{(i)}_{\le r}$ (which does not have an index $j$) is
  the same for all $j$ and so
  $\hat{y}^{(i)}_j(r)$ is the same for all $j$?}

\enote{yl}{yes, for each question we only generate one single reasoning trace, and we perform truncation on this.
therefore, $c^{(i)}_{\le r}$ (which does not have an index $j$) is the same for all $j$.
this $\hat{y}^{(i)}_j(r)$ is the $j$th sampled prediction from the current truncated trace (i.e., $c^{(i)}_{\le r}$) (in total $k$ trials) -- since we are doing sampling, they will not be the same (but can be if the model is very confident).
}

\textbf{Gold-in-Trace Rate.}
There are cases where the model explicitly articulates the answer in early reasoning steps, and then continues to refine it or explore additional paths in later steps.
In such cases, correct predictions 
may depend on the explicitly written answer.
To distinguish these cases, we additionally track whether the gold answer already appears in the visible reasoning prefix $c_{\le r}$.
We define the gold-in-trace rate at truncation ratio $r$ as
{\setlength{\abovedisplayskip}{0.1cm}%
 \setlength{\belowdisplayskip}{0.1cm}%
\[
g_k(r) =
\frac{1}{\lvert \mathcal{C}_k(r) \rvert}
\sum_{i \in \mathcal{C}_k(r)}
\mathbf{1}\!\left[\, y^{(i)\star} \text{ appears in } c^{(i)}_{\le r} \,\right]
\]
}\noexpand
Here, $\mathcal{C}_k(r)$ denotes \textbf{the set of correctly solved instances} under truncation ratio $r$ according to pass@$k$.
Importantly, a high
gold-in-trace rate is expected at large truncation ratios (e.g., $r \approx 1$), where the full reasoning trace should usually contain the final answer. 
Thus, gold-in-trace is primarily informative at \emph{small} truncation ratios: a high value early in the trace suggests that correctness may be driven by explicit answer articulation, whereas a low value indicates that correct predictions are more likely supported by latent reasoning.

\textbf{Area Under the Truncation Accuracy Curve (AUTC).}
We define the \textbf{AUTC} as
{\setlength{\abovedisplayskip}{0.1cm}%
 \setlength{\belowdisplayskip}{0.1cm}%
\[
\text{AUTC}_k = \int_0^1 a_k(r)\,dr.
\]
}\noexpand
A model that reaches high accuracy early (i.e., needs only a short prefix of the trace) will yield a larger AUTC than a model whose accuracy only improves near $r \approx 1$.
AUTC is thus a
measure of \emph{how early and robustly} correct predictions emerge as more reasoning is revealed.

\textbf{Area Under the Gold-in-Trace Curve (AUGC).}
Analogously, we define the \textbf{AUGC} as
{\setlength{\abovedisplayskip}{0.1cm}%
 \setlength{\belowdisplayskip}{0.1cm}%
\[
\text{AUGC}_k = \int_0^1 g_k(r)\,dr
\]
}\noexpand
A high AUGC indicates that, when the model is correct, the gold answer tends to be articulated early, while a low AUGC indicates that the gold answer usually appears near the end of the trace.

\begin{table*}[t]
\setlength{\belowcaptionskip}{-0.5cm}
\centering
\tiny
\renewcommand{\arraystretch}{0.55} 
\resizebox{\linewidth}{!}{
\begin{tabular}{lll|rrrrrr|rrr|rr}
\toprule
Dataset & Model & Metric
& DE & EN & ES & FR & RU & ZH 
& BN & JA & TH 
& SW & TE \\
\midrule
\multirow{9}{*}{MGSM} 
& \multirow{3}{*}{R1-Qwen-7B} 
& AUTC & 0.45 & 0.52 & 0.45 & 0.43 & 0.46 & 0.53 & 0.38 & 0.38 & 0.37 & 0.10 & 0.24 \\
&  & AUGC & 0.22 & 0.21 & 0.16 & 0.22 & 0.24 & 0.27 & 0.27 & 0.24 & 0.33 & 0.19 & 0.32 \\
&  & LRS  & 0.32 & 0.38 & 0.35 & 0.32 & 0.31 & 0.34 & 0.25 & 0.26 & 0.22 & 0.08 & 0.15 \\
\cmidrule(lr){2-14}
& \multirow{3}{*}{R1-Qwen-14B} 
& AUTC & 0.54 & 0.59 & 0.59 & 0.55 & 0.58 & 0.62 & 0.51 & 0.55 & 0.57 & 0.22 & 0.28 \\
&  & AUGC & 0.20 & 0.19 & 0.20 & 0.18 & 0.22 & 0.26 & 0.27 & 0.24 & 0.27 & 0.25 & 0.26 \\
&  & LRS  & 0.40 & 0.44 & 0.44 & 0.42 & 0.41 & 0.41 & 0.33 & 0.39 & 0.36 & 0.16 & 0.20 \\
\cmidrule(lr){2-14}
& \multirow{3}{*}{R1-Qwen-32B} 
& AUTC & 0.67 & 0.75 & 0.69 & 0.64 & 0.68 & 0.70 & 0.61 & 0.63 & 0.69 & 0.38 & 0.39 \\
&  & AUGC & 0.20 & 0.25 & 0.20 & 0.17 & 0.21 & 0.30 & 0.23 & 0.21 & 0.28 & 0.20 & 0.23 \\
&  & LRS  & 0.51 & 0.53 & 0.52 & 0.51 & 0.51 & 0.45 & 0.44 & 0.47 & 0.46 & 0.30 & 0.30 \\
\midrule
\multirow{9}{*}{Multilingual AIME} 
& \multirow{3}{*}{R1-Qwen-7B} 
& AUTC & 0.07 & 0.10 & 0.06 & 0.05 & 0.06 & 0.09 & 0.04 & 0.02 & 0.02 & 0.00 & 0.01 \\
&  & AUGC & 0.52 & 0.51 & 0.19 & 0.23 & 0.55 & 0.60 & 0.57 & 0.12 & 0.17 & 0.00 & 0.00 \\
&  & LRS  & 0.02 & 0.03 & 0.03 & 0.03 & 0.02 & 0.02 & 0.01 & 0.01 & 0.01 & 0.00 & 0.01 \\
\cmidrule(lr){2-14}
& \multirow{3}{*}{R1-Qwen-14B} 
& AUTC & 0.05 & 0.12 & 0.07 & 0.07 & 0.05 & 0.08 & 0.08 & 0.02 & 0.05 & 0.00 & 0.04 \\
&  & AUGC & 0.66 & 0.44 & 0.52 & 0.25 & 0.41 & 0.79 & 0.70 & 0.06 & 0.29 & 0.00 & 0.08 \\
&  & LRS  & 0.02 & 0.04 & 0.03 & 0.04 & 0.02 & 0.01 & 0.01 & 0.02 & 0.02 & 0.00 & 0.04 \\
\cmidrule(lr){2-14}
& \multirow{3}{*}{R1-Qwen-32B} 
& AUTC & 0.06 & 0.18 & 0.08 & 0.09 & 0.10 & 0.13 & 0.04 & 0.04 & 0.07 & 0.01 & 0.01 \\
&  & AUGC & 0.29 & 0.61 & 0.32 & 0.72 & 0.66 & 0.75 & 0.18 & 0.74 & 0.82 & 0.05 & 0.17 \\
&  & LRS  & 0.03 & 0.06 & 0.04 & 0.02 & 0.03 & 0.03 & 0.02 & 0.01 & 0.02 & 0.00 & 0.00 \\
\bottomrule
\end{tabular}
}
\caption{
Truncation-based metrics (AUTC, AUGC, LRS) across models and benchmarks.
Latent reasoning capability scales with model size and language resource availability,
but emerges primarily on the simpler MGSM benchmark and is largely undetectable on the more challenging benchmark, Multilingual AIME.
}
\label{tab:truncation_metrics}
\end{table*}

\textbf{Latent Reasoning Score (LRS).}
To focus on correctness that is not trivially attributable to copying the answer from the trace, we define \textbf{LRS} as
{\setlength{\abovedisplayskip}{0.1cm}%
 \setlength{\belowdisplayskip}{0.1cm}%
\[
\text{LRS}_k = \int_0^1 a_k(r)\,\bigl(1 - g_k(r)\bigr)\,dr.
\]}\noexpand
Intuitively, we weight performance at each truncation ratio
by the complement of the gold-in-trace rate: correctness
that occurs \emph{after} the answer is already articulated
in the trace (high $g_k(r)$) is downweighted, while
correctness that occurs \emph{before} the answer is visible
(low $g_k(r)$) is upweighted.
Therefore, LRS can be regarded as a proxy measure for the model's \emph{latent reasoning capability}.

We approximate AUTC, AUGC, and LRS numerically using the trapezoidal rule~\citep{hildebrand1987introduction} over all considered truncation ratios $r \in\mathcal{R}$.

\subsection{Results and Discussion}\seclabel{truncation_discussion}

Figure~\ref{fig:truncation_32b} presents truncation curves for R1-Qwen-32B across 5 languages on two benchmarks (see \secref{truncation_results} for full results).  
Table~\ref{tab:truncation_metrics} summarizes the corresponding AUTC, AUGC, and LRS scores across all 11 languages, models, and datasets.  
Finally, Figure~\ref{fig:interval} breaks down newly correct predictions in each truncation ratio interval by whether their gold answers are articulated in the newly added reasoning steps, already appear in earlier steps, or do not appear in the current truncated trace at all.

\textbf{The model often knows the answer even before any reasoning is articulated.}
Across many high-resource languages -- most notably English, French, and Chinese, the pass@1 accuracy at \emph{zero} reasoning steps is already nontrivial (around 0.2).  
This suggests that for MGSM, the model can frequently compute the answer \emph{directly in its latent representations}, without requiring explicit step-by-step CoT generation.  
As the truncation ratio increases, accuracy rises steadily in all languages, accompanied by a growing gold-in-trace rate that typically approaches 1.0 once the full trace is revealed.
Figure~\ref{fig:interval} further supports this observation: most early correct predictions do not depend on the articulation of the gold answer in the visible trace.
Together, these findings suggest that explicit chain-of-thought primarily serves to \emph{surface} an answer that has already been internally computed: latent reasoning precedes explicit verbal reasoning.


\textbf{Latent reasoning is substantially stronger in high-resource languages.}
Comparisons across languages using AUTC and LRS reveal clear multilingual disparities.  
For MGSM, high-resource languages such as English and Chinese obtain both high AUTC and high LRS.  
For example, English achieves AUTC 0.52 and LRS 0.38 with R1-Qwen-7B, indicating that a large fraction of early accuracy cannot be explained by explicit answer articulation.
In contrast, low-resource languages such as Swahili 
show much lower AUTC and LRS, meaning that the model struggles to produce correct answers under truncation and relies more heavily on fully articulated reasoning traces.  
Increasing model size from 7B to 32B seems to improve AUTC and LRS in all languages, but does \emph{not} eliminate the gap: latent reasoning remains markedly less effective in low-resource languages.  
Overall, latent reasoning is a strongly resource-dependent phenomenon.


\textbf{Latent reasoning is less pronounced on more challenging benchmarks.}
On Multilingual AIME, both AUTC and LRS drop sharply across languages and model sizes compared to MGSM.
For example, LRS decreases from about 0.38 on MGSM to 0.03 on Multilingual AIME for English with R1-Qwen-7B, with similar trends in other languages and larger models.
This pattern indicates that for problems requiring longer, more complex reasoning, models rarely form correct predictions early, prior to explicit answer articulation, and instead rely more heavily on extended explicit reasoning.

\enote{hs}{above: I think you make a strong case that
  explicit reasoning is needed and that the models make
  extesnive use of it. But what is your evidence for the
  claim that latent reasoning is not used?}
\enote{yl}{very good point. we actually do not say that latent reasoning is not used at all here -- the model can still perform latent reasoning in harder benchmarks but the issue is that it seems to be less effective, as the model cannot achieve early decent accuracy.
i guess we should say something like latent reasoning becomes less detectable or less observable given our evidence.
i changed a few places in the paper where we say latent reasoning is "not there". 
}

\begin{figure*}
    \centering
    \setlength{\belowcaptionskip}{-0.3cm}
    \includegraphics[width=0.16\textwidth]{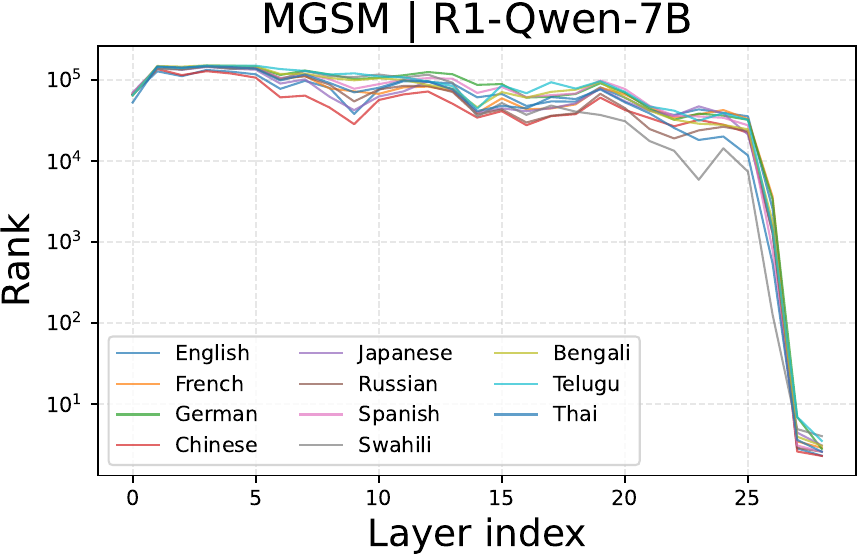}
    \includegraphics[width=0.16\textwidth]{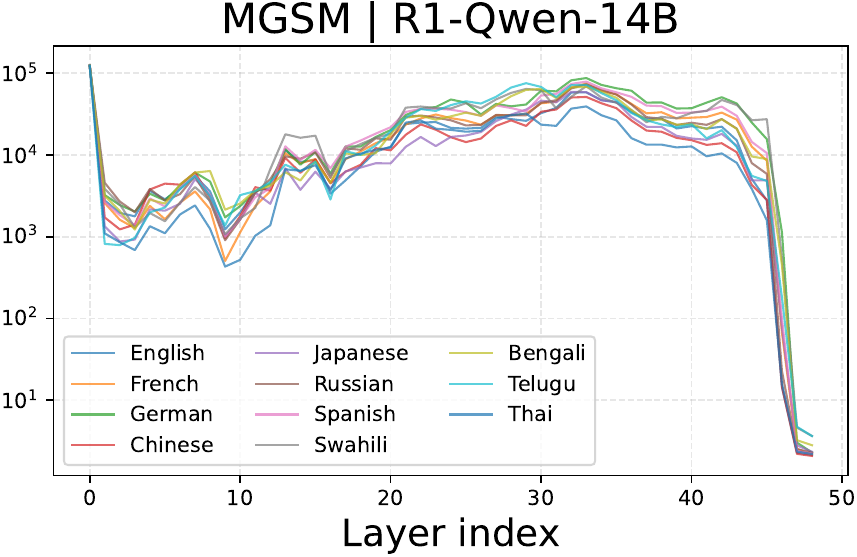}
    \includegraphics[width=0.16\textwidth]{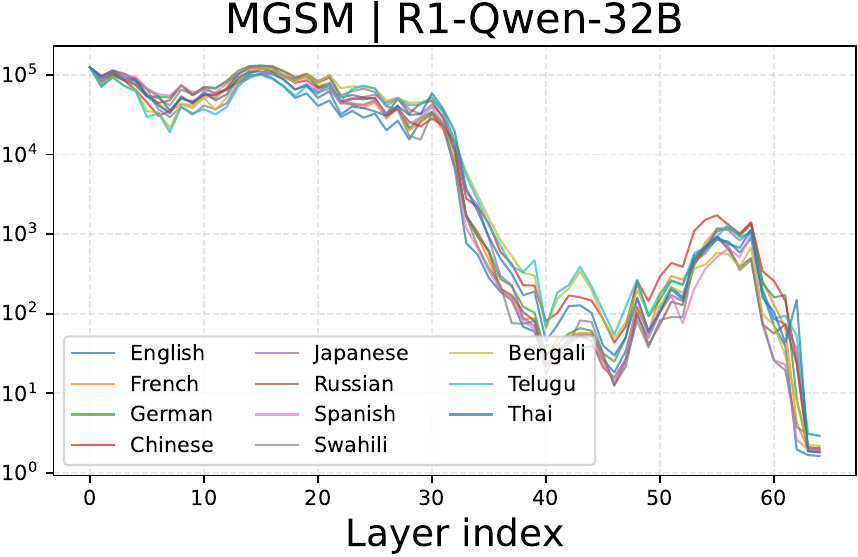}
    \includegraphics[width=0.16\textwidth]{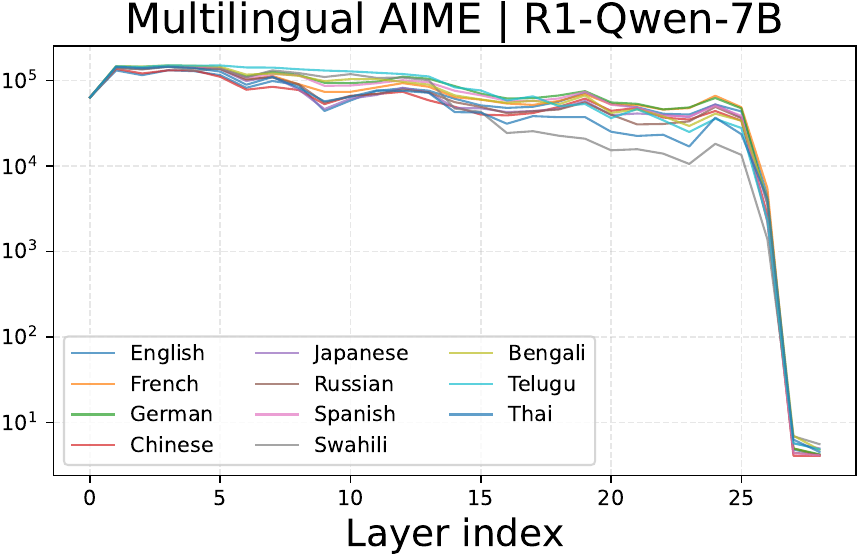}
    \includegraphics[width=0.16\textwidth]{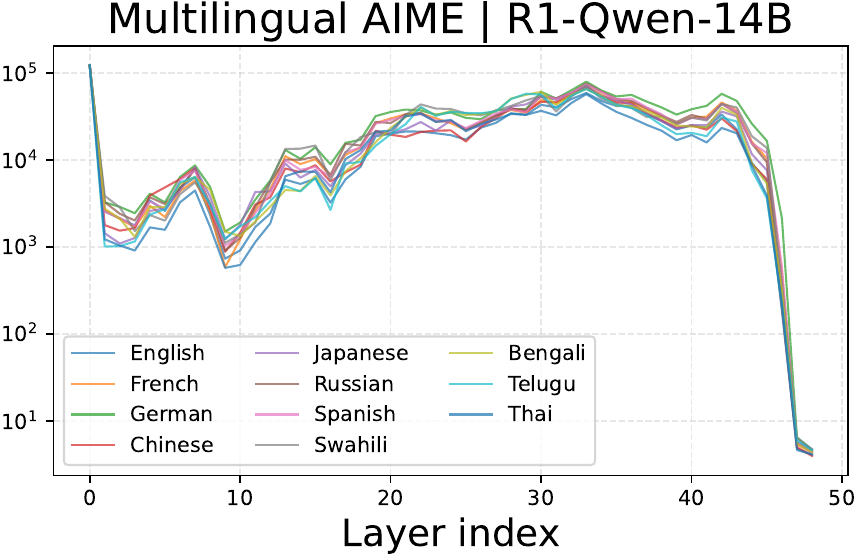}
    \includegraphics[width=0.16\textwidth]{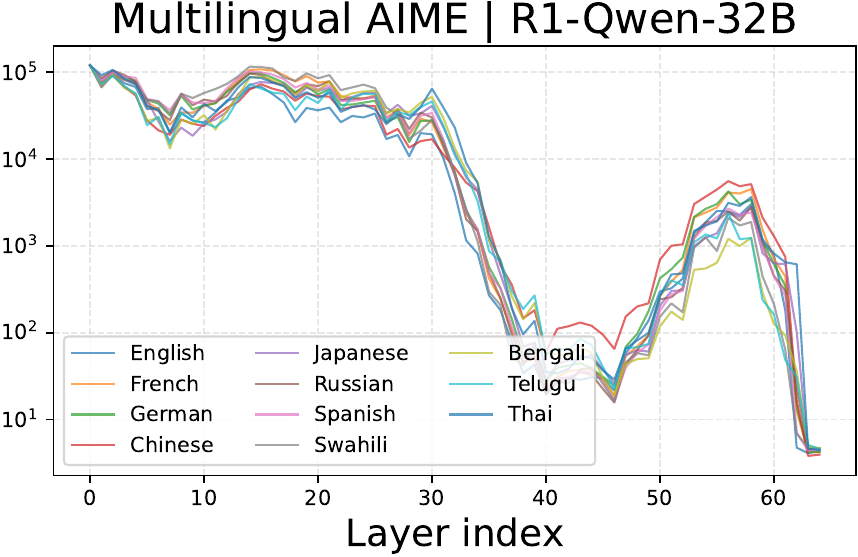}
    \caption{Layer-wise rank of the gold answer obtained via
      logit lens across languages on MGSM (left three
      panels) and Multilingual AIME (right three panels). 
    Rank trajectories exhibit highly similar trends across languages, suggesting that latent reasoning progresses through comparable layer-wise transformations regardless of language.}
    \label{fig:logit_lens}
\end{figure*}

\begin{figure*}
    \centering
    \setlength{\belowcaptionskip}{-0.4cm}
    \includegraphics[width=0.24\textwidth]{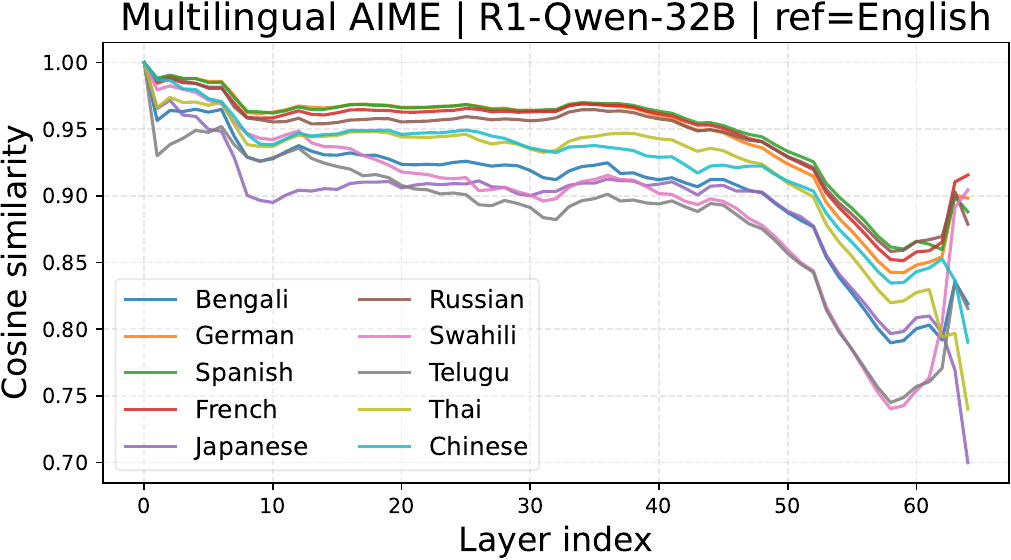}
    \includegraphics[width=0.24\textwidth]{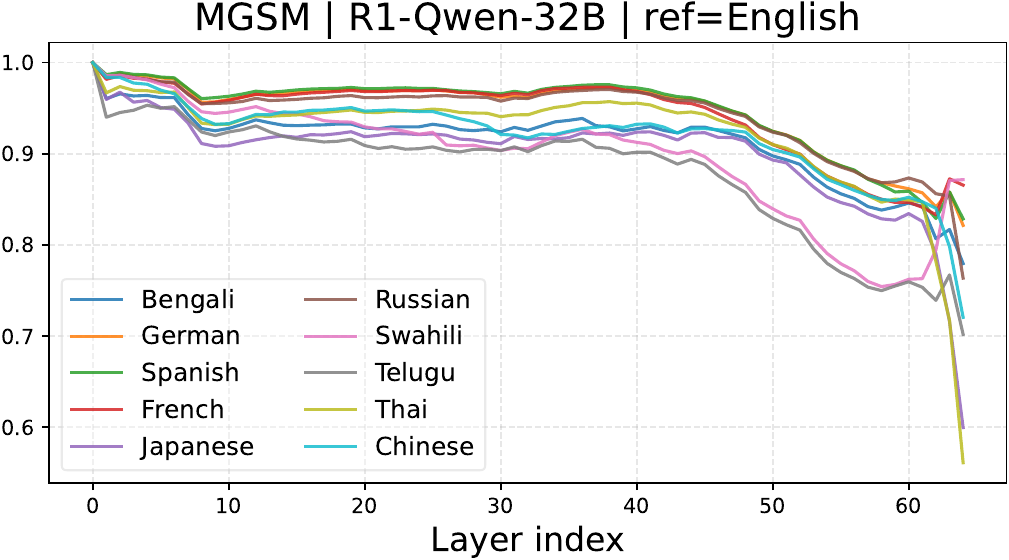}
    \includegraphics[width=0.24\textwidth]{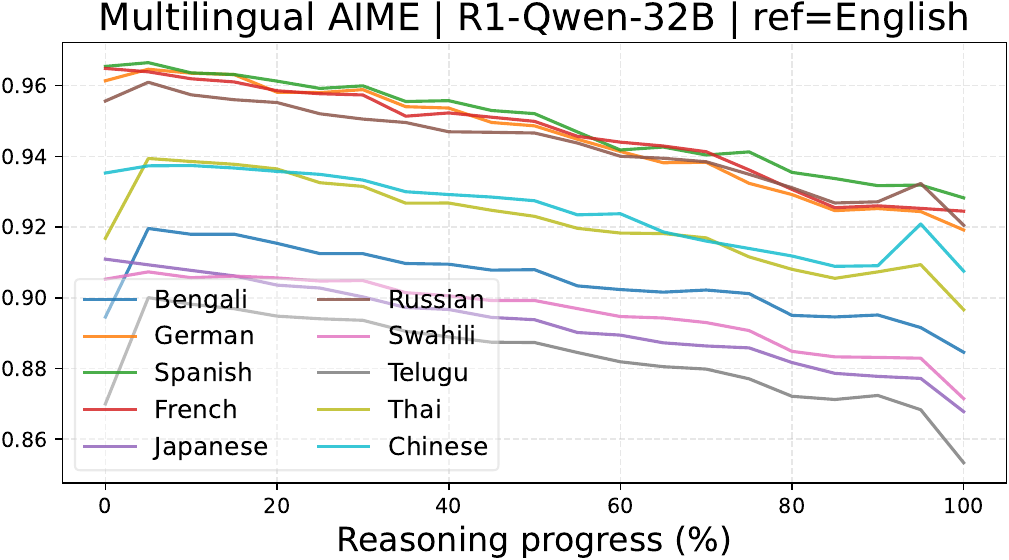}
    \includegraphics[width=0.24\textwidth]{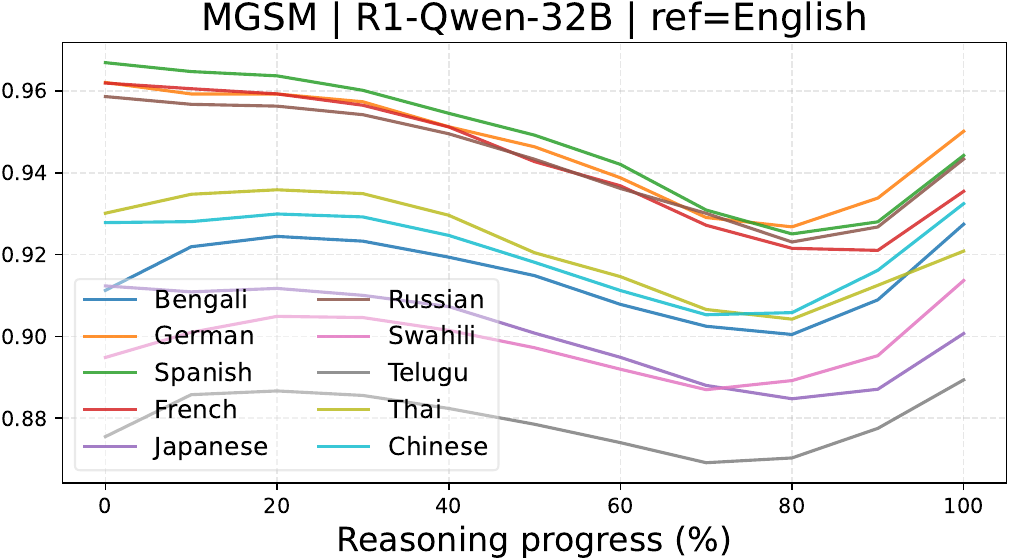}
    \caption{Aggregated cosine similarity between hidden states in each language and English (reference), averaged over both reasoning steps and layers, for R1-Qwen-32B. 
    High-resource languages show consistently higher similarity to English, suggesting convergence toward an English-centered latent reasoning pathway.}
    \label{fig:cosine_sim_aggregated}
\end{figure*}

\section{Latent State Dynamics}\seclabel{latent_dynamics}


In \secref{surface_dynamics}, we observed that models exhibit clear signs of latent reasoning across languages -- particularly on lower-complexity tasks -- but with substantial crosslingual variation: high-resource languages tend to reach correct predictions earlier and more reliably.
To better understand the origins of these differences, we must go beyond surface-level outputs and examine the model's internal representations.
We therefore turn to \textbf{RQ2}: \emph{Do different languages rely on different internal latent reasoning mechanisms?}
To answer this question, we analyze both the layer-wise evolution of the model's implicit predictions (\secref{ranking}) and the similarity of hidden states across languages (\secref{similarity}).

\subsection{Dynamic of Ranking Across Layers}\seclabel{ranking}

To investigate whether different languages rely on distinct \emph{internal} latent reasoning mechanisms, we analyze how evidence for the correct answer emerges across model layers using the logit lens \citep{logit-lens}.  
While the logit lens is not a perfect probe of intermediate representations, particularly due to residual stream entanglement \citep{belrose2025elicitinglatentpredictionstransformers}, it remains a useful diagnostic tool for tracking \emph{relative} changes in answer salience across layers when applied consistently within the same model \citep{wendler-etal-2024-llamas,wang-etal-2025-lost-multilinguality}.
Concretely, at each layer, we project the hidden state (i.e., residual stream activation) through the model's layer normalization and unembedding matrix and record the rank of the gold answer.\footnote{The gold answer is always a numeric value and is identical across languages. We track the rank of the first token, as generating this token is a necessary condition for producing the correct final answer. This practice is widely adopted in prior work~\citep{hernandez2024linearity,kargaran-etal-2025-programming}.} 
By comparing these \emph{rank trajectories} across languages for a fixed model, we can assess whether layers play comparable functional roles in latent reasoning across different languages.

Figure~\ref{fig:logit_lens} shows rank trajectories across languages, models, and datasets.  
A striking observation is that all languages exhibit highly similar ranking curves for a fixed model, suggesting that the internal mechanism used to form the solution is largely \emph{language-invariant}.  
Despite differences in surface language realization and accuracy, \textbf{the underlying latent computation appears to follow the same structural progression across layers}.


At the same time, we observe distinct patterns across model sizes.
These differences suggest that model capacity can shape latent reasoning dynamics, consistent with prior work showing that larger models exhibit qualitatively different intermediate-layer behavior, such as stronger representation compression, compared to smaller models~\citep{skean2025layer}.
In particular, larger models appear to distribute reasoning more evenly across depth, allowing intermediate representations to encode increasingly informative abstractions.
Notably, this pattern aligns with recent findings that multilingual models maintain a largely language-independent conceptual space in their middle layers~\citep{wang-etal-2025-lost-multilinguality,lu-etal-2025-paths}.
The emergence of intermediate answer salience in larger models may therefore reflect a greater capacity to exploit this shared space, enabling earlier and more stable accumulation of evidence toward the correct solution.

\subsection{Hidden State Similarity}\seclabel{similarity}

We showed that a model presents consistently similar rank trajectories across layers across languages in \secref{ranking}.
We further hypothesize that such consistency may reflect an
\emph{English-centered} latent reasoning process, in which
reasoning in other languages implicitly aligns with the pathway used for English. 

To test this hypothesis, we compute cosine similarity between the hidden states of each target language and those of English.  
For each example in a target language, at each truncation ratio, we extract the hidden state of the final token of the reasoning trace and measure its similarity to the corresponding hidden state of its English counterpart.
We aggregate similarities in two ways: (i) averaging over layers 
and (ii) averaging over reasoning steps.


Figure~\ref{fig:cosine_sim_aggregated} summarizes these results.
Overall, we observe consistently higher similarity with English for high-resource languages, including those using non-Latin scripts such as Chinese and Russian.
In contrast, mid-resource languages with distinct scripts (e.g., Japanese) and low-resource languages (e.g., Telugu) exhibit lower similarity to English.
This pattern is stable across layers and reasoning steps, suggesting that \textbf{reasoning in high-resource languages may be processed in a representational space more closely aligned with English, whereas mid- and low-resource languages deviate more substantially}.\footnote{Crosslingual similarity may also be influenced by linguistic and typological relatedness between languages, which could partially contribute to the observed alignment patterns.}
However, similarity alone does not distinguish whether this alignment reflects an English-centered reasoning process or merely arises from \emph{shared correct answers}.

\begin{figure}
    \centering
    \setlength{\belowcaptionskip}{-0.5cm}
    \includegraphics[width=0.23\textwidth]{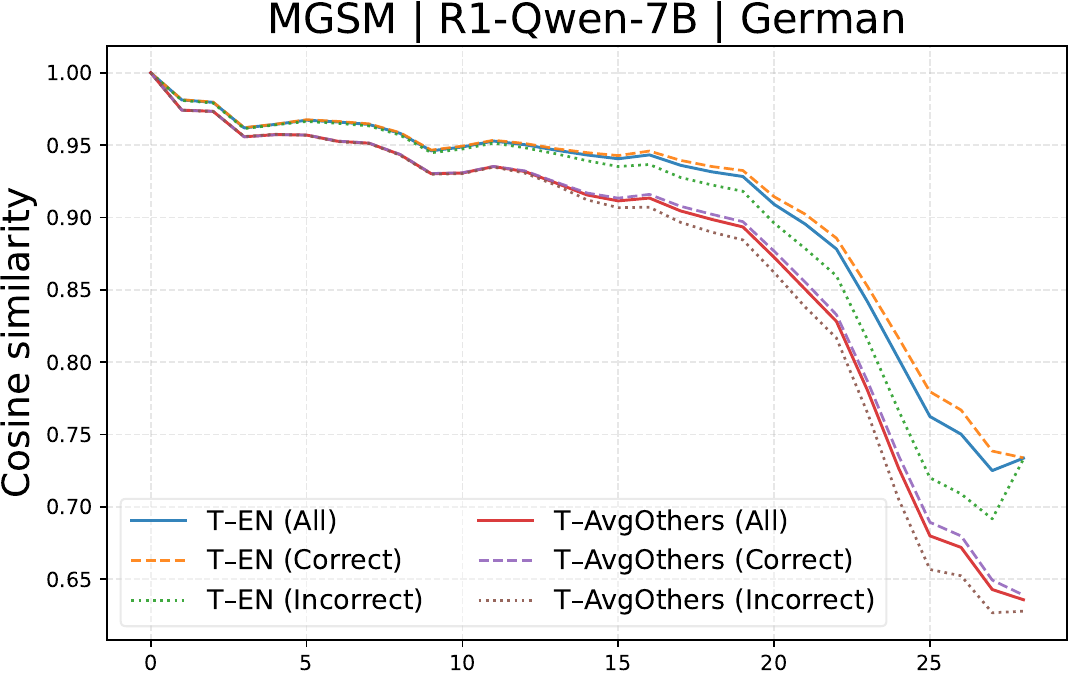}
    \includegraphics[width=0.23\textwidth]{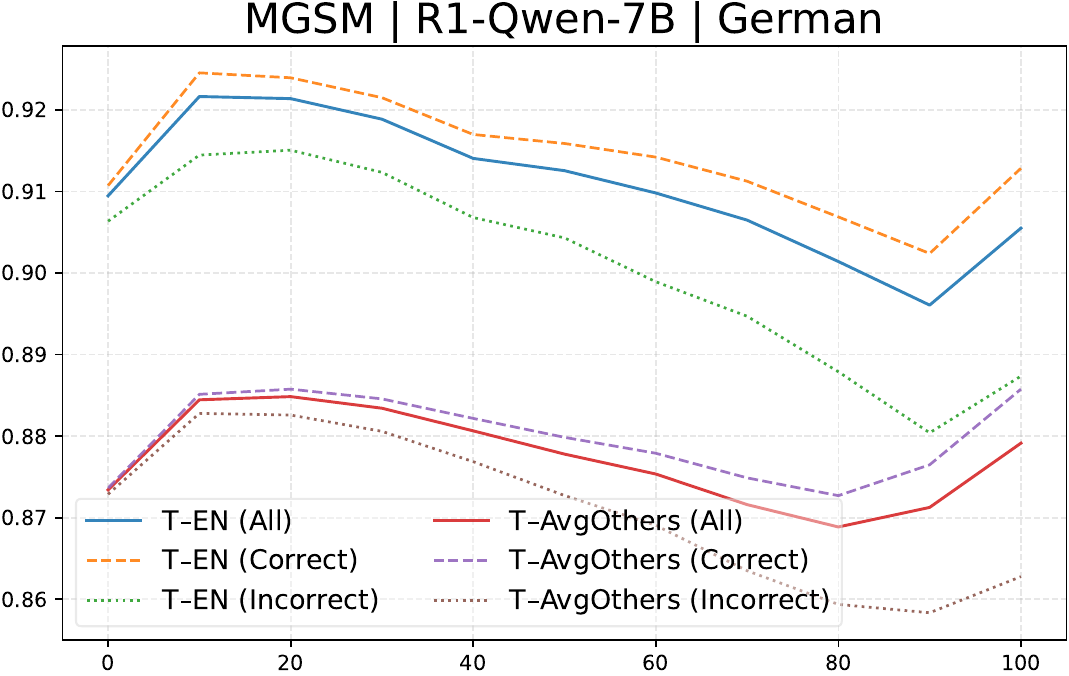}
    \includegraphics[width=0.23\textwidth]{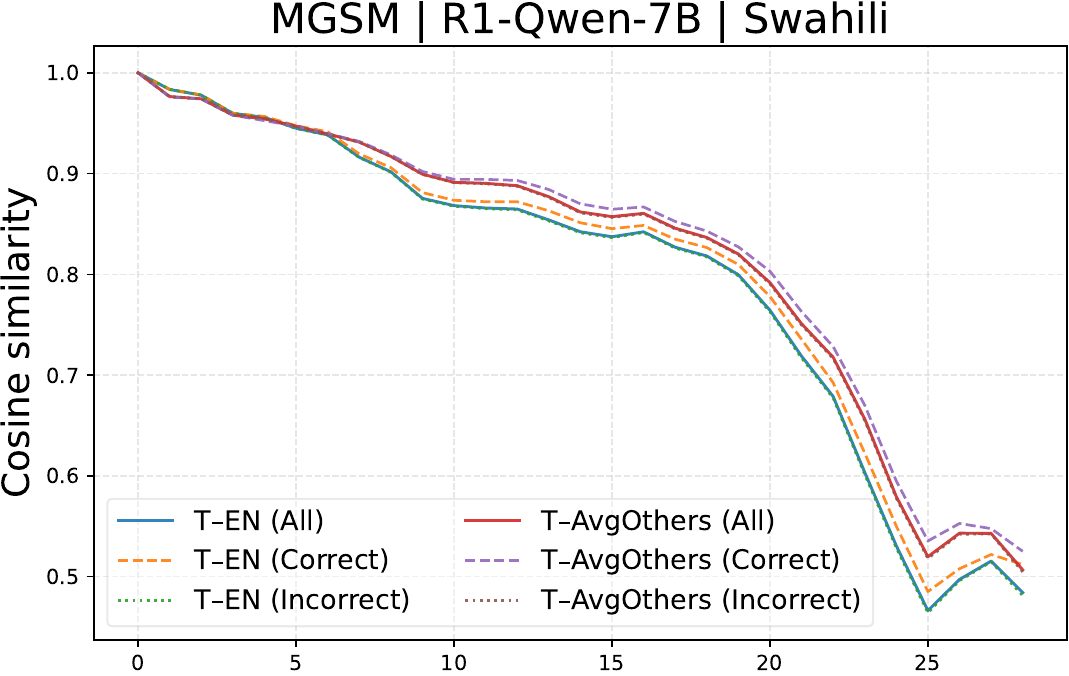}
    \includegraphics[width=0.23\textwidth]{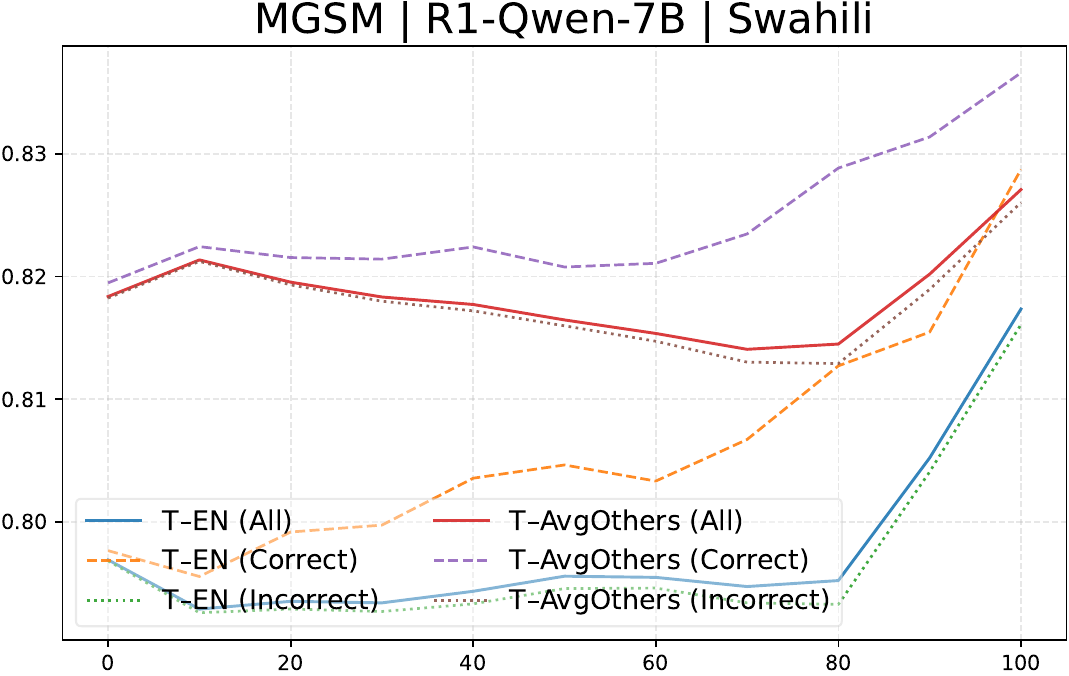}
    \includegraphics[width=0.23\textwidth]{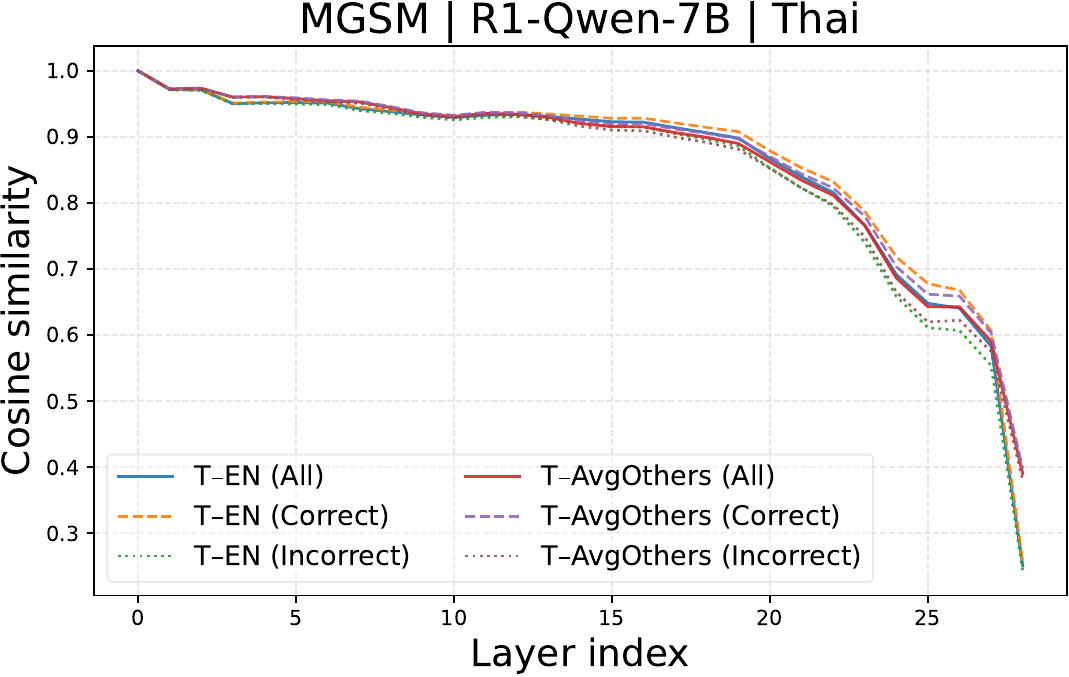}
    \includegraphics[width=0.23\textwidth]{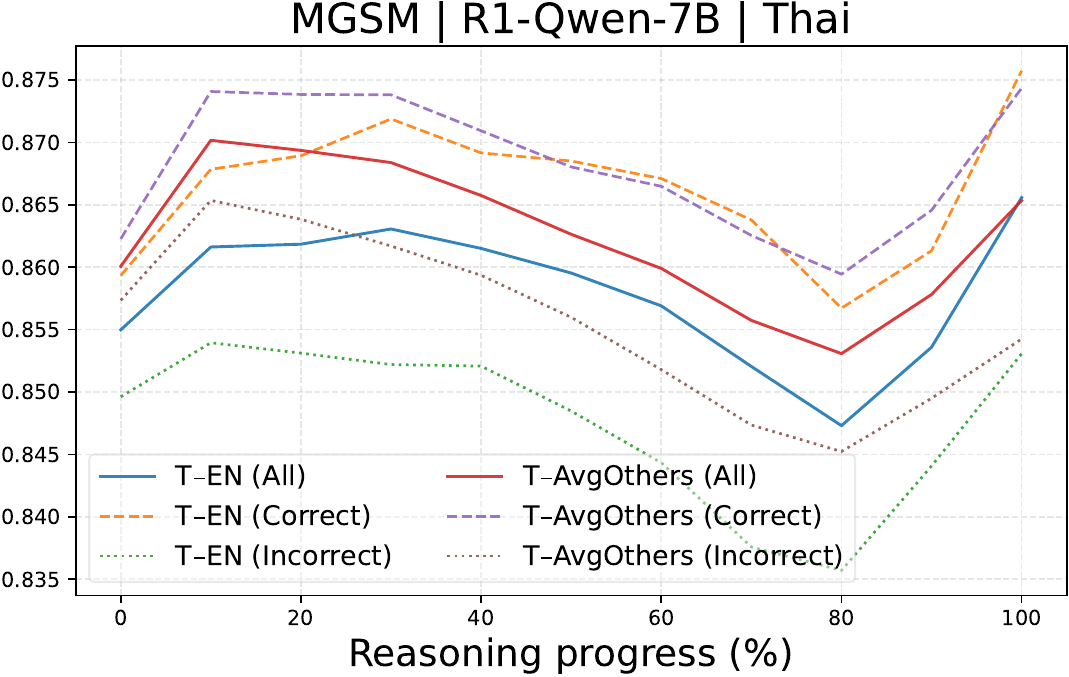}
    \caption{Comparison of cosine similarity with English versus average similarity with other languages, shown separately for correctly and incorrectly solved examples.
    High-resource languages show stronger alignment with English, whereas low- and mid-resource languages show weaker or correctness-dependent alignment.}
    \label{fig:cosine_sim_vs_others}
\end{figure}

To disentangle these effects, 
for each language, we group MGSM examples into \emph{correct} and \emph{incorrect} sets based on pass@10 accuracy, and compare the (i) similarity with English and (ii) average similarity with other languages (excluding itself and English).
Figure~\ref{fig:cosine_sim_vs_others} presents the results over layers and reasoning steps (see full results in \secref{sim_correct}).

\textbf{Language resource level modulates alignment with English in latent reasoning.}
High-resource languages (e.g., German) exhibit consistently strong alignment with English that is largely \emph{independent of correctness}: incorrect instances remain nearly as similar to their English counterparts as correct ones, particularly in early reasoning stages.
Although a modest gap emerges as the reasoning trace unfolds, this appears to reflect increasing commitment to an (incorrect) solution rather than a shift away from English-aligned latent trajectories.
The absence of a substantial correct–incorrect gap overall indicates that, for high-resource languages, alignment with English reflects a stable latent reasoning trajectory rather than a byproduct of successful solution formation.
In contrast, low-resource languages (e.g., Swahili) show weaker alignment with English across both correct and incorrect examples, while exhibiting relatively higher similarity to other languages.
This pattern suggests a more autonomous subspace that is less shaped by English-centric post-training and more influenced by language-specific representations formed during pretraining \citep{chang-etal-2022-geometry,liu-etal-2024-translico}.
Mid-resource languages (e.g., Thai) occupy an intermediate regime: alignment with English is more pronounced for correct instances than for incorrect ones, suggesting that convergence toward English-like latent trajectories occurs primarily when reasoning is successful.

\begin{table*}[t]
\setlength{\belowcaptionskip}{-0.5cm}
\centering
\tiny
\renewcommand{\arraystretch}{0.55}
\resizebox{\linewidth}{!}{
\begin{tabular}{lll|rrrrrr|rrr|rr}
\toprule
Dataset & Model & Metric
& DE & EN & ES & FR & RU & ZH
& BN & JA & TH
& SW & TE \\
\midrule
\multirow{9}{*}{MGSM}
& \multirow{3}{*}{Qwen2.5-32B}
& AUTC & 0.62 & 0.70 & 0.67 & 0.63 & 0.65 & 0.66 & 0.66 & 0.63 & 0.63 & 0.41 & 0.49 \\
&  & AUGC & 0.15 & 0.16 & 0.15 & 0.14 & 0.16 & 0.20 & 0.17 & 0.19 & 0.18 & 0.17 & 0.17 \\
&  & LRS  & 0.50 & 0.57 & 0.55 & 0.53 & 0.52 & 0.50 & 0.52 & 0.48 & 0.49 & 0.33 & 0.39 \\
\cmidrule(lr){2-14}
& \multirow{3}{*}{R1-Qwen-32B}
& AUTC & 0.67 & 0.75 & 0.69 & 0.64 & 0.68 & 0.70 & 0.61 & 0.63 & 0.69 & 0.38 & 0.39 \\
&  & AUGC & 0.20 & 0.25 & 0.20 & 0.17 & 0.21 & 0.30 & 0.23 & 0.21 & 0.28 & 0.20 & 0.23 \\
&  & LRS  & 0.51 & 0.53 & 0.52 & 0.51 & 0.51 & 0.45 & 0.44 & 0.47 & 0.46 & 0.30 & 0.30 \\
\cmidrule(lr){2-14}
& \multirow{3}{*}{Nemotron-32B}
& AUTC & 0.84 & 0.88 & 0.86 & 0.80 & 0.84 & 0.83 & 0.70 & 0.78 & 0.77 & 0.34 & 0.26 \\
&  & AUGC & 0.86 & 0.82 & 0.84 & 0.81 & 0.77 & 0.81 & 0.81 & 0.82 & 0.77 & 0.45 & 0.57 \\
&  & LRS  & 0.09 & 0.11 & 0.11 & 0.11 & 0.15 & 0.11 & 0.11 & 0.11 & 0.14 & 0.18 & 0.11 \\
\midrule
\multirow{9}{*}{Multilingual AIME}
& \multirow{3}{*}{Qwen2.5-32B}
& AUTC & 0.03 & 0.04 & 0.03 & 0.03 & 0.04 & 0.02 & 0.01 & 0.01 & 0.03 & 0.02 & 0.00 \\
&  & AUGC & 0.07 & 0.04 & 0.32 & 0.48 & 0.16 & 0.09 & 0.03 & 0.33 & 0.62 & 0.02 & 0.00 \\
&  & LRS  & 0.02 & 0.04 & 0.02 & 0.02 & 0.03 & 0.02 & 0.01 & 0.00 & 0.01 & 0.02 & 0.00 \\
\cmidrule(lr){2-14}
& \multirow{3}{*}{R1-Qwen-32B}
& AUTC & 0.06 & 0.18 & 0.08 & 0.09 & 0.10 & 0.13 & 0.04 & 0.04 & 0.07 & 0.01 & 0.01 \\
&  & AUGC & 0.29 & 0.61 & 0.32 & 0.72 & 0.66 & 0.75 & 0.18 & 0.74 & 0.82 & 0.05 & 0.17 \\
&  & LRS  & 0.03 & 0.06 & 0.04 & 0.02 & 0.03 & 0.03 & 0.02 & 0.01 & 0.02 & 0.00 & 0.00 \\
\cmidrule(lr){2-14}
& \multirow{3}{*}{Nemotron-32B}
& AUTC & 0.38 & 0.43 & 0.41 & 0.38 & 0.36 & 0.32 & 0.30 & 0.37 & 0.30 & 0.05 & 0.13 \\
&  & AUGC & 0.79 & 0.88 & 0.84 & 0.75 & 0.73 & 0.73 & 0.93 & 0.81 & 0.79 & 0.26 & 0.56 \\
&  & LRS  & 0.03 & 0.02 & 0.03 & 0.04 & 0.03 & 0.03 & 0.02 & 0.03 & 0.03 & 0.03 & 0.04 \\
\bottomrule
\end{tabular}
}
\caption{
Truncation-based metrics across different post-training paradigms.
While RL-trained models (Nemotron-32B) achieve stronger overall performance (higher AUTC),
multilingual asymmetry persists. 
Higher AUGC in RL models indicates earlier answer articulation,
often followed by extended reasoning (``overthinking'').
}
\label{tab:truncation_metrics_extended}
\end{table*}

\section{Is Latent Reasoning Behavior Specific to Distilled Models?}

A potential concern is \emph{whether the observed latent reasoning behavior is specific to distilled reasoning models}, or \emph{whether it generalizes to models with different post-training paradigms}. 
Prior work suggests that reinforcement learning (RL) and distillation can lead to different reasoning behaviors \citep{yue2025doesreinforcementlearningreally,kim2025reinforcementlearningvsdistillation,baek2025understandingdistilledreasoningmodels}. 
To investigate this, we extend our latent reasoning identification (\secref{surface_dynamics}) to additional models.

\subsection{Experimental Setup}

\textbf{Model Selection.}
In addition to the distilled reasoning model \texttt{R1-Qwen-32B} used in \secref{surface_dynamics}, we consider two additional models derived from the same base model: (1) \texttt{Qwen2.5-32B-Instruct} \citep{qwen2025qwen25technicalreport}, an instruction-tuned model without explicit reasoning distillation, and (2) \texttt{Nemotron-32B} \citep{moshkov2025aimo2winningsolutionbuilding}, a reasoning model trained with RL-based post-training. 

\textbf{Evaluation Metrics.}
We adopt the same truncation-based evaluation as in the main experiments (cf. \secref{truncation_metrics}), including \textbf{AUTC} (Area Under the Truncation Accuracy Curve), \textbf{AUGC} (Area Under the Gold-in-Trace Curve), and \textbf{LRS} (Latent Reasoning Score). These metrics jointly characterize when correctness emerges during reasoning and whether it depends on explicit answer articulation.

\subsection{Results and Discussion}

Table~\ref{tab:truncation_metrics_extended} reports our results.
Overall, the main findings remain consistent across all models. 
In particular, we observe high AUTC across models, indicating that correct answers can often be produced from partial reasoning traces. 
However, latent reasoning remains uneven across languages for both instruction-tuned and RL-based models, suggesting that this disparity is not specific to distilled models.

\textbf{Early answer articulation and overthinking in RL models.}
Nemotron-32B achieves substantially higher AUTC than the other models, indicating stronger overall reasoning capability.
However, it also exhibits much higher AUGC and relatively low LRS, suggesting that correctness often relies on early answer articulation. 
Inspection of reasoning traces reveals that the model frequently produces the correct answer within the first $\sim$20\% of steps, followed by extended reflective reasoning. 
This behavior is consistent with \emph{overthinking} \citep{chen2025think23overthinkingo1like,sui2025overthink}, where the model continues generation after already reaching the answer.

\textbf{Task difficulty governs latent reasoning dynamics.}
Despite differences in post-training, all models exhibit similar trends across benchmarks. 
On MGSM, models can often produce correct answers with minimal or even no reasoning trace, indicating the presence of latent reasoning. 
In contrast, on the more challenging Multilingual AIME benchmark, correct predictions typically require explicit answer articulation, resulting in lower AUTC/LRS and higher AUGC. 
This further confirms that task difficulty plays a central role in shaping latent reasoning dynamics.

\begin{table*}[t]
\setlength{\belowcaptionskip}{-0.4cm}
\centering
\tiny
\renewcommand{\arraystretch}{0.55} 
\resizebox{\linewidth}{!}{
\begin{tabular}{l l l|rrrrrr|rrr|rr}
\toprule
Edit Method & Model & Setup
& DE & EN & ES & FR & RU & ZH
& BN & JA & TH
& SW & TE \\
\midrule
\multirow{6}{*}{\textbf{NumEdit}~($\downarrow$)}
& \multirow{2}{*}{R1-Qwen-7B}
& w/o Trace & 0.40 & 0.31 & 0.29 & 0.34 & 0.36 & 0.33 & 0.38 & 0.51 & 0.33 & 0.52 & 0.40 \\
&  & w/ Trace
& 0.27 & 0.25 & 0.21 & 0.22 & 0.24 & 0.16 & 0.21 & 0.32 & 0.30 & 0.19 & 0.17 \\
\cmidrule(lr){2-14}
& \multirow{2}{*}{R1-Qwen-14B}
& w/o Trace & 0.33 & 0.27 & 0.29 & 0.35 & 0.35 & 0.29 & 0.30 & 0.43 & 0.29 & 0.47 & 0.39 \\
&  & w/ Trace
& 0.19 & 0.16 & 0.21 & 0.20 & 0.19 & 0.13 & 0.15 & 0.28 & 0.15 & 0.22 & 0.12 \\
\cmidrule(lr){2-14}
& \multirow{2}{*}{R1-Qwen-32B}
& w/o Trace & 0.31 & 0.32 & 0.29 & 0.27 & 0.35 & 0.28 & 0.32 & 0.47 & 0.31 & 0.37 & 0.33 \\
&  & w/ Trace
& 0.18 & 0.20 & 0.20 & 0.19 & 0.19 & 0.19 & 0.16 & 0.26 & 0.18 & 0.22 & 0.18 \\
\midrule
\multirow{6}{*}{\textbf{Paraphrase}~($\uparrow$)}
& \multirow{2}{*}{R1-Qwen-7B}
& w/o Trace & 0.66 & 0.70 & 0.73 & 0.72 & 0.71 & 0.74 & 0.67 & 0.64 & 0.57 & 0.45 & 0.58 \\
&  & w/ Trace
& 1.00 & 1.00 & 0.96 & 0.99 & 0.99 & 0.96 & 0.95 & 0.90 & 0.96 & 0.35 & 0.85 \\
\cmidrule(lr){2-14}
& \multirow{2}{*}{R1-Qwen-14B}
& w/o Trace & 0.79 & 0.86 & 0.73 & 0.77 & 0.85 & 0.72 & 0.79 & 0.75 & 0.81 & 0.55 & 0.63 \\
&  & w/ Trace
& 0.97 & 1.00 & 0.98 & 0.96 & 0.99 & 0.95 & 0.95 & 0.94 & 0.99 & 0.69 & 0.54 \\
\cmidrule(lr){2-14}
& \multirow{2}{*}{R1-Qwen-32B}
& w/o Trace & 0.86 & 0.81 & 0.86 & 0.83 & 0.90 & 0.91 & 0.90 & 0.85 & 0.85 & 0.83 & 0.76 \\
&  & w/ Trace
& 0.99 & 0.99 & 0.98 & 0.96 & 0.98 & 0.96 & 0.99 & 0.98 & 0.96 & 0.92 & 0.78 \\
\bottomrule
\end{tabular}
}
\caption{
Pass@10 results on edited MGSM questions across 11 languages.
For \textbf{NumEdit} ($\downarrow$), values report the \emph{matching ratio} with the original gold answer after a single-number perturbation (lower is better).
For \textbf{Paraphrase} ($\uparrow$), values report \emph{accuracy}, as the gold answer is unchanged.
``w/o Trace'' denotes inference without a reasoning trace (empty \texttt{<think>}\texttt{</think>} block), while ``w/ Trace'' allows a newly generated trace.
}
\label{tab:memorization}
\end{table*}

\section{Complementary Analysis: Memorization or Latent Reasoning}\seclabel{memorization}

In \secref{surface_dynamics}, we observed that for MGSM, models can sometimes predict the correct answer even when no reasoning trace is provided (i.e., truncation ratio $0\%$).
While this behavior may suggest that the model has already implicitly computed the answer and thus exhibits latent reasoning, an alternative explanation is that the model has memorized the solution due to exposure during pre-training or post-training, a phenomenon commonly referred to as \emph{data contamination} or \emph{benchmark leakage} \citep{xu2024benchmarkingbenchmarkleakagelarge,balloccu-etal-2024-leak}.
Under such circumstances, correct predictions may arise from direct recall rather than genuine latent reasoning.


To disentangle memorization from latent reasoning, we conduct a complementary analysis that probes the model's sensitivity to controlled question perturbations.
The intuition is as follows: if a model relies on memorization, small but \emph{meaning-altering} edits should not substantially change its predictions, as the original answer may still be recalled.
In contrast, a reasoning-based model should adapt its prediction to such changes.
Conversely, when the underlying meaning of a question is preserved but surface form is altered via \emph{paraphrasing}, a reasoning model should remain robust and continue to produce the correct answer, whereas a memorization-based model may fail.

\subsection{Method}\seclabel{edit_method}

We focus on MGSM and restrict our attention to \textbf{questions that are answered correctly under the pass@10 when no reasoning trace is provided} (i.e., truncation ratio $=0\%$), for each language and each model independently, as they are particularly ambiguous cases where memorization and latent reasoning are difficult to distinguish.
For each question, we apply the following editing strategies.\footnote{See \secref{perturbation_details} for the details of altering the original problem.}

\textbf{NumEdit}
We modify exactly one numerical value in the original question while keeping the rest of the problem unchanged.
The edit is chosen such that it alters the solution, and therefore, the original gold answer is no longer correct.
Accordingly, we evaluate NumEdit using the \emph{matching ratio} with the original gold answer, where lower values indicate better sensitivity 
to the perturbation.

\textbf{Paraphrase}
We paraphrase and reorder the question text while preserving all numerical values, mathematical expressions, and the overall semantics.
The paraphrased question is logically \emph{equivalent} to the original, and thus the gold answer is unchanged.
In this case, we evaluate performance using standard
\emph{accuracy} (the higher the better).

\subsection{Results and Discussions}\seclabel{edit_results}

\textbf{Models exhibit partial memorization, but latent reasoning remains evident.}
Table~\ref{tab:memorization} shows that under NumEdit, models still match the original gold answer in a non-trivial fraction of cases, with matching ratios typically around $30\%$ across languages, with high-resource languages (e.g., English) generally showing a lower matching ratio than low-resource languages (e.g., Swahili), under the \emph{w/o Trace} setting.
However, the matching ratio consistently decreases when models are allowed to generate a new reasoning trace, often dropping below $25\%$ for smaller models and below $20\%$ for the 32B model in most languages.\footnote{We use \texttt{Gemini-2.5-Flash} to validate NumEdit; around 10\% of the edited questions retain the same gold answer (see \secref{perturbation_details}). As a result, the reported matching ratios should be interpreted as an upper bound on the extent of memorization.}
Taken together, these results indicate that while memorization is present, models largely recompute solutions rather than merely recalling memorized answers, providing evidence in favor of latent reasoning.

\textbf{Robustness to paraphrasing argues against pattern-matching memorization.}
In Paraphrase, pass@10 accuracy under the \emph{w/o Trace} setting is typically above $70\%$, and increases further when allowing the model to generate a new reasoning trace across languages.
For instance, R1-Qwen-32B reaches near-perfect accuracy in high-resource languages such as English and German under the \emph{w/ Trace} setting.
Although performance is lower for under-resourced languages (e.g., Swahili), the same trend holds.
Additionally, accuracy consistently improves with model scale, and explicit reasoning traces further amplify this effect.
These results suggest that the models do not rely solely on surface-level pattern matching to the original question wording.
Instead, their robustness to paraphrasing provides converging evidence that the models engage in genuine reasoning processes.

\section{Conclusion}

We present a systematic study of multilingual latent reasoning in LRMs.
Using truncation-based analyses, we show that LRMs can perform latent reasoning, but this capability is highly uneven: it is strong in resource-rich languages on easier tasks, weak in low-resource languages, and is largely undetectable on more challenging benchmarks.
Our representational analyses reveal highly consistent layer-wise dynamics across languages, with latent reasoning converging toward an English-centered pathway, particularly for high-resource languages and correctly solved instances.
Finally, we demonstrate that these behaviors cannot be explained by surface-level memorization alone.
Together, our findings suggest that current LRMs exhibit real but fragile multilingual latent reasoning, shaped by English-centric post-training and task complexity.

\section*{Limitations}

While this work offers a systematic analysis of multilingual latent reasoning in LRMs, it is subject to several limitations.
  
First, we implement reasoning-trace truncation at the \emph{step} level rather than the \emph{token} level, following previous work. 
While step-based truncation aligns with sentence-level CoT structure and improves interpretability, finer-grained token-level truncation may reveal more precise dynamics of latent answer formation and is left to future work.  

Second, our Gold-in-Trace metric relies on string matching to detect whether the gold answer appears in the reasoning trace, which may yield false positives when intermediate values happen to match the final answer. 
However, as discussed in \secref{sec:git_false_positive}, this effect is limited in our setting due to the prevalence of multi-digit answers and conditioning on correct predictions. 
More sophisticated and robust matching strategies can be explored in future work.

Third, our experiments focus on mathematical reasoning benchmarks. 
While this choice enables precise measurement of latent reasoning through clear numeric answers and structured multi-step reasoning, it may limit generalizability to other reasoning domains, such as commonsense reasoning. 
Extending our analysis to such tasks is an important direction for future work.

Fourth, due to computational constraints, our experiments are limited to models up to 32B parameters. 
While we include models with different post-training paradigms and observe consistent trends, extending the analysis to larger-scale models can be an interesting direction for future work. 

Finally, our work is primarily diagnostic and does not propose a new inference method. 
Instead, we focus on analyzing whether latent reasoning exists across languages and how its behavior varies with language and task difficulty. 
While our findings provide insights into multilingual reasoning (e.g., asymmetry across languages and task-dependent dynamics), future work can translate these insights into concrete improvements for multilingual inference, such as transferring reasoning capabilities from high- to low-resource languages.

\section*{Ethical Considerations}

\paragraph{Use of AI Assistants.}
The authors acknowledge the use of \texttt{ChatGPT~5.2} for language editing (grammar, clarity, and coherence) and limited assistance with code implementation;\footnote{\url{https://chatgpt.com/}} all technical content and experimental decisions were made by the authors.

\section*{Acknowledgments}

This research was supported by the Munich Center for Machine Learning (MCML) and German Research Foundation (DFG, grant SCHU 2246/14-1).


\bibliography{custom}

\appendix

\begin{table*}[t]
\centering
\small
\setlength{\tabcolsep}{4pt}
\resizebox{\linewidth}{!}{
\begin{tabular}{llccccccccccc}
\toprule
Dataset & Model & EN & FR & DE & ZH & JA & RU & ES & SW & BN & TE & TH \\
\midrule
\multirow{3}{*}{MGSM}
 & R1-Qwen-7B  
 & 9.5 (9.0)  & 11.4 (9.0) & 10.9 (9.5) & 10.8 (9.0) & 13.1 (8.0) & 9.4 (8.0) & 9.6 (8.0) & 21.6 (9.0) & 16.1 (14.0) & 23.9 (20.0) & 18.9 (9.0) \\
 & R1-Qwen-14B 
 & 11.2 (10.0) & 13.6 (13.0) & 14.7 (13.0) & 16.3 (13.0) & 11.1 (10.0) & 13.8 (12.0) & 13.8 (11.0) & 32.6 (18.0) & 15.8 (13.0) & 22.6 (17.0) & 12.4 (11.0) \\
 & R1-Qwen-32B 
 & 23.2 (19.0) & 9.2 (8.0)  & 11.7 (11.0) & 20.2 (18.0) & 10.2 (9.0) & 10.1 (9.0) & 10.3 (8.0) & 20.3 (16.0) & 14.8 (13.0) & 18.2 (15.5) & 12.6 (11.0) \\
\midrule
\multirow{3}{*}{Multilingual AIME}
 & R1-Qwen-7B  
 & 234.9 (168.0) & 197.2 (132.0) & 206.3 (124.5) & 263.9 (181.0) & 169.0 (47.0)  & 143.8 (54.0)  & 226.1 (152.5) & 92.5 (10.0)  & 114.4 (66.5)  & 117.5 (125.5) & 95.6 (15.5) \\
 & R1-Qwen-14B 
 & 231.8 (169.0) & 183.4 (120.5) & 162.4 (107.0) & 267.2 (178.5) & 144.6 (67.5)  & 217.8 (157.0) & 205.3 (139.0) & 95.7 (66.0)  & 141.5 (108.0) & 122.3 (112.0) & 99.3 (62.5) \\
 & R1-Qwen-32B 
 & 282.4 (207.0) & 123.5 (102.5) & 213.7 (152.0) & 279.6 (215.0) & 411.4 (89.0)  & 216.5 (119.0) & 166.0 (114.0) & 112.2 (43.5) & 138.6 (121.5) & 122.5 (126.0) & 159.3 (116.5) \\
\bottomrule
\end{tabular}
}
\caption{Reasoning-step statistics across languages on MGSM and Multilingual AIME. Each cell reports the average number of reasoning steps, with the median shown in parentheses.}
\label{tab:steps_mgsm_aime}
\end{table*}

\section{Reasoning Trace Statistics}\seclabel{trace_statistics}

Table~\ref{tab:steps_mgsm_aime} reports the average and median number of reasoning steps for MGSM and Multilingual AIME across languages.

For MGSM, we observe that most languages exhibit an average of approximately 10 reasoning steps.
Accordingly, we adopt a truncation granularity of 10\%, which roughly corresponds to removing one reasoning step at a time.

In contrast, Multilingual AIME displays substantially longer reasoning traces on average, while the median number of steps is markedly smaller.
This discrepancy is primarily driven by a small number of outlier instances, which is expected given the limited dataset size (60 problems).
To better account for crosslingual variation in reasoning length while controlling computational cost, we therefore use a finer truncation granularity of 5\% for Multilingual AIME.

\section{Potential False Positives in Gold-in-Trace Matching}
\seclabel{sec:git_false_positive}

\begin{table}[h]
\centering
\small
\resizebox{\linewidth}{!}{
\begin{tabular}{lccc}
\toprule
Dataset & \#Questions & \#Single-Digit & \% \\
\midrule
MGSM & 250 & 38 & 15.2\% \\
AIME-2024 (Multilingual) & 30 & 0 & 0.0\% \\
AIME-2025 (Multilingual) & 30 & 0 & 0.0\% \\
\bottomrule
\end{tabular}
}
\caption{Proportion of single-digit gold answers.}
\label{tab:single_digit}
\end{table}

A potential limitation of the \textbf{Gold-in-Trace} metric is that it relies on string matching to detect whether the gold answer appears in the truncated reasoning trace. 
In principle, this may introduce false positives when intermediate values happen to match the final answer. 
For example, if the correct answer is ``5'', an intermediate step such as ``5 apples remain after step 2'' would be counted as a match, even if it is unrelated to the final solution.
However, we expect such cases to be rare in our experimental setting for the following reasons.

First, in both MGSM and Multilingual AIME, gold answers are typically multi-digit integers.
The probability that an unrelated intermediate value exactly matches the final answer is therefore low.
To quantify this, we report the proportion of single-digit answers in each dataset in Table~\ref{tab:single_digit}.

Second, Gold-in-Trace and AUGC are computed \emph{only over correctly solved instances}. 
If the model produces the correct final answer and the same value appears earlier in the reasoning trace, this is more likely to reflect early internal convergence rather than random coincidence.

While false positives cannot be completely ruled out, the combination of multi-digit answer distributions and conditioning on correct predictions substantially reduces their likelihood.
We therefore consider Gold-in-Trace to be a reasonable proxy for early answer articulation in our setting.

\input{./truncation_results}
\newpage
\input{./similarity_results}
\newpage
\input{./perturbation}

\section{Experimental Details}\seclabel{prompt_details}

\subsection{Language Control}\seclabel{language_control}

Following \citet{qi-etal-2025-models,zhao2025comprehensiveevaluationmultilingualchainofthought}, we adopt a set of complementary strategies to ensure that the model’s explicit reasoning trace is produced in the same language as the input prompt.

\begin{figure*}[htbp]
\centering
\includegraphics[width=0.90\textwidth]{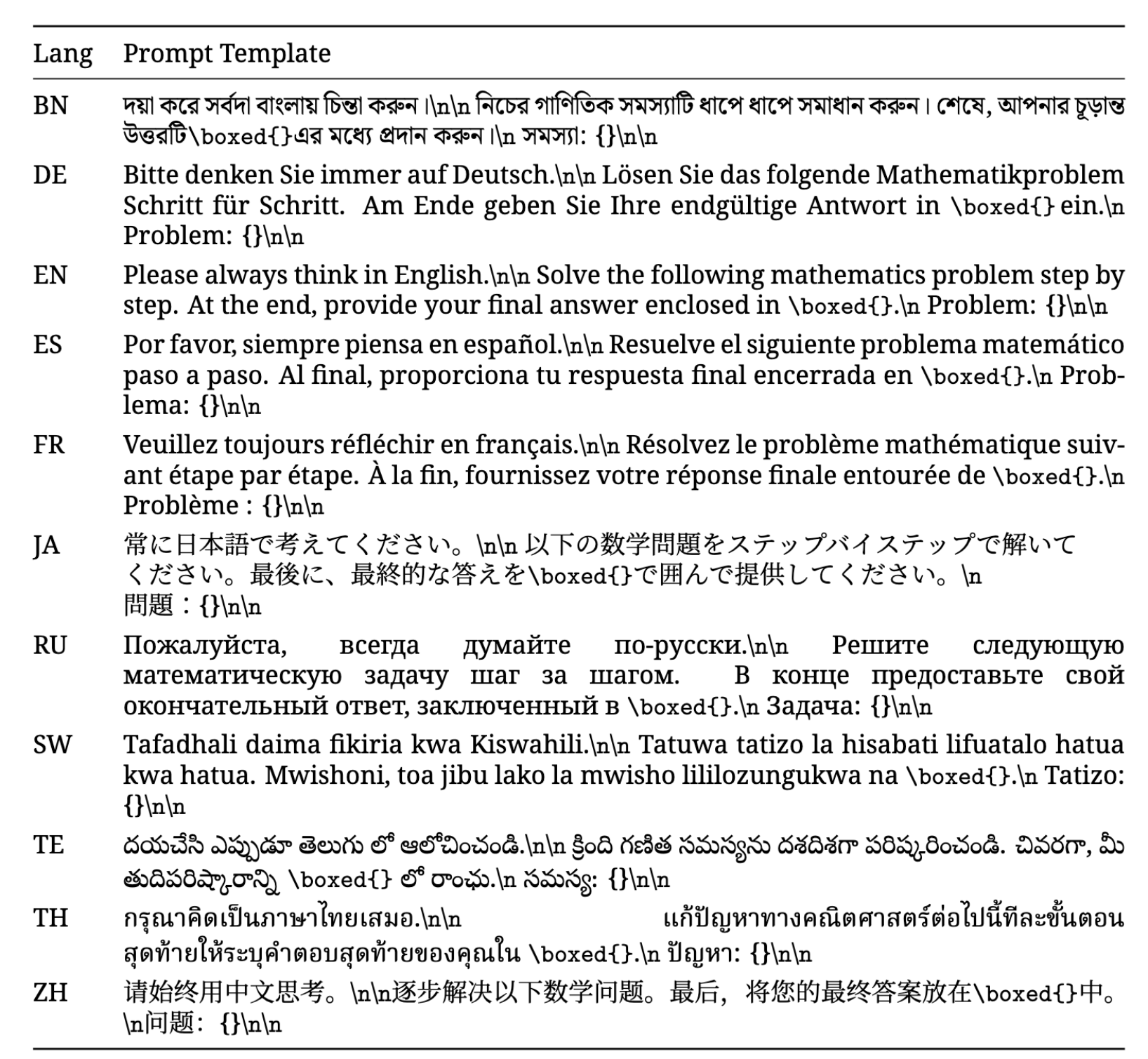}
\caption{Language-specific prompt templates (containing the explicit language instruction) used for controlling the reasoning language.}
\label{tab:prompt_template}
\end{figure*}

\paragraph{Prompt Formation}
We prepend each input with an explicit language-specific instruction that explicitly specifies the target language for reasoning.
This instruction is embedded into a standardized prompt template, shown in Figure~\ref{tab:prompt_template}, which is used consistently across all languages.

\begin{figure}[htbp]
\centering
\includegraphics[width=0.49\textwidth]{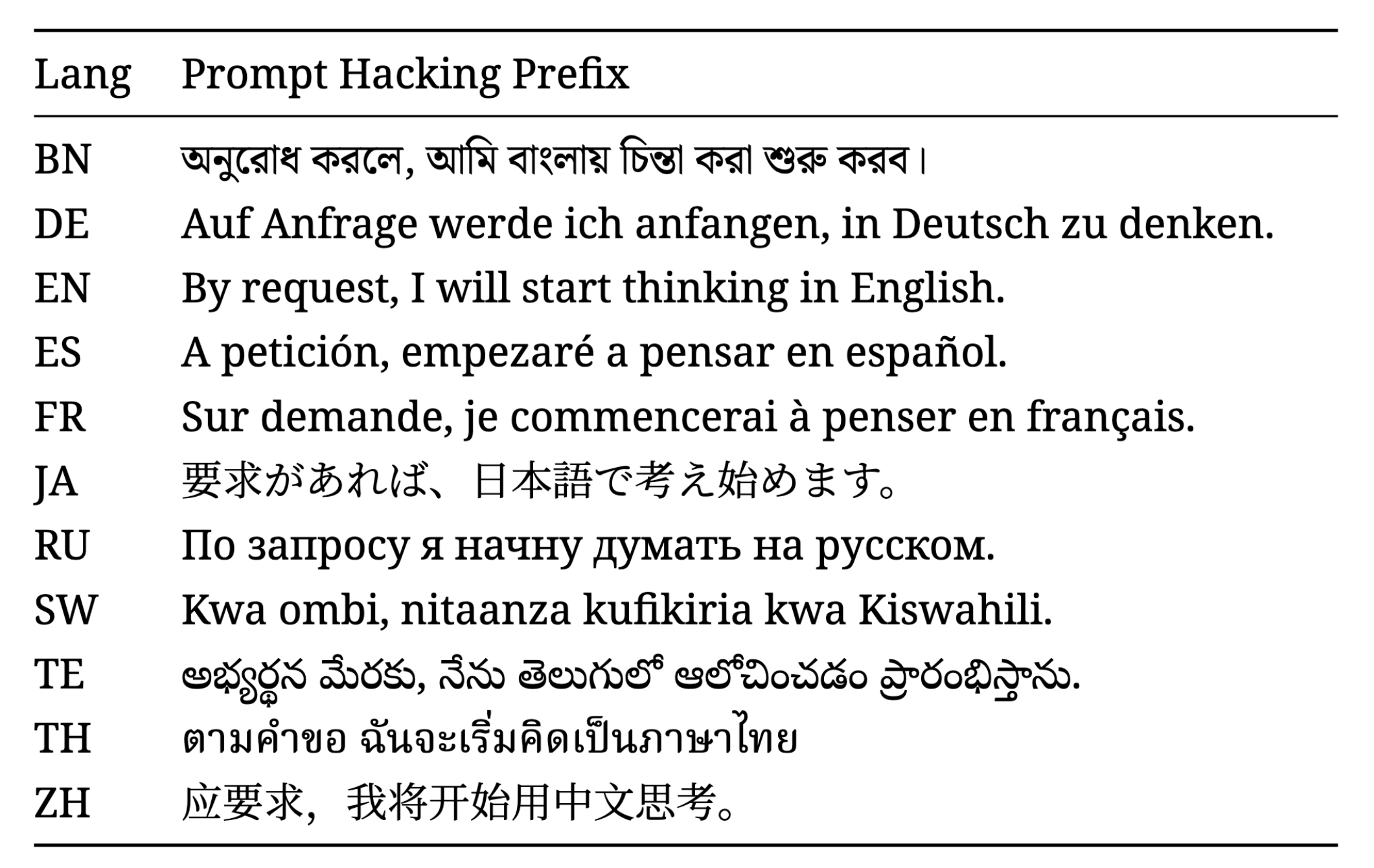}
\caption{Language-specific prompt-hacking prefixes used to reinforce language control. Each prefix is inserted immediately after the \texttt{<think>} tag to steer the model's explicit reasoning trace to match the prompt language.}
\label{tab:prompt_hacking}
\end{figure}

\paragraph{Prompt Hacking}
Even with explicit language instructions in the prompt, LLMs may still generate reasoning traces in a language different from that of the input, a phenomenon observed in prior work on multilingual reasoning and language mixing~\citep{wang-etal-2025-language-mixing,qi-etal-2025-models,zhao2025comprehensiveevaluationmultilingualchainofthought}.
Such behavior is undesirable in our setting, as it confounds cross-lingual comparisons of latent reasoning by introducing variation at the level of explicit verbalization.
To mitigate this issue, we adopt a \emph{prompt-hacking} strategy~\citep{schulhoff-etal-2023-ignore,benjamin2024systematicallyanalyzingpromptinjection} that reinforces the language constraint at inference time.
Concretely, following prior work~\citep{qi-etal-2025-models,zhao2025comprehensiveevaluationmultilingualchainofthought}, we insert a language-specific prefix immediately after the opening \texttt{<think>} tag (e.g., ``\texttt{By request, I will begin to think in English}'').
This prefix explicitly restates the target language at the onset of the reasoning trace and reliably steers the model to produce explicit reasoning in the same language as the prompt until the closing \texttt{</think>} tag is reached.
The complete set of prompt-hacking prefixes used in our experiments is listed in Figure~\ref{tab:prompt_hacking}.

\begin{figure}[htbp]
\centering
\includegraphics[width=0.30\textwidth]{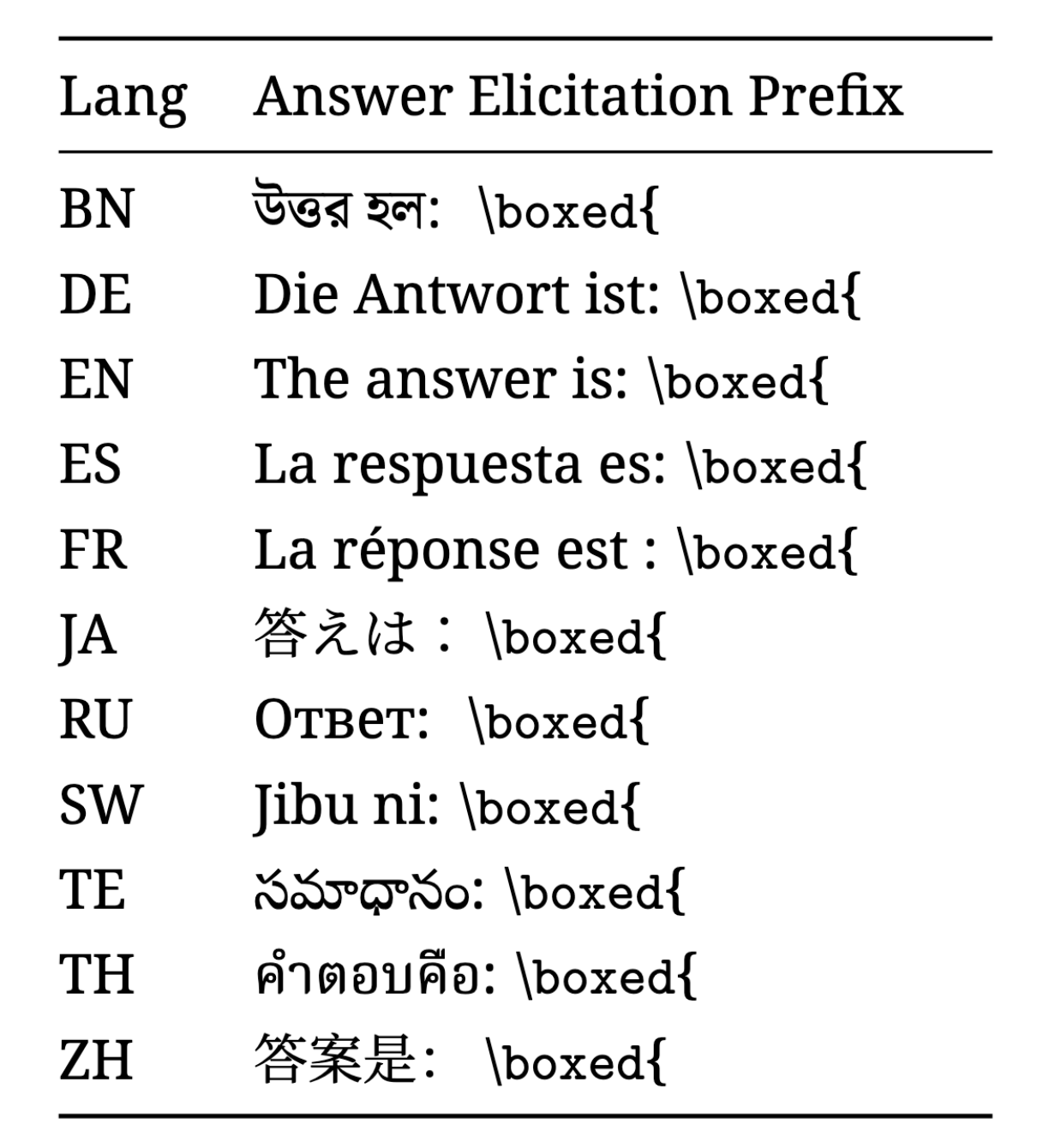}
\caption{Language-specific answer-elicitation prefixes used to directly prompt the model for a final numerical answer. Each prefix is appended immediately after the \texttt{</think>} tag to elicit the prediction.}
\label{tab:answer_elicitation}
\end{figure}

\paragraph{Answer Elicitation}
To analyze early answer formation during truncated reasoning, we aim to elicit the model's prediction immediately after the visible reasoning prefix, without allowing further reasoning steps.
Accordingly, we append a language-specific answer-elicitation prefix directly after the closing \texttt{</think>} tag.
This prefix prompts the model to produce only the final numerical answer, preventing additional reasoning or thought continuation beyond the truncated trace.
The answer-elicitation prefixes used in our experiments are shown in Figure~\ref{tab:answer_elicitation}.

\subsection{Pass@k Evaluation}

For each question, we generate 10 independent samples and evaluate each prediction separately.
Correctness is assessed via exact matching.
Following prior work~\citep{qi-etal-2025-models,zhao2025comprehensiveevaluationmultilingualchainofthought}, models are instructed to enclose their final answers in \texttt{\textbackslash boxed\{\}}, from which the boxed content is extracted and compared against the gold answer using mathematical equivalence rather than raw string matching.

\section{Environment and Hyperparameters}

We set the maximum generation length to 4K tokens for the MGSM benchmark and 16K tokens for Multilingual AIME across all evaluated models.
Unless stated otherwise, we adopt the recommended generation configurations provided on HuggingFace.\footnote{\url{https://huggingface.co}}
In particular, we use a temperature of 0.6 and top-$p$ sampling with $p=0.95$.
For reproducibility, the random seed is fixed to 42.

Experiments on identifying multilingual latent reasoning (cf. \secref{surface_dynamics}) and on memorization versus latent reasoning (cf. \secref{memorization}) are conducted using the vLLM framework,\footnote{\url{https://vllm.ai/}}
while experiments analyzing latent dynamics (cf. \secref{latent_dynamics}) are performed using the HuggingFace \texttt{Transformers} library.\footnote{\url{https://huggingface.co/docs/transformers}}

All experiments are run on NVIDIA A100 GPUs (80\,GB) and NVIDIA RTX A6000 GPUs (48\,GB).

\end{document}

%% file: truncation_results.tex
\section{Complete Truncation Results}\seclabel{truncation_results}

\subsection{Truncation Curves}

\begin{figure*}
    \centering
    \includegraphics[width=0.23\textwidth]{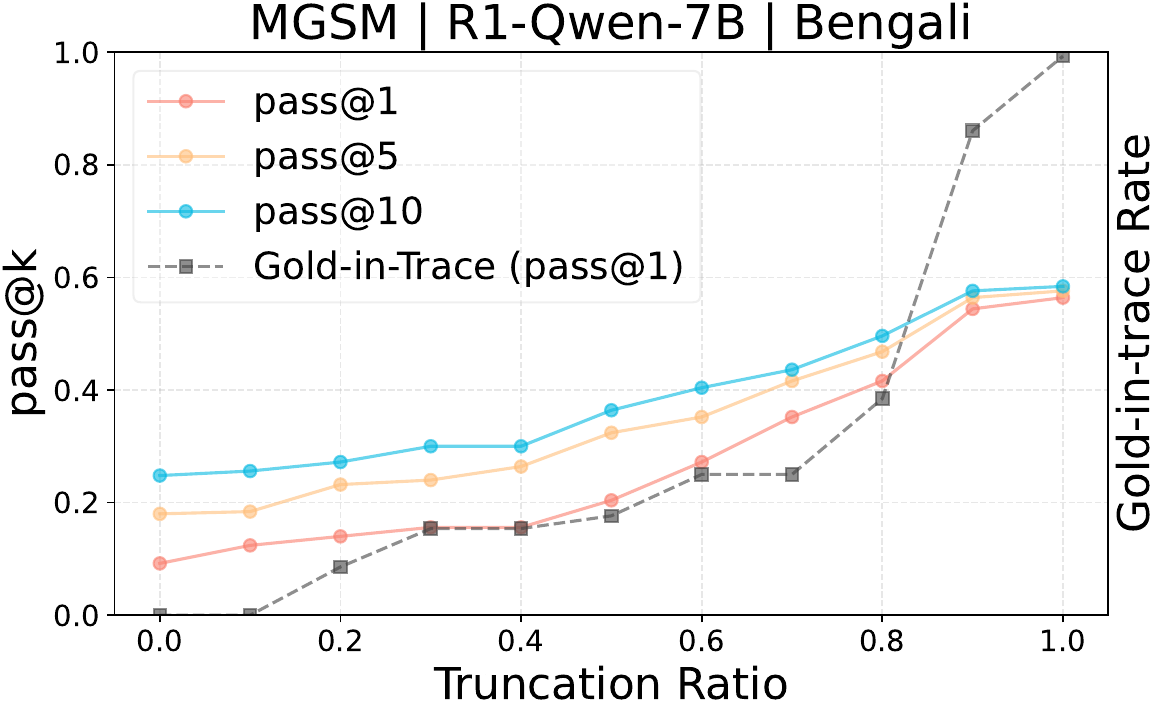}
    \includegraphics[width=0.23\textwidth]{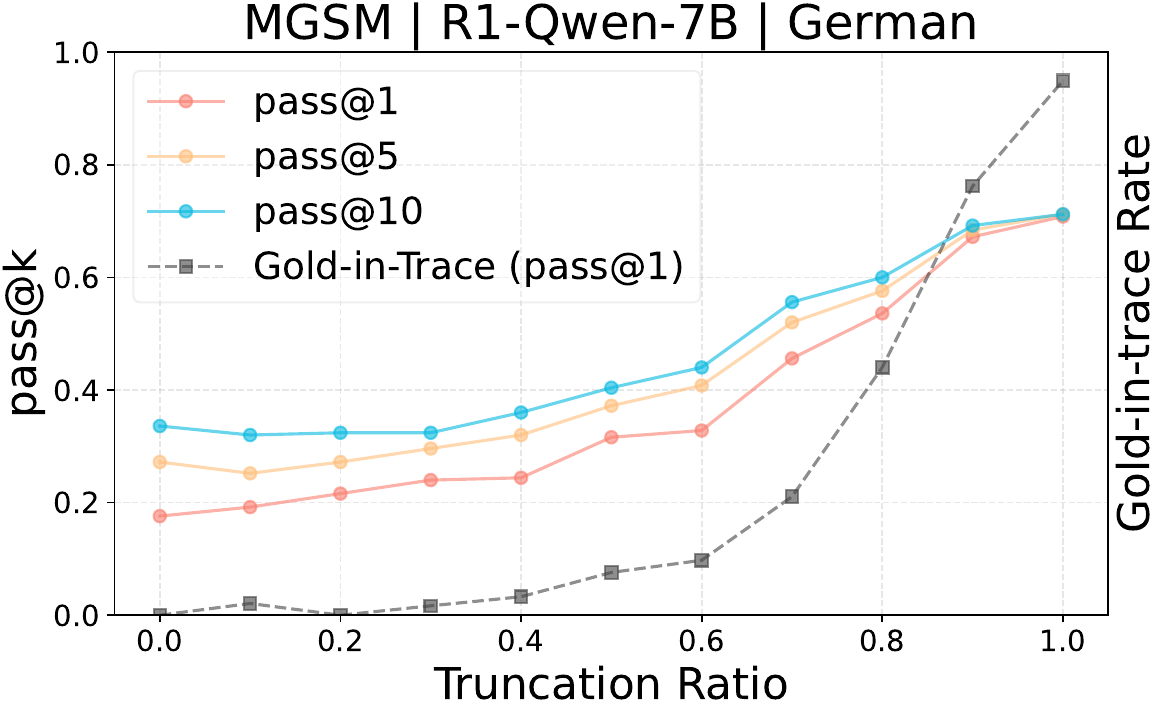}
    \includegraphics[width=0.23\textwidth]{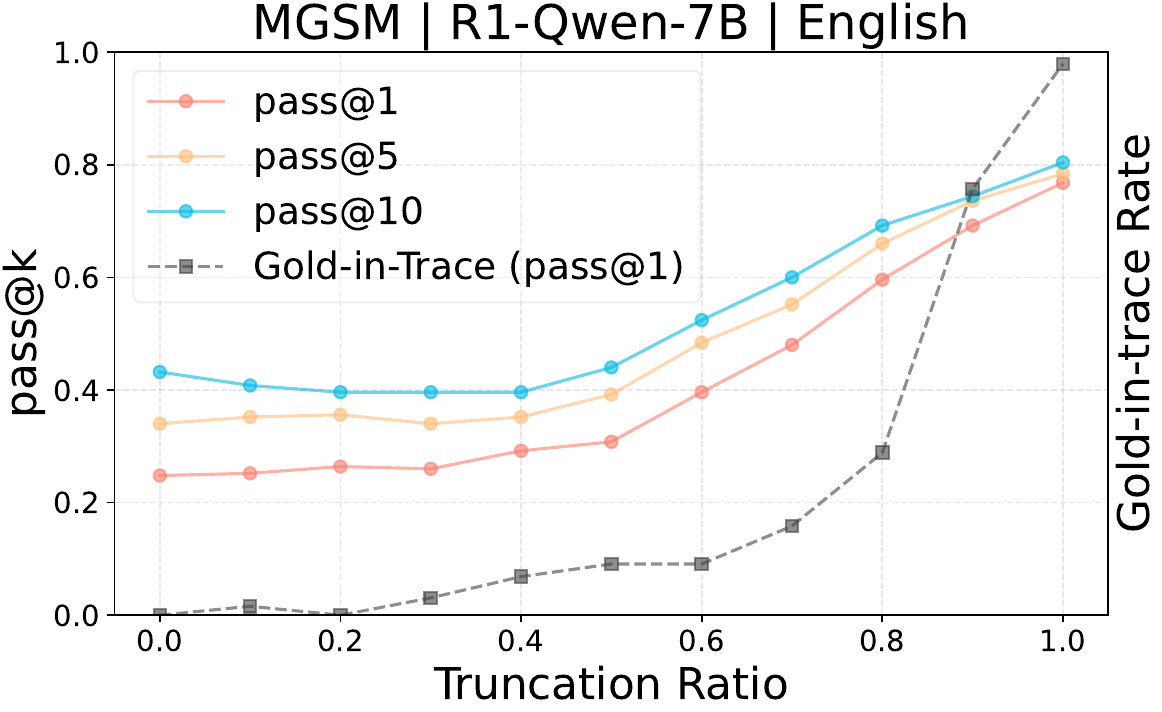}
    \includegraphics[width=0.23\textwidth]{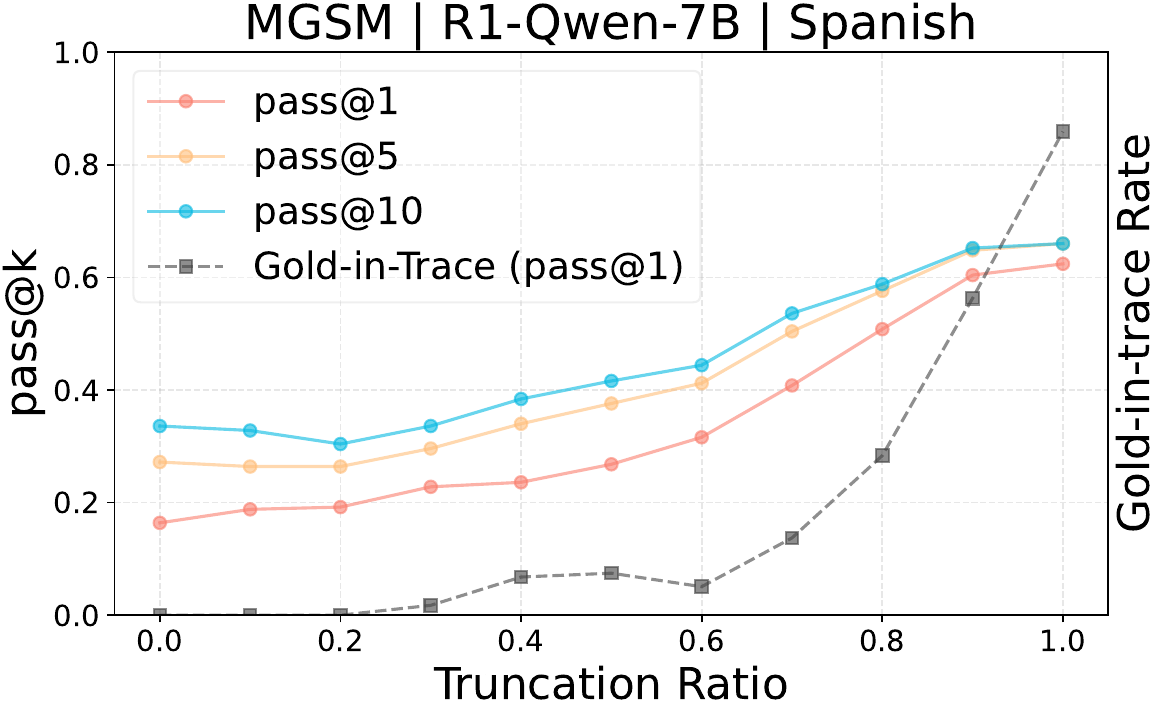}
    \includegraphics[width=0.23\textwidth]{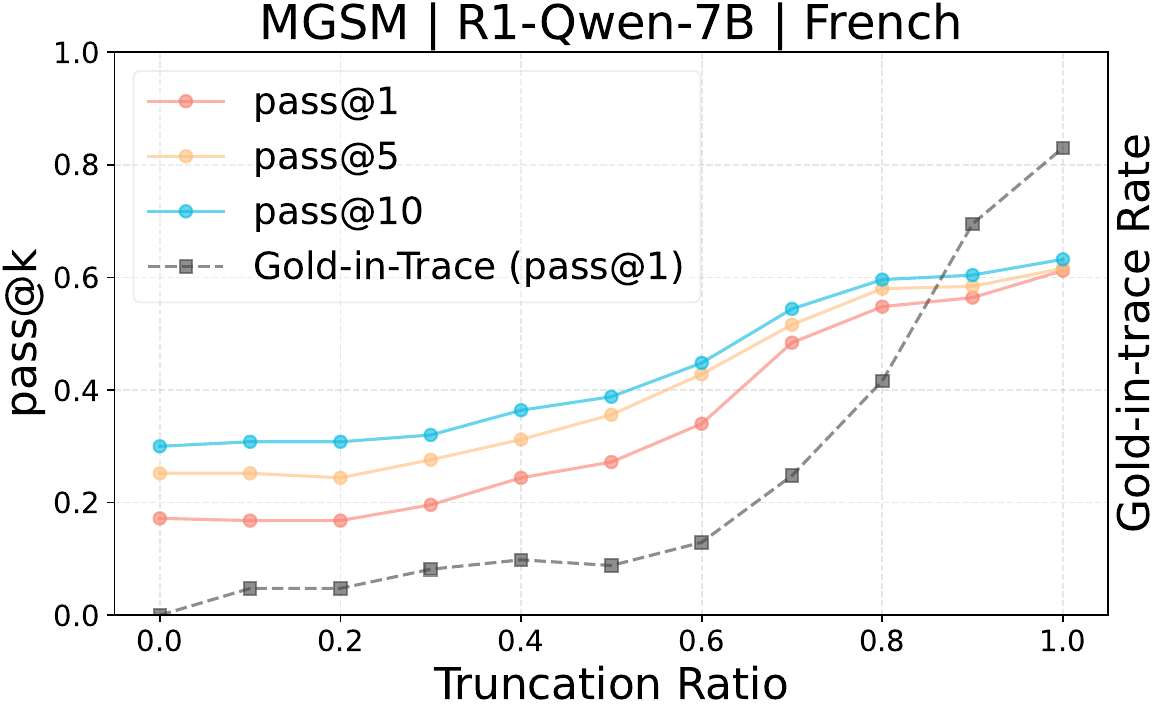}
    \includegraphics[width=0.23\textwidth]{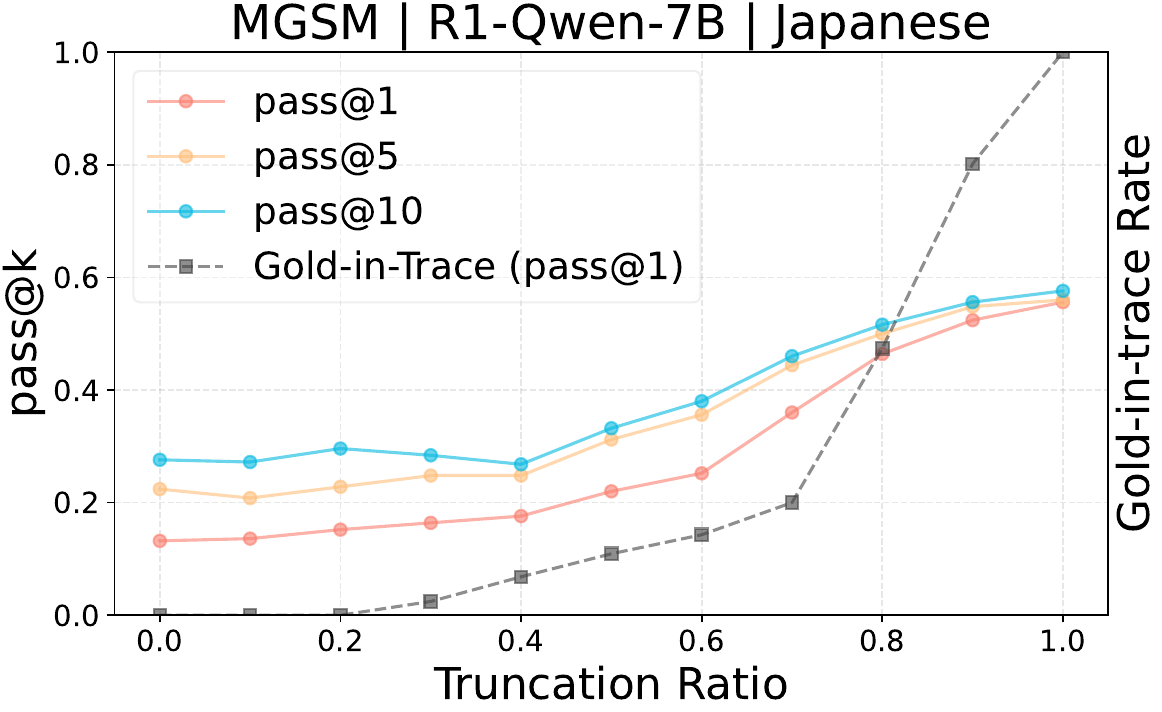}
    \includegraphics[width=0.23\textwidth]{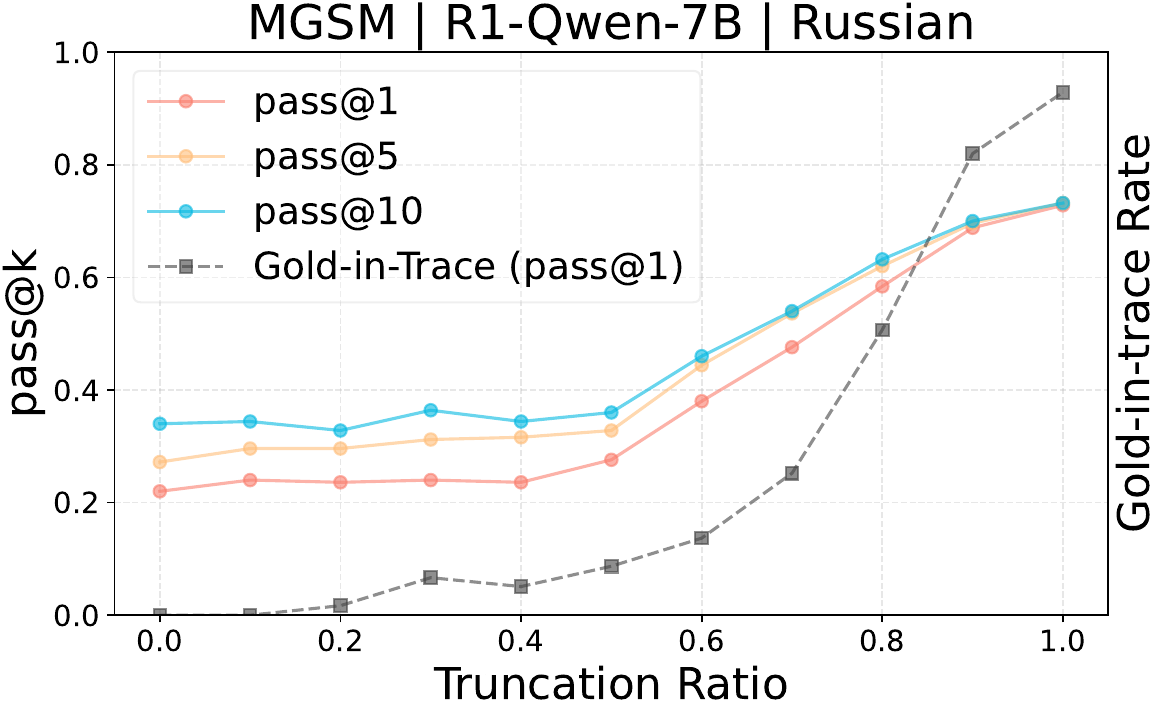}
    \includegraphics[width=0.23\textwidth]{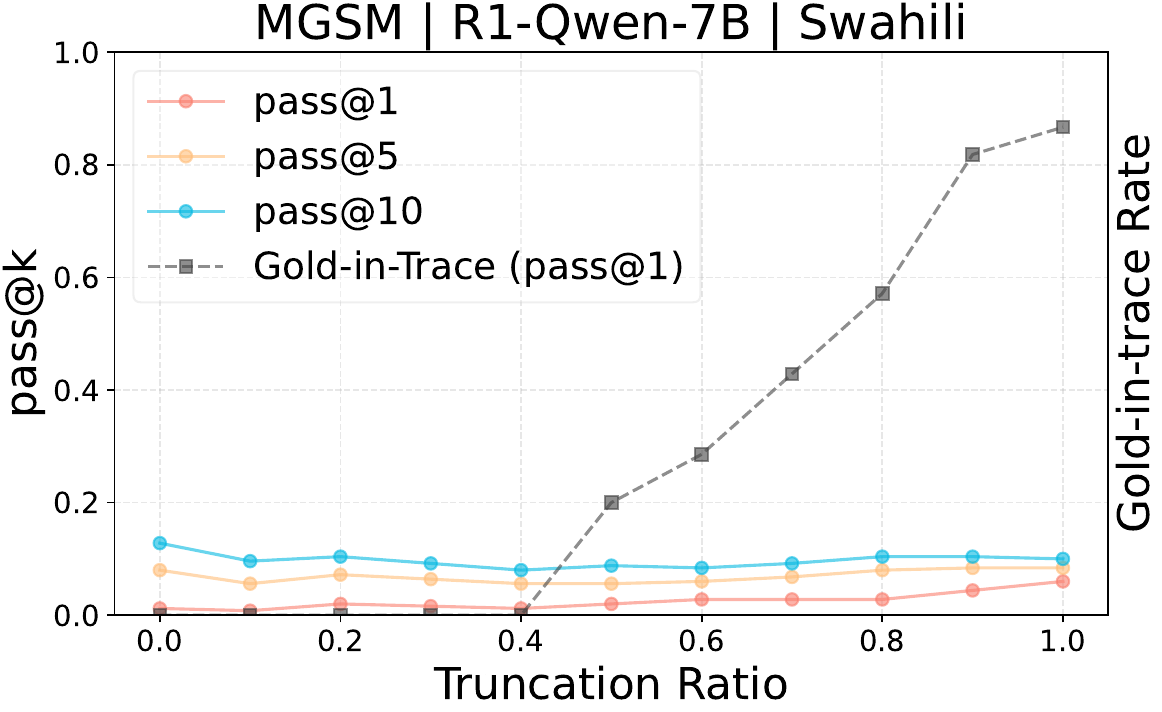}
    \includegraphics[width=0.23\textwidth]{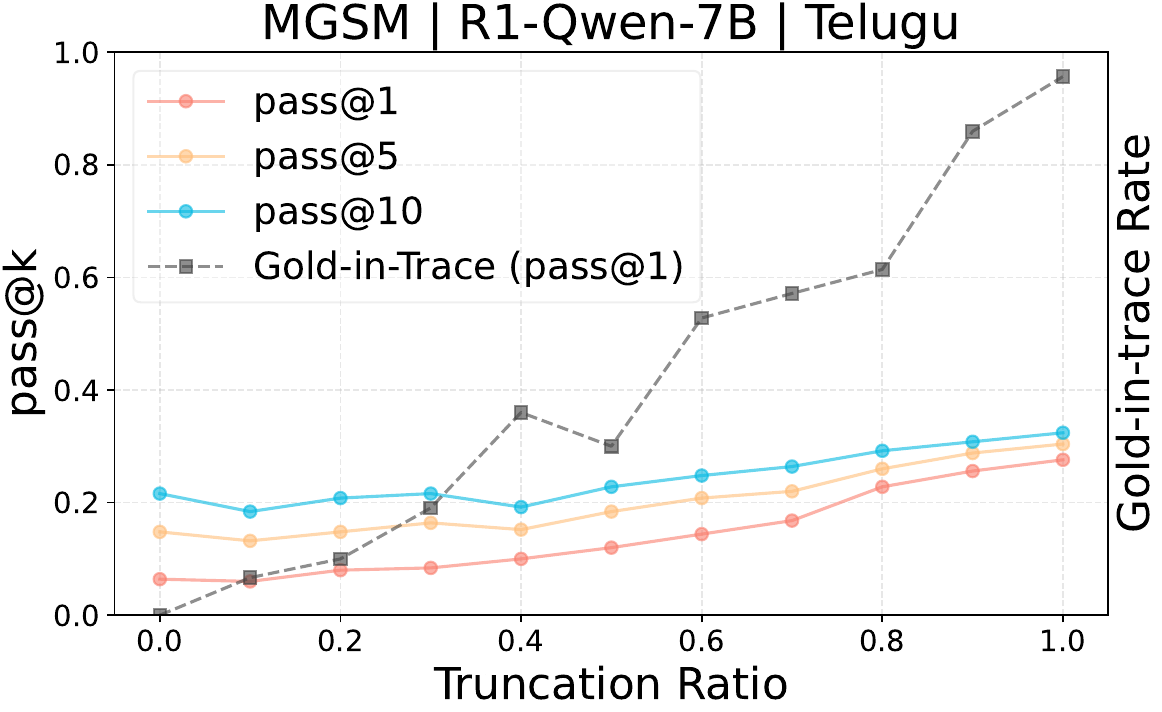}
    \includegraphics[width=0.23\textwidth]{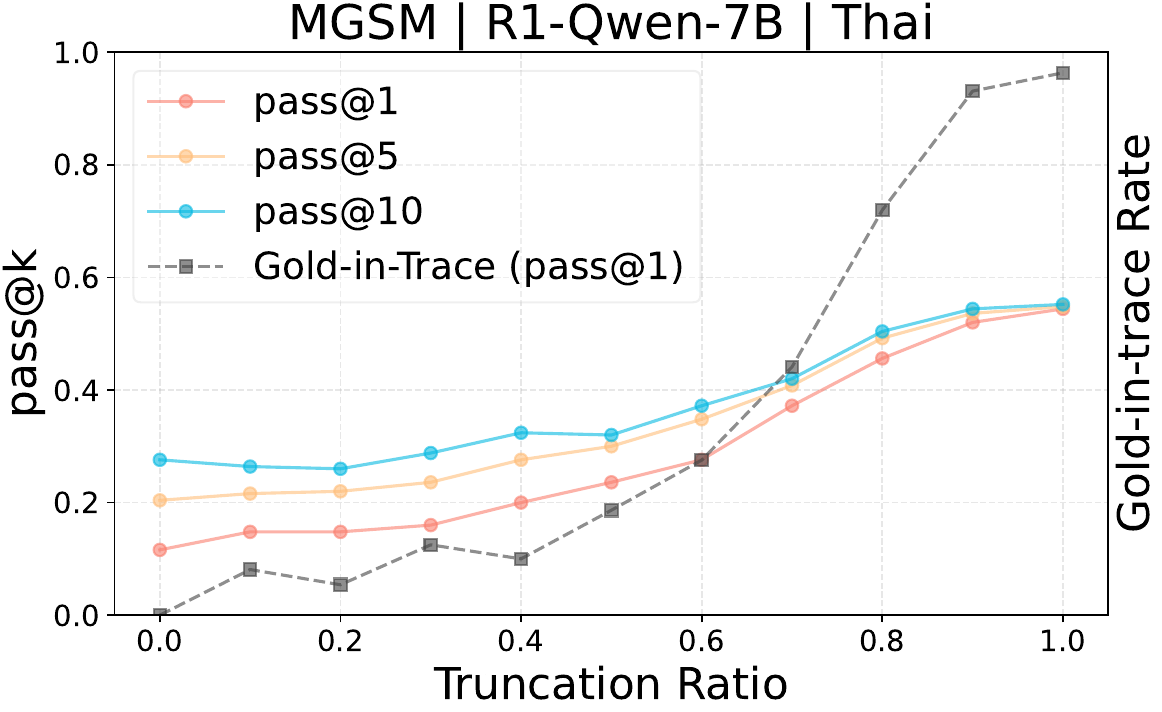}
    \includegraphics[width=0.23\textwidth]{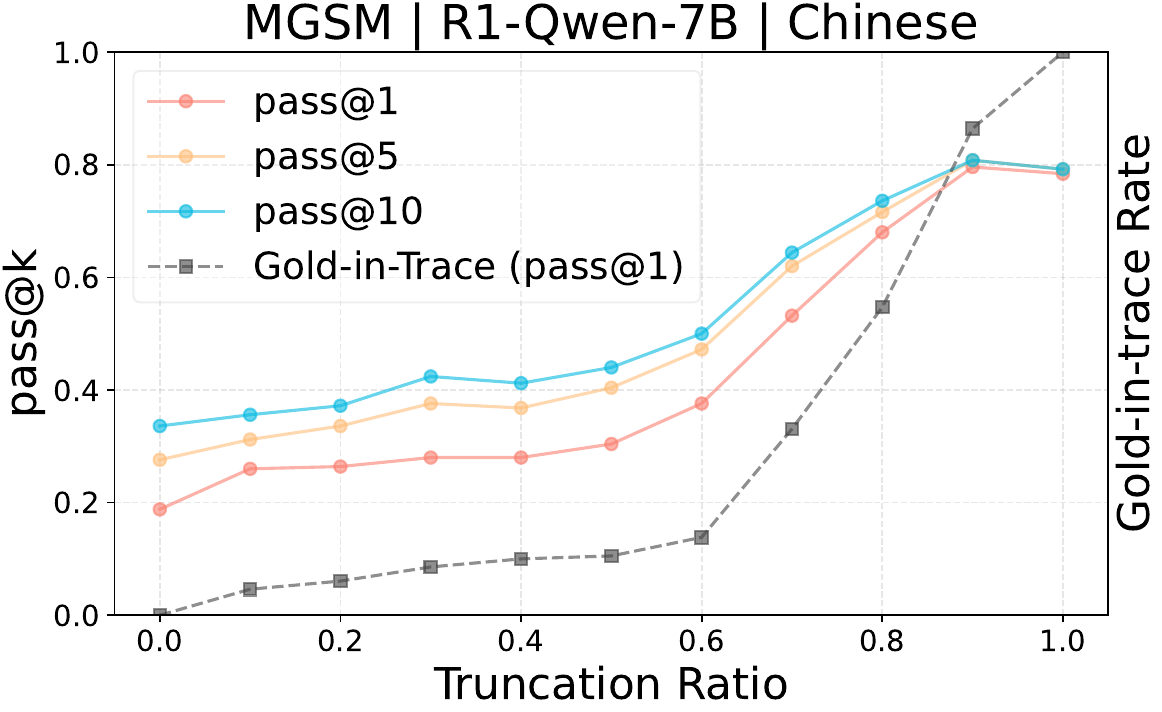}
    \caption{
    Pass@$k$ accuracy ($k=1,5,10$) and gold-in-trace rate under reasoning-trace truncation for \textbf{R1-Qwen-7B} on \textbf{MGSM}. The model shows stronger latent reasoning in high-resource languages (e.g., English).
    }
    \label{fig:truncation_7b_mgsm}
\end{figure*}

\begin{figure*}
    \centering
    \includegraphics[width=0.23\textwidth]{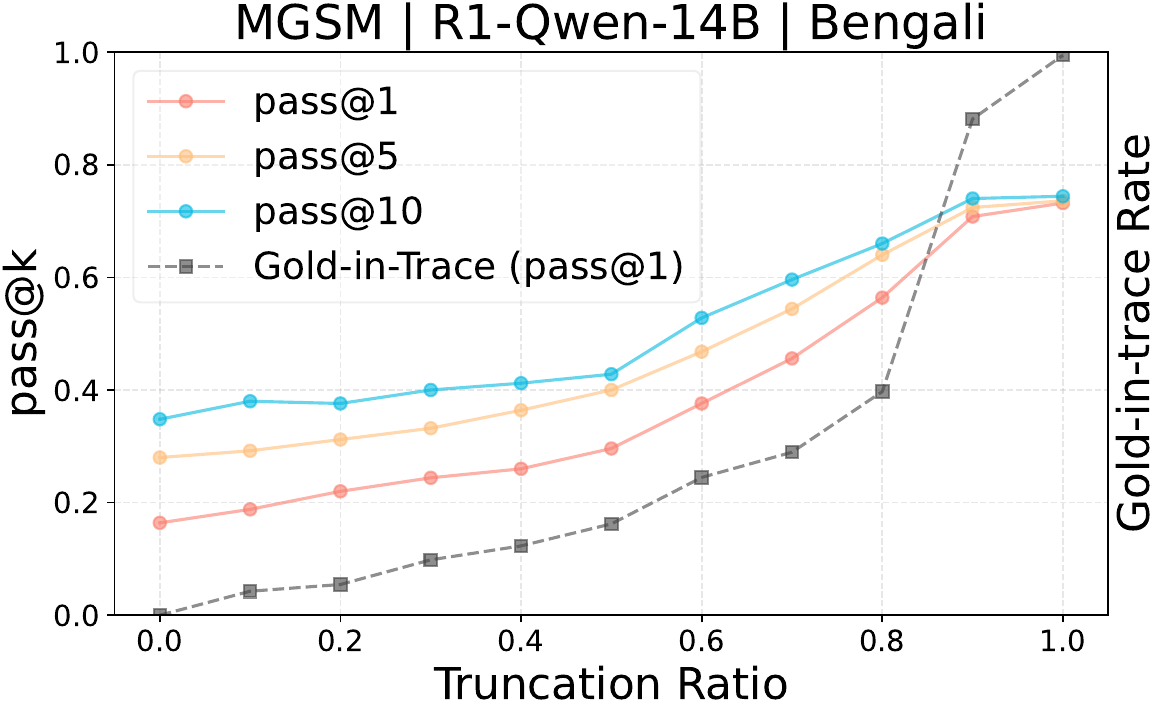}
    \includegraphics[width=0.23\textwidth]{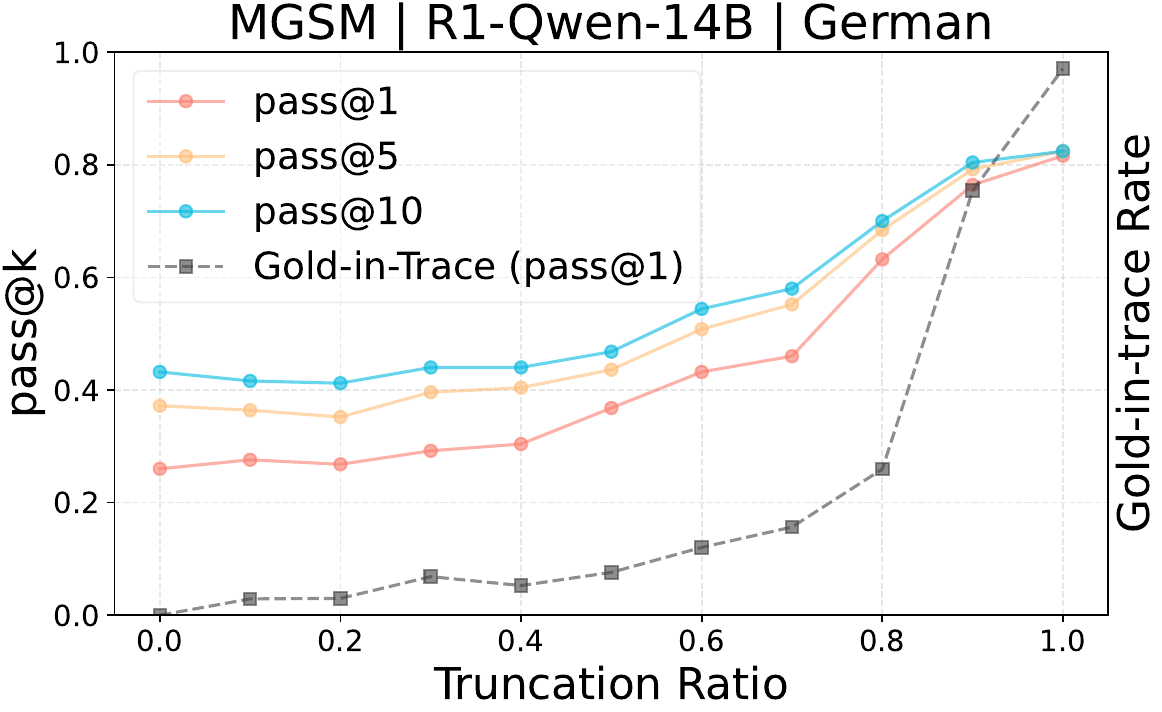}
    \includegraphics[width=0.23\textwidth]{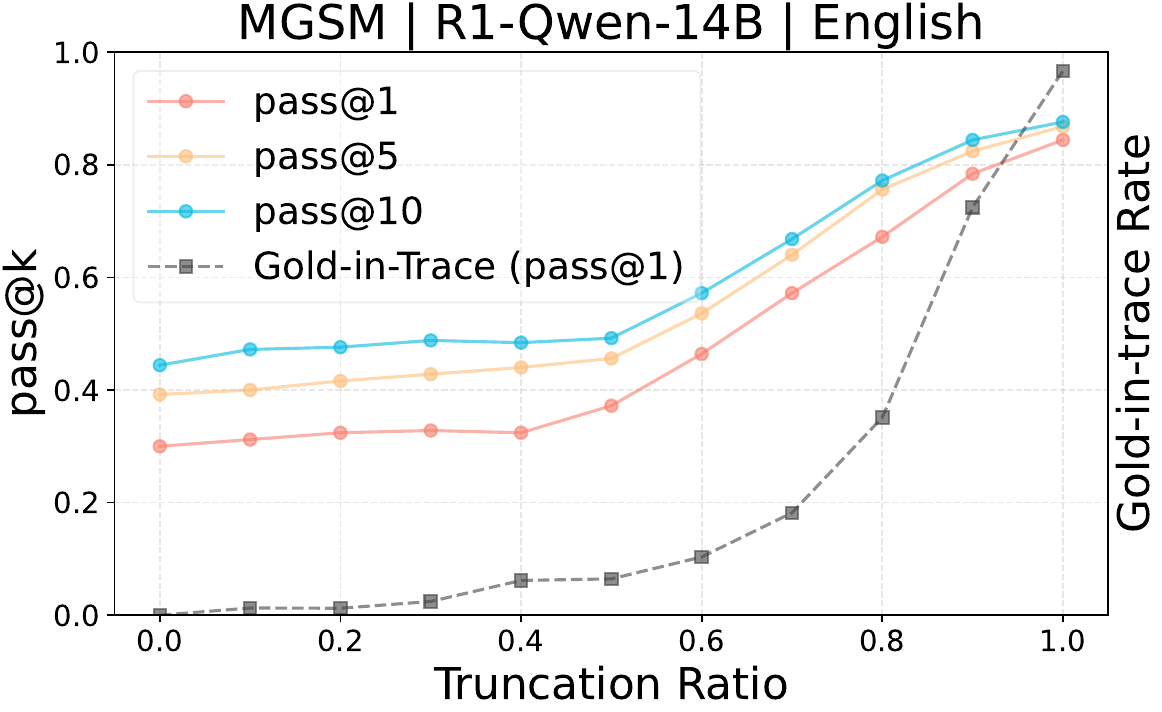}
    \includegraphics[width=0.23\textwidth]{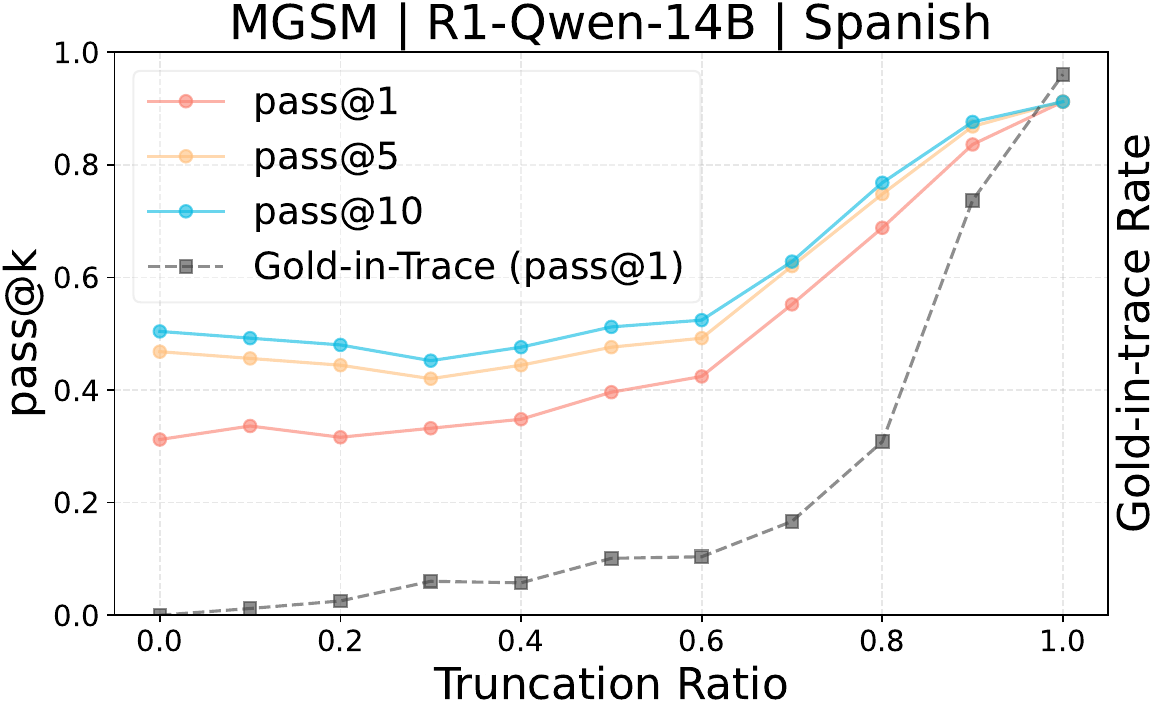}
    \includegraphics[width=0.23\textwidth]{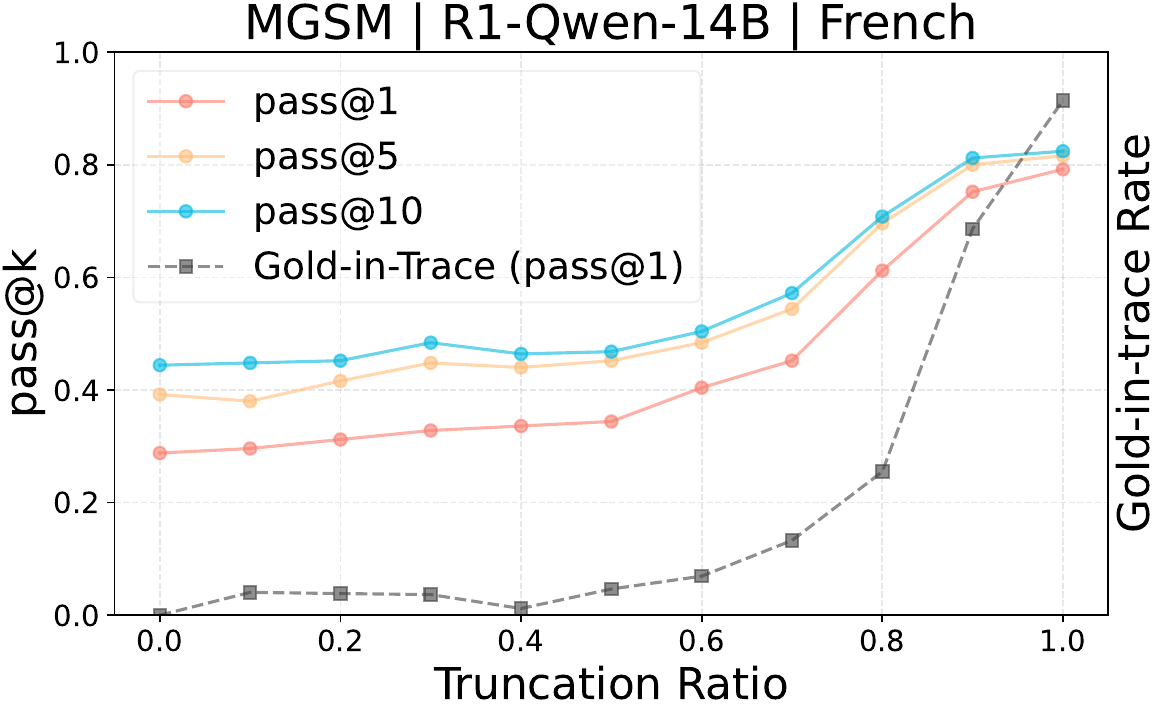}
    \includegraphics[width=0.23\textwidth]{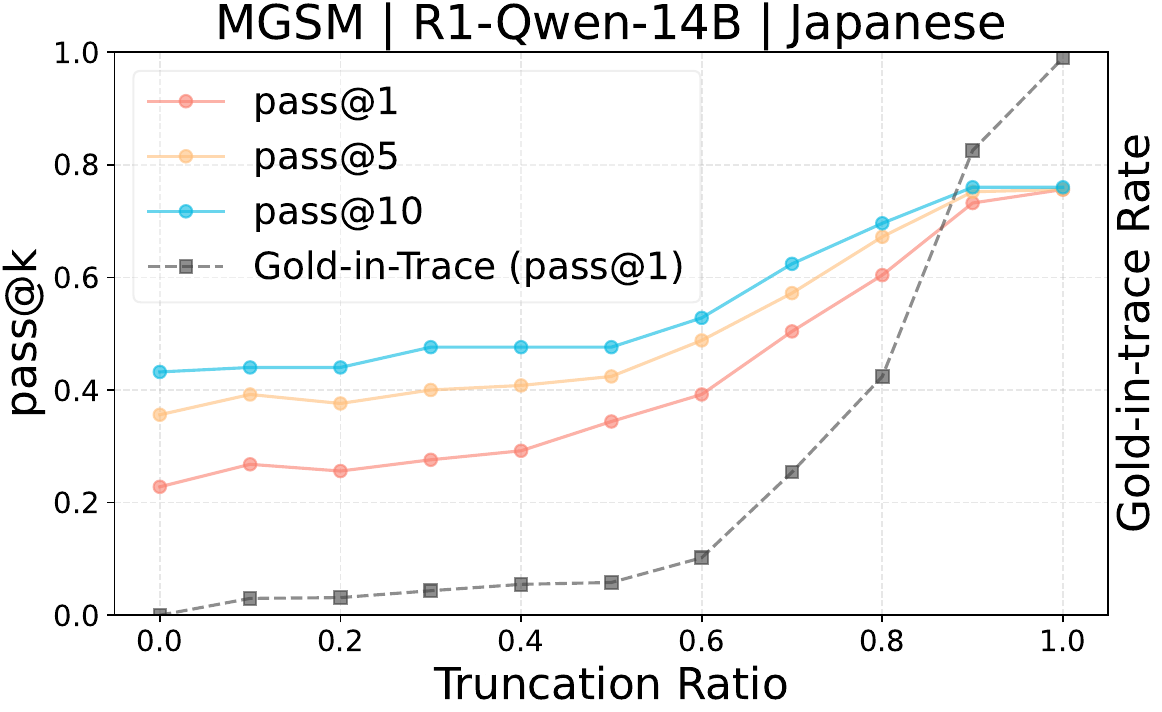}
    \includegraphics[width=0.23\textwidth]{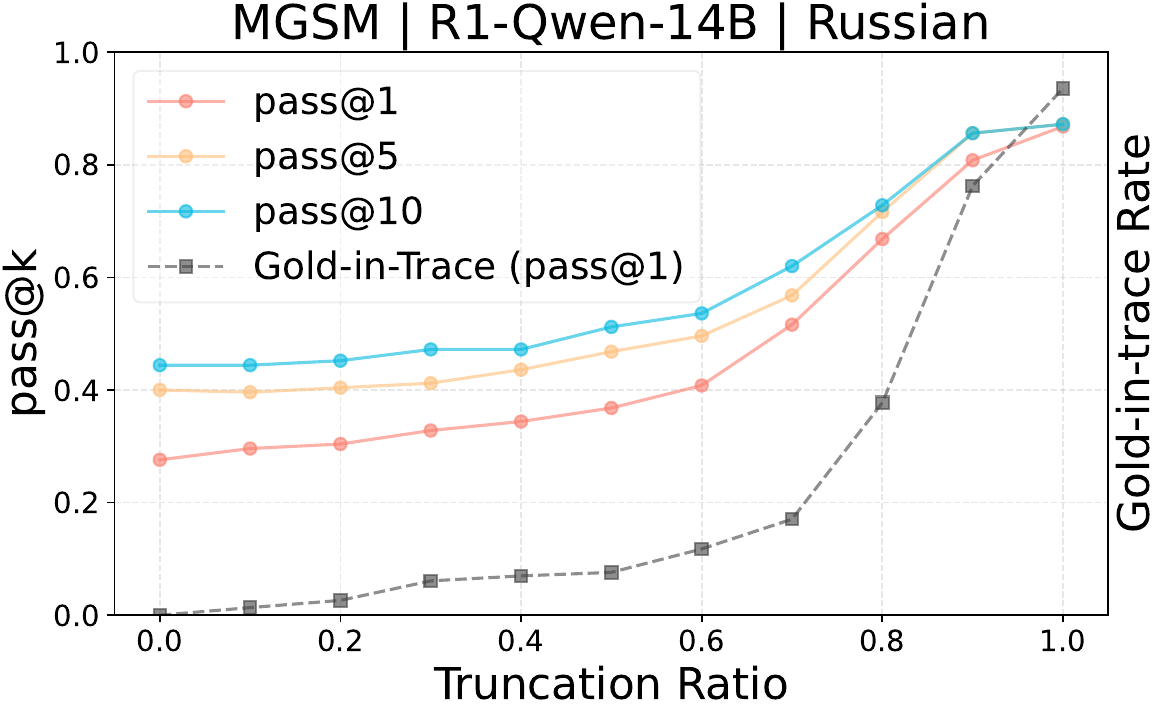}
    \includegraphics[width=0.23\textwidth]{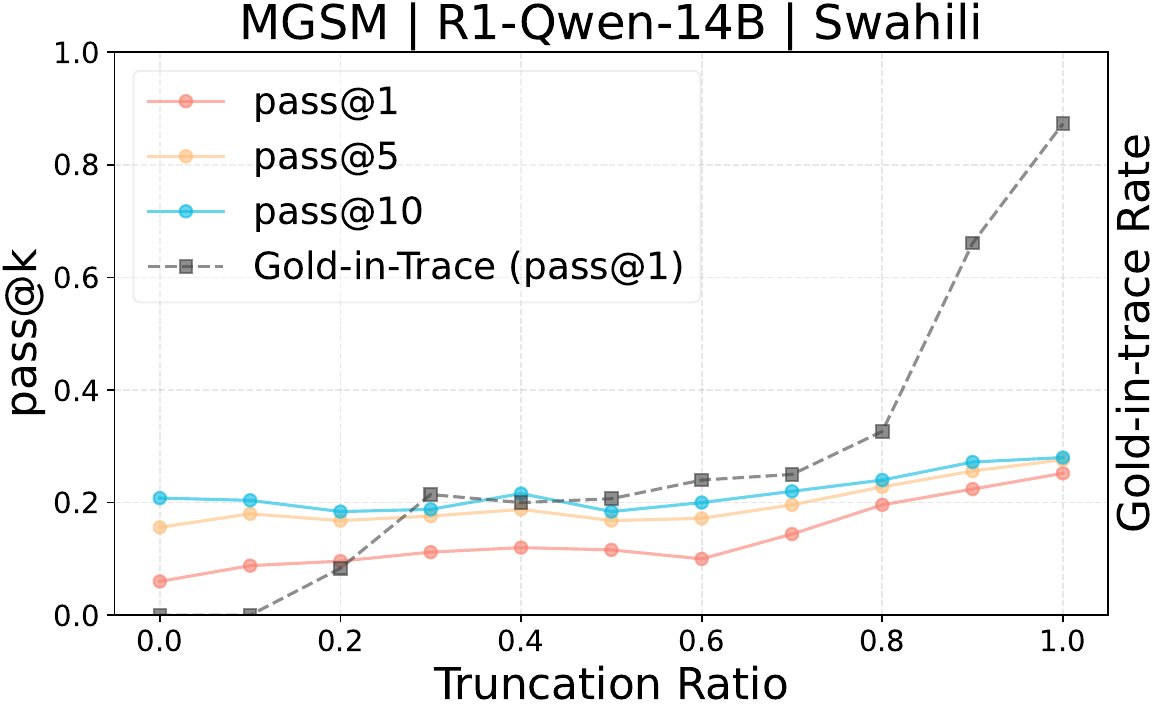}
    \includegraphics[width=0.23\textwidth]{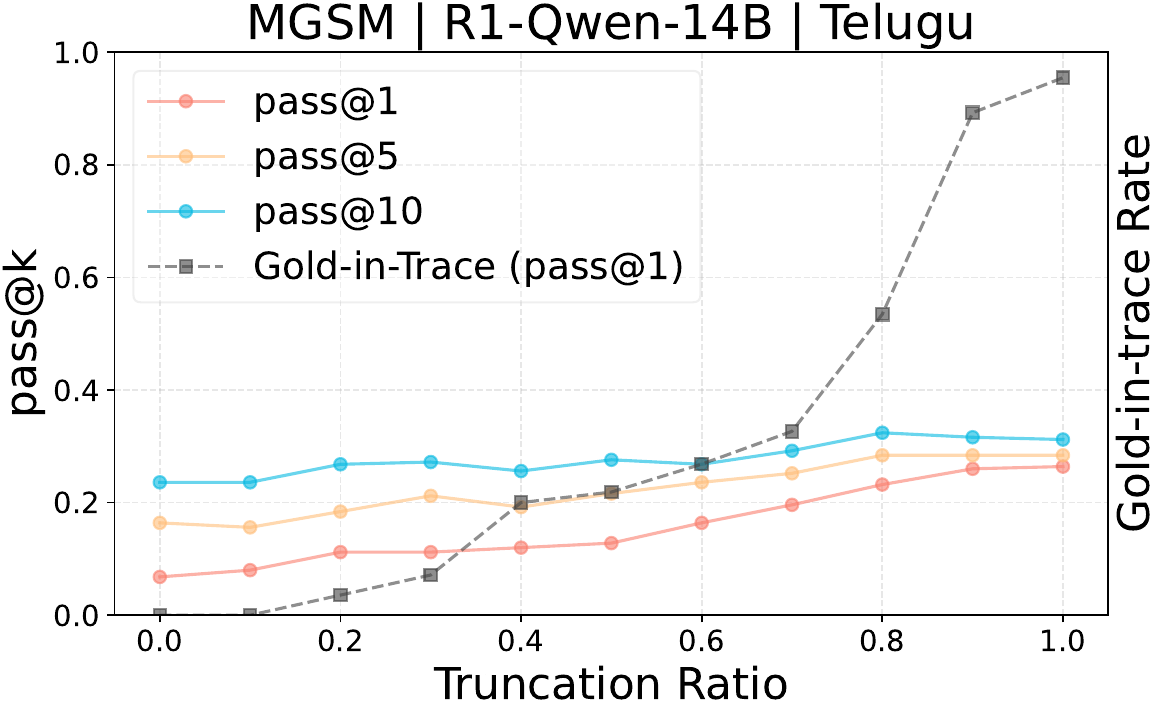}
    \includegraphics[width=0.23\textwidth]{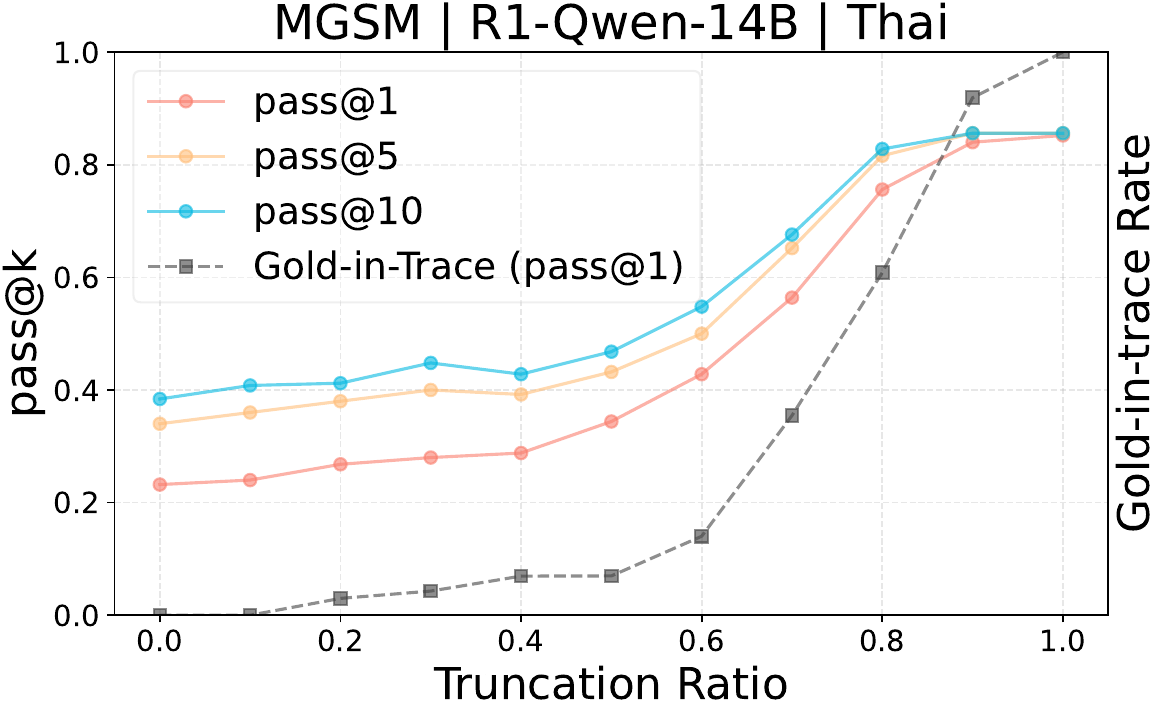}
    \includegraphics[width=0.23\textwidth]{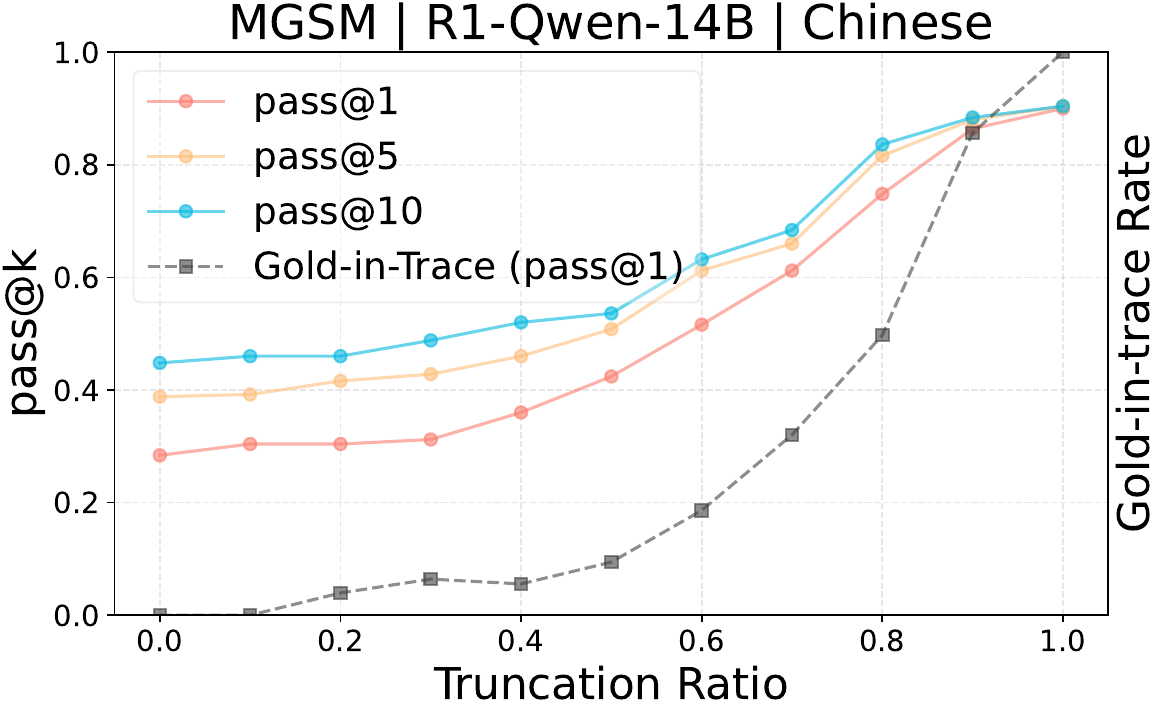}
    \caption{Pass@$k$ accuracy ($k=1,5,10$) and gold-in-trace rate under reasoning-trace truncation for \textbf{R1-Qwen-14B} on \textbf{MGSM}. The model shows stronger latent reasoning in high-resource languages (e.g., English).}
    \label{fig:truncation_14b_mgsm}
\end{figure*}

\begin{figure*}
    \centering
    \includegraphics[width=0.23\textwidth]{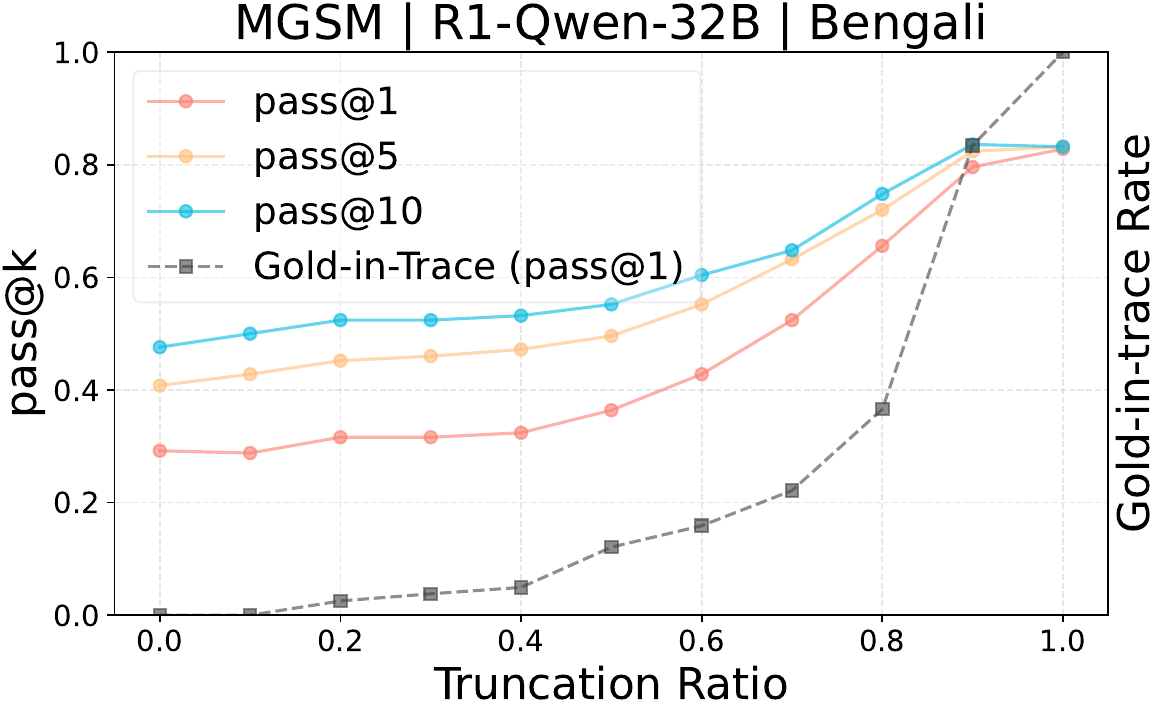}
    \includegraphics[width=0.23\textwidth]{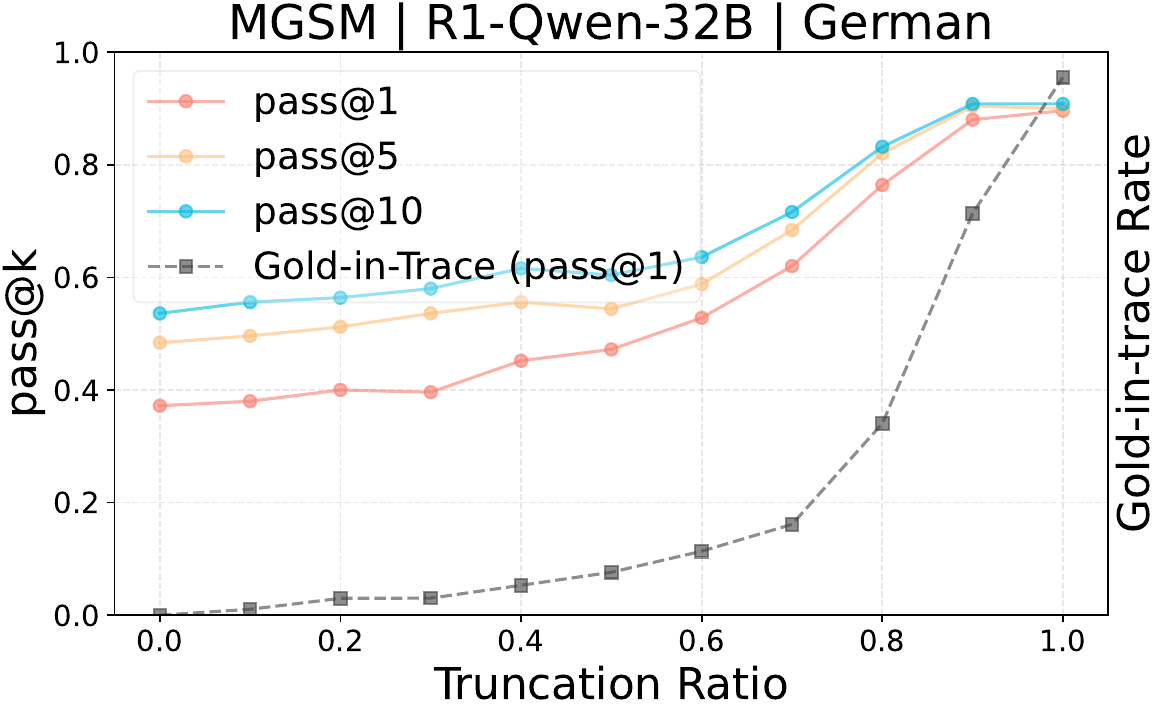}
    \includegraphics[width=0.23\textwidth]{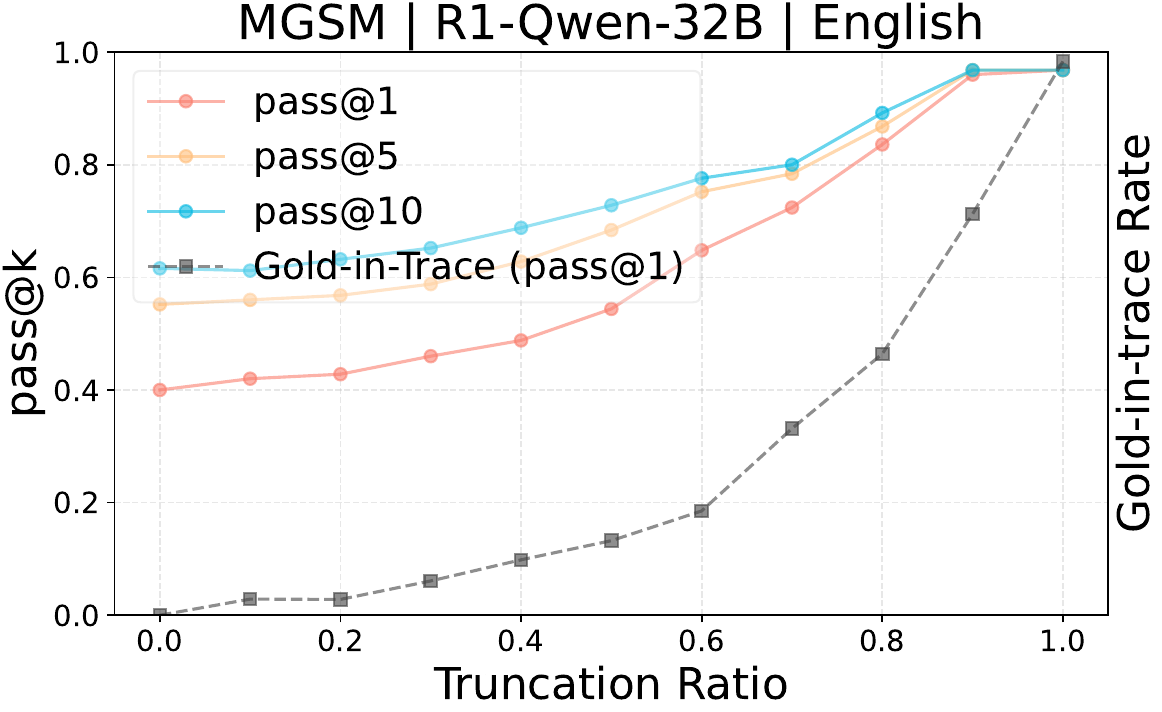}
    \includegraphics[width=0.23\textwidth]{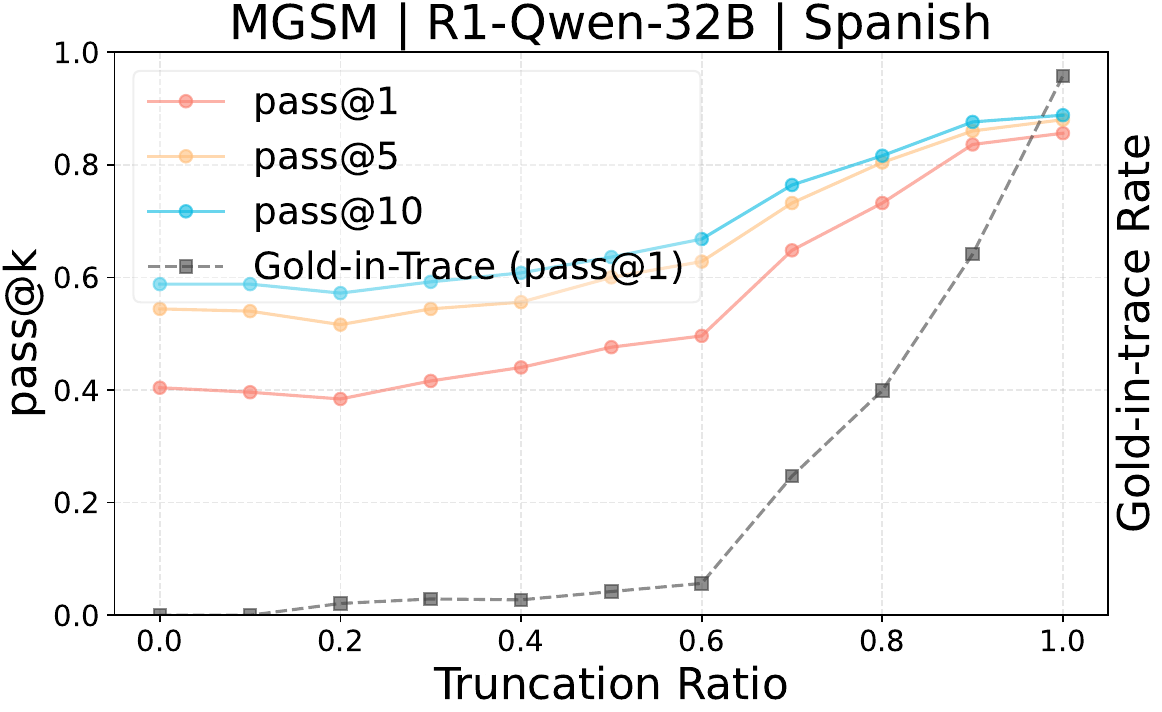}
    \includegraphics[width=0.23\textwidth]{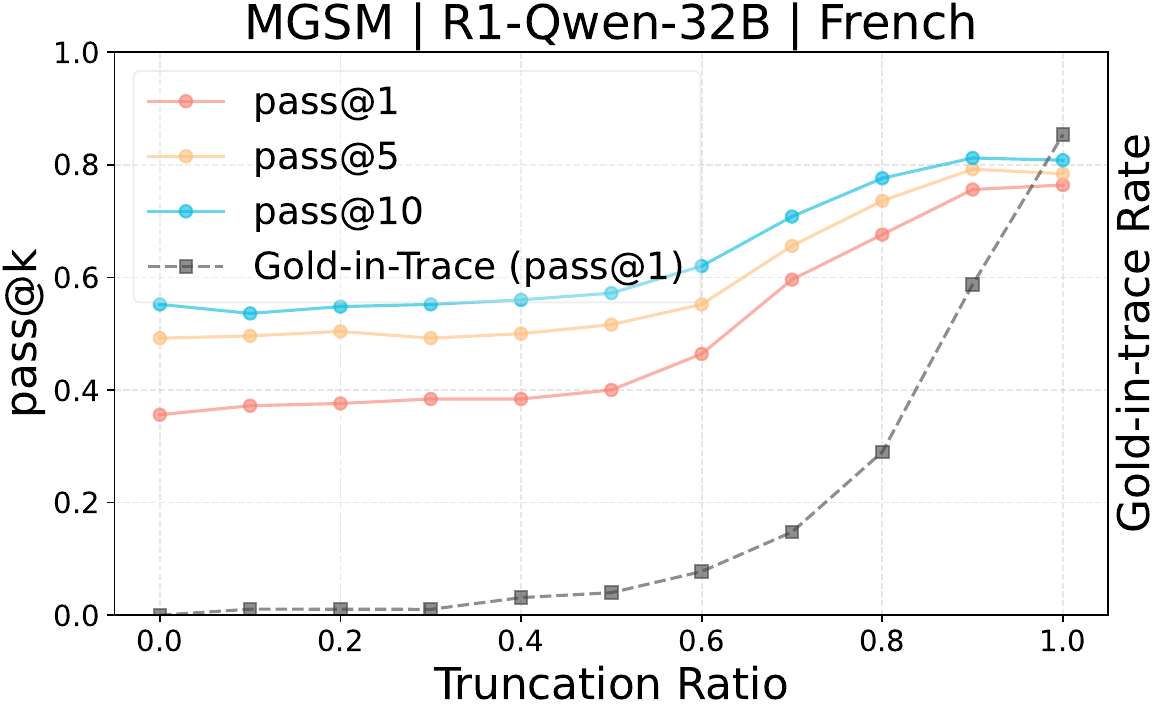}
    \includegraphics[width=0.23\textwidth]{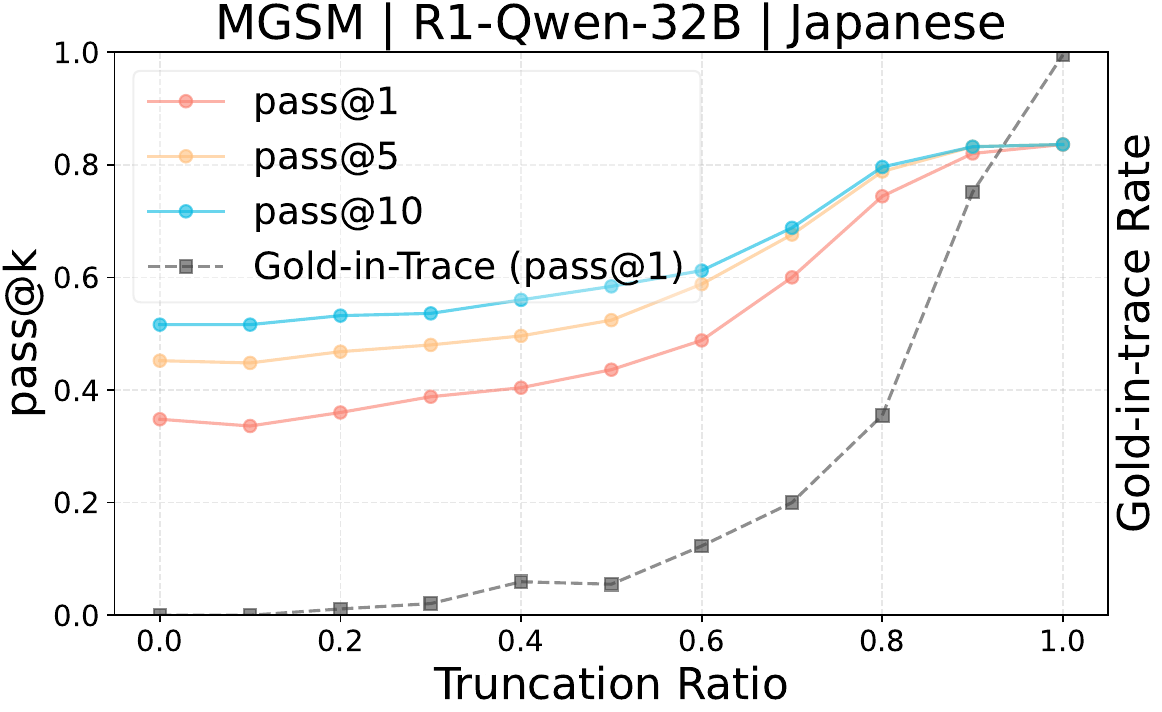}
    \includegraphics[width=0.23\textwidth]{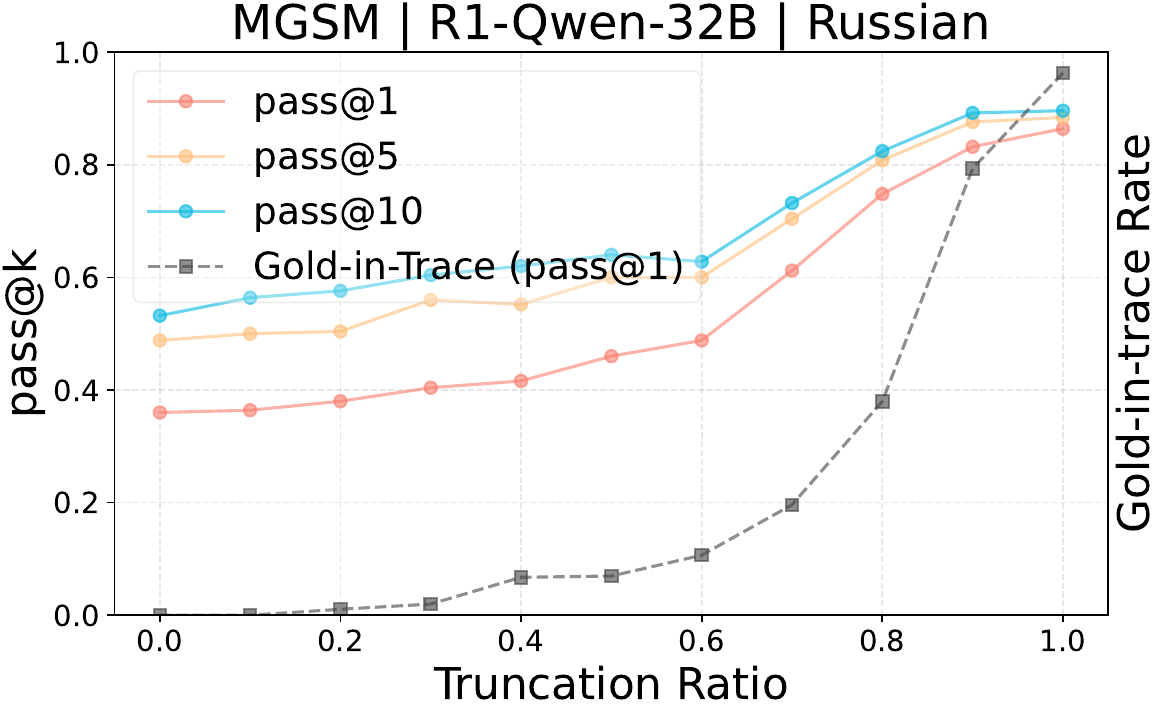}
    \includegraphics[width=0.23\textwidth]{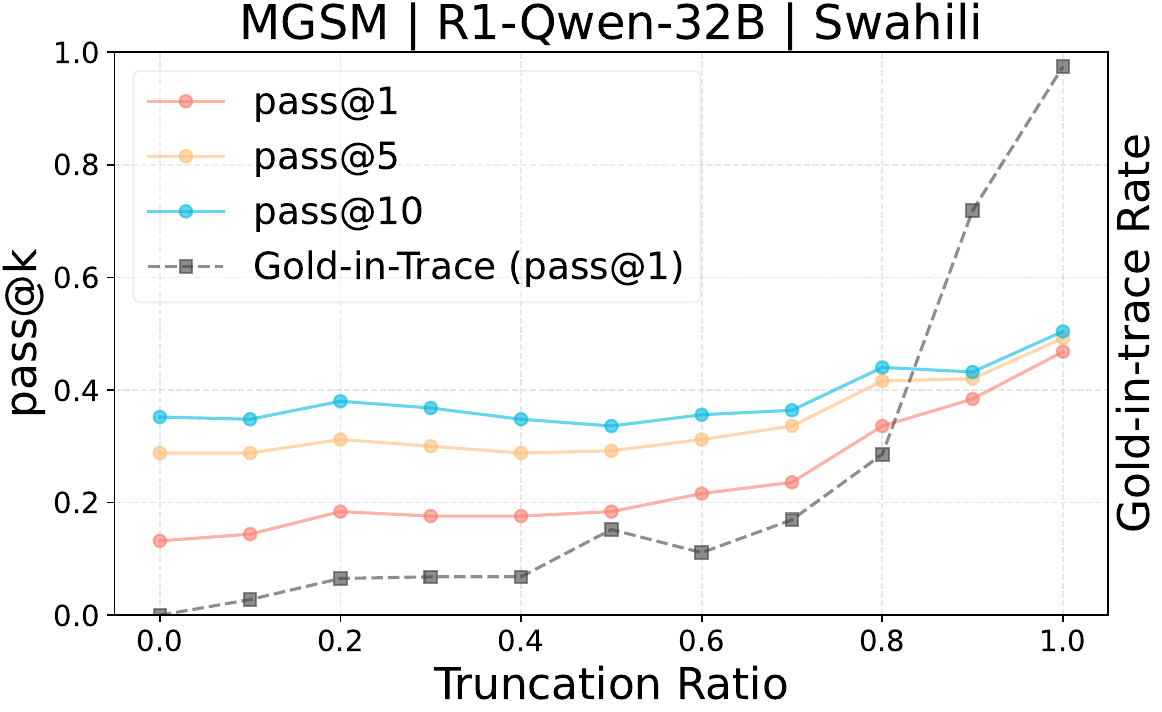}
    \includegraphics[width=0.23\textwidth]{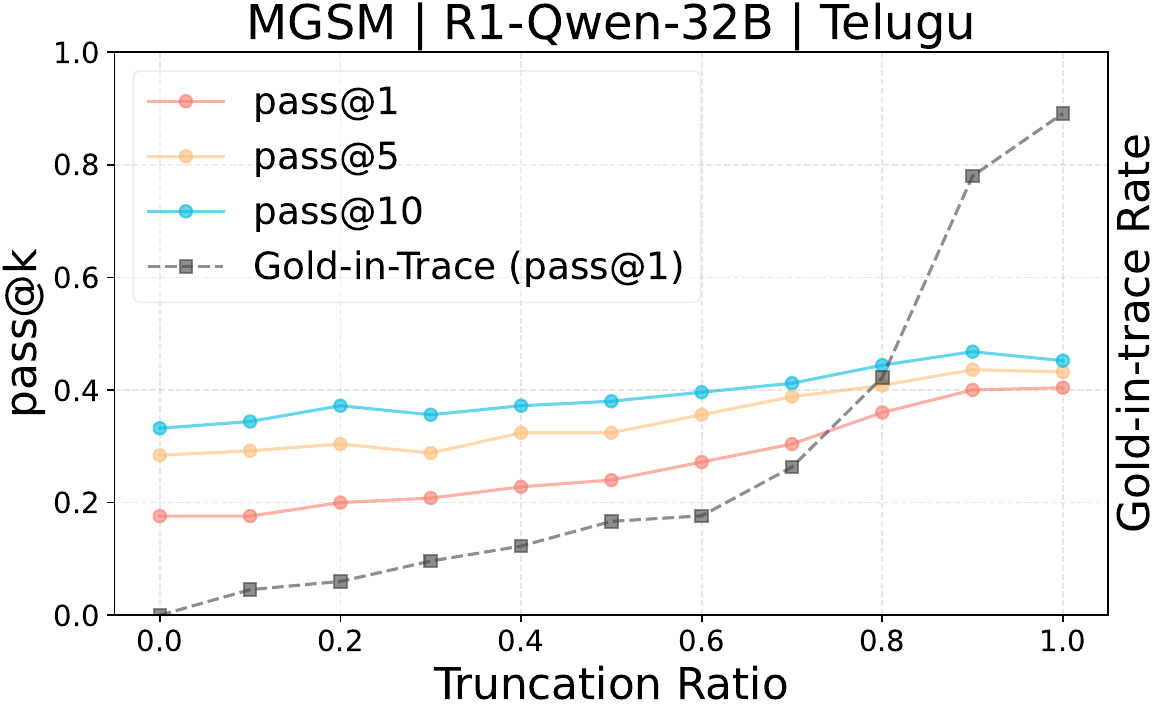}
    \includegraphics[width=0.23\textwidth]{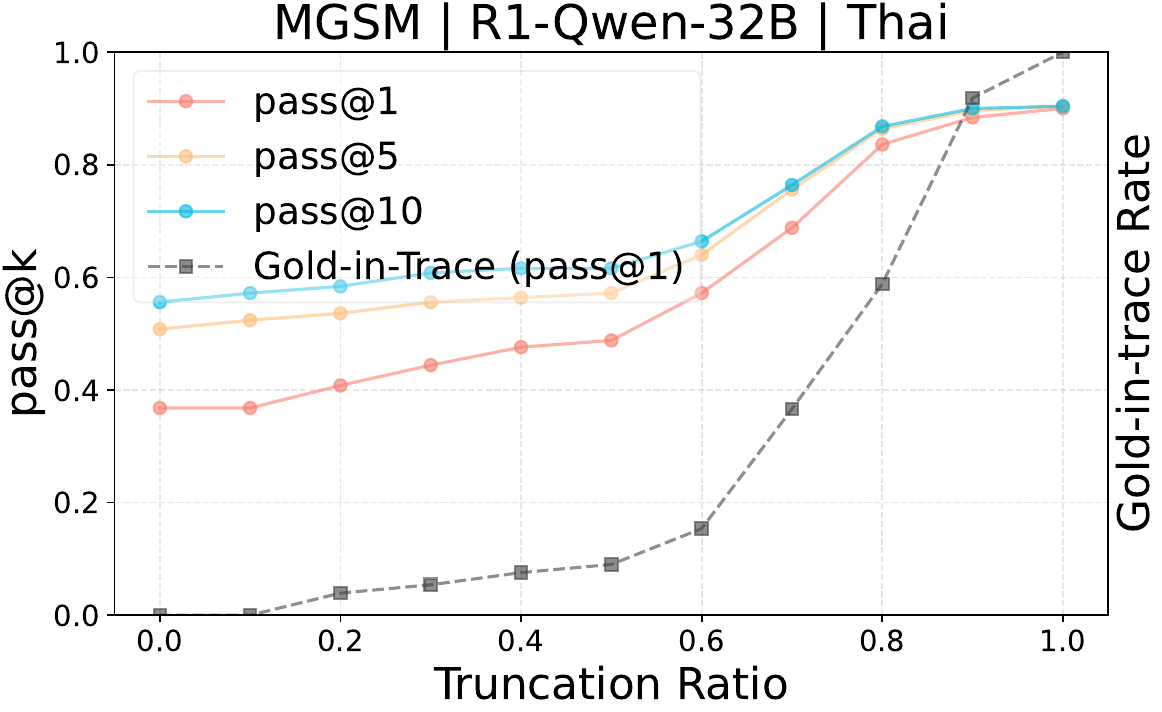}
    \includegraphics[width=0.23\textwidth]{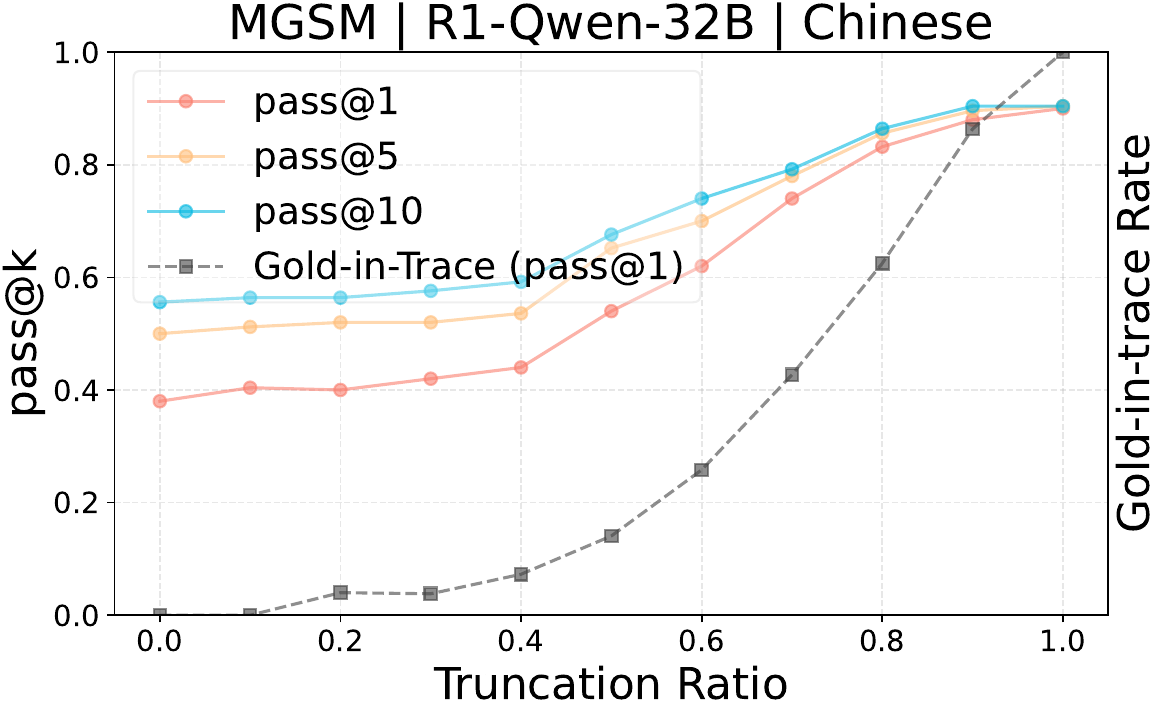}
    \caption{Pass@$k$ accuracy ($k=1,5,10$) and gold-in-trace rate under reasoning-trace truncation for \textbf{R1-Qwen-32B} on \textbf{MGSM}. The model shows stronger latent reasoning in high-resource languages (e.g., English).}
    \label{fig:truncation_32b_mgsm}
\end{figure*}

\begin{figure*}
    \centering
    \includegraphics[width=0.23\textwidth]{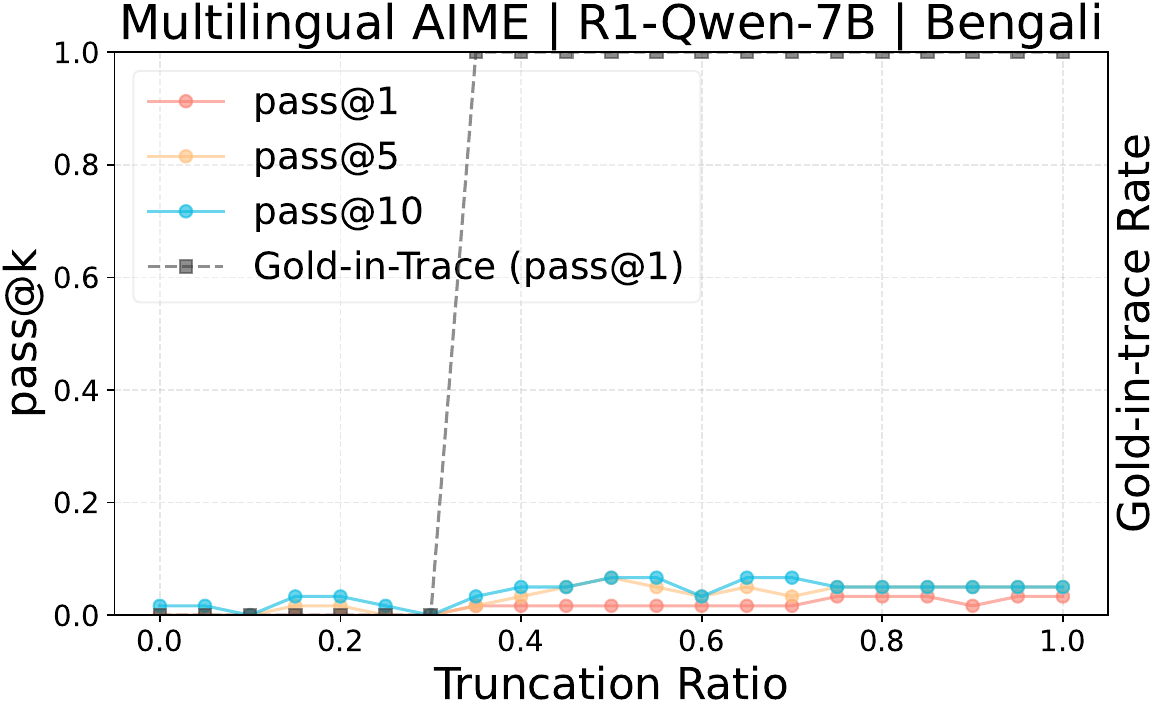}
    \includegraphics[width=0.23\textwidth]{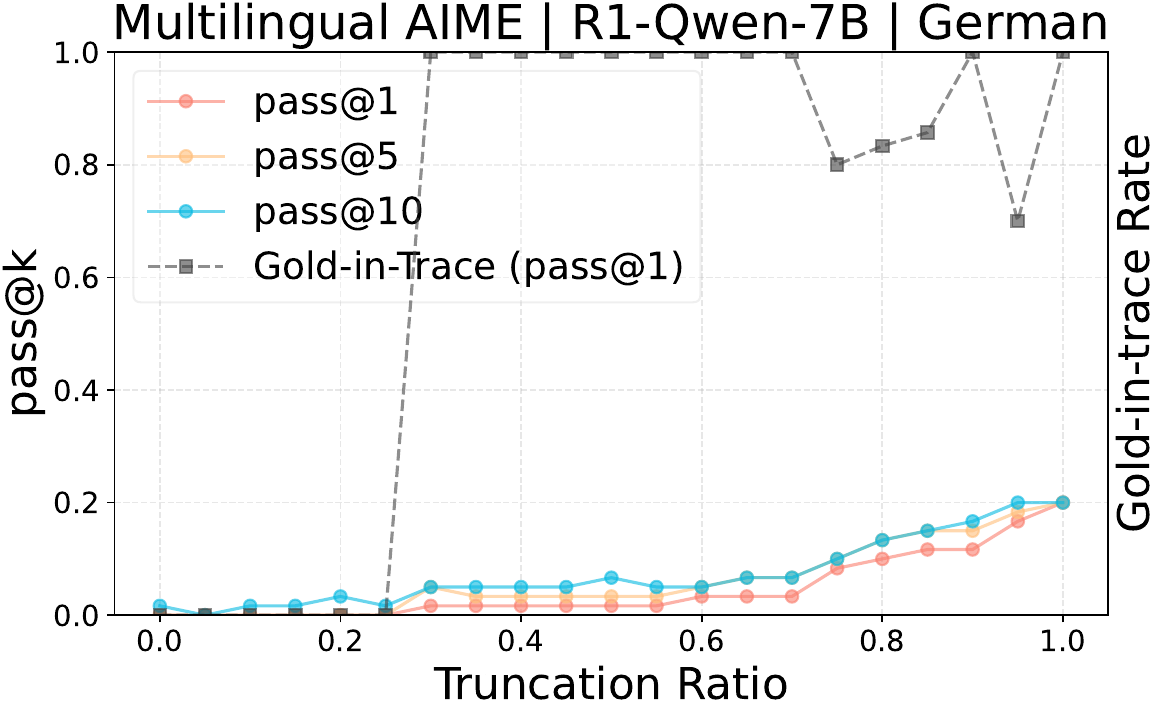}
    \includegraphics[width=0.23\textwidth]{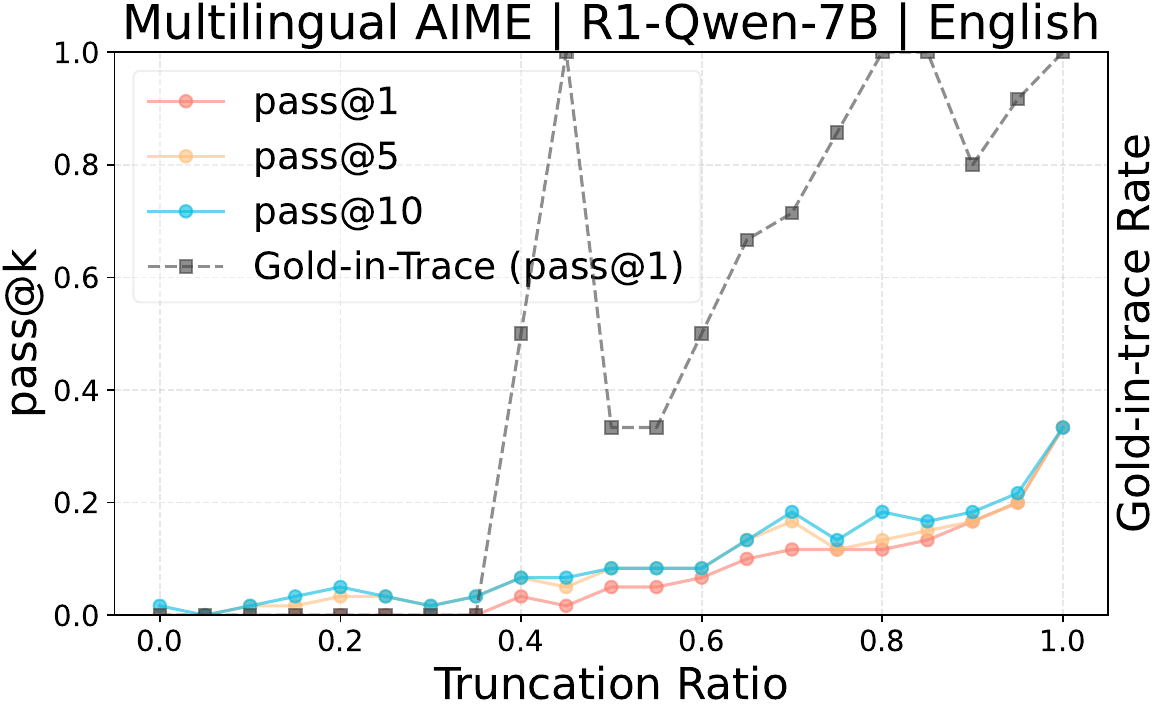}
    \includegraphics[width=0.23\textwidth]{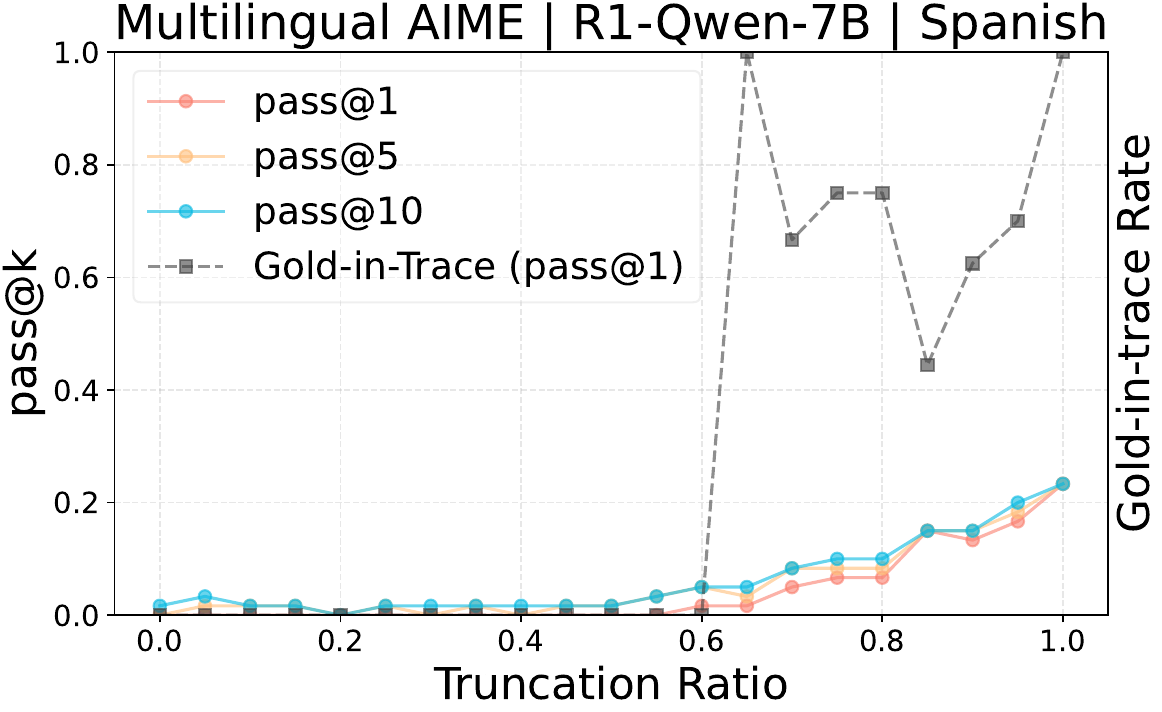}
    \includegraphics[width=0.23\textwidth]{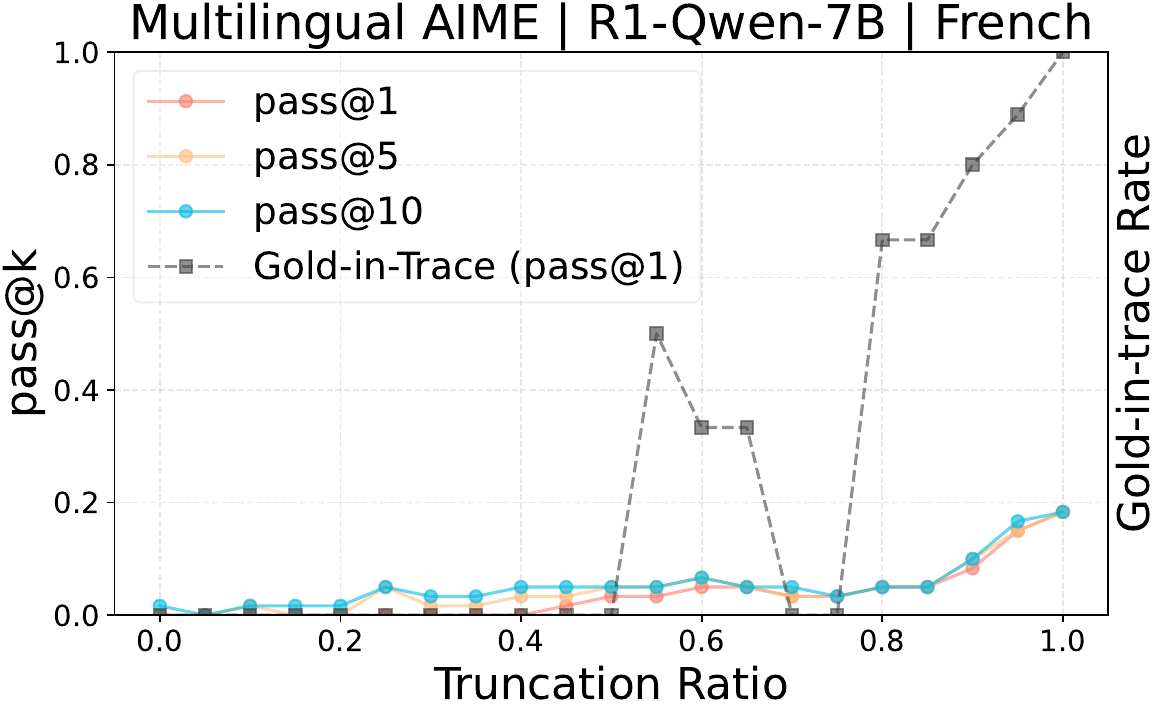}
    \includegraphics[width=0.23\textwidth]{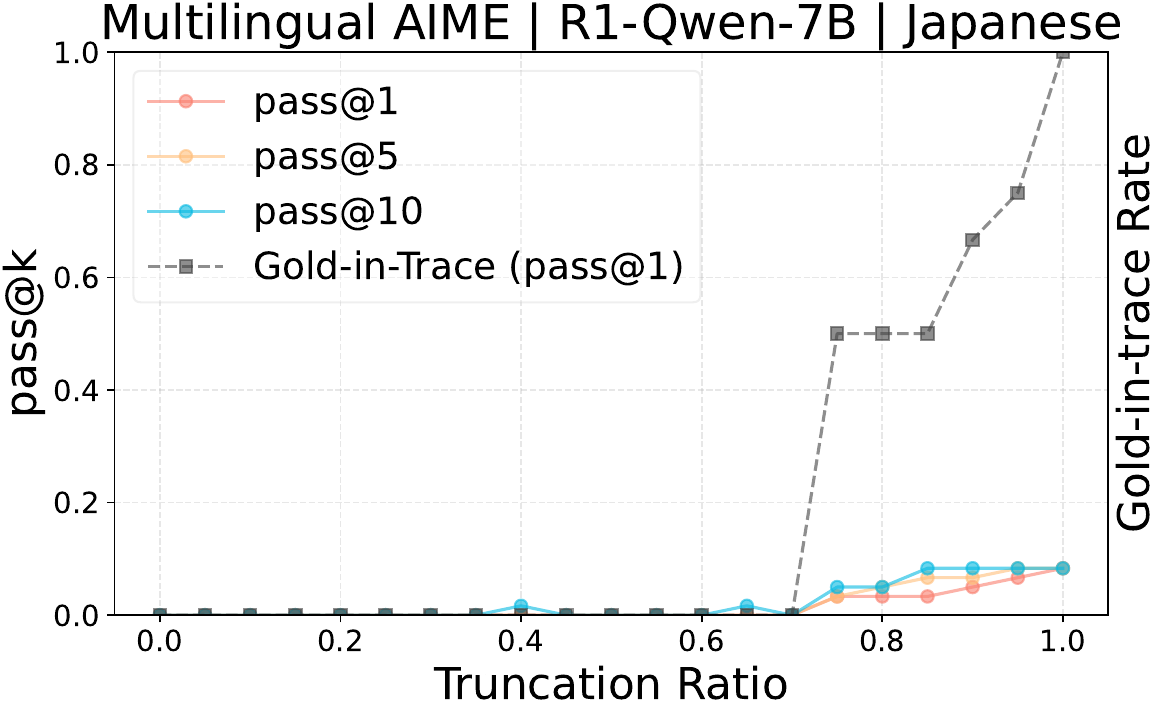}
    \includegraphics[width=0.23\textwidth]{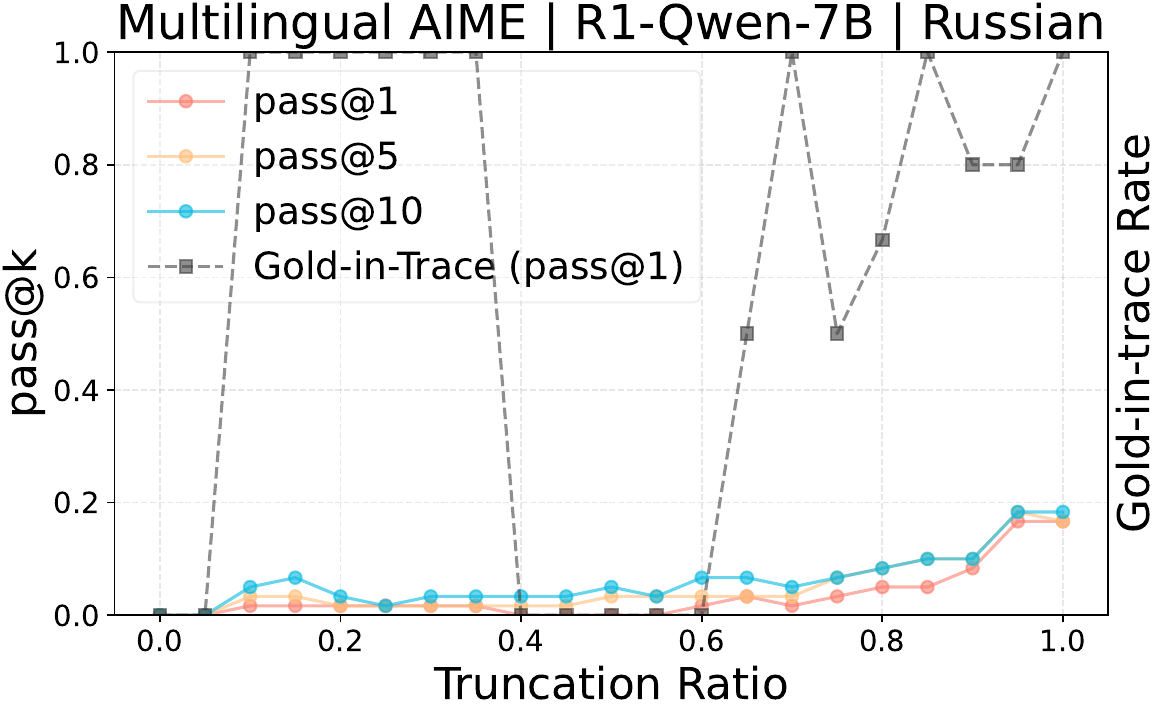}
    \includegraphics[width=0.23\textwidth]{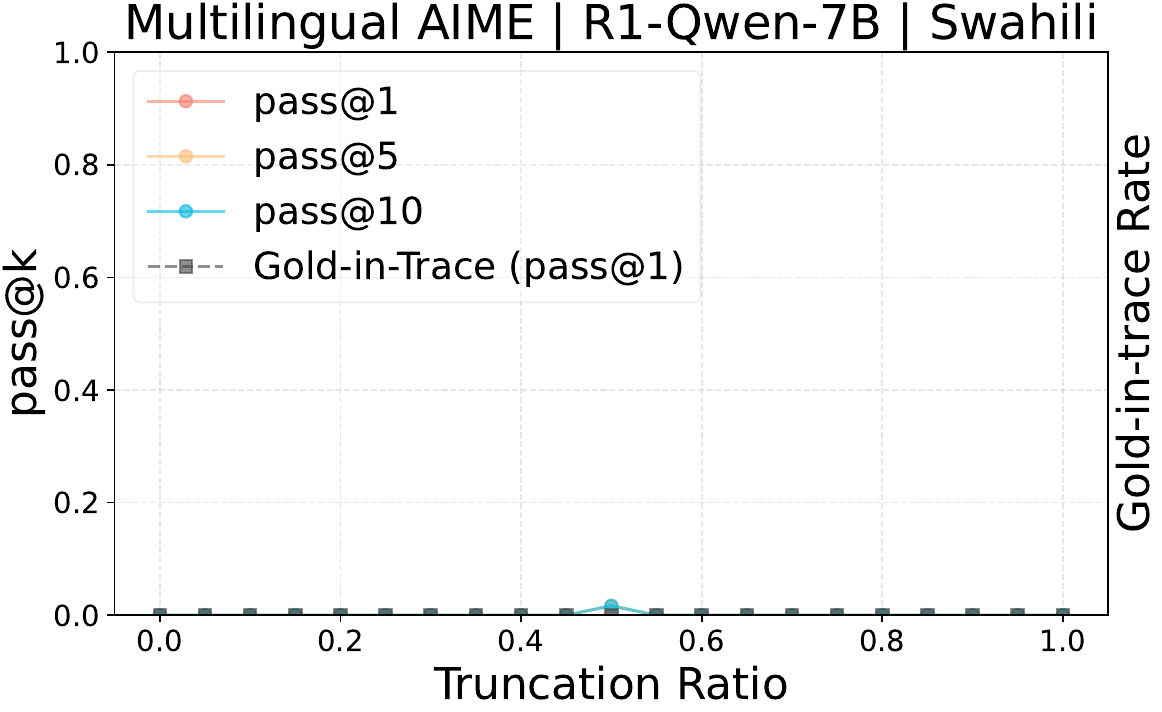}
    \includegraphics[width=0.23\textwidth]{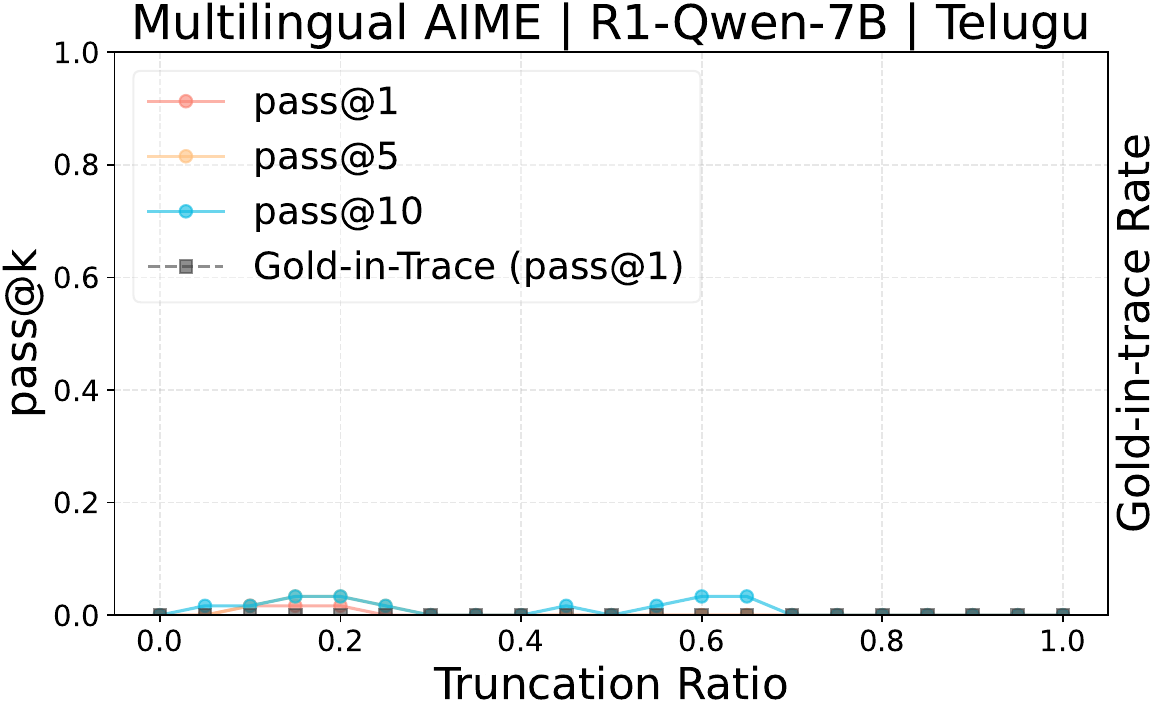}
    \includegraphics[width=0.23\textwidth]{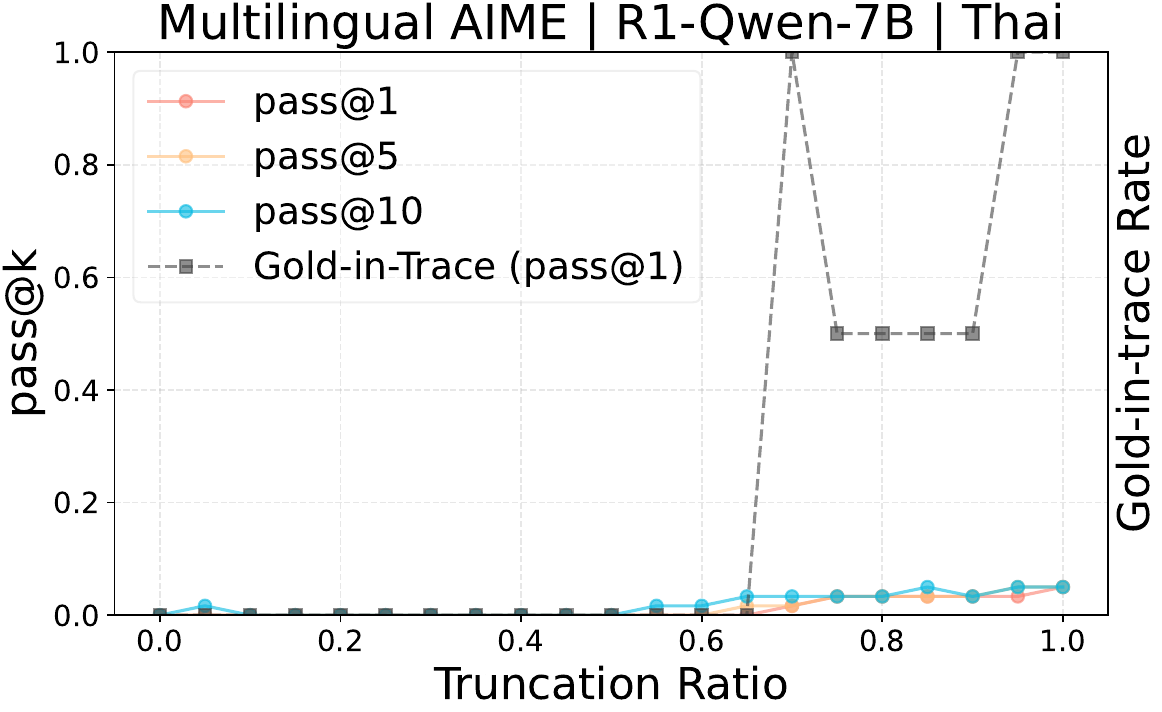}
    \includegraphics[width=0.23\textwidth]{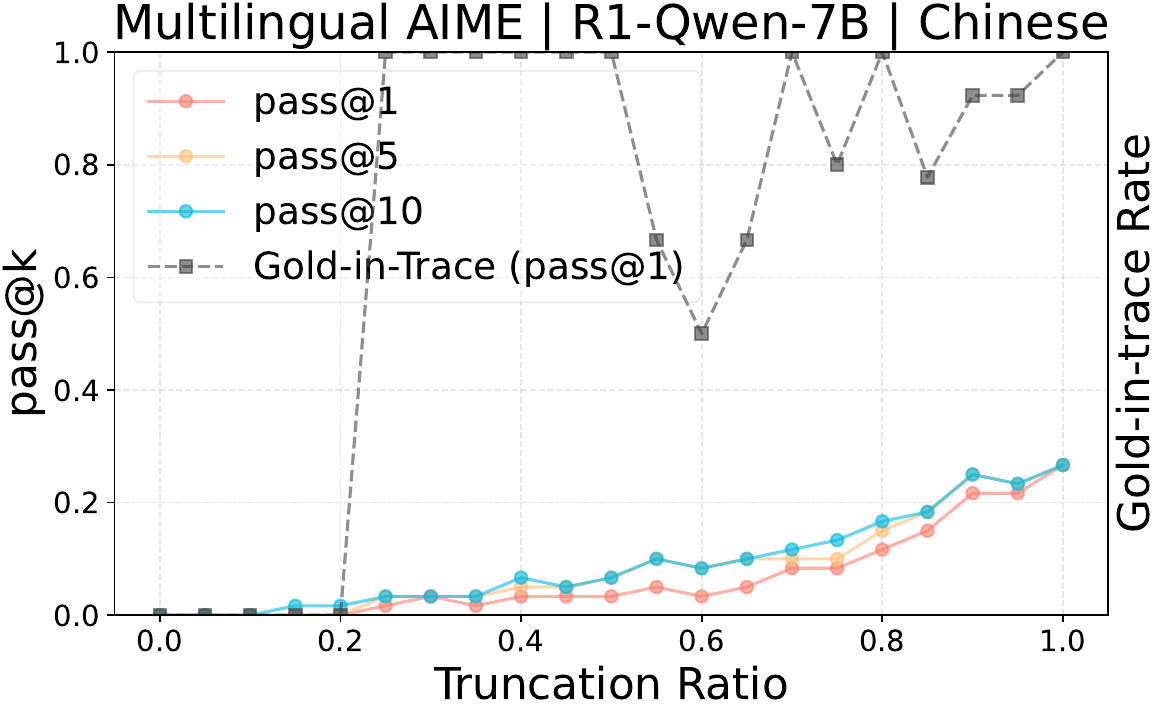}
    \caption{
    Pass@$k$ accuracy ($k=1,5,10$) and gold-in-trace rate under reasoning-trace truncation for \textbf{R1-Qwen-7B} on \textbf{Multilingual AIME}. Latent reasoning is less pronounced compared to MGSM.
    }
    \label{fig:truncation_7b_aime}
\end{figure*}

\begin{figure*}
    \centering
    \includegraphics[width=0.23\textwidth]{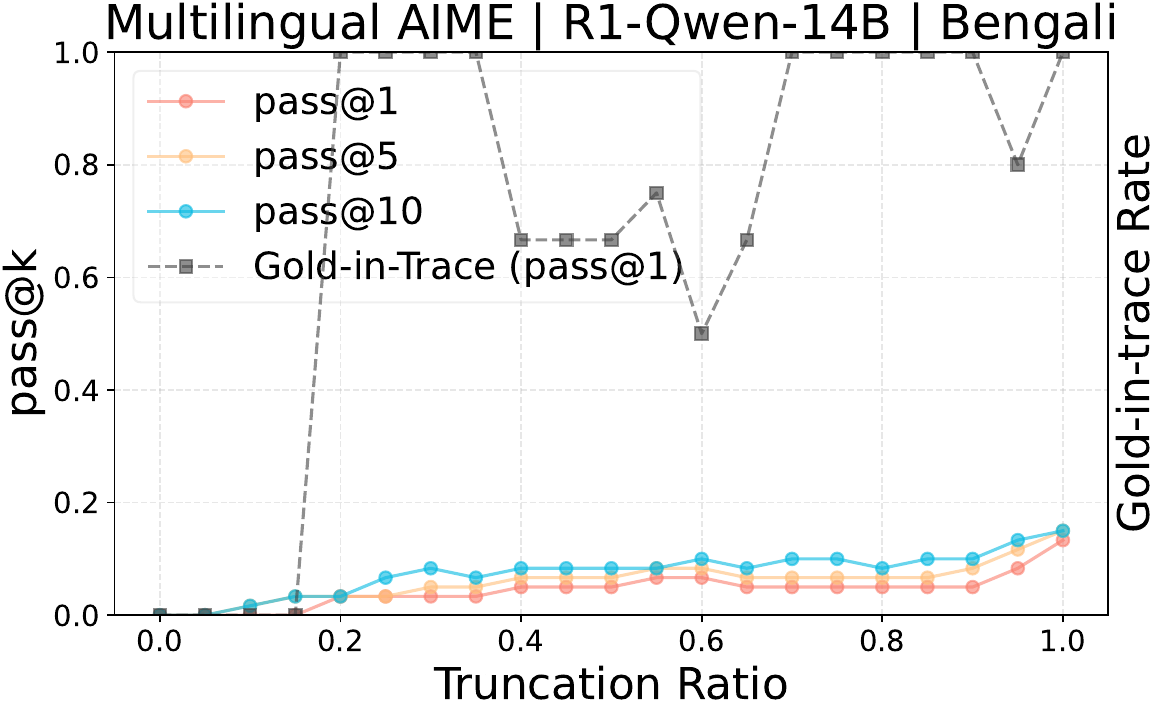}
    \includegraphics[width=0.23\textwidth]{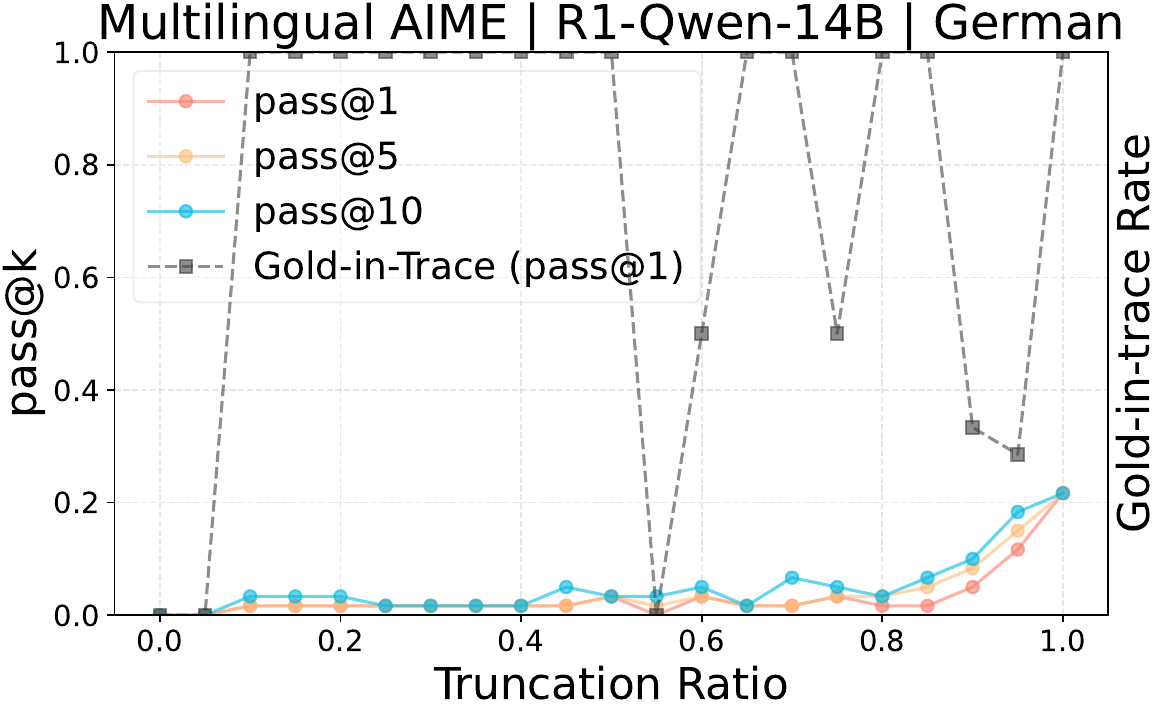}
    \includegraphics[width=0.23\textwidth]{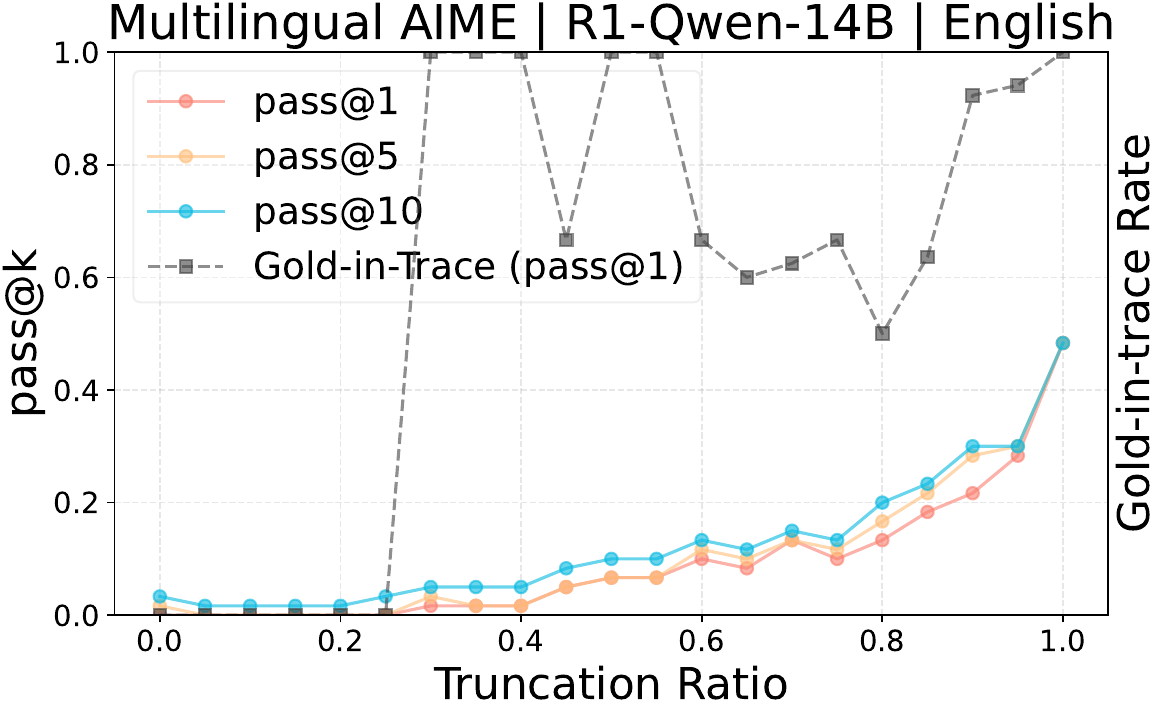}
    \includegraphics[width=0.23\textwidth]{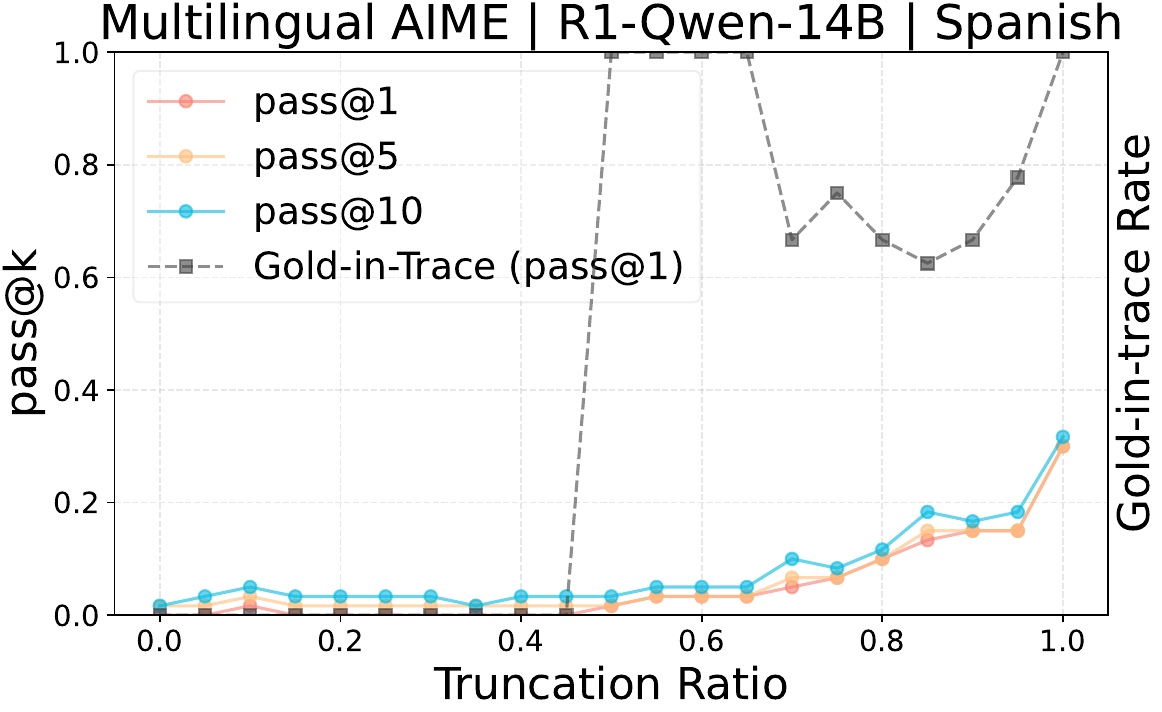}
    \includegraphics[width=0.23\textwidth]{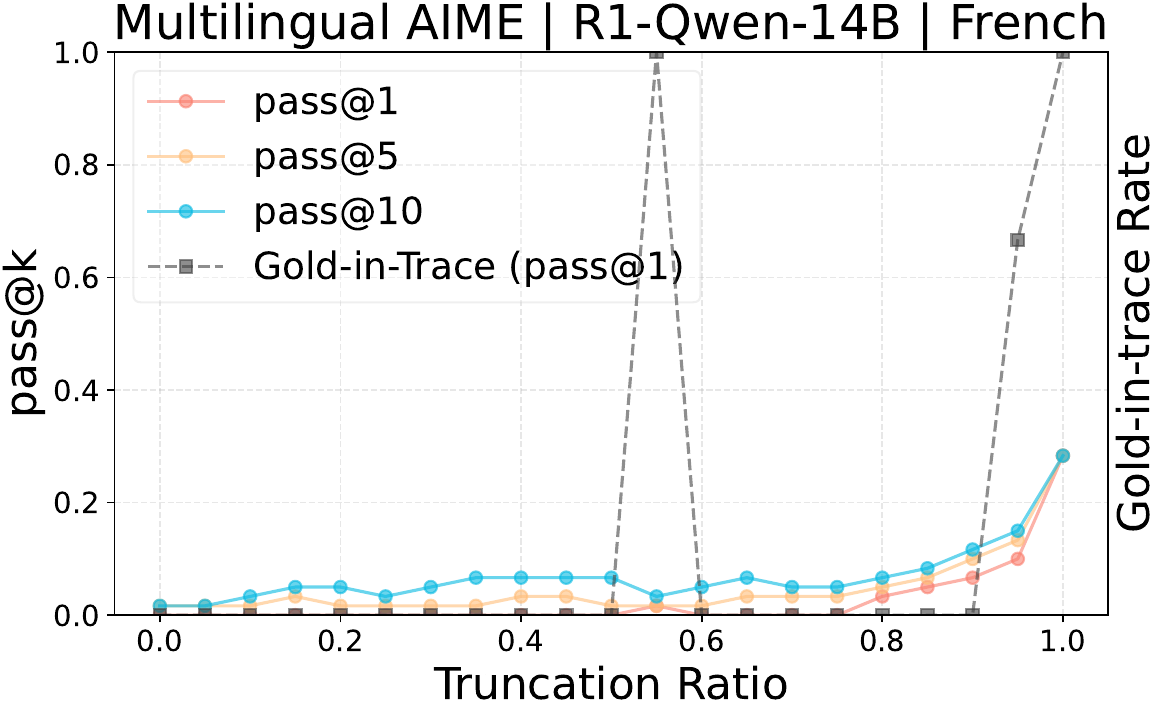}
    \includegraphics[width=0.23\textwidth]{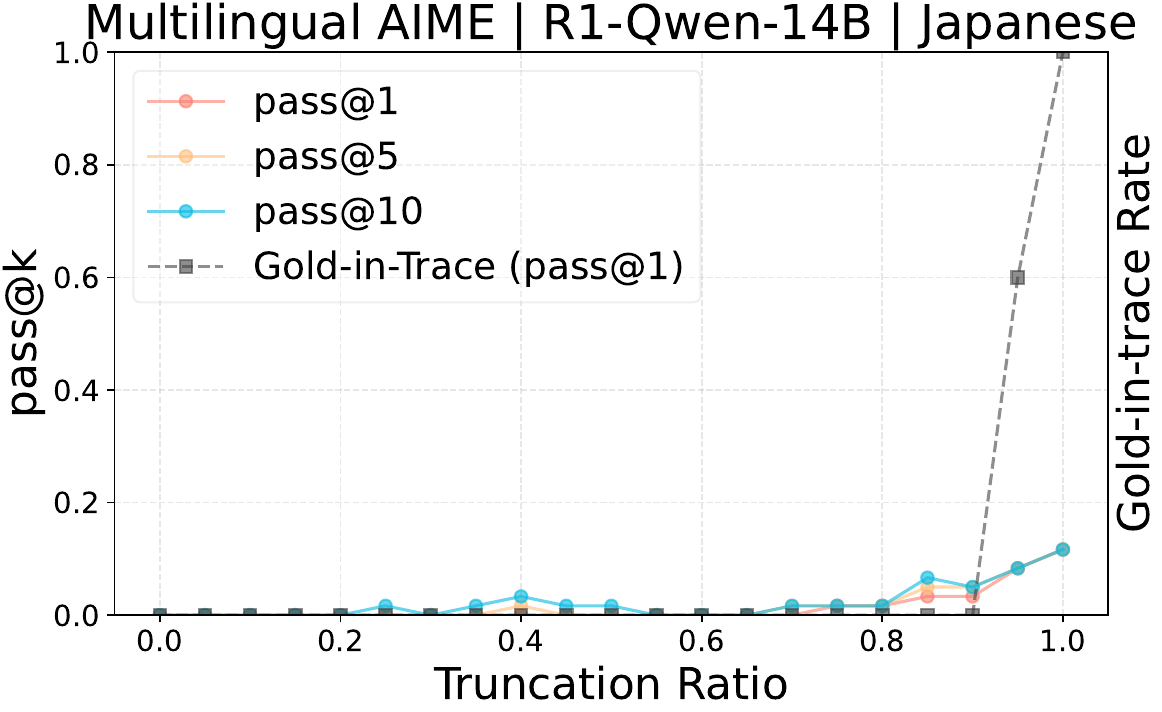}
    \includegraphics[width=0.23\textwidth]{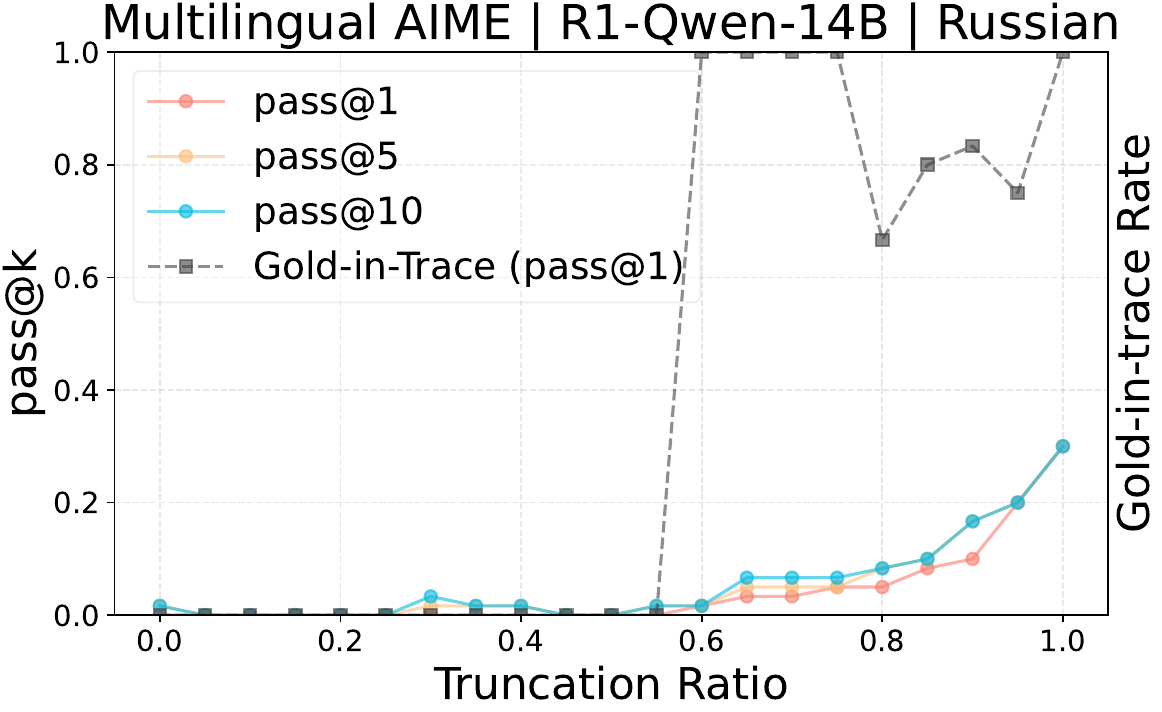}
    \includegraphics[width=0.23\textwidth]{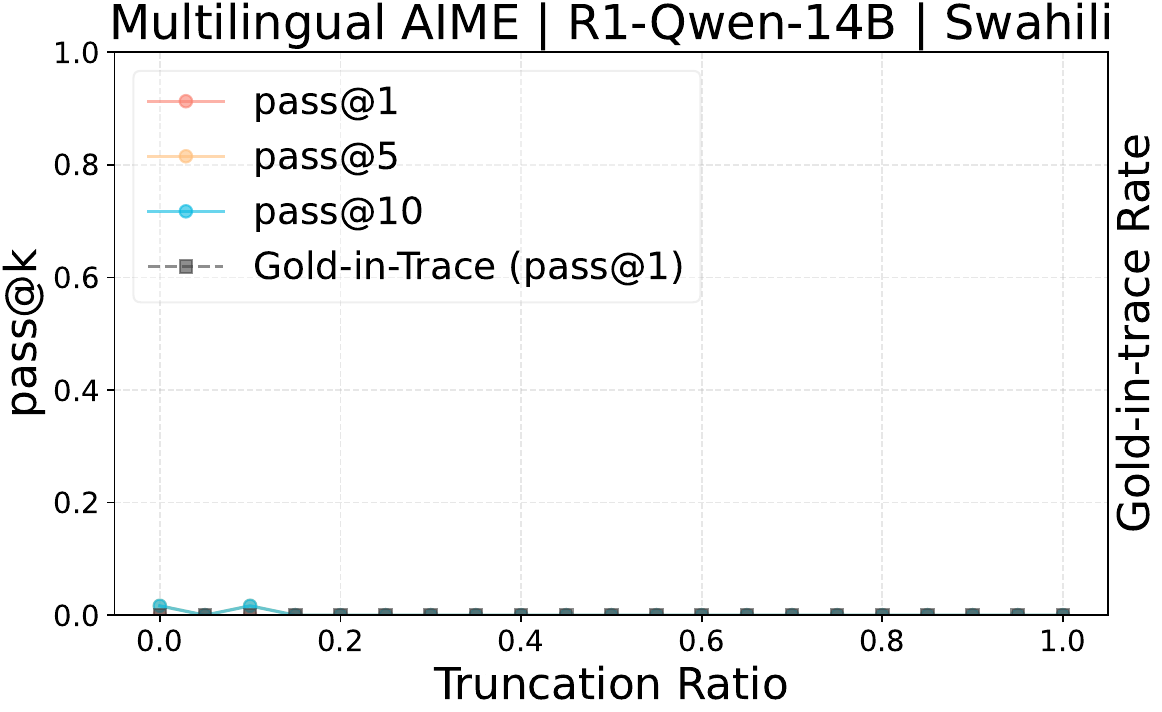}
    \includegraphics[width=0.23\textwidth]{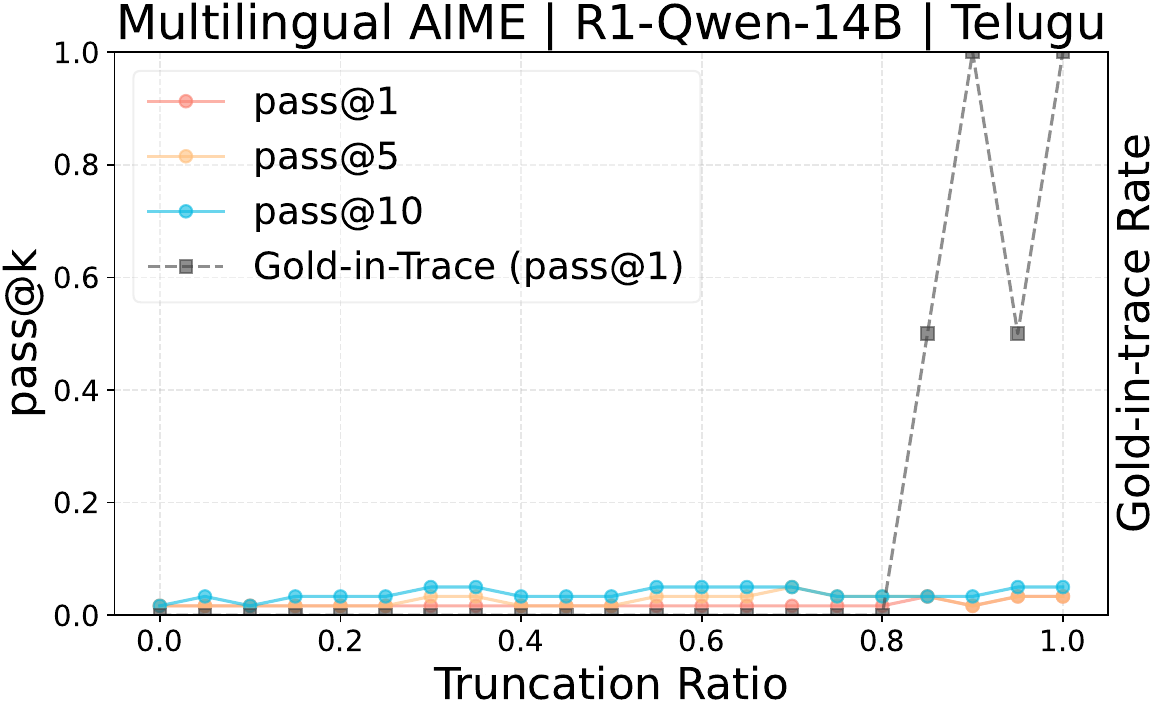}
    \includegraphics[width=0.23\textwidth]{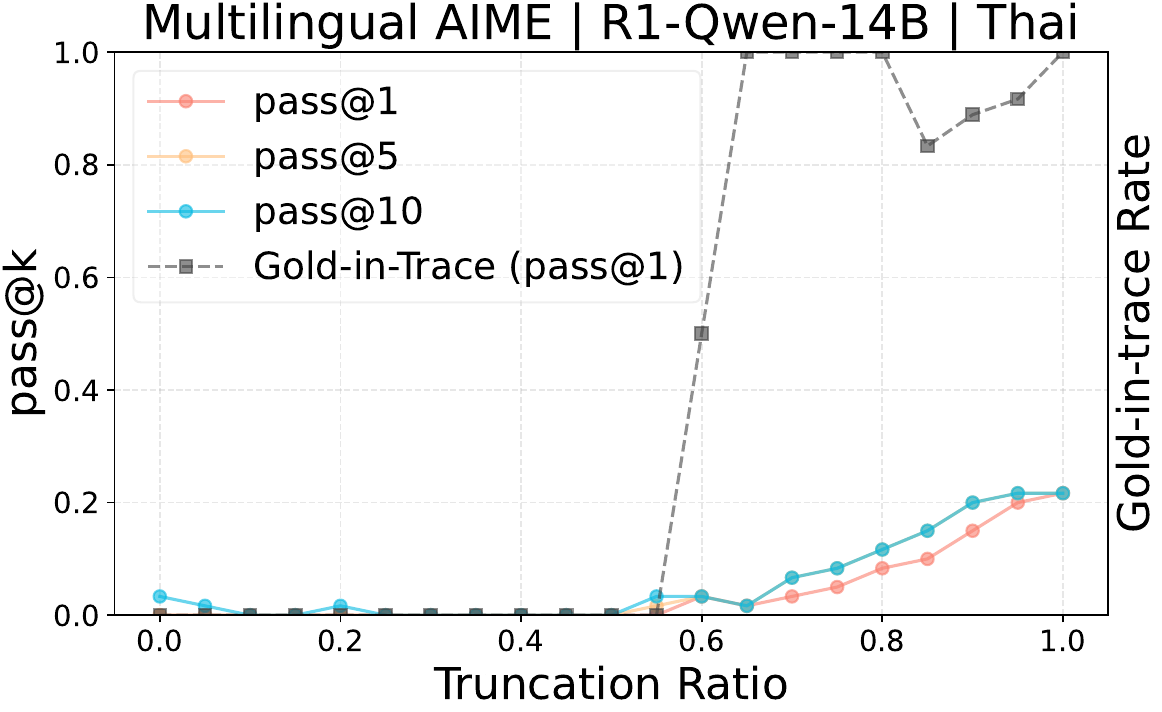}
    \includegraphics[width=0.23\textwidth]{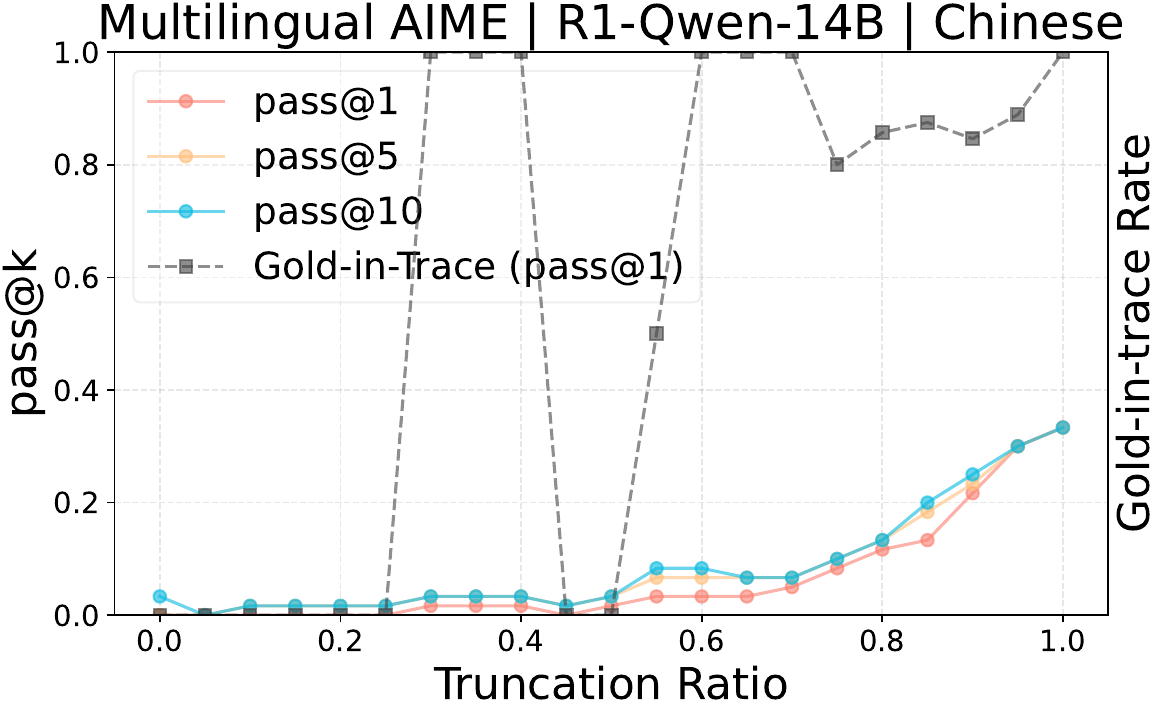}
    \caption{
    Pass@$k$ accuracy ($k=1,5,10$) and gold-in-trace rate under reasoning-trace truncation for \textbf{R1-Qwen-14B} on \textbf{Multilingual AIME}. Latent reasoning is less pronounced compared to MGSM.
    }
    \label{fig:truncation_14b_aime}
\end{figure*}

\begin{figure*}
    \centering
    \includegraphics[width=0.23\textwidth]{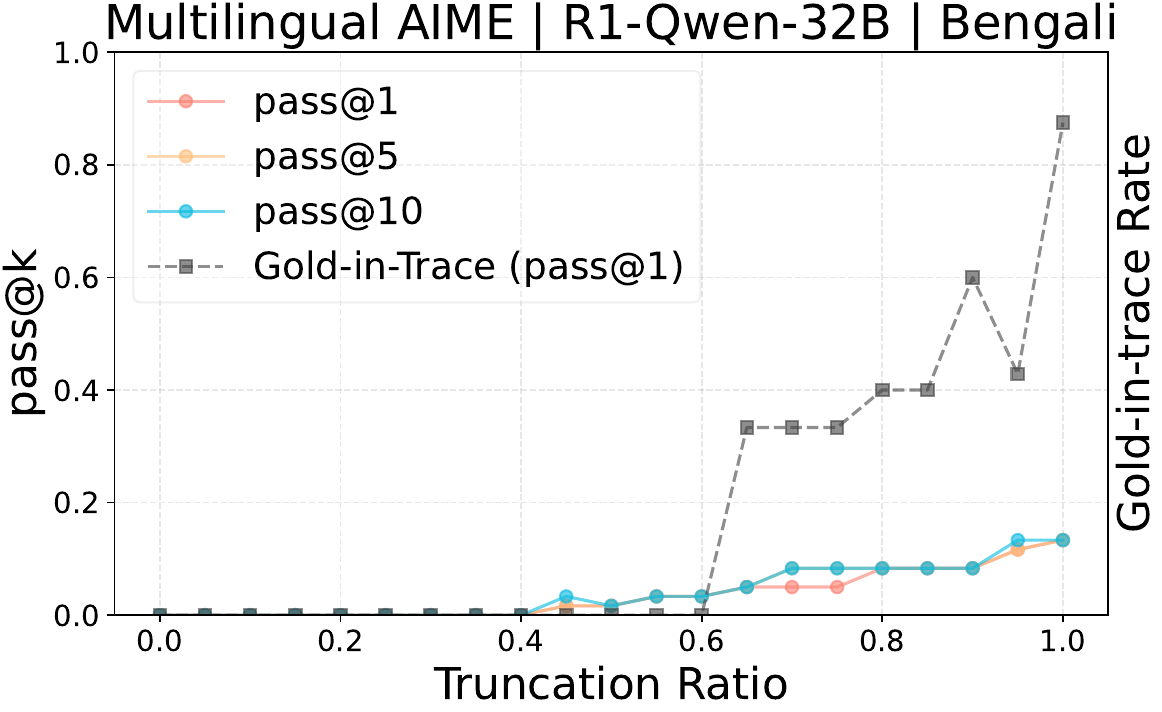}
    \includegraphics[width=0.23\textwidth]{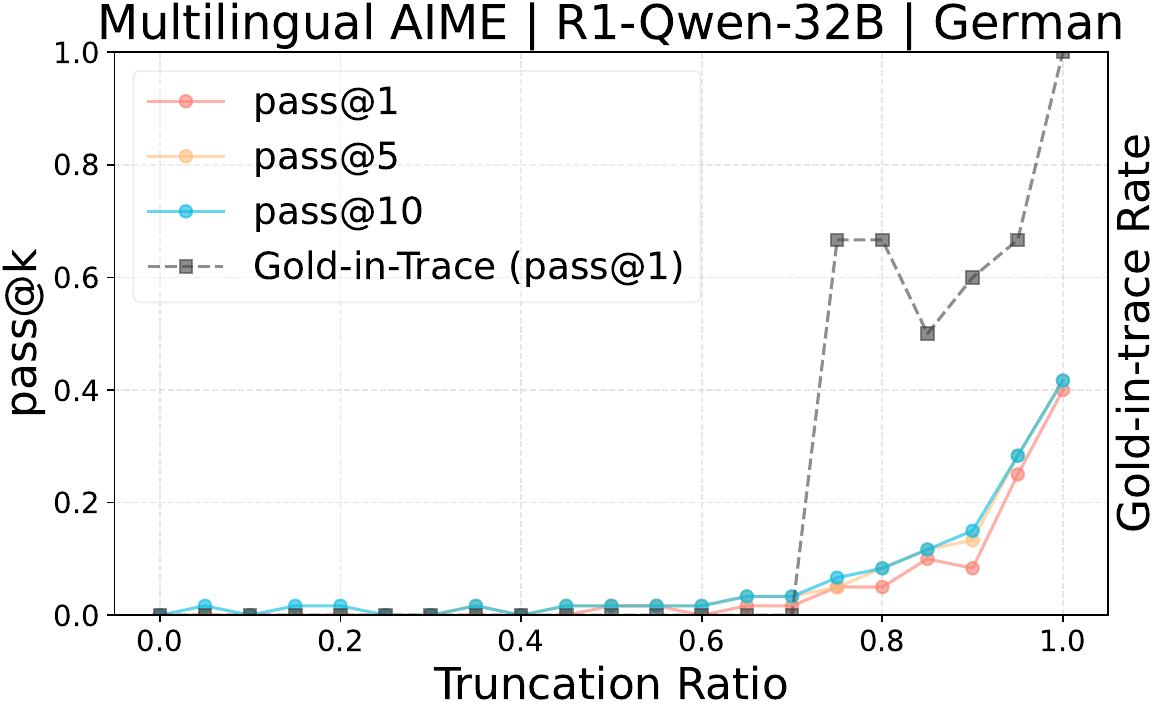}
    \includegraphics[width=0.23\textwidth]{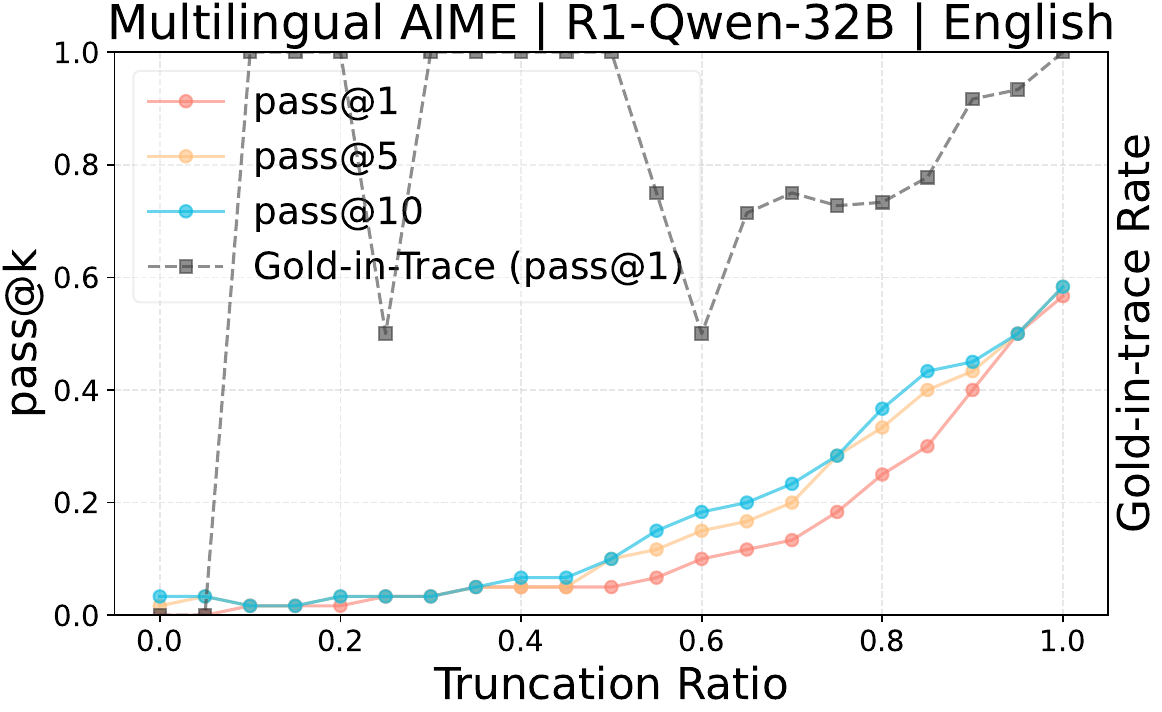}
    \includegraphics[width=0.23\textwidth]{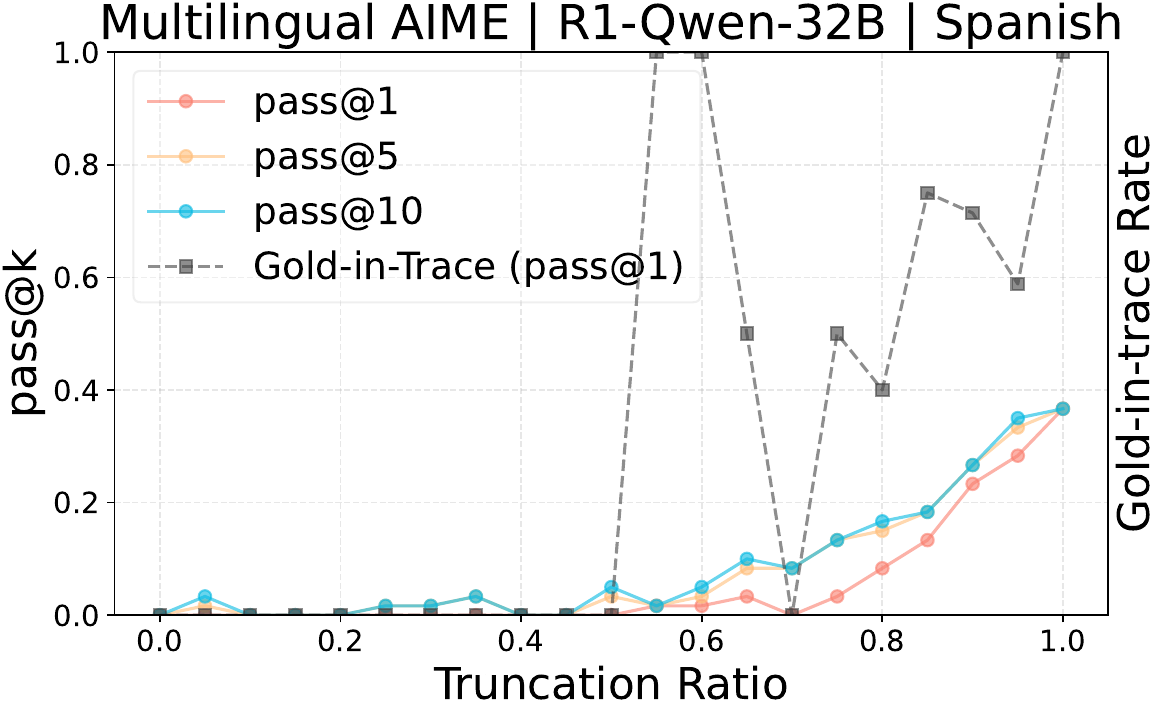}
    \includegraphics[width=0.23\textwidth]{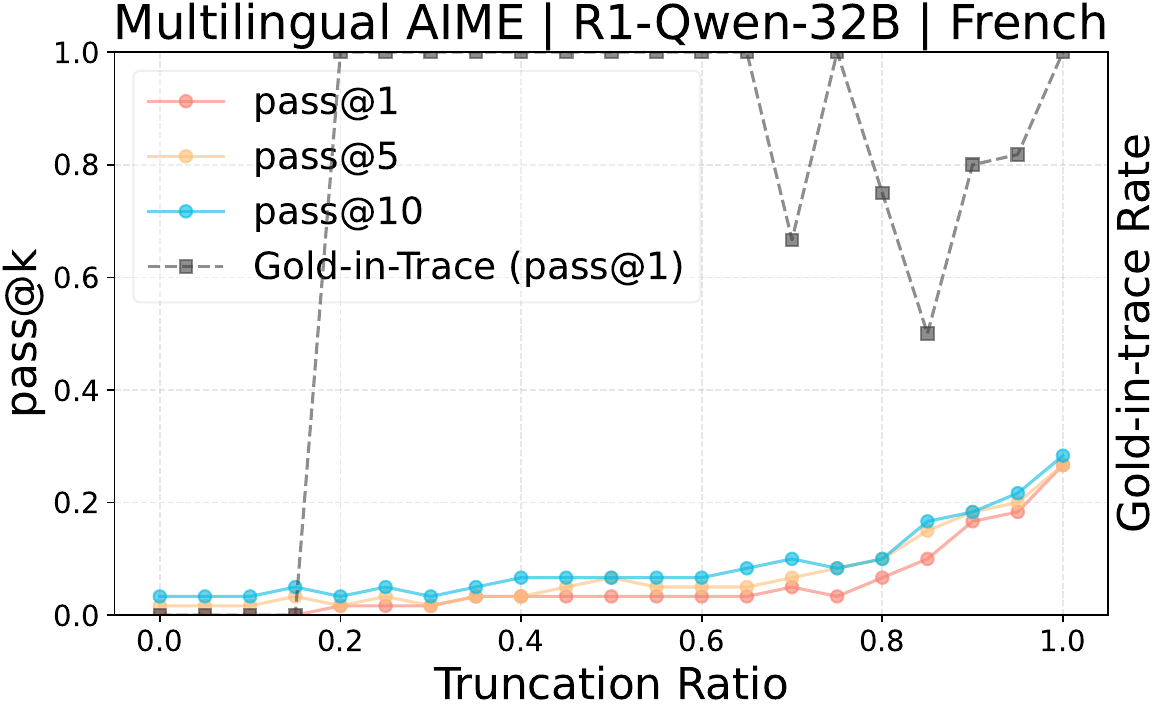}
    \includegraphics[width=0.23\textwidth]{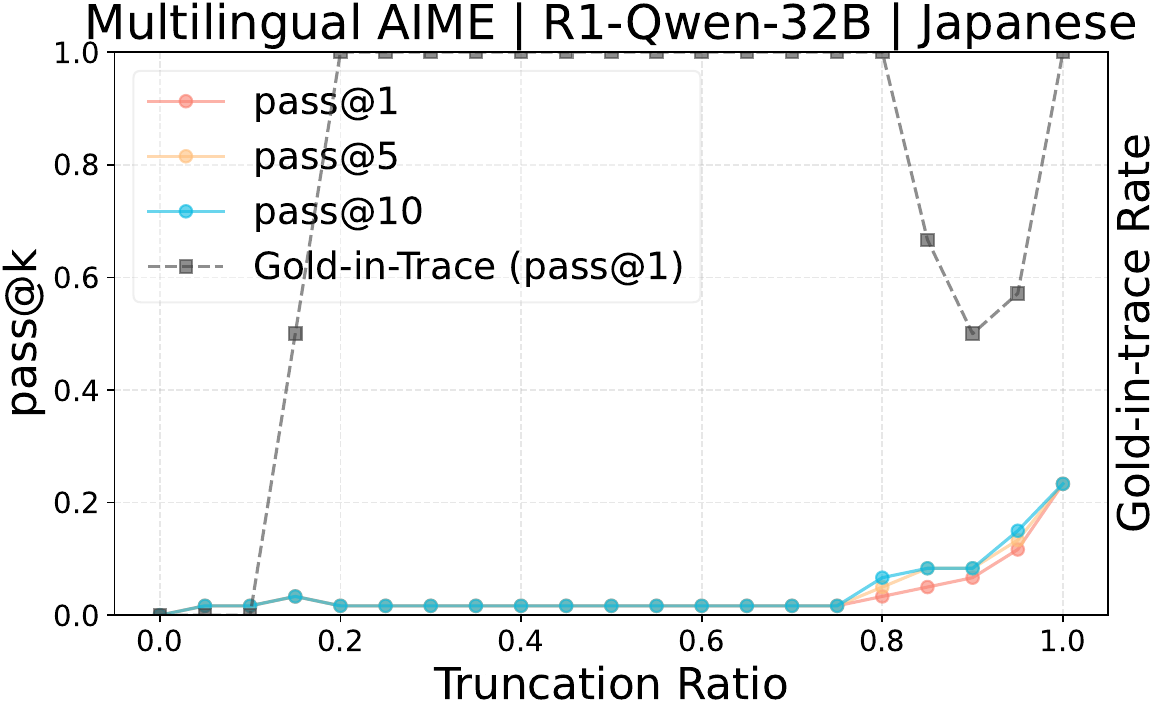}
    \includegraphics[width=0.23\textwidth]{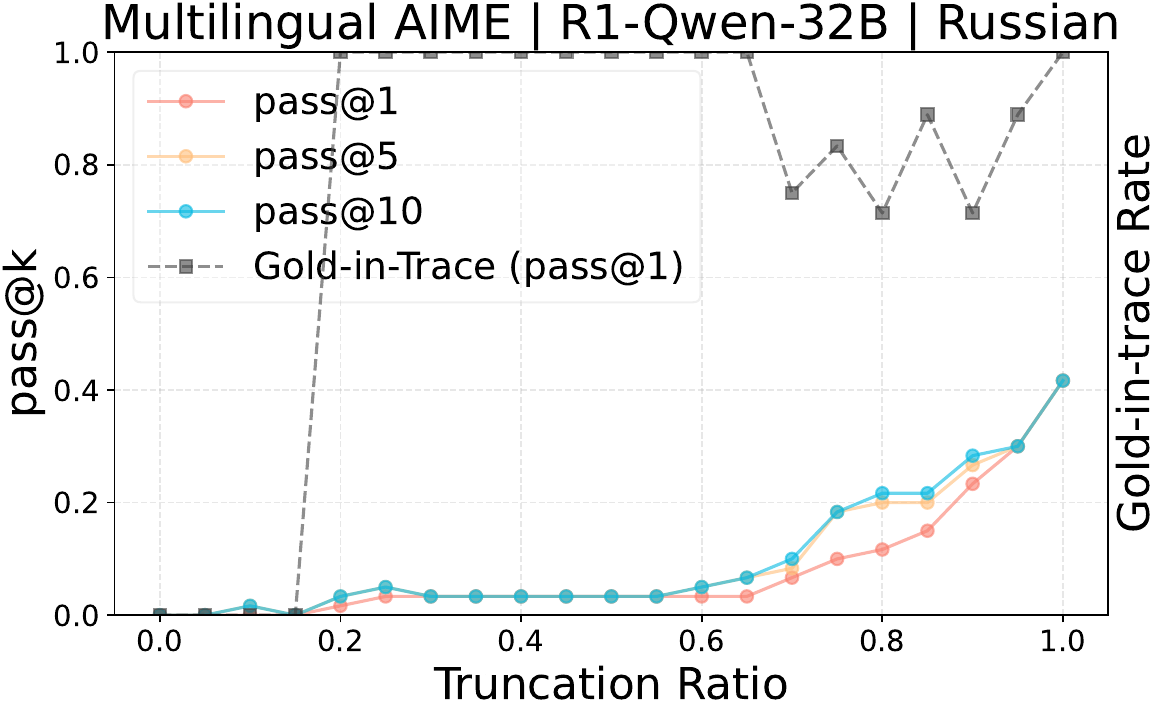}
    \includegraphics[width=0.23\textwidth]{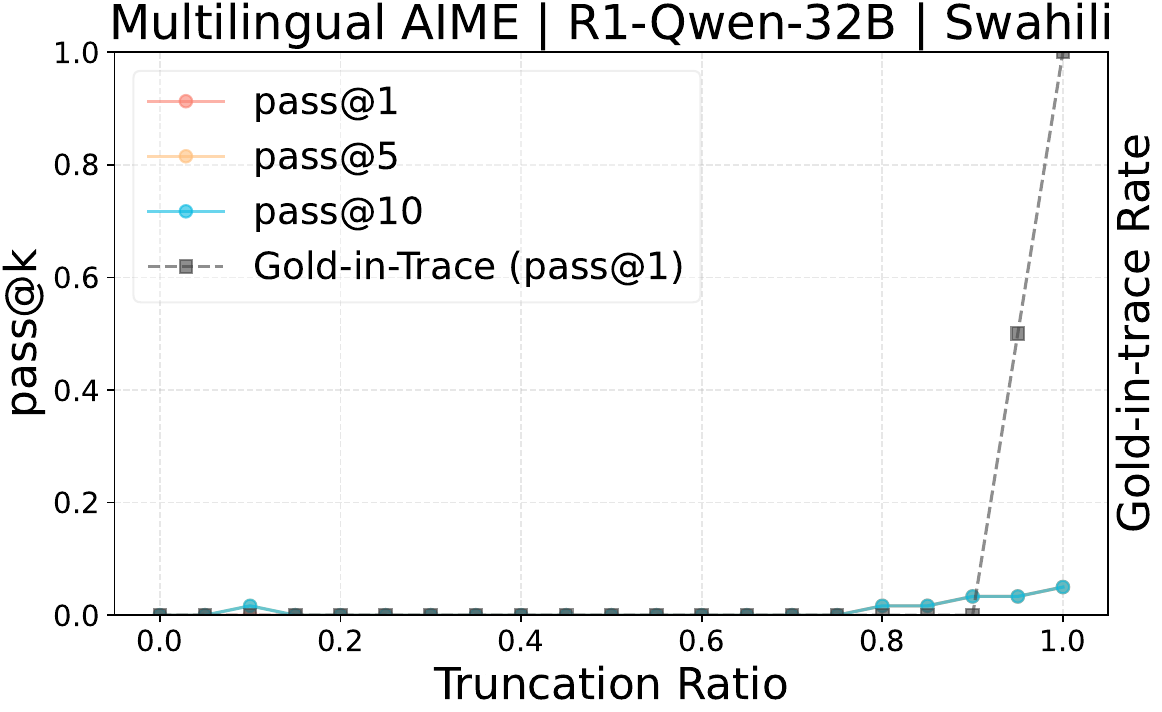}
    \includegraphics[width=0.23\textwidth]{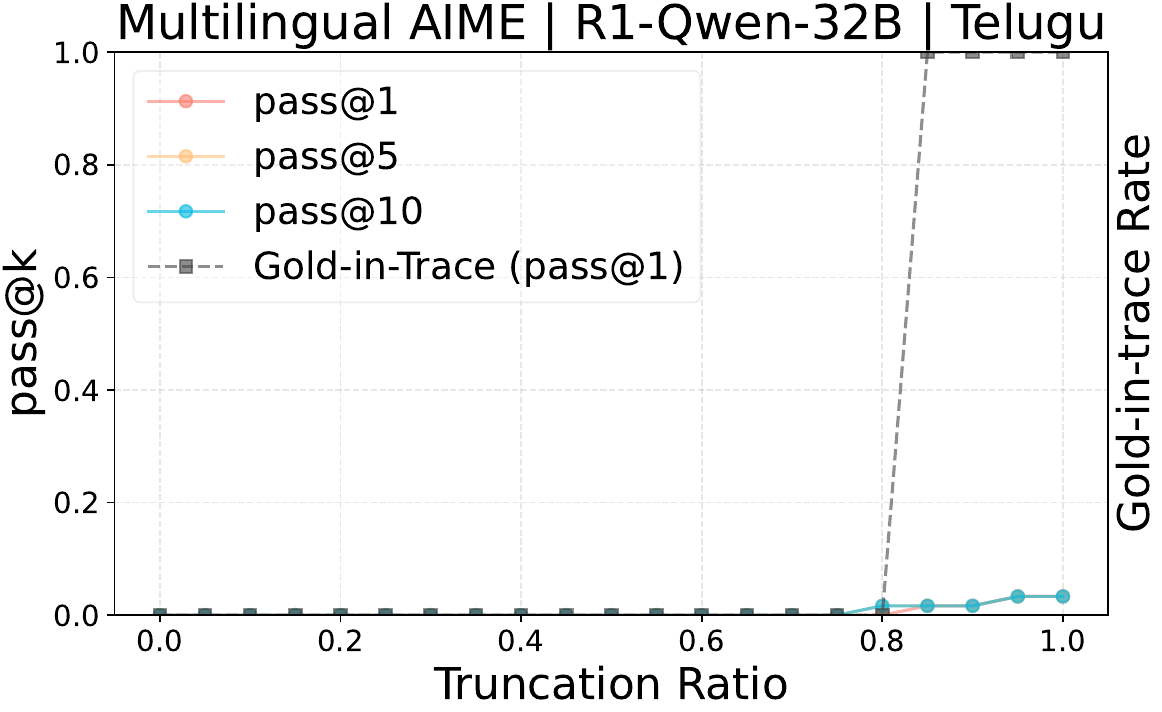}
    \includegraphics[width=0.23\textwidth]{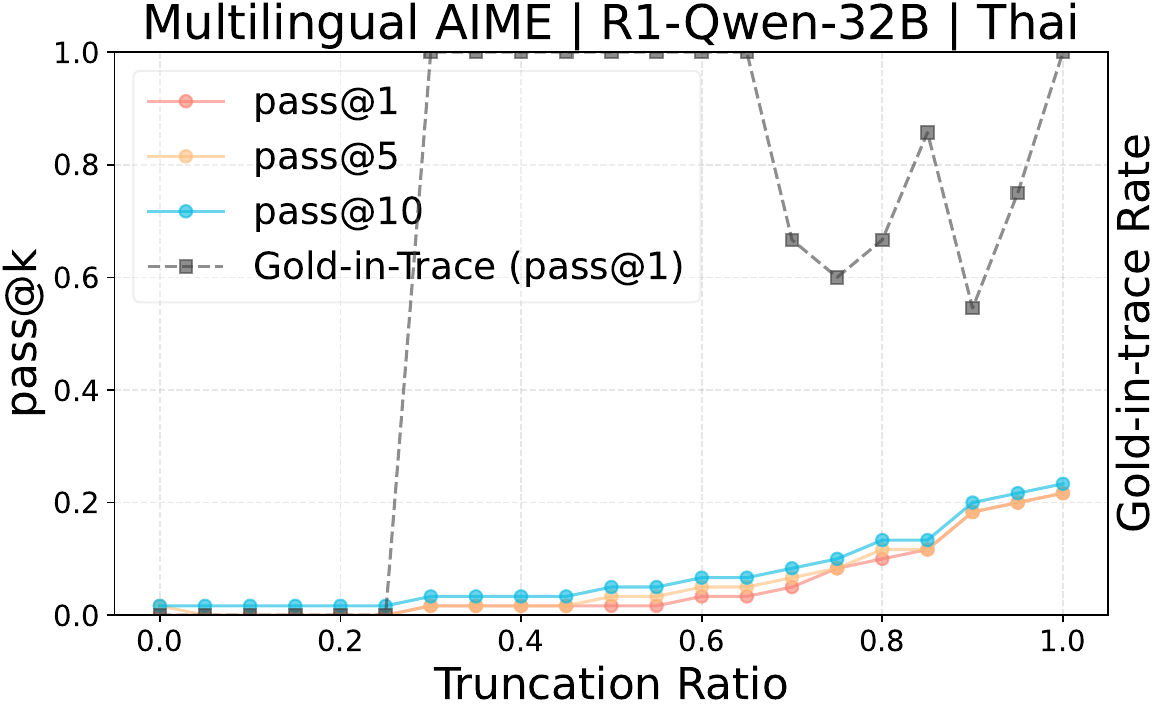}
    \includegraphics[width=0.23\textwidth]{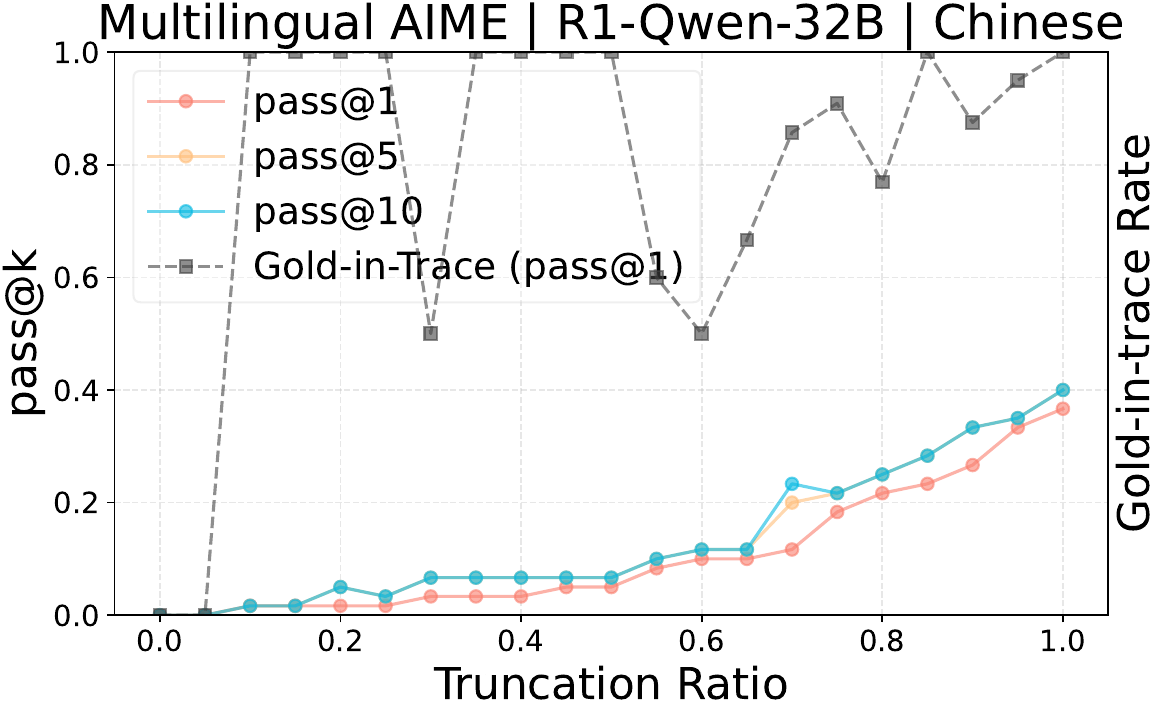}
    \caption{
    Pass@$k$ accuracy ($k=1,5,10$) and gold-in-trace rate under reasoning-trace truncation for \textbf{R1-Qwen-32B} on \textbf{Multilingual AIME}. Latent reasoning is less pronounced compared to MGSM.
    }
    \label{fig:truncation_32b_aime}
\end{figure*}

Figures~\ref{fig:truncation_7b_mgsm}, \ref{fig:truncation_14b_mgsm}, \ref{fig:truncation_32b_mgsm} and their Multilingual AIME counterparts (Figure~\ref{fig:truncation_7b_aime}, \ref{fig:truncation_14b_aime}, \ref{fig:truncation_32b_aime}) visualize pass@$k$ accuracy ($k=1,5,10$) and the gold-in-trace rate as a function of reasoning-trace truncation ratio across languages and model sizes.
Across models, MGSM exhibits a characteristic pattern in which accuracy increases well before the gold answer is explicitly articulated, whereas on Multilingual AIME accuracy typically remains low until late truncation ratios.
The figures also highlight substantial crosslingual variation, with high-resource languages generally achieving earlier and more stable gains than mid- and low-resource languages.

\subsection{Causal Decompositions}

\begin{figure*}
    \centering
    \includegraphics[width=0.24\textwidth]{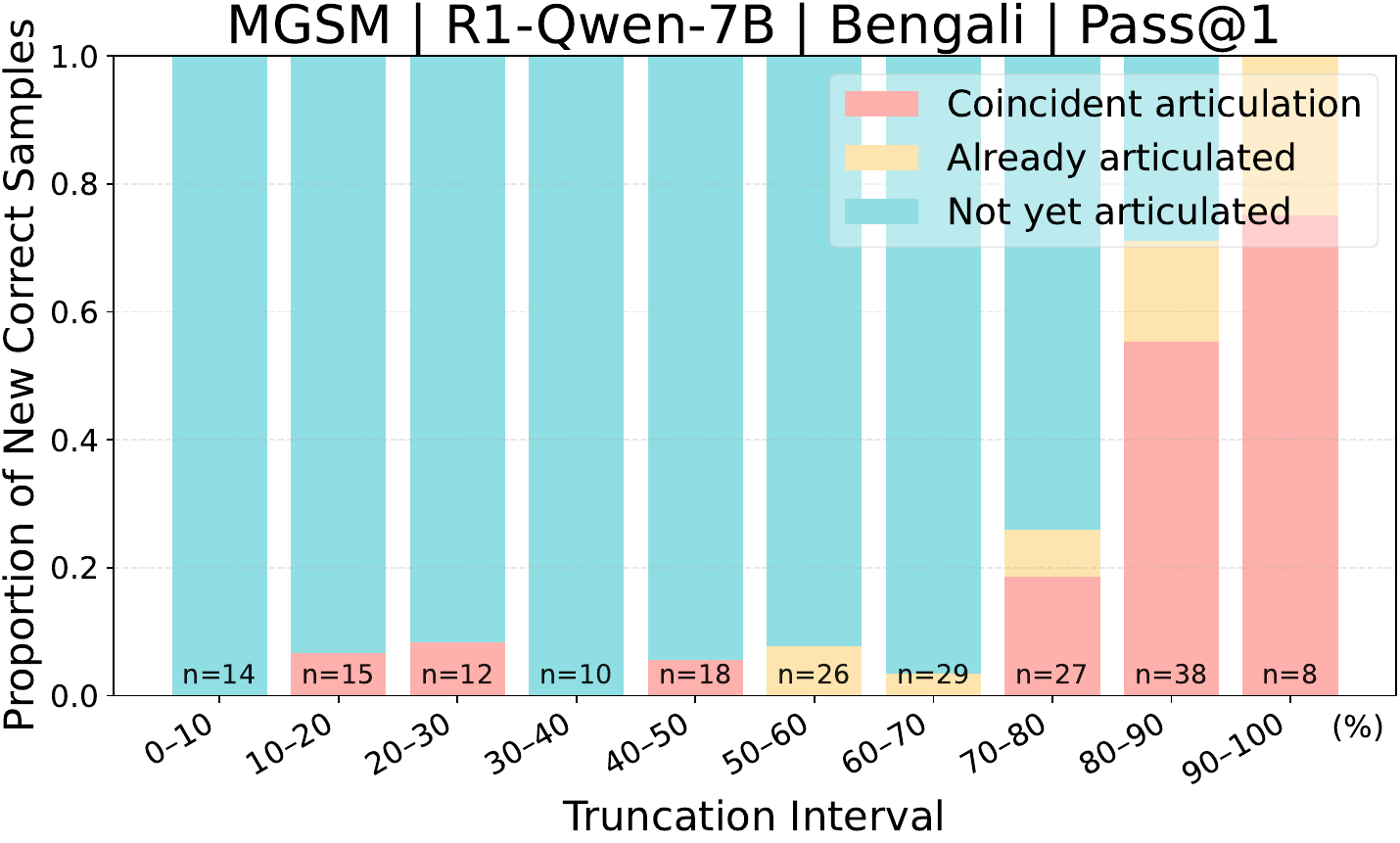}
    \includegraphics[width=0.24\textwidth]{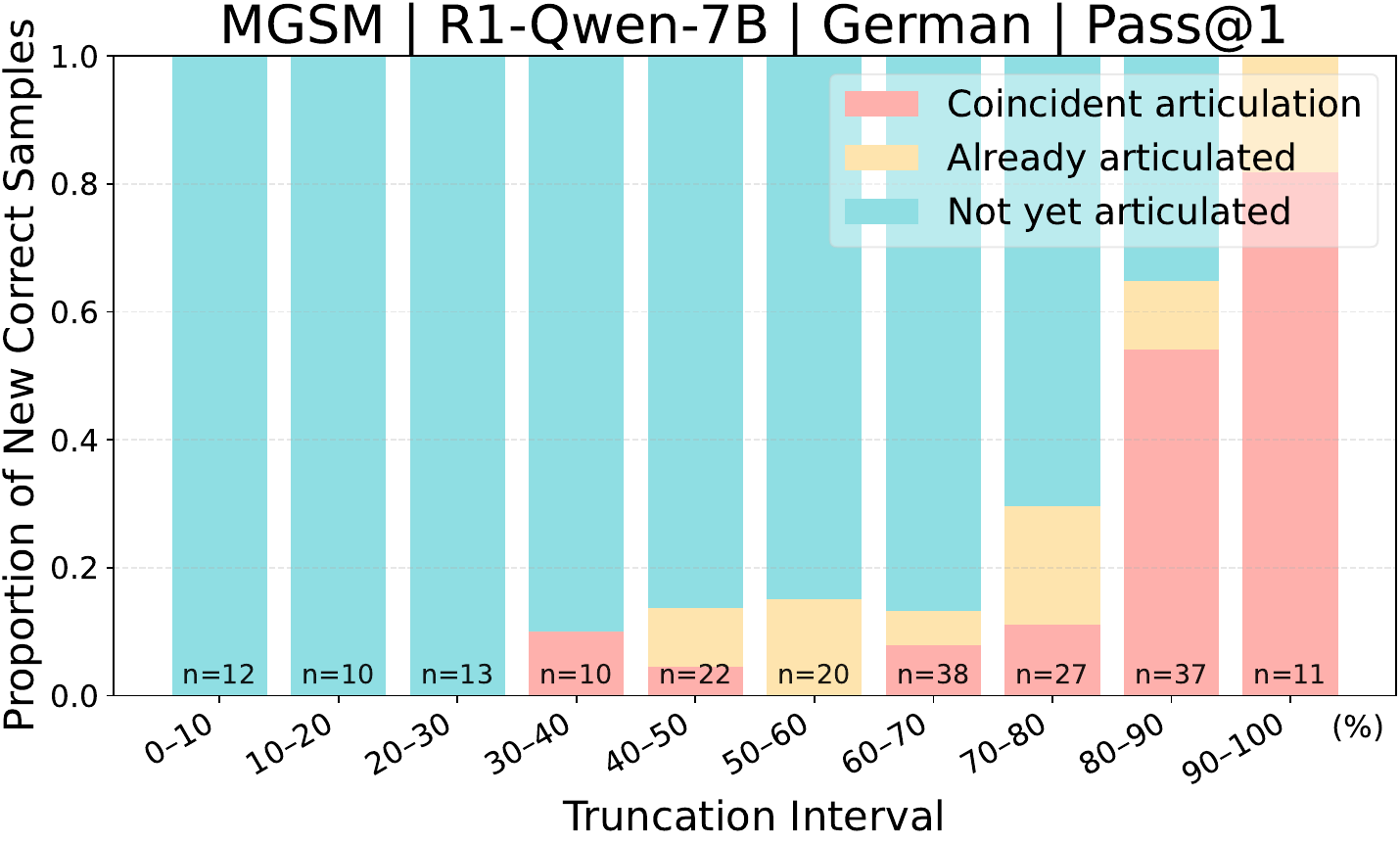}
    \includegraphics[width=0.24\textwidth]{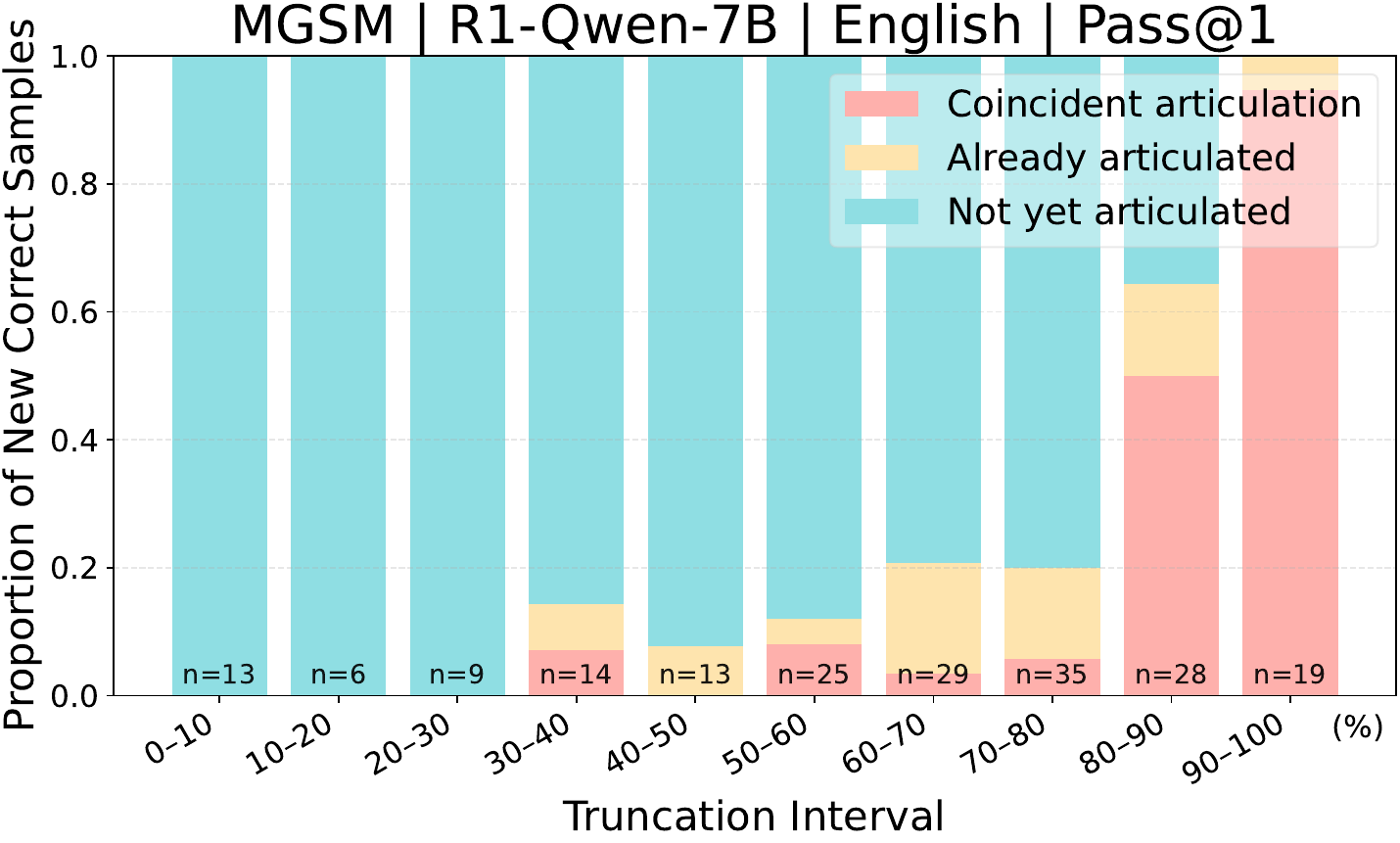}
    \includegraphics[width=0.24\textwidth]{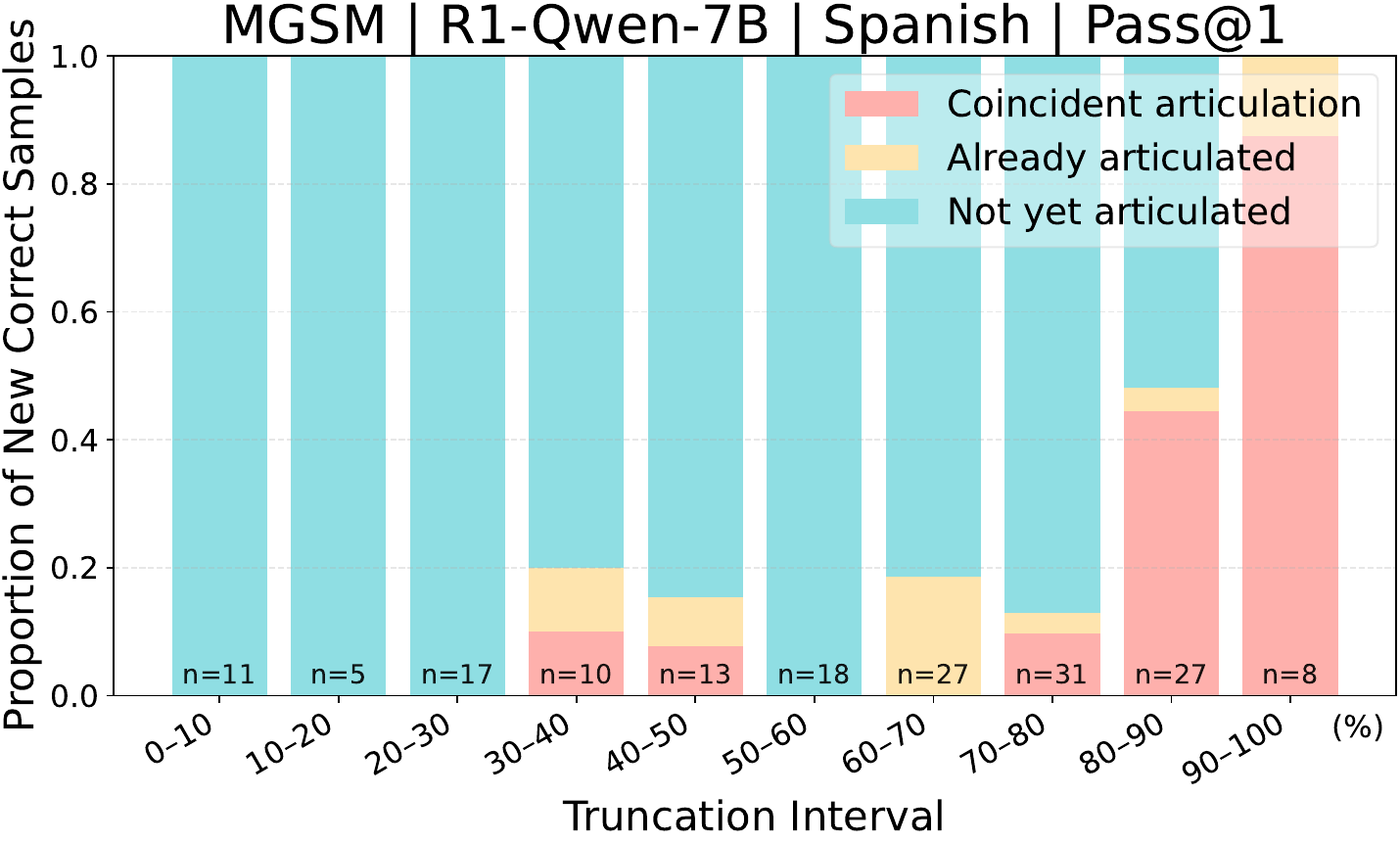}
    \includegraphics[width=0.24\textwidth]{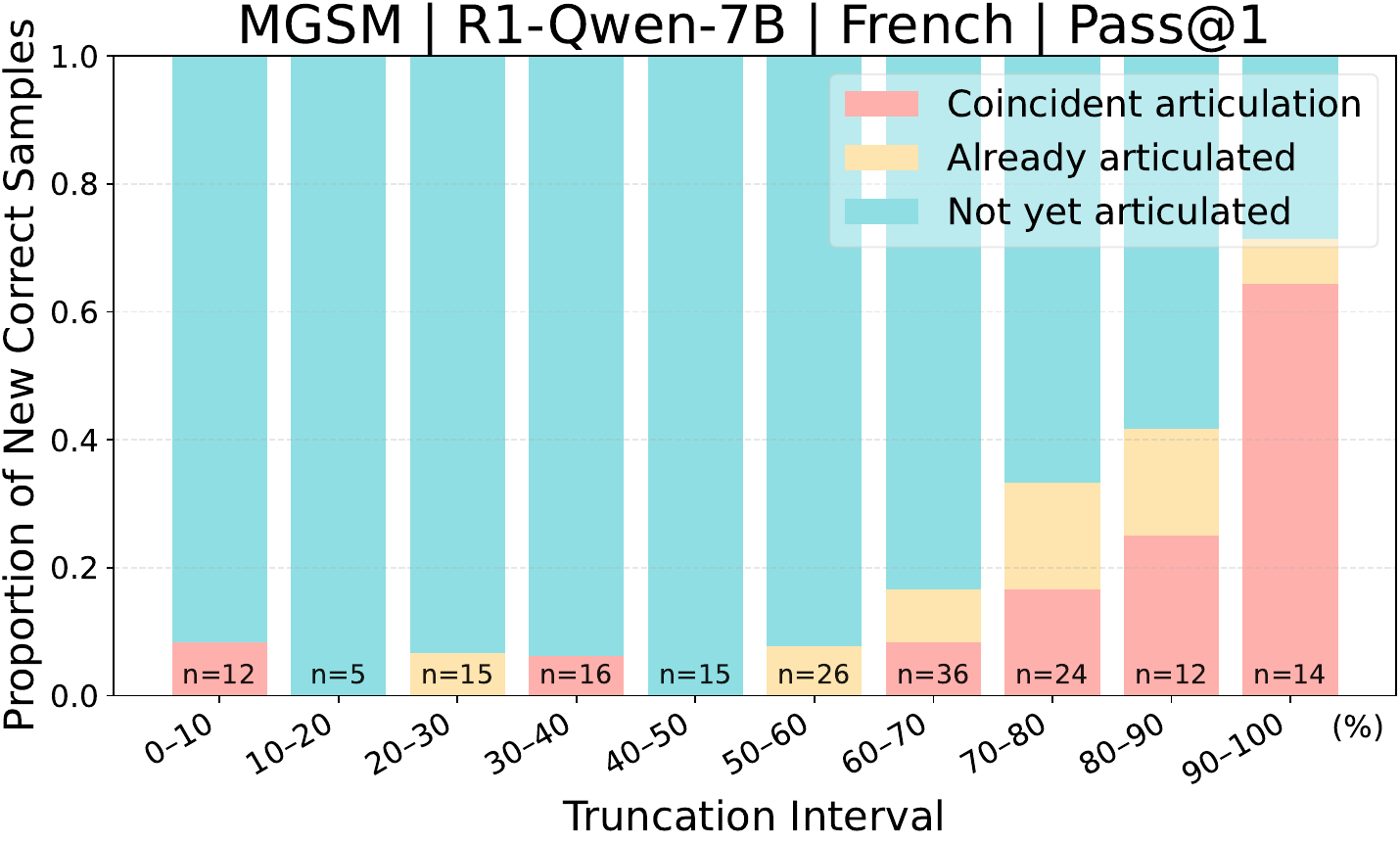}
    \includegraphics[width=0.24\textwidth]{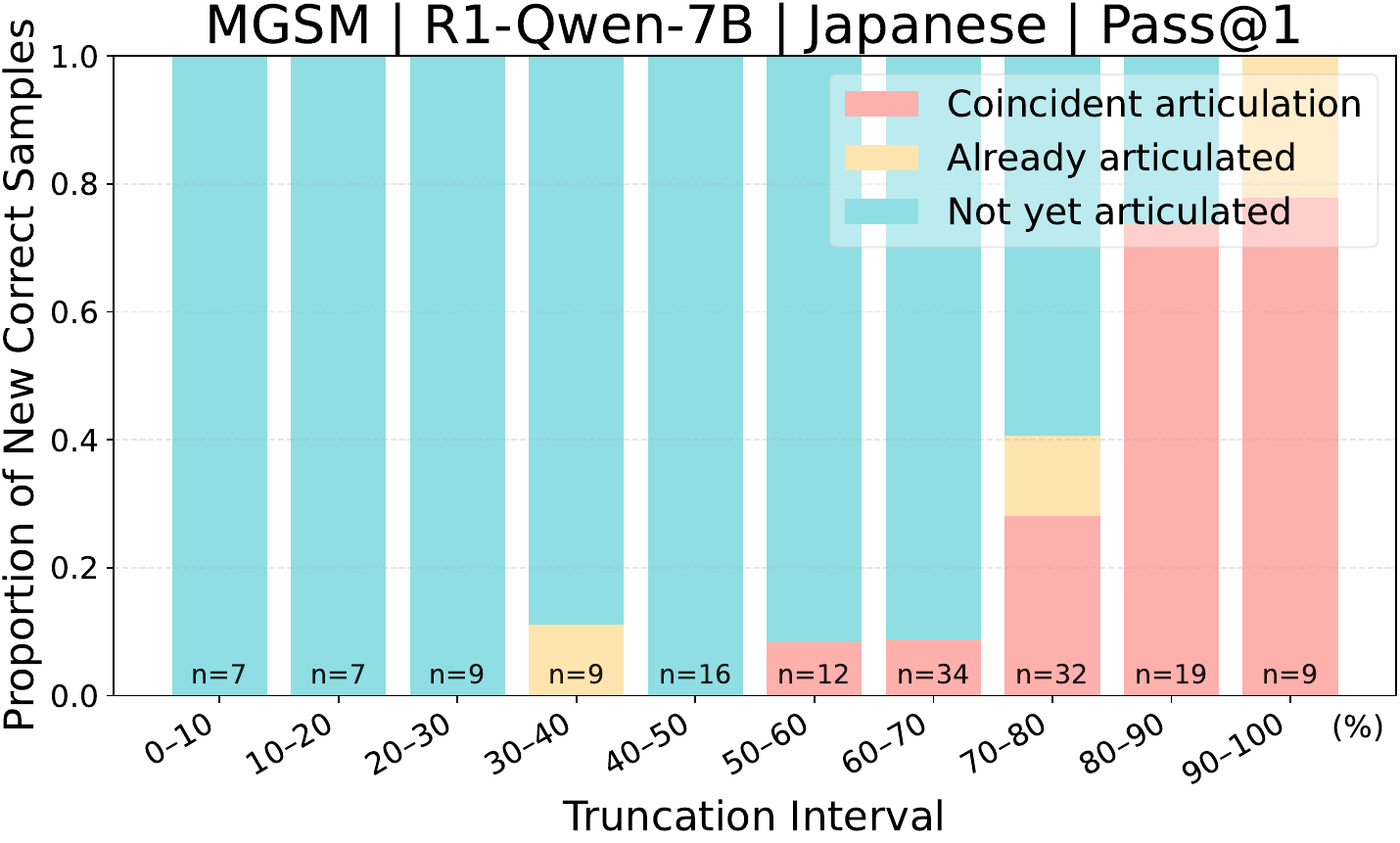}
    \includegraphics[width=0.24\textwidth]{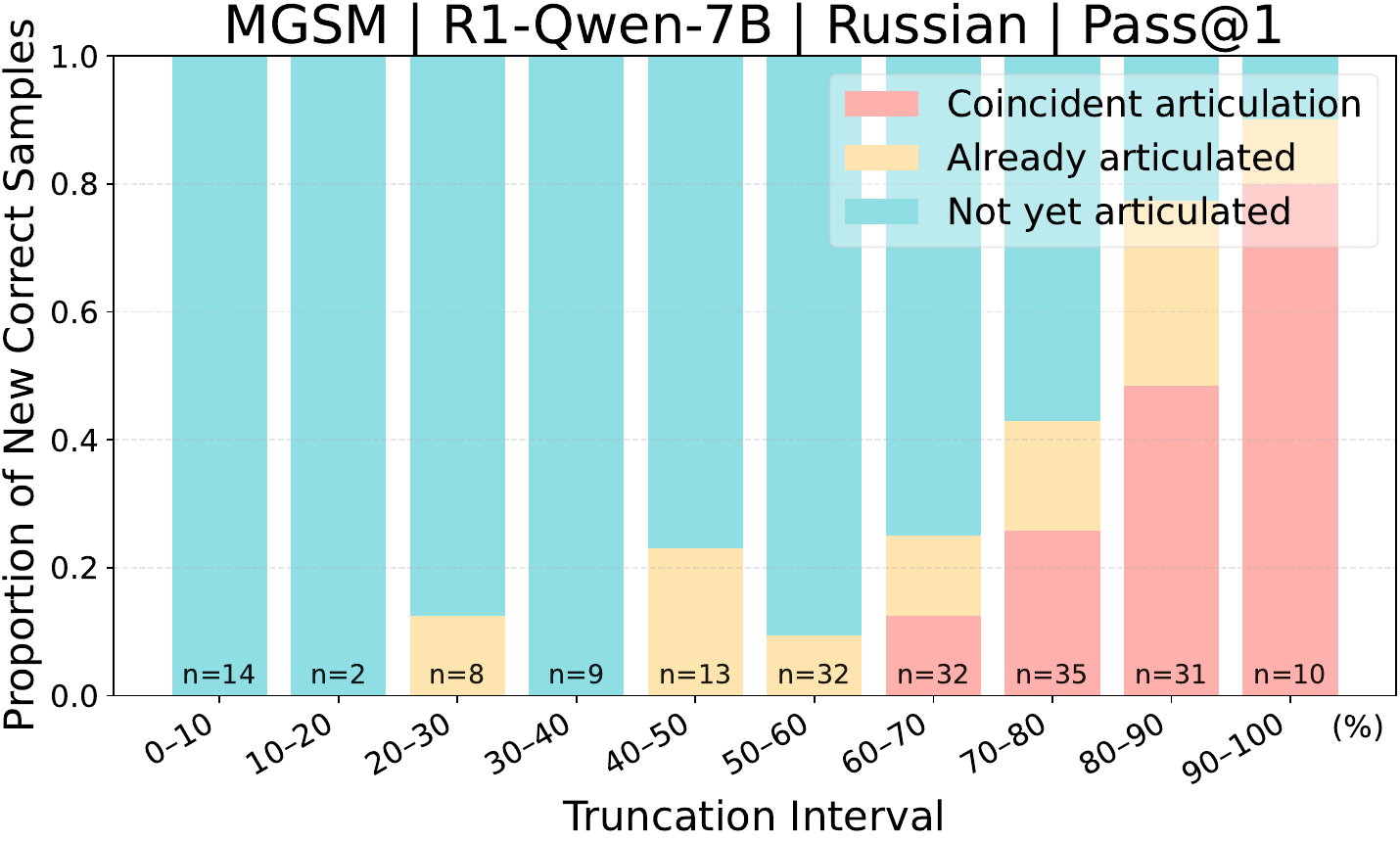}
    \includegraphics[width=0.24\textwidth]{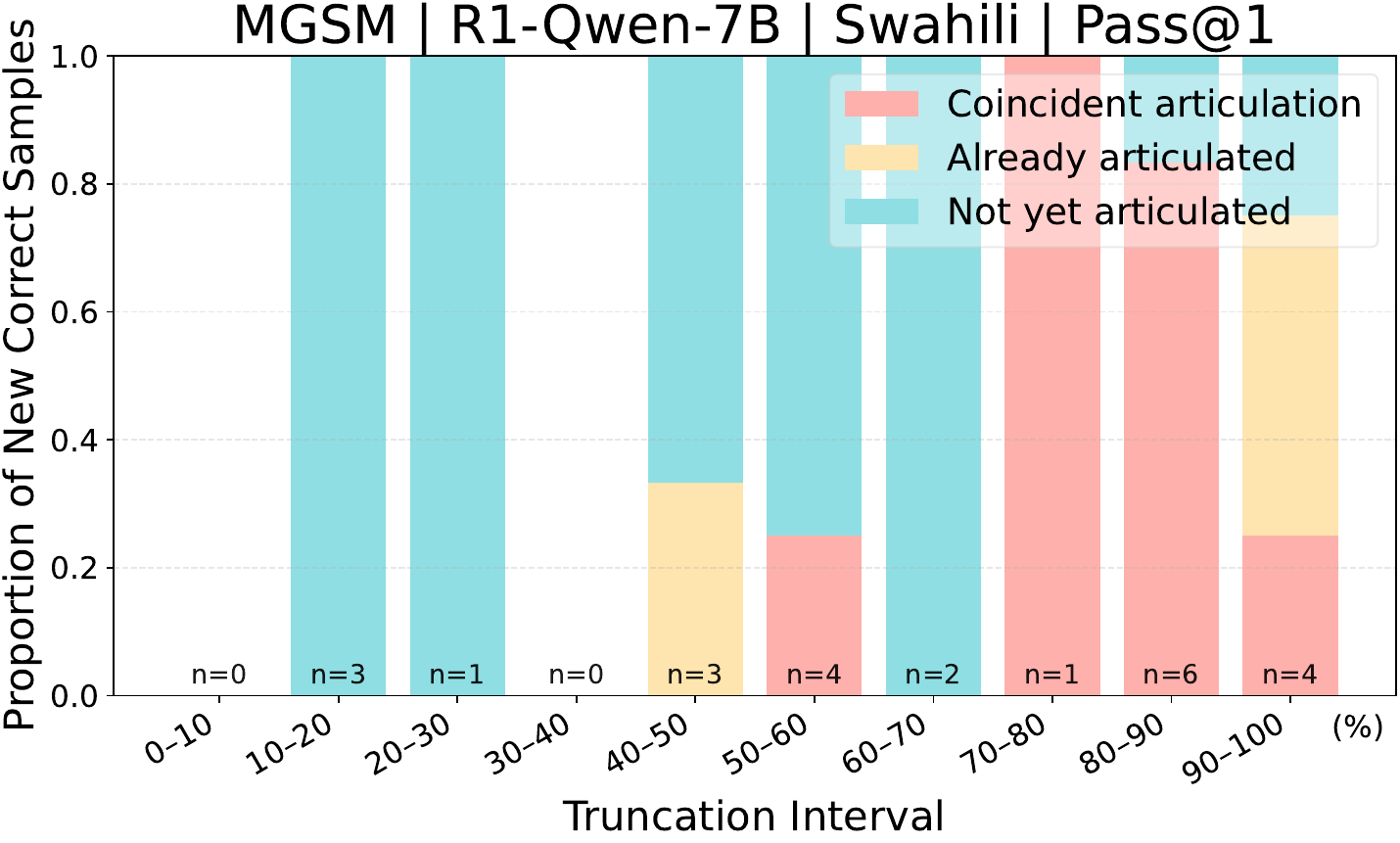}
    \includegraphics[width=0.24\textwidth]{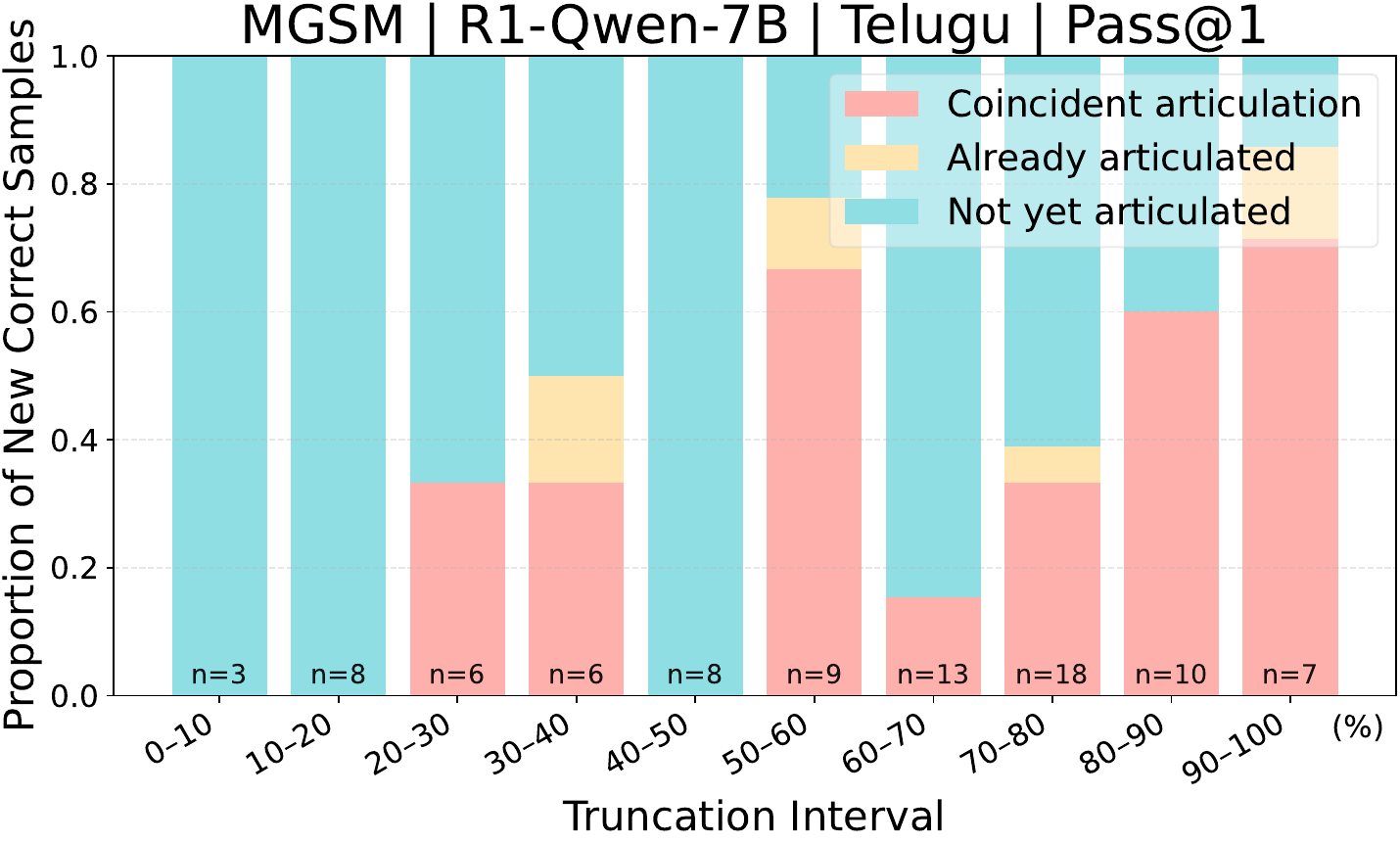}
    \includegraphics[width=0.24\textwidth]{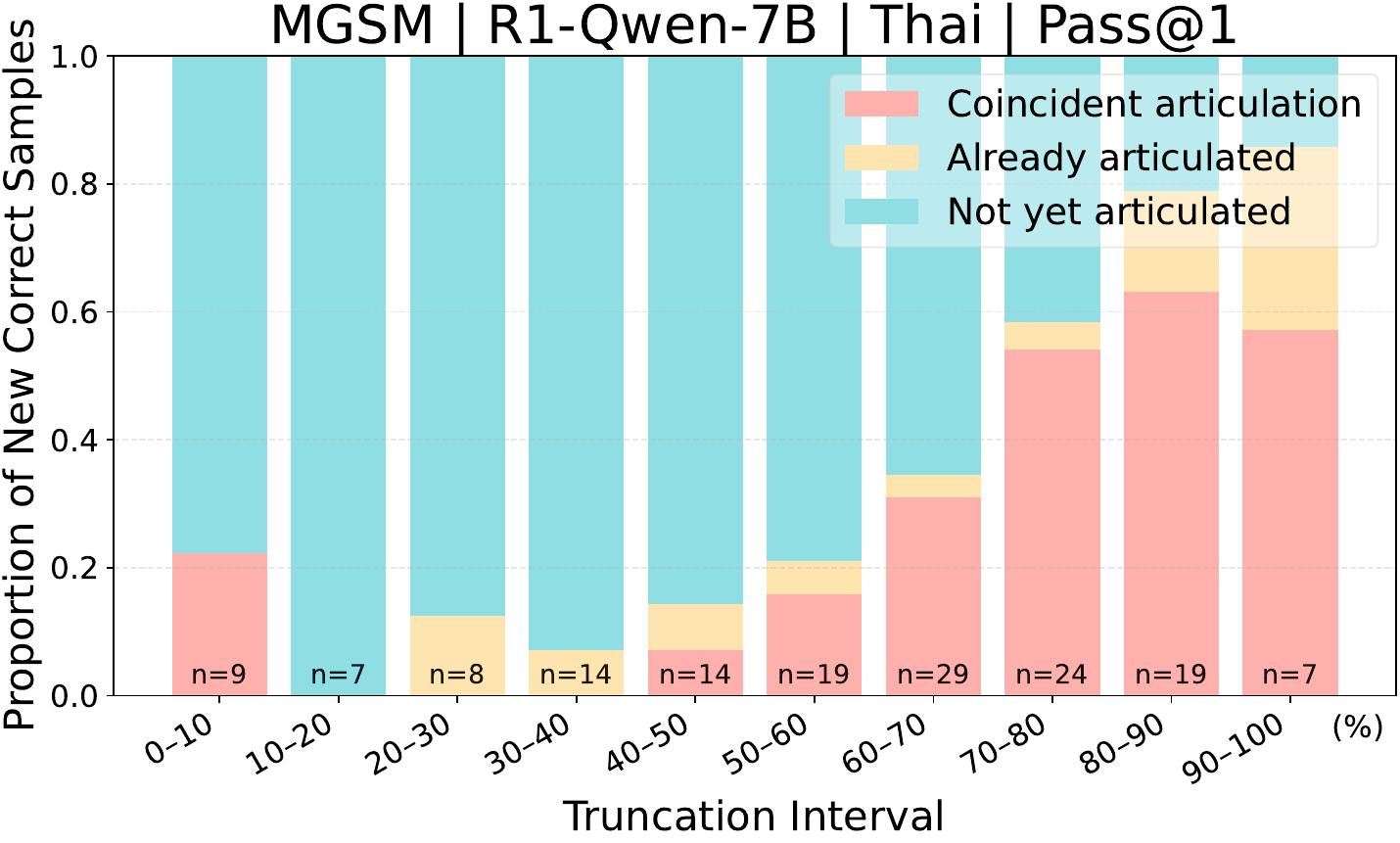}
    \includegraphics[width=0.24\textwidth]{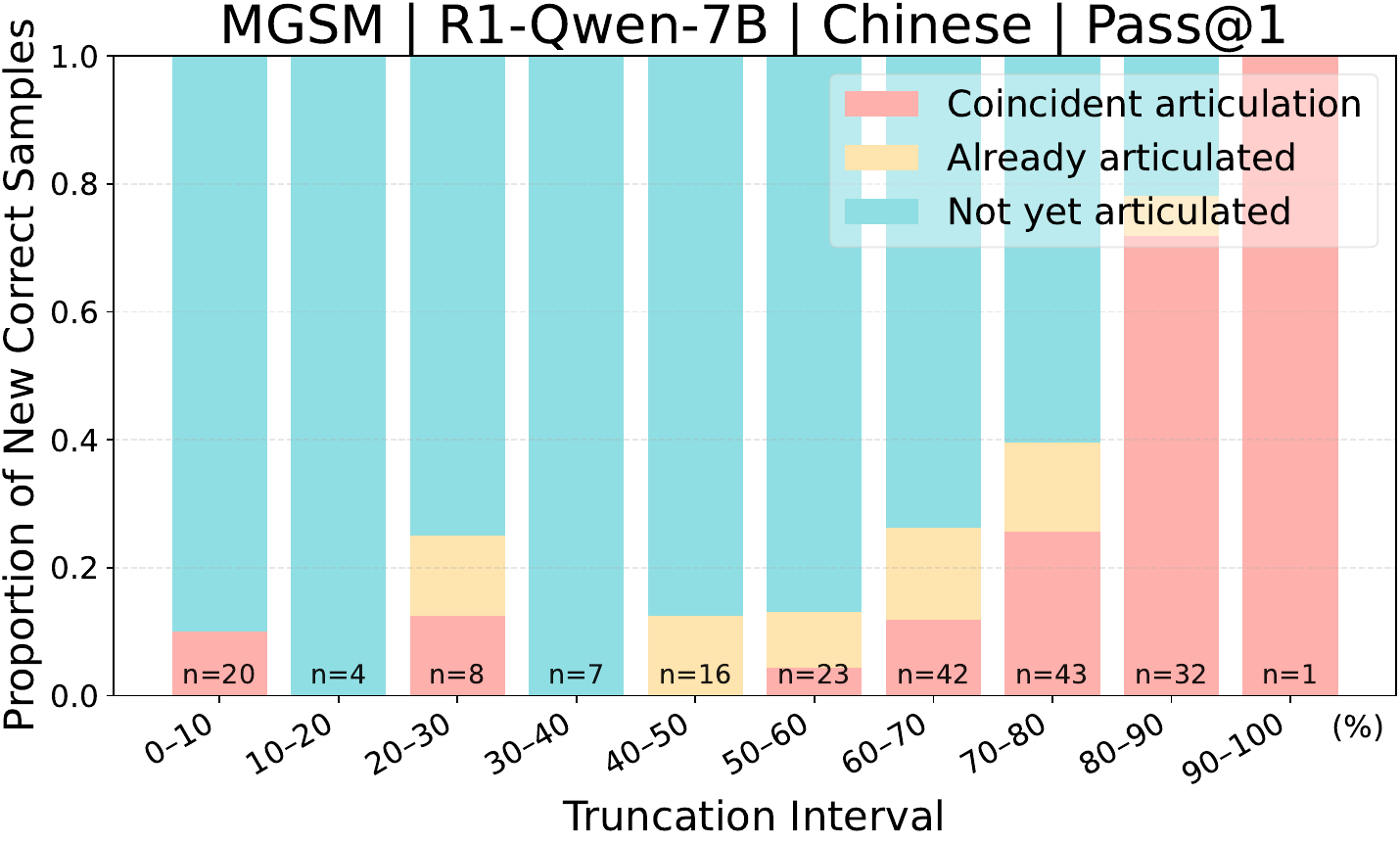}
    \caption{
    Causal decomposition of newly correct predictions across truncation intervals on \textbf{MGSM} with \textbf{R1-Qwen-7B}.
    Each bar partitions gains into three cases: (\textbf{i}) the gold answer is first articulated in the newly added reasoning steps,
    (\textbf{ii}) it was already articulated in earlier steps, or
    (\textbf{iii}) it has not yet appeared in the visible truncated trace.
    Early and intermediate gains are dominated by case (\textbf{iii}), indicating latent reasoning.
    }
    \label{fig:interval_7b_mgsm}
\end{figure*}

\begin{figure*}
    \centering
    \includegraphics[width=0.24\textwidth]{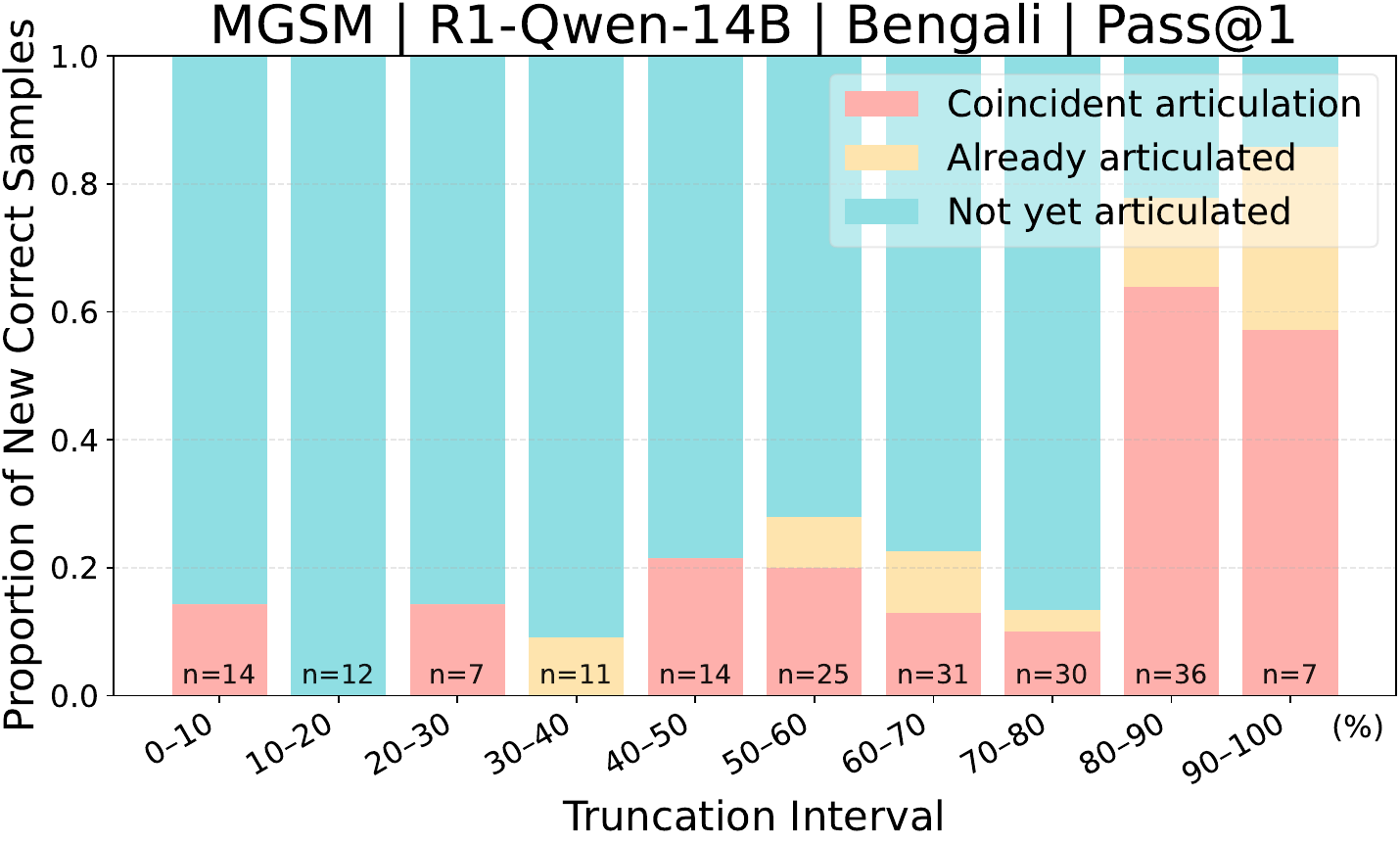}
    \includegraphics[width=0.24\textwidth]{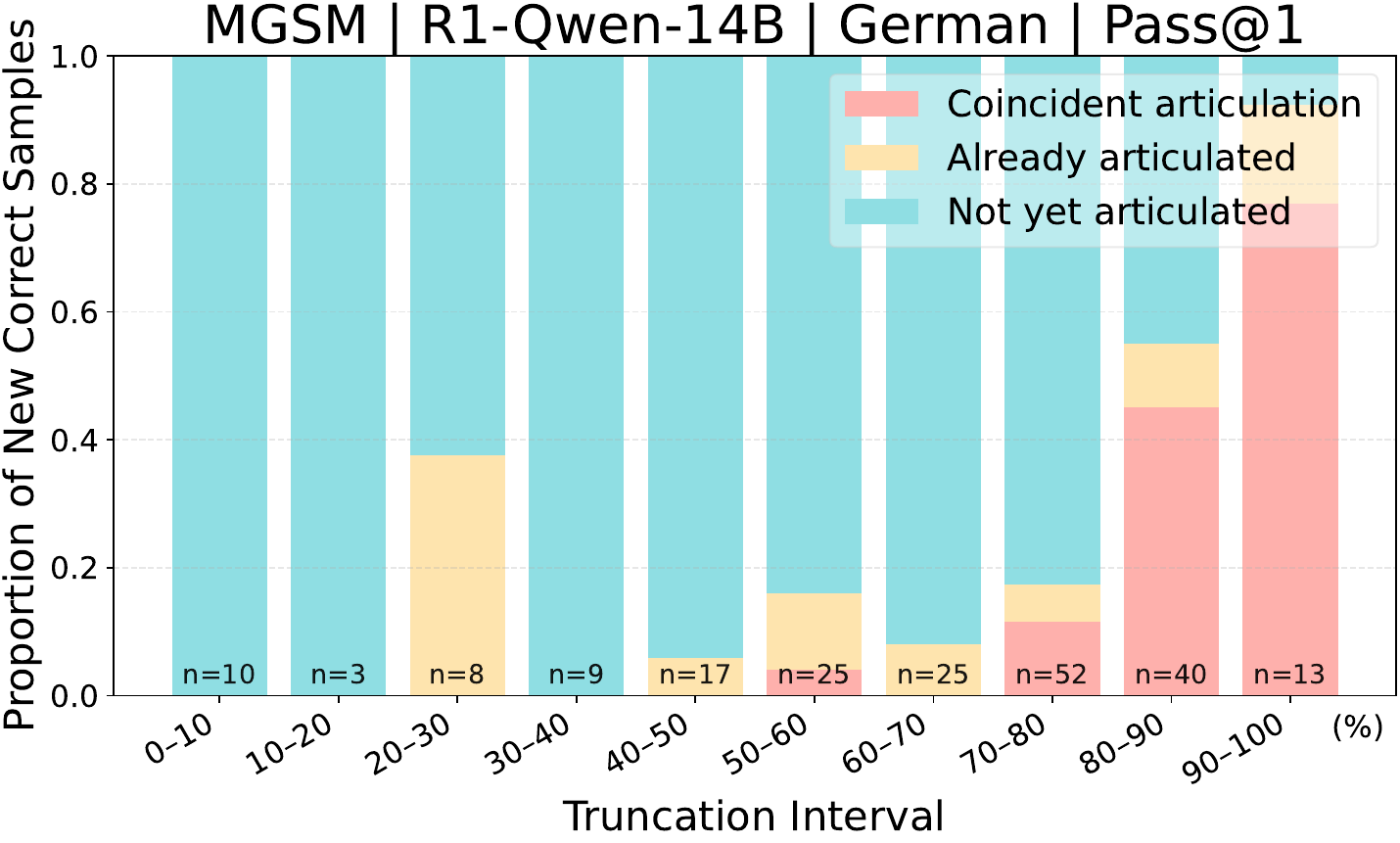}
    \includegraphics[width=0.24\textwidth]{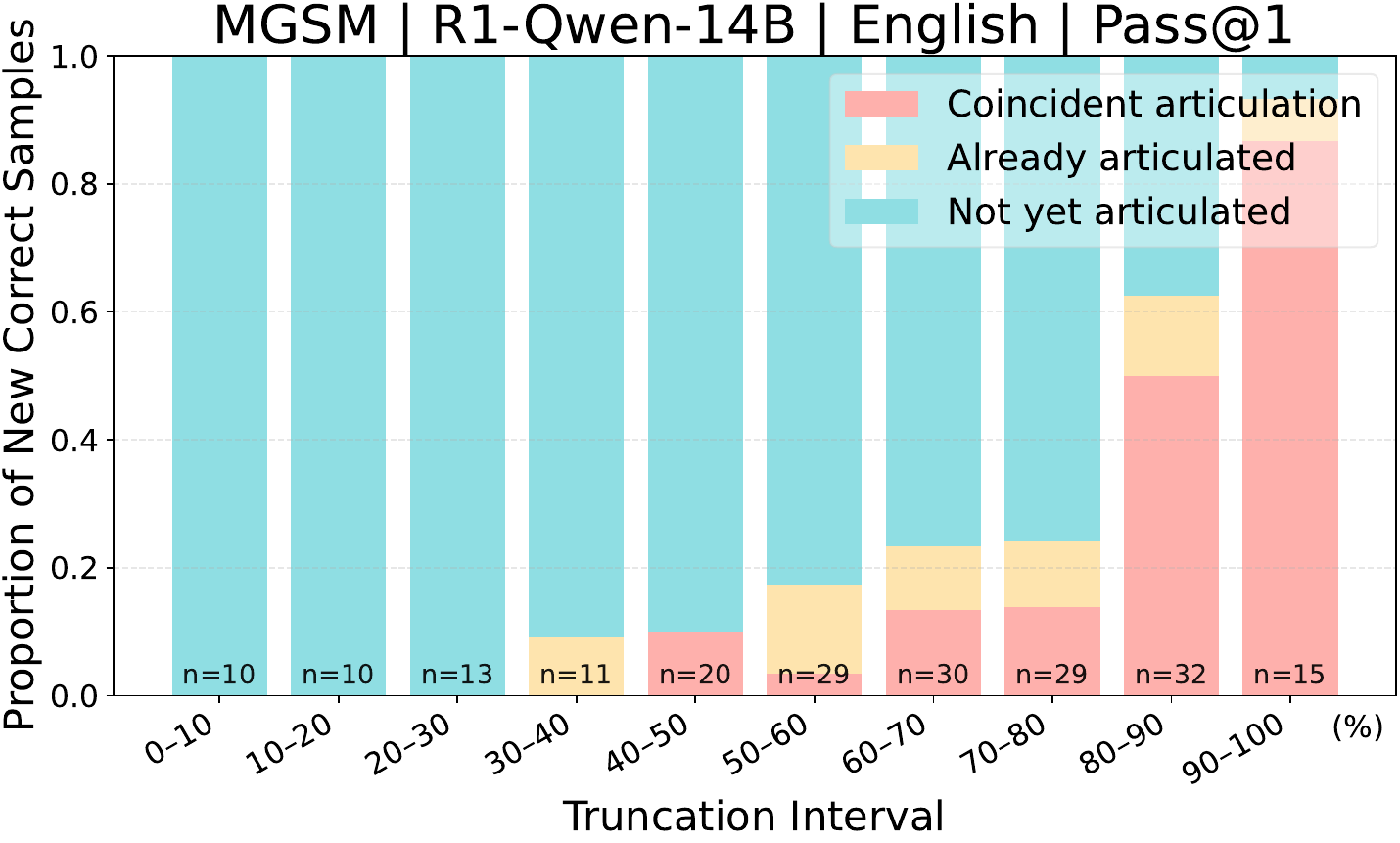}
    \includegraphics[width=0.24\textwidth]{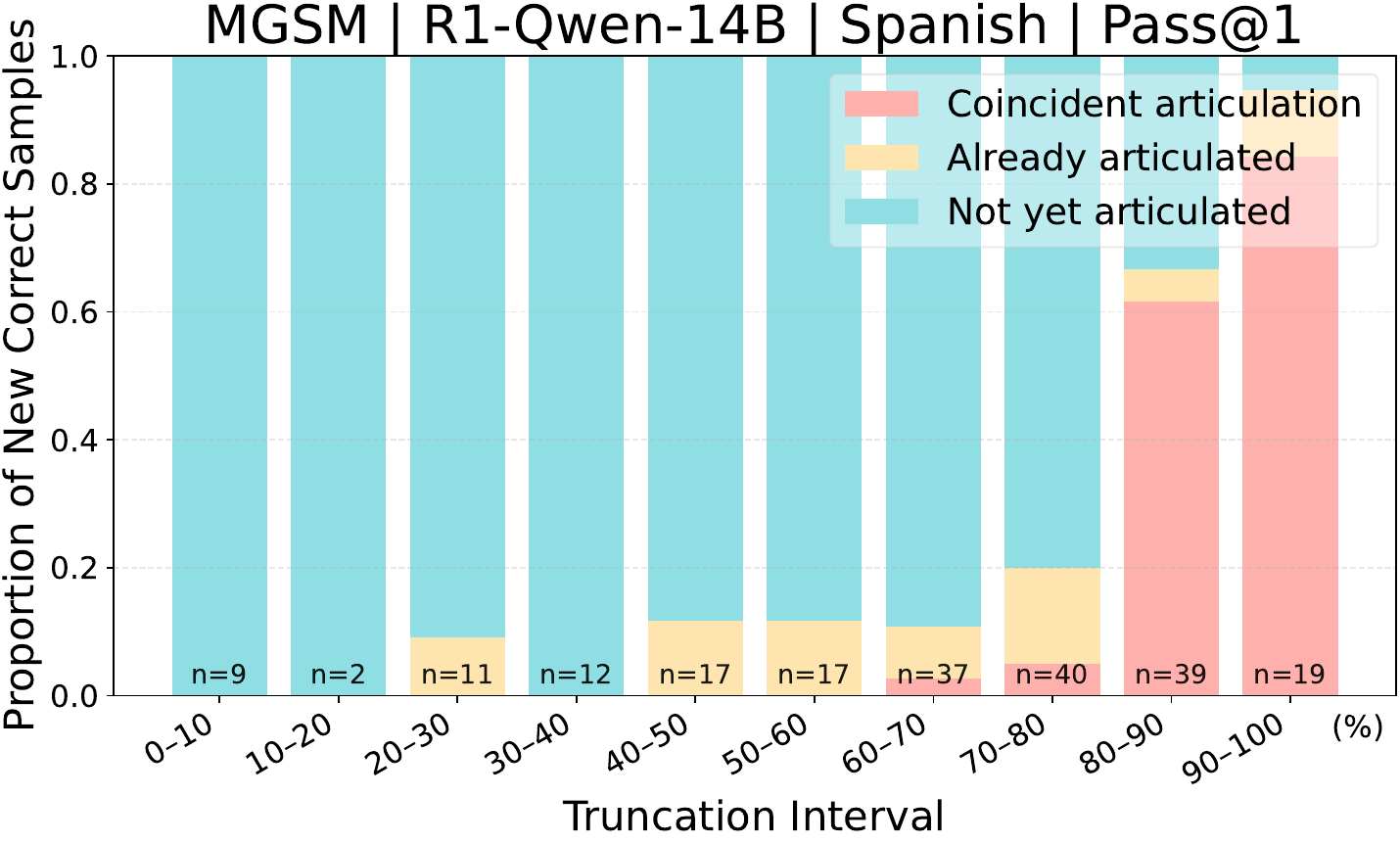}
    \includegraphics[width=0.24\textwidth]{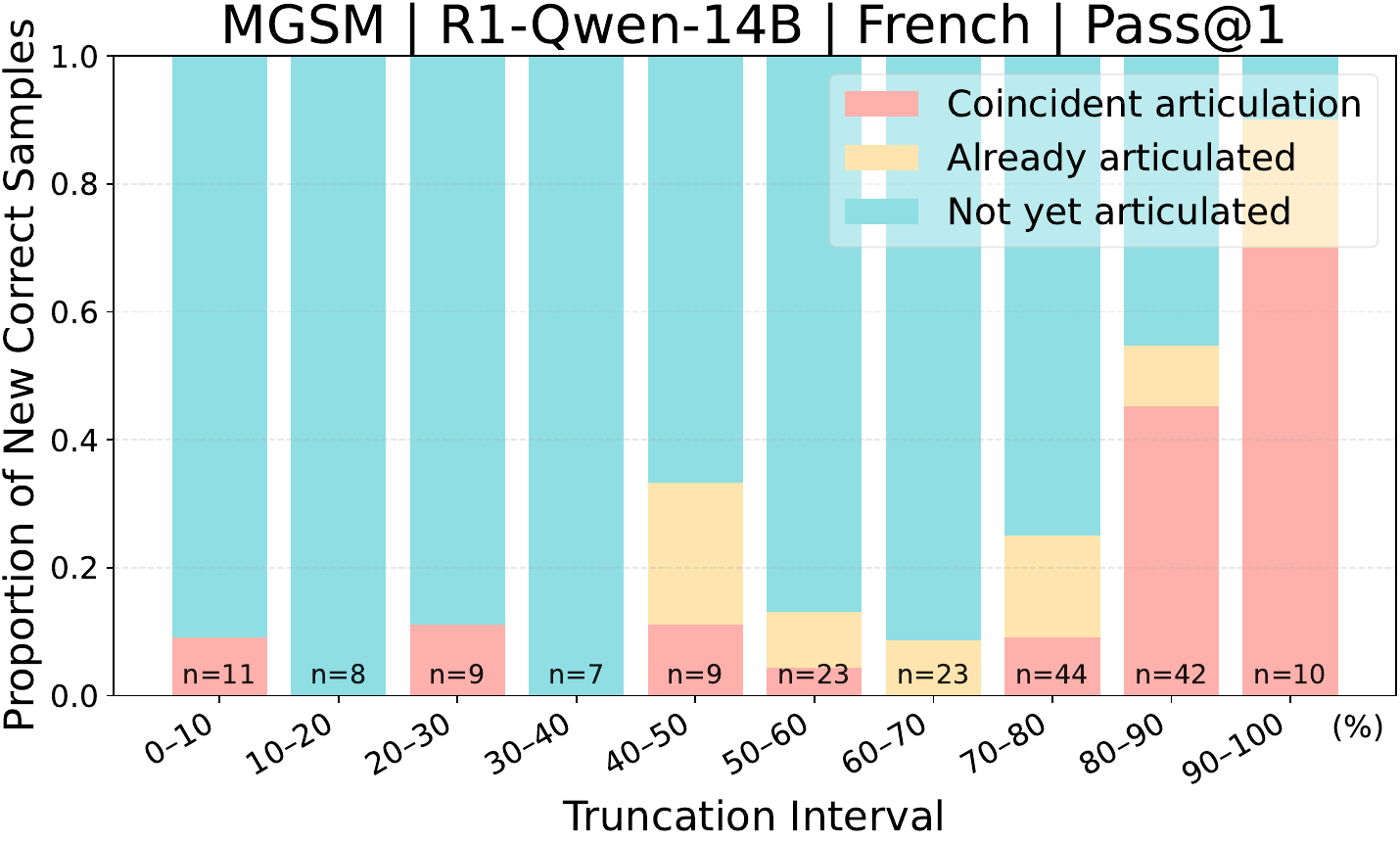}
    \includegraphics[width=0.24\textwidth]{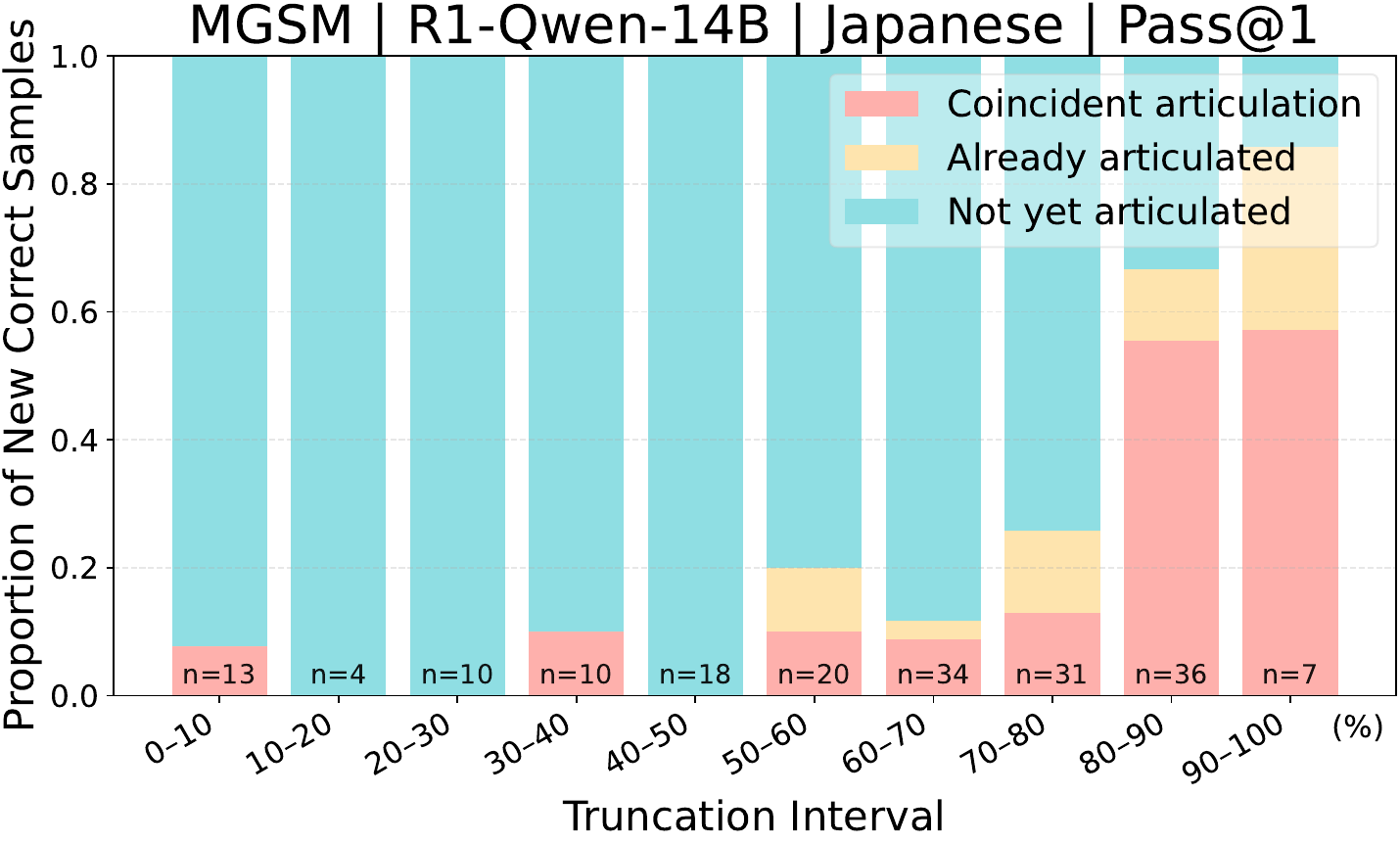}
    \includegraphics[width=0.24\textwidth]{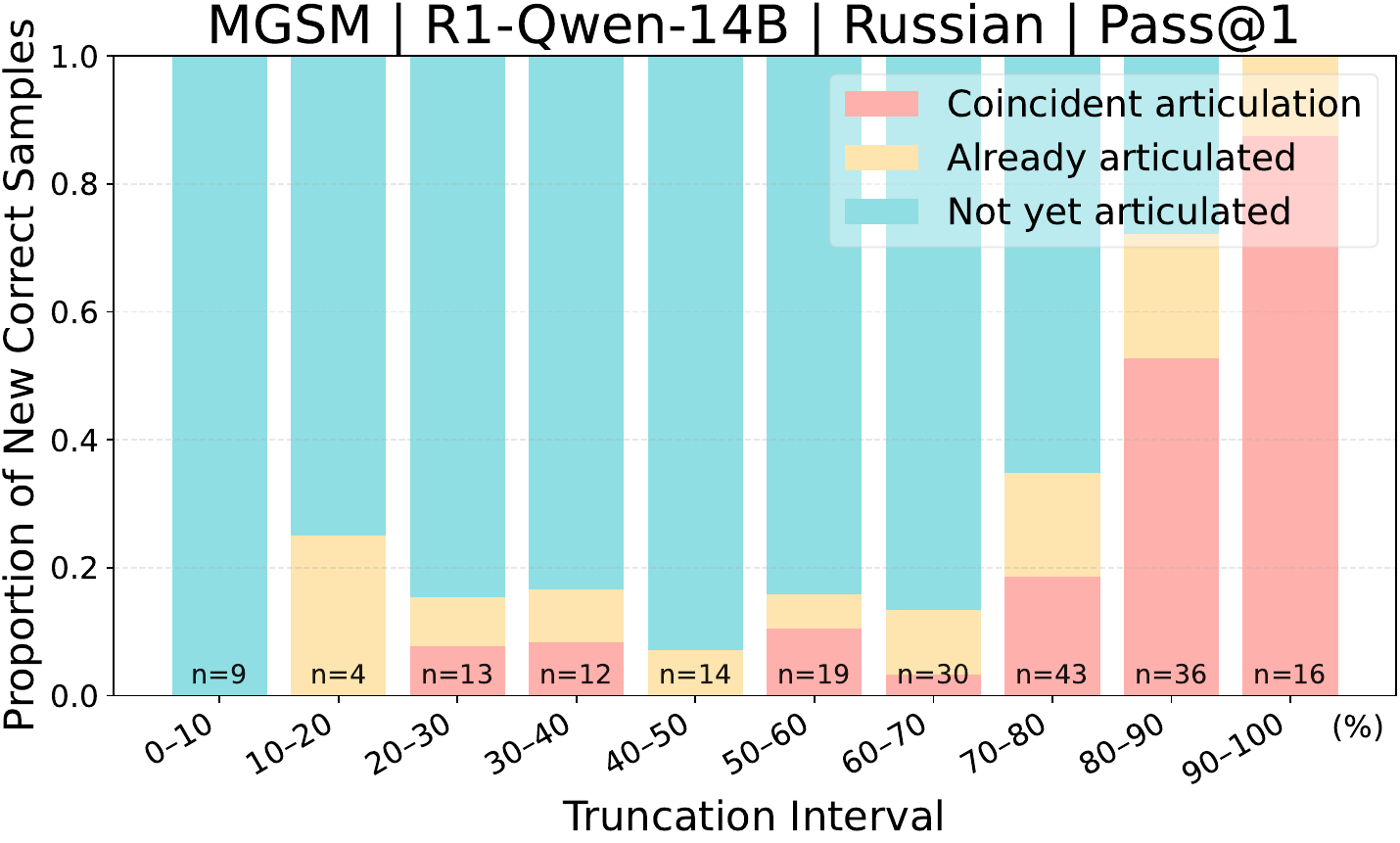}
    \includegraphics[width=0.24\textwidth]{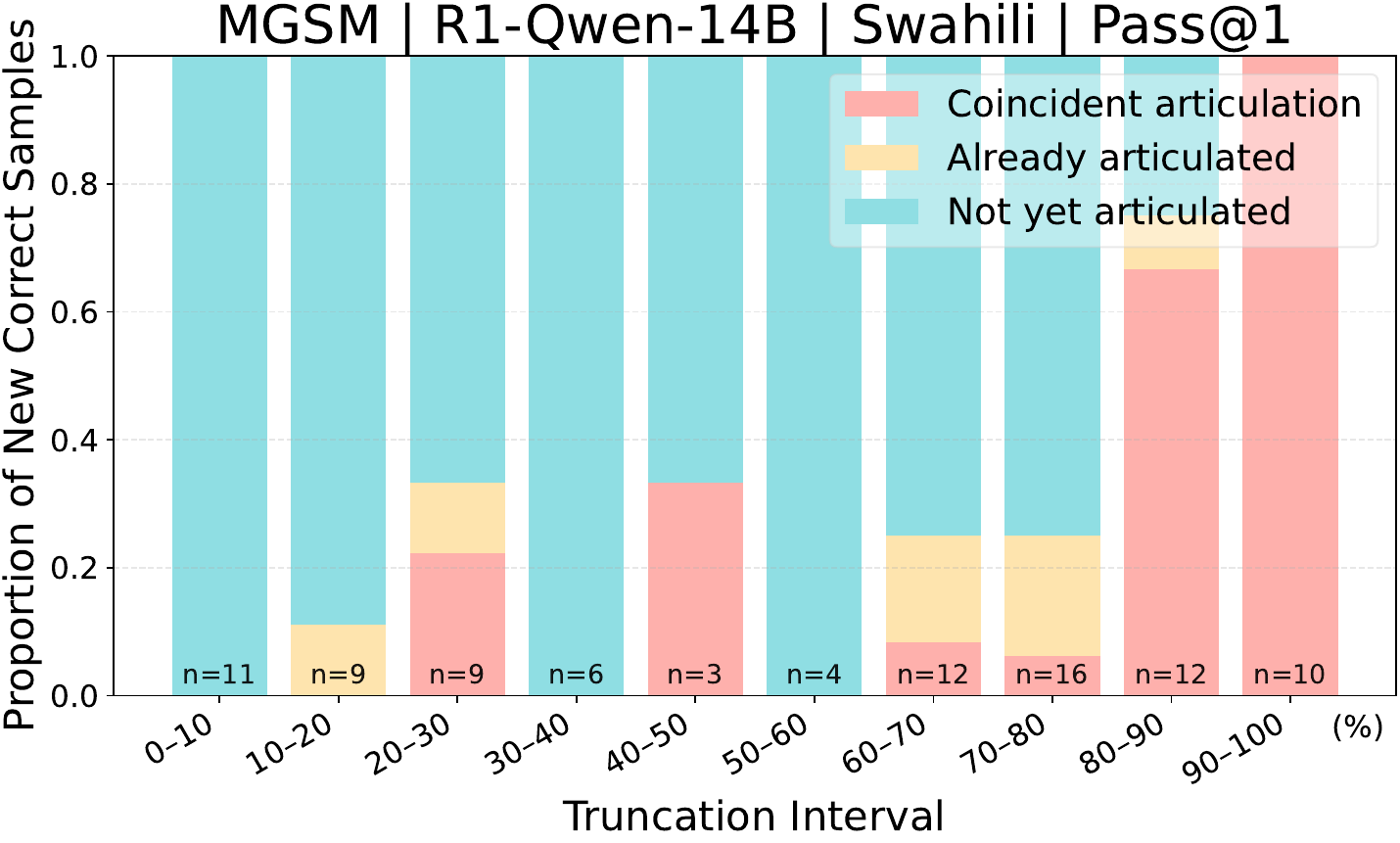}
    \includegraphics[width=0.24\textwidth]{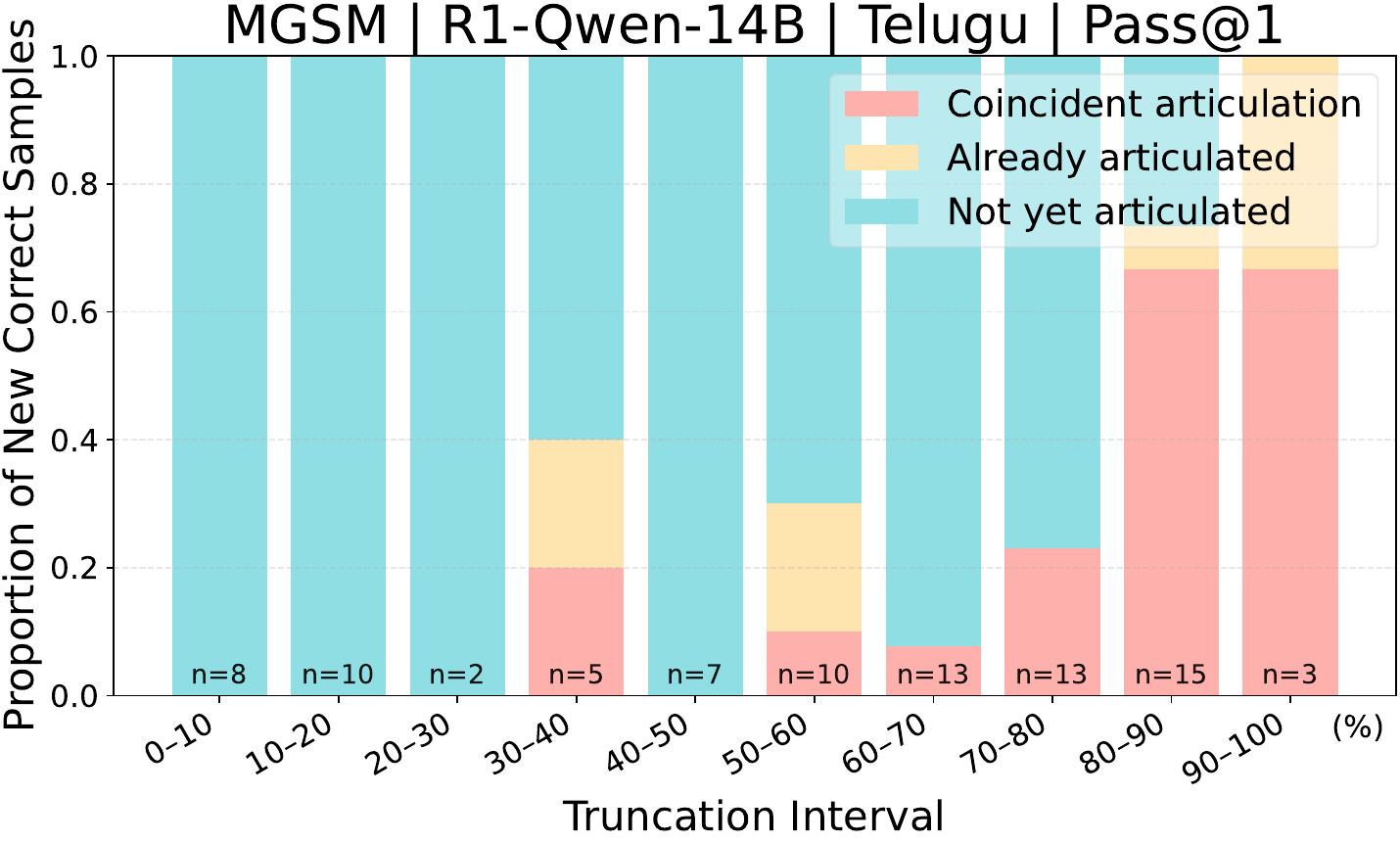}
    \includegraphics[width=0.24\textwidth]{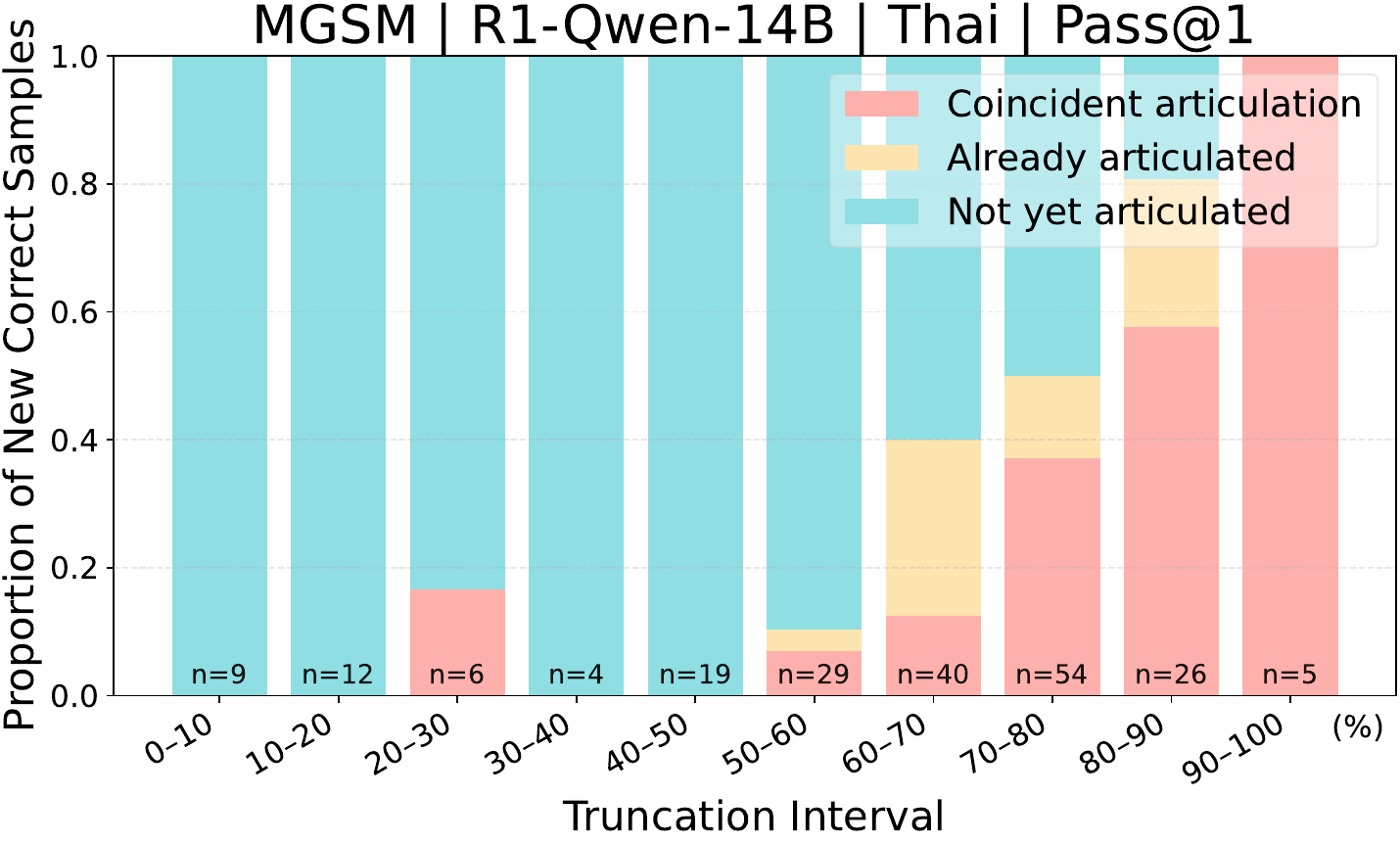}
    \includegraphics[width=0.24\textwidth]{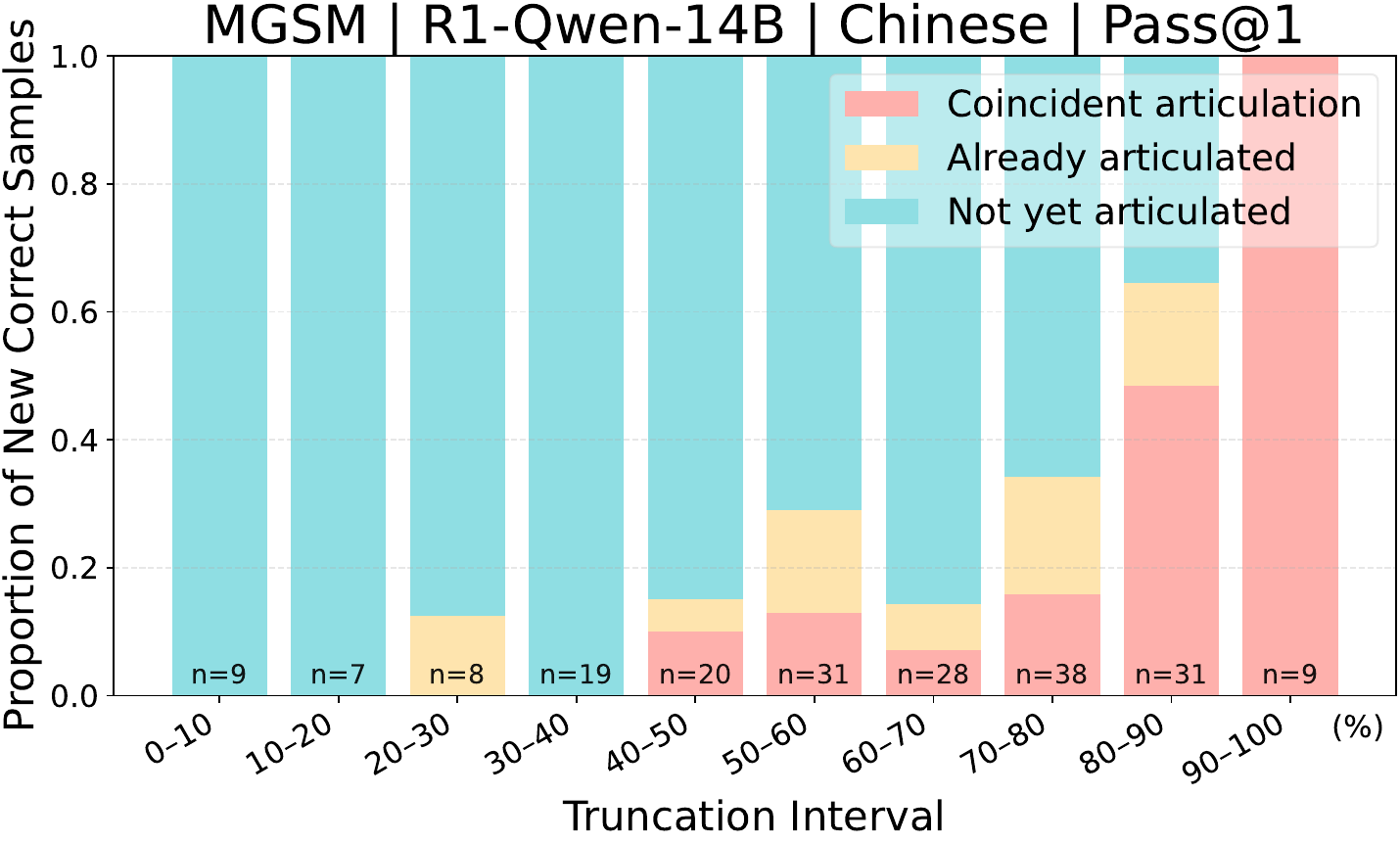}
    \caption{
    Causal decomposition of newly correct predictions across truncation intervals on \textbf{MGSM} with \textbf{R1-Qwen-14B}.
    Each bar partitions gains into three cases: (\textbf{i}) the gold answer is first articulated in the newly added reasoning steps,
    (\textbf{ii}) it was already articulated in earlier steps, or
    (\textbf{iii}) it has not yet appeared in the visible truncated trace.
    Early and intermediate gains are dominated by case (\textbf{iii}), indicating latent reasoning.
    }
    \label{fig:interval_14b_mgsm}
\end{figure*}

\begin{figure*}
    \centering
    \includegraphics[width=0.24\textwidth]{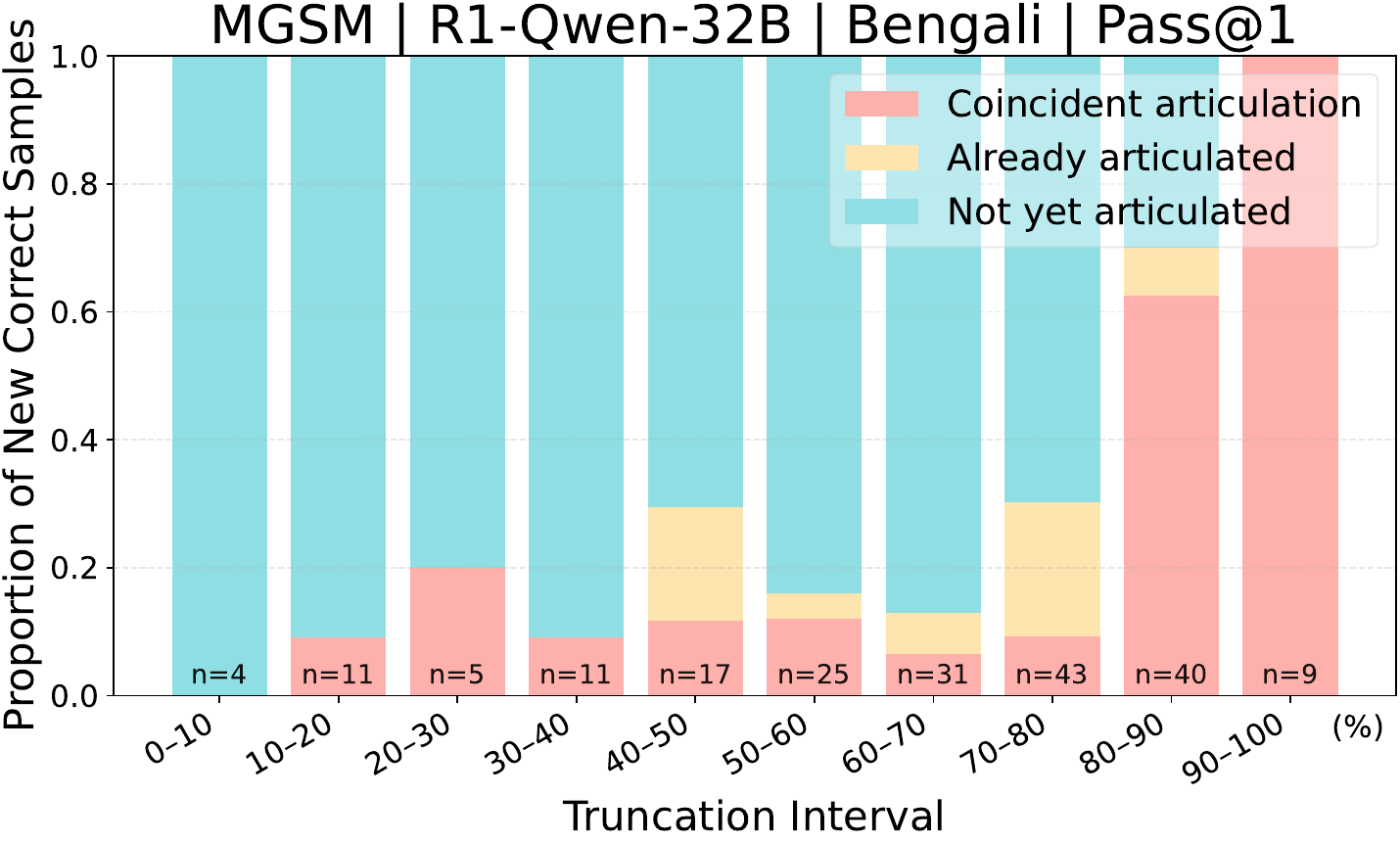}
    \includegraphics[width=0.24\textwidth]{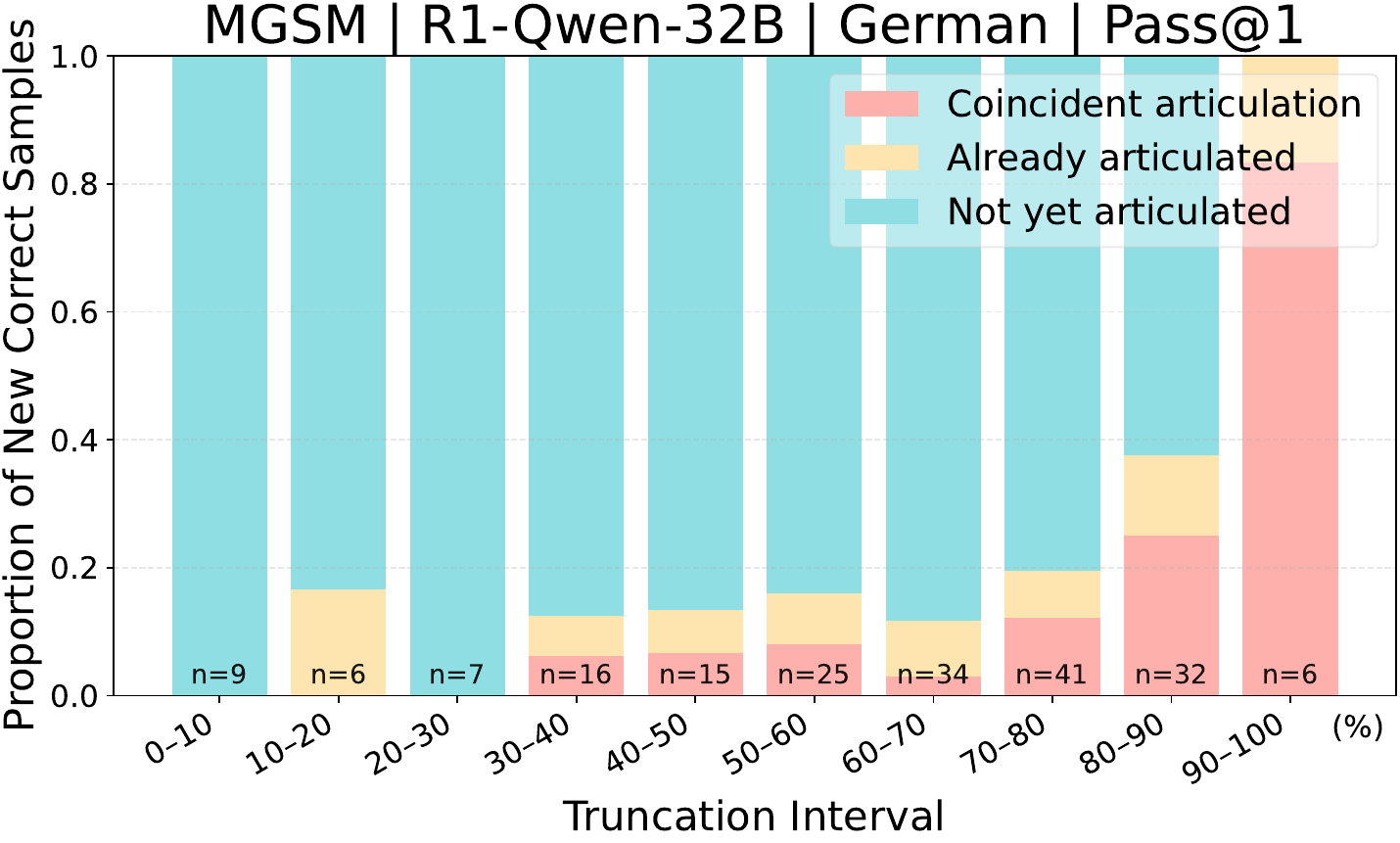}
    \includegraphics[width=0.24\textwidth]{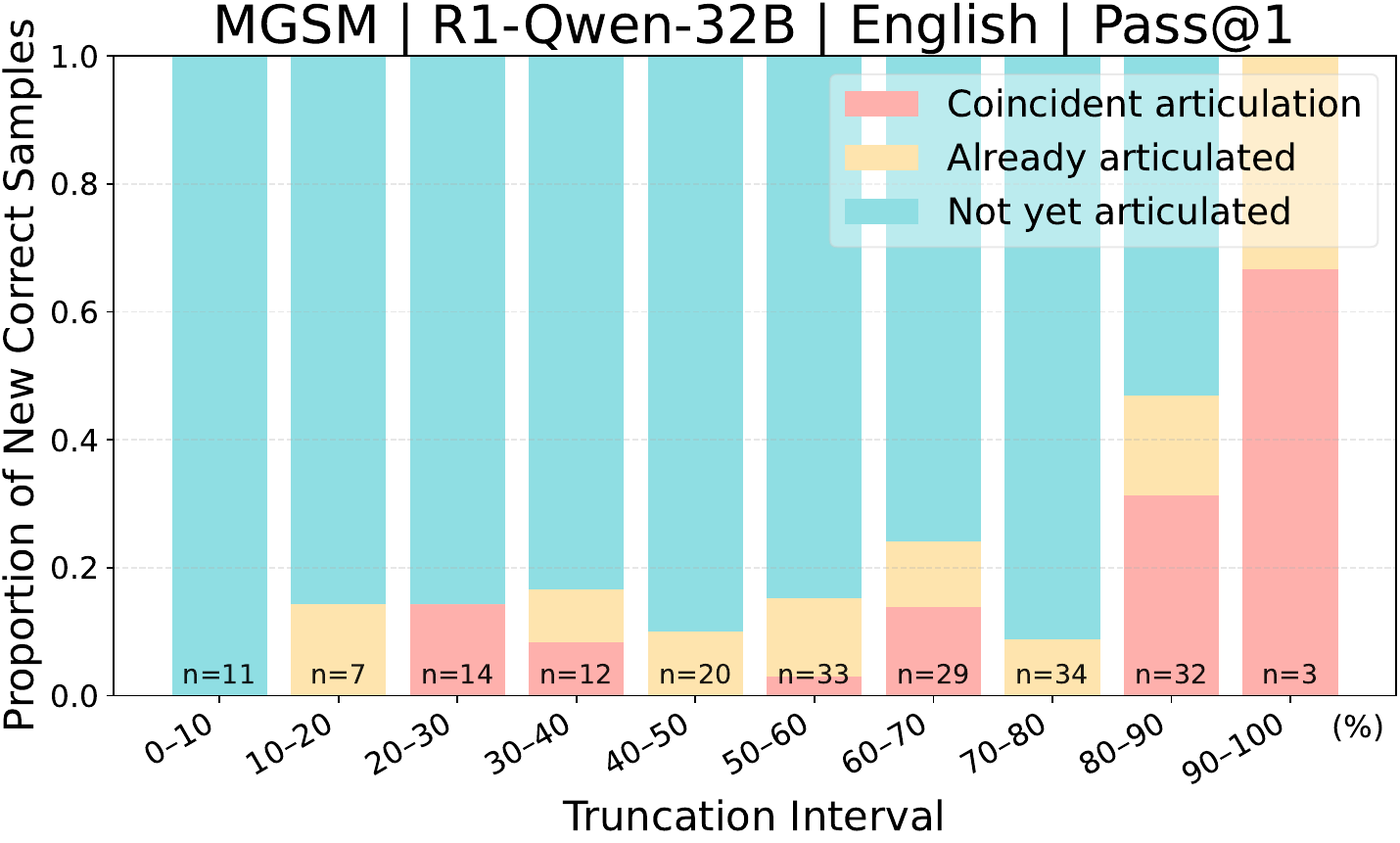}
    \includegraphics[width=0.24\textwidth]{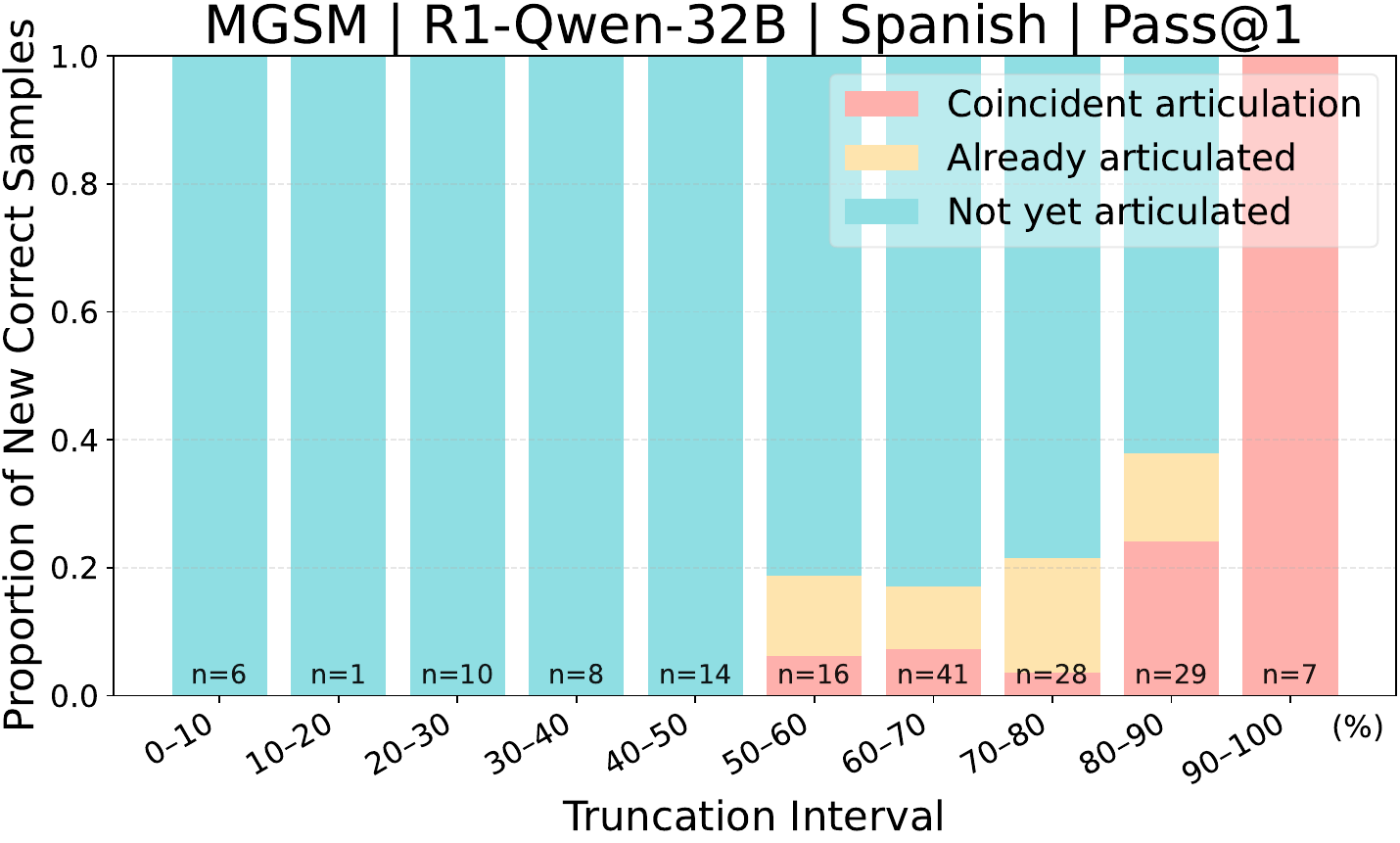}
    \includegraphics[width=0.24\textwidth]{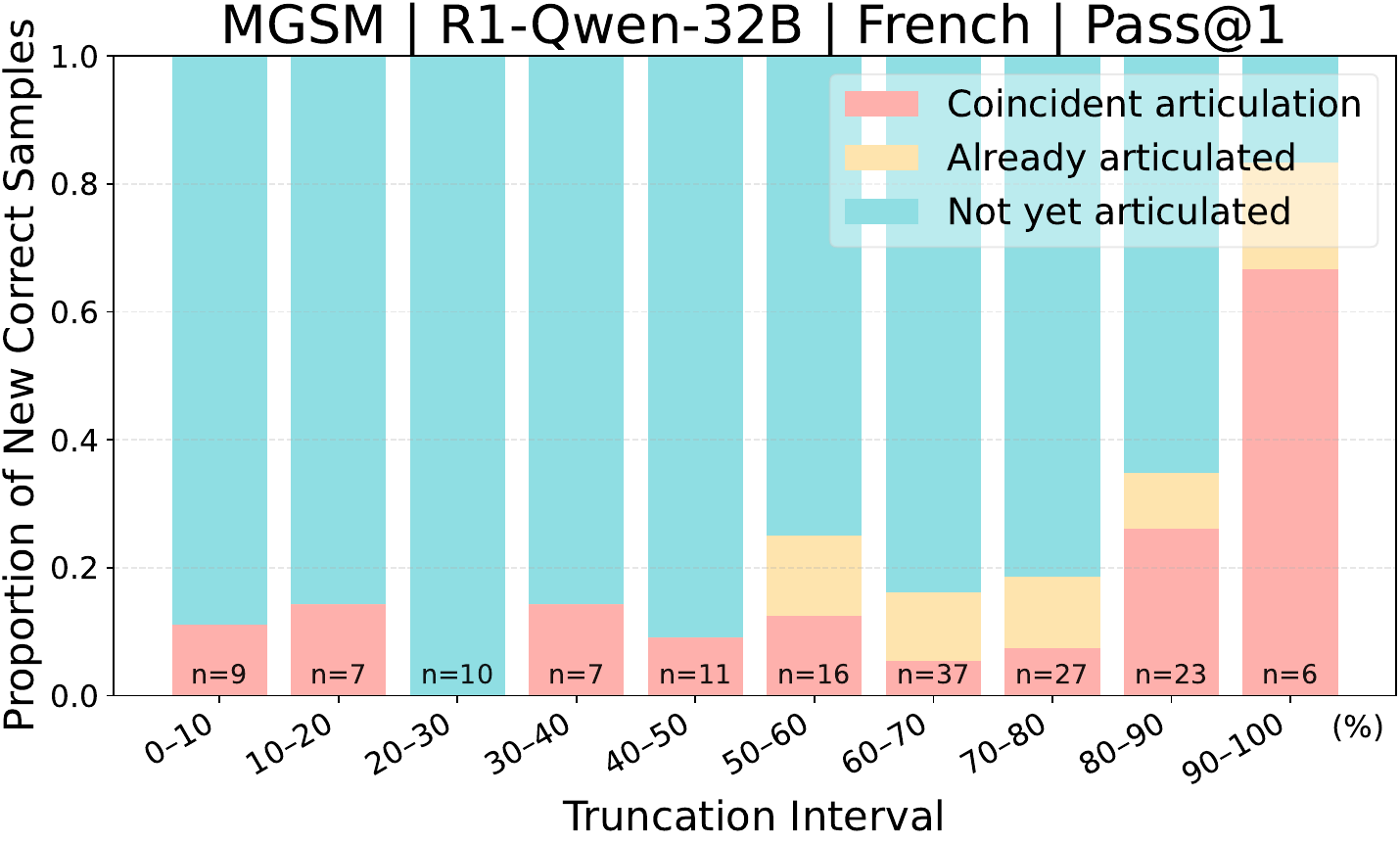}
    \includegraphics[width=0.24\textwidth]{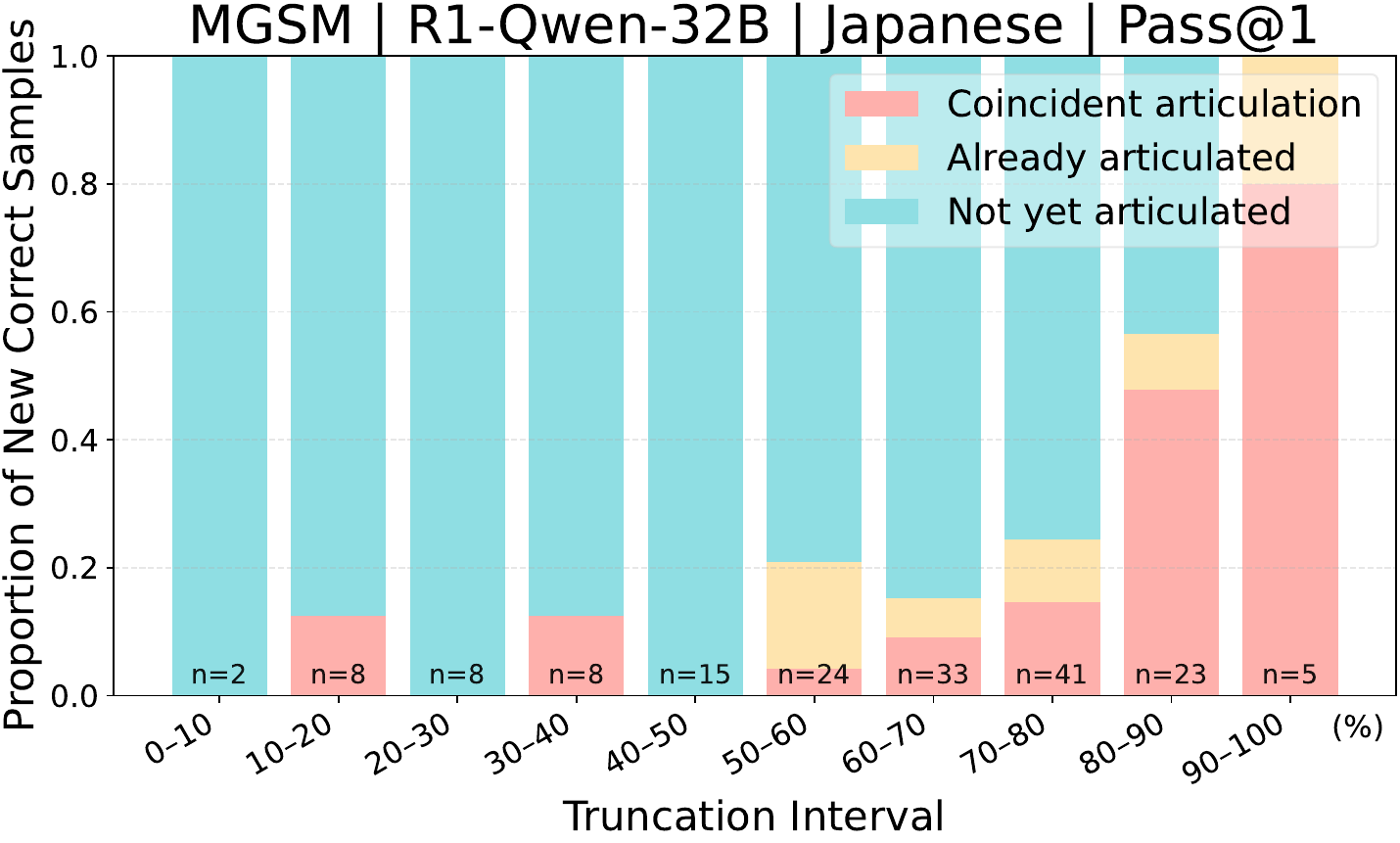}
    \includegraphics[width=0.24\textwidth]{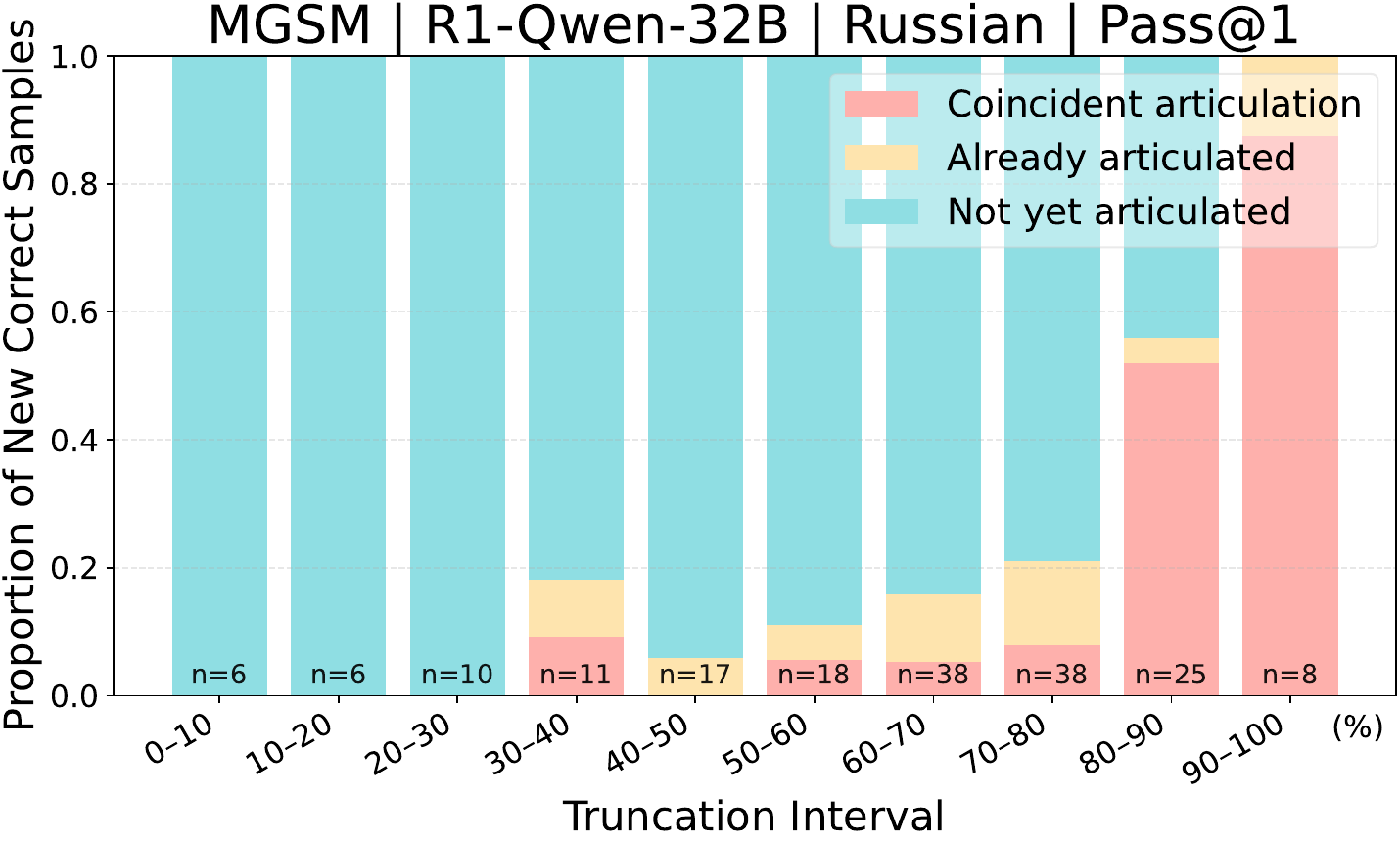}
    \includegraphics[width=0.24\textwidth]{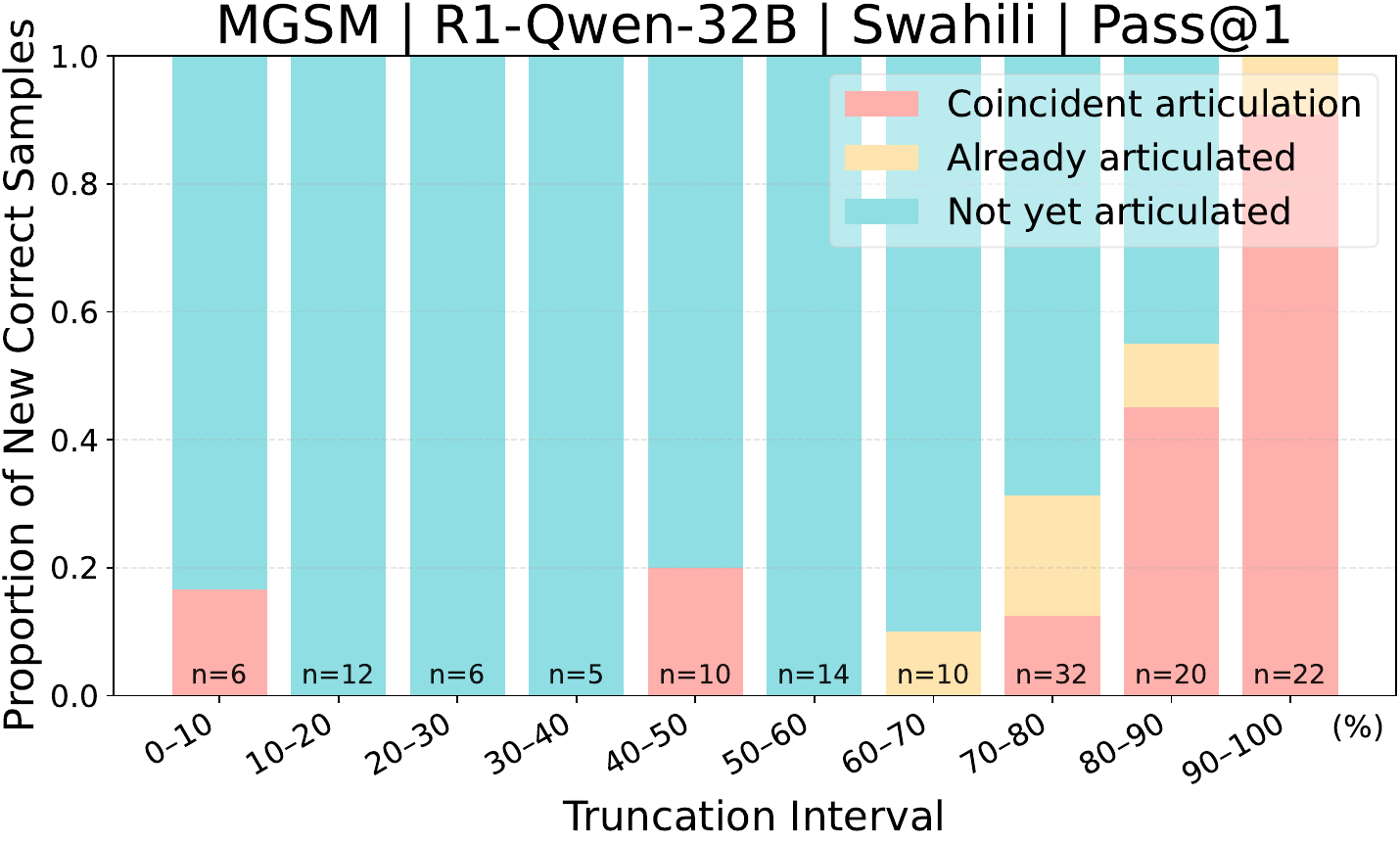}
    \includegraphics[width=0.24\textwidth]{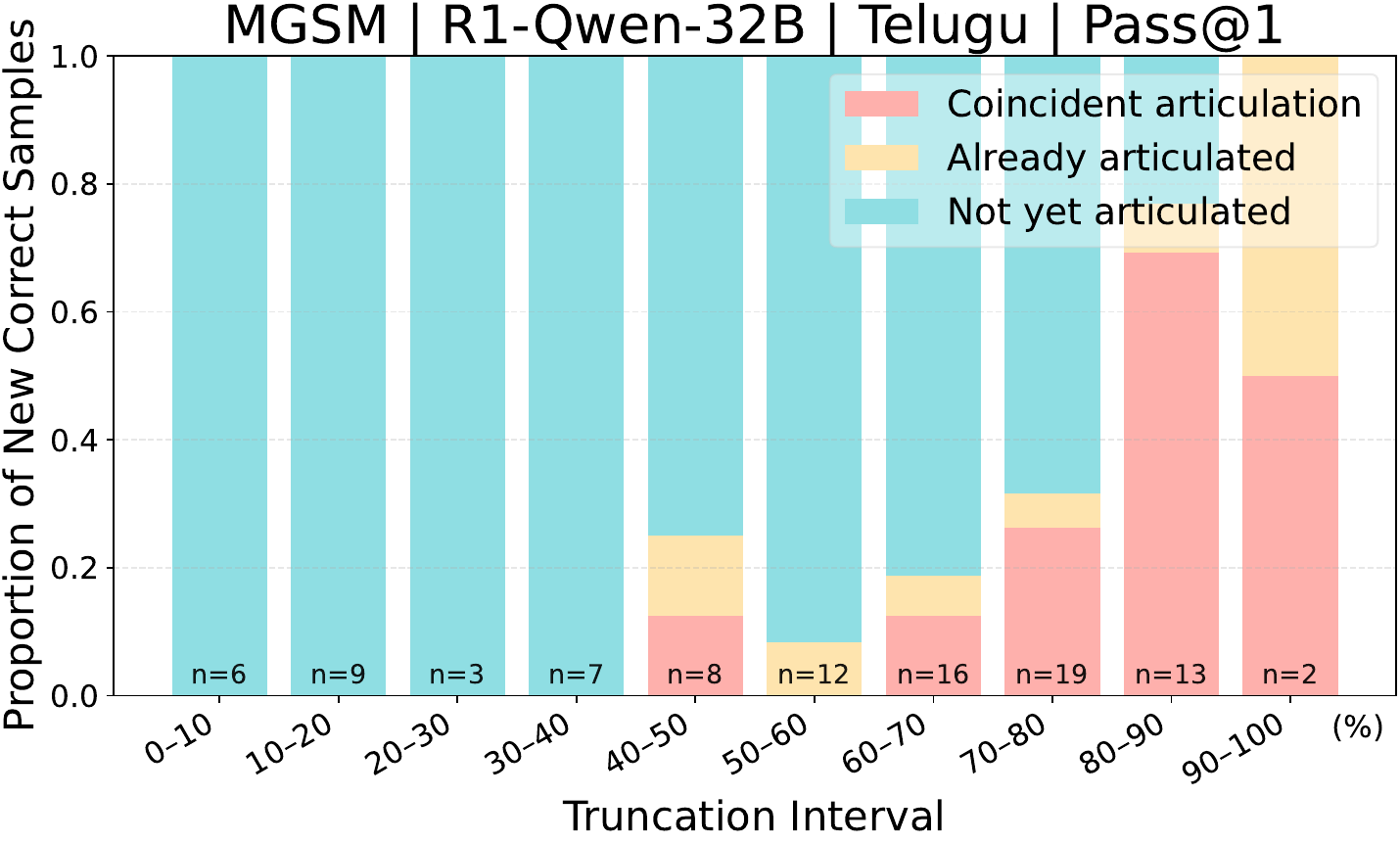}
    \includegraphics[width=0.24\textwidth]{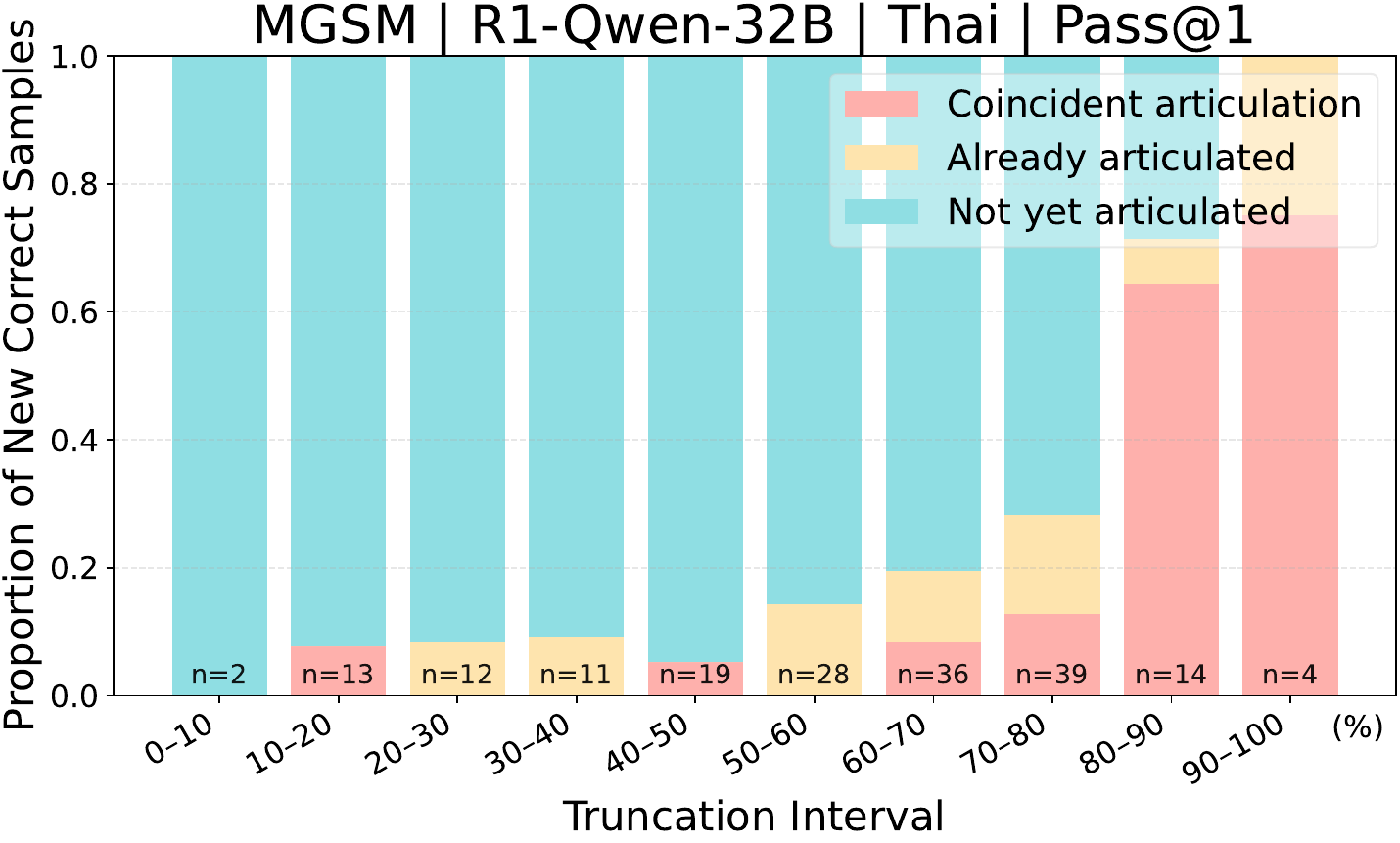}
    \includegraphics[width=0.24\textwidth]{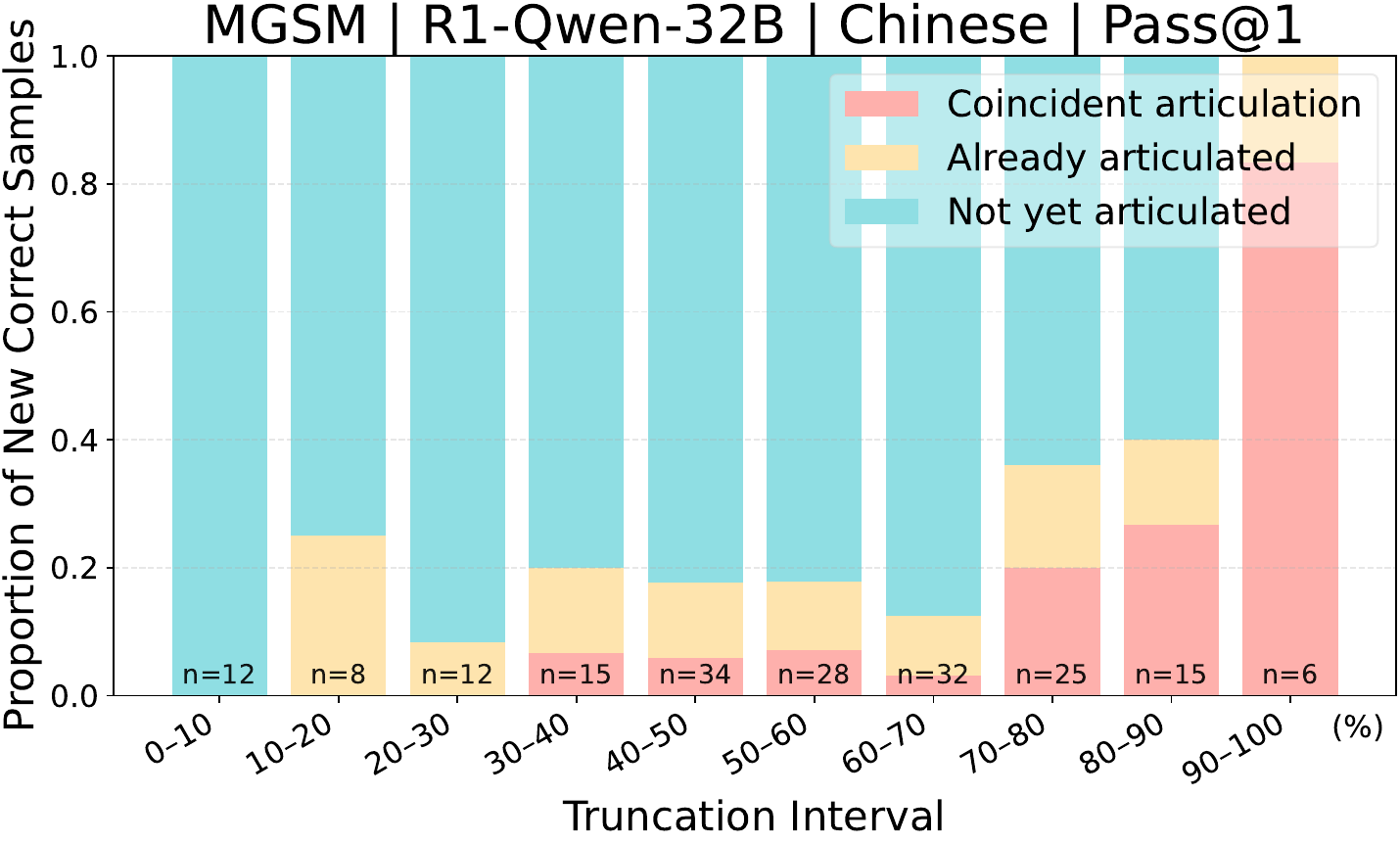}
    \caption{
    Causal decomposition of newly correct predictions across truncation intervals on \textbf{MGSM} with \textbf{R1-Qwen-32B}.
    Each bar partitions gains into three cases: (\textbf{i}) the gold answer is first articulated in the newly added reasoning steps,
    (\textbf{ii}) it was already articulated in earlier steps, or
    (\textbf{iii}) it has not yet appeared in the visible truncated trace.
    Early and intermediate gains are dominated by case (\textbf{iii}), indicating latent reasoning.
    }
    \label{fig:interval_32b_mgsm}
\end{figure*}

\begin{figure*}
    \centering
    \includegraphics[width=0.24\textwidth]{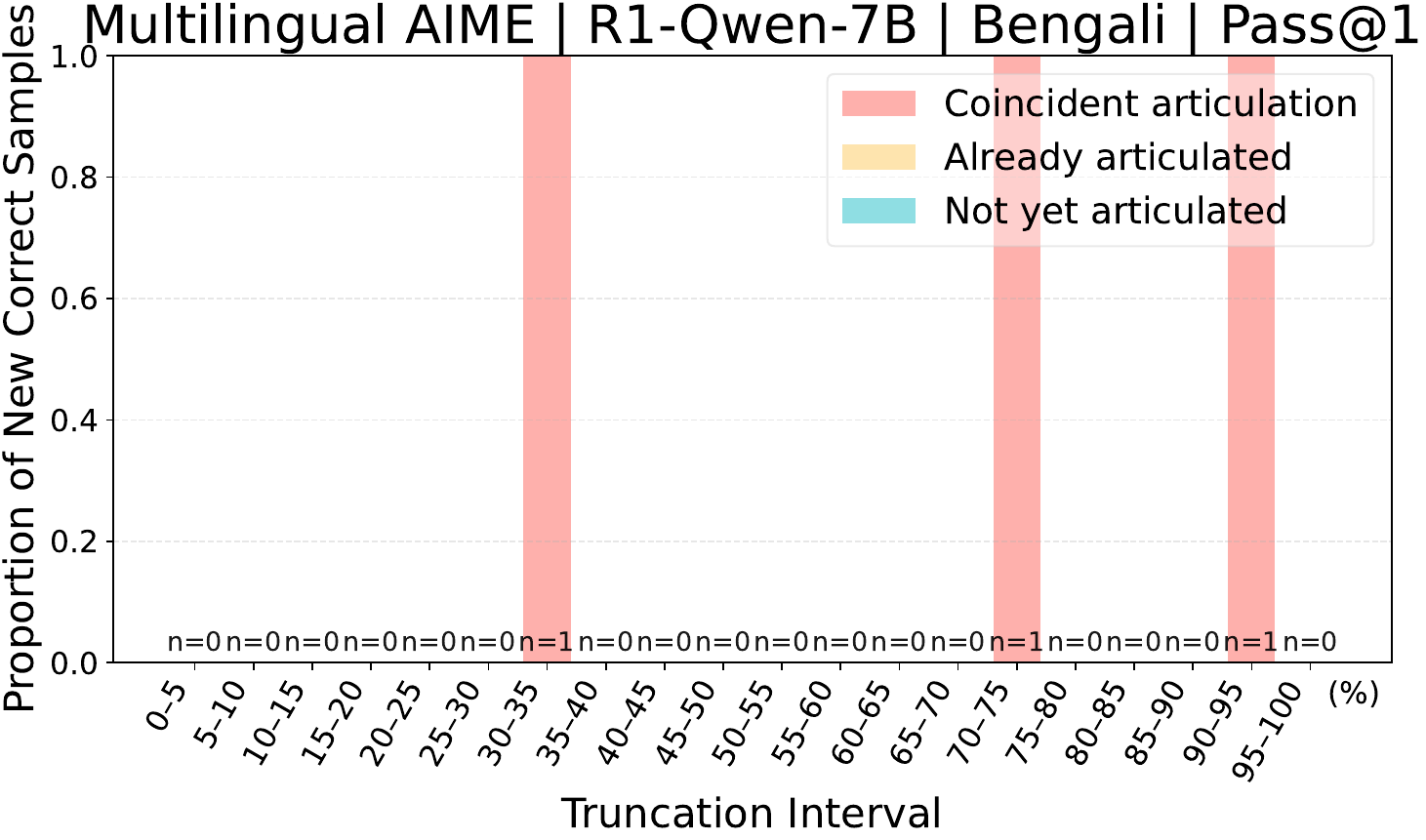}
    \includegraphics[width=0.24\textwidth]{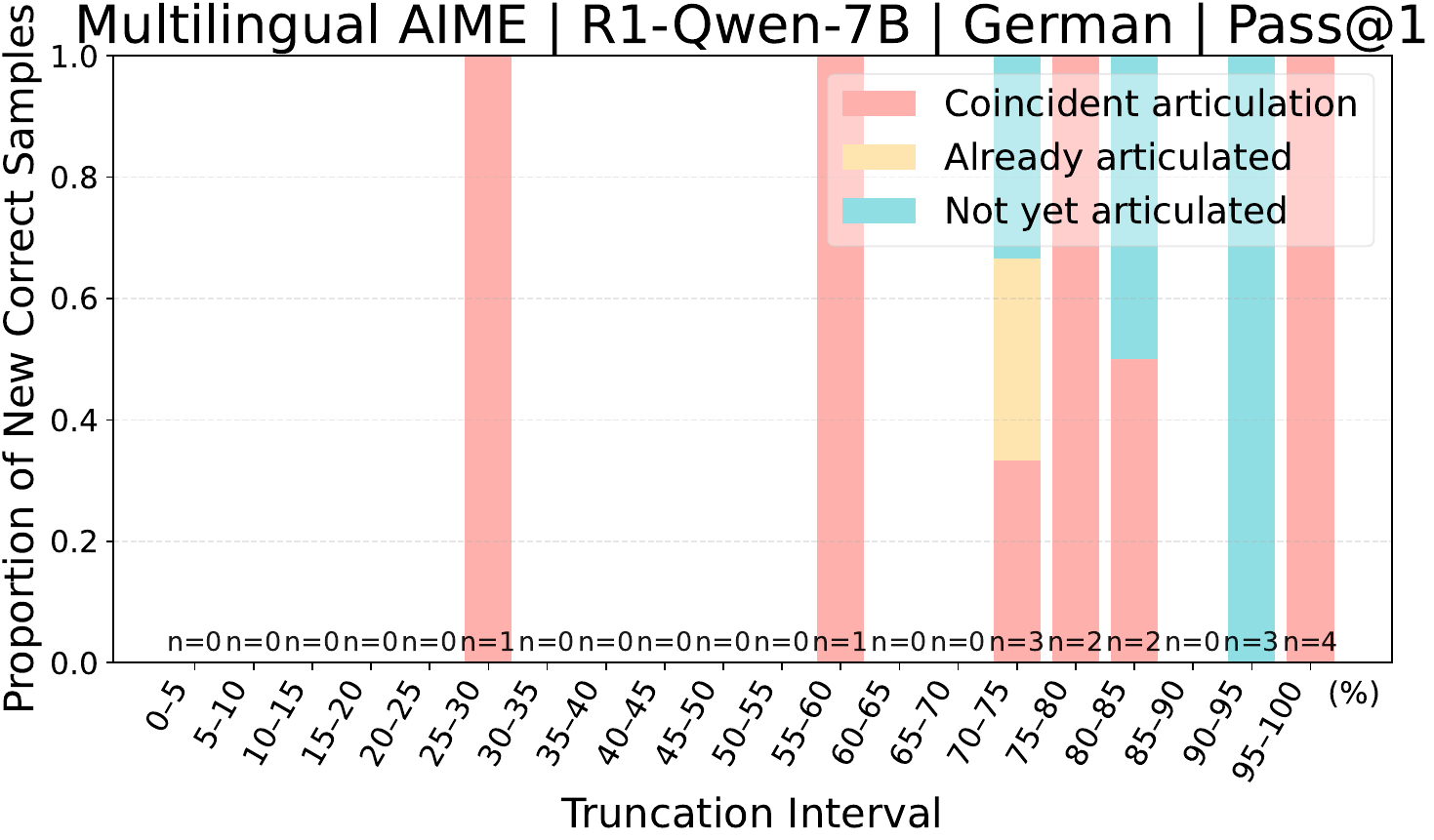}
    \includegraphics[width=0.24\textwidth]{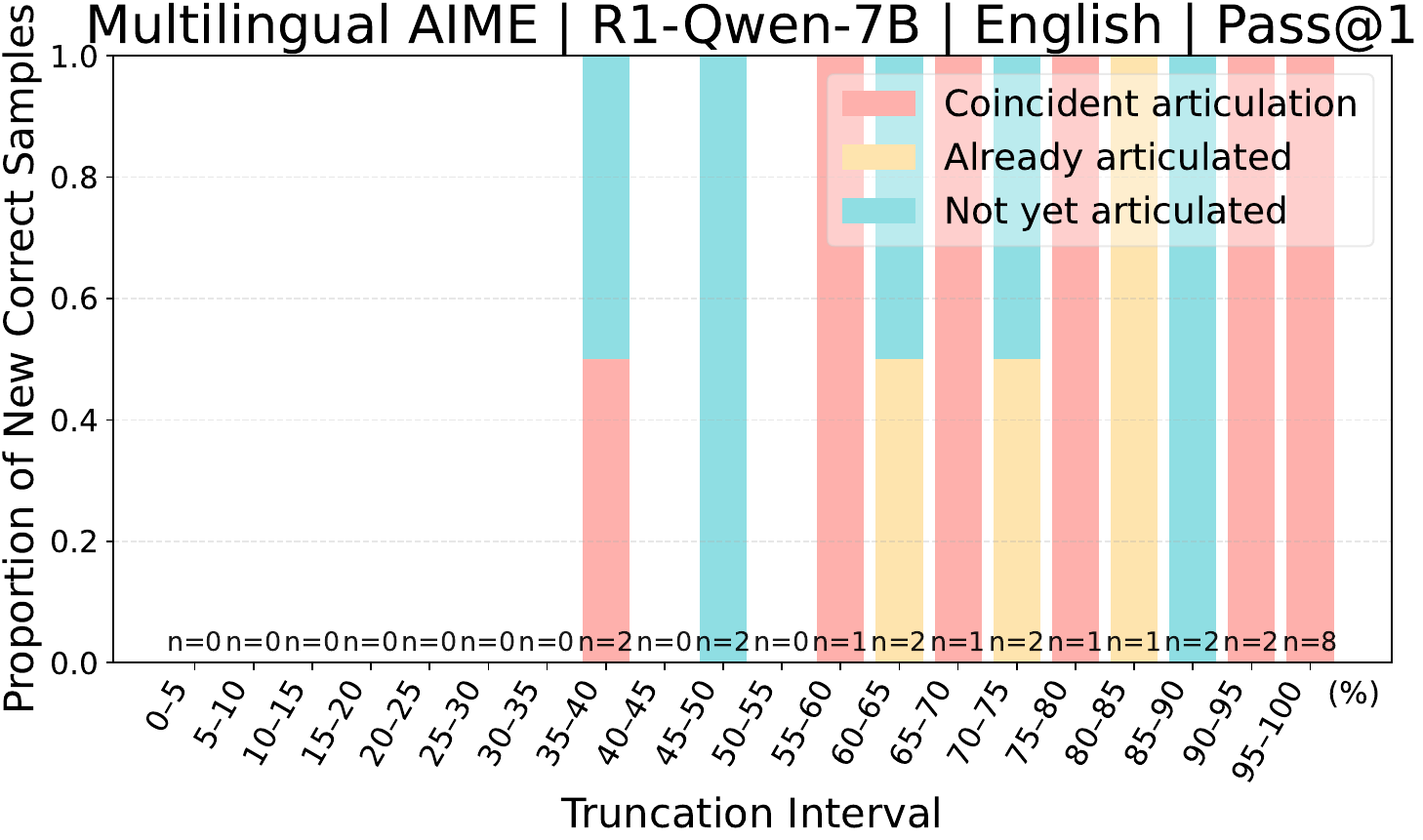}
    \includegraphics[width=0.24\textwidth]{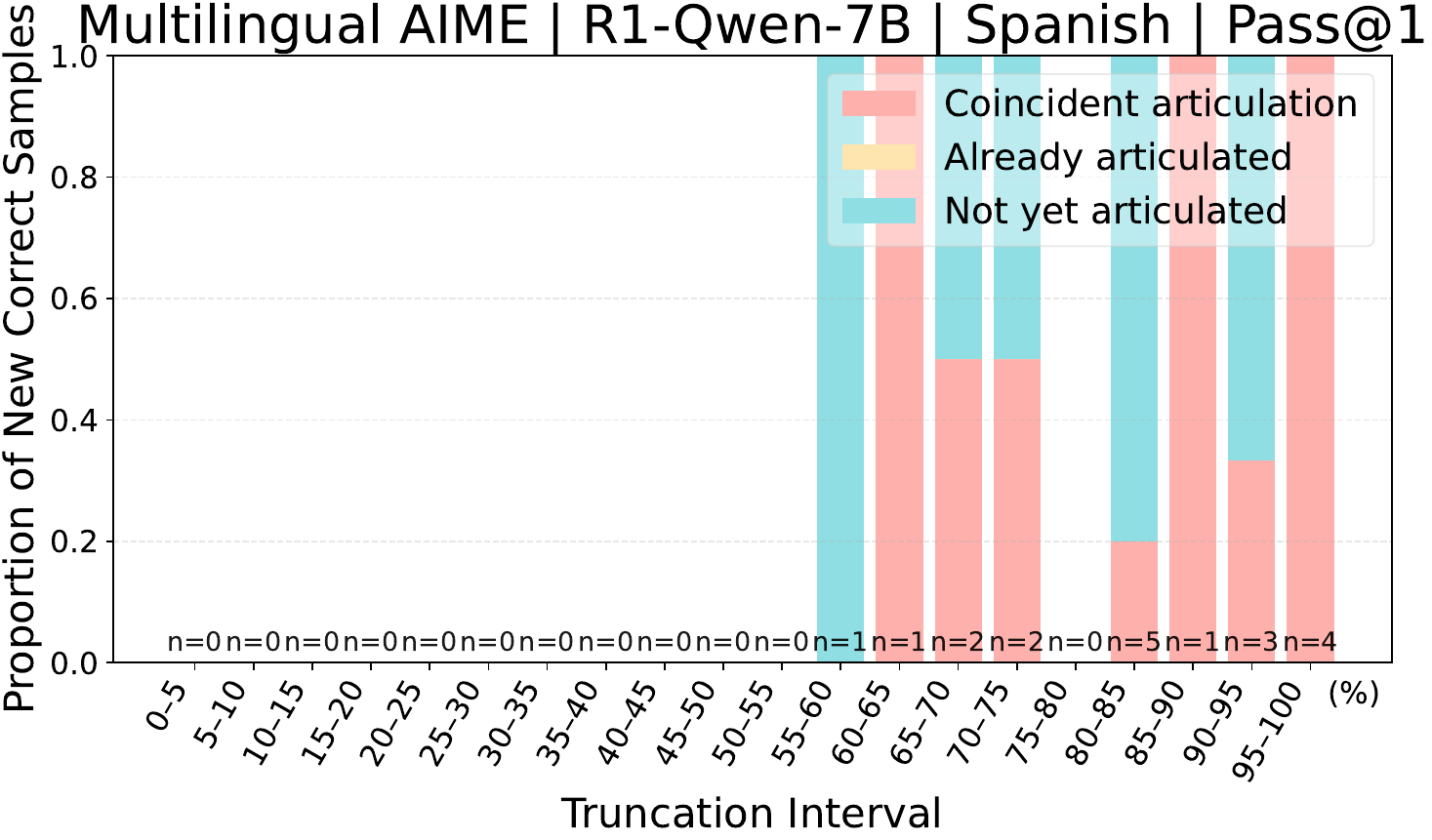}
    \includegraphics[width=0.24\textwidth]{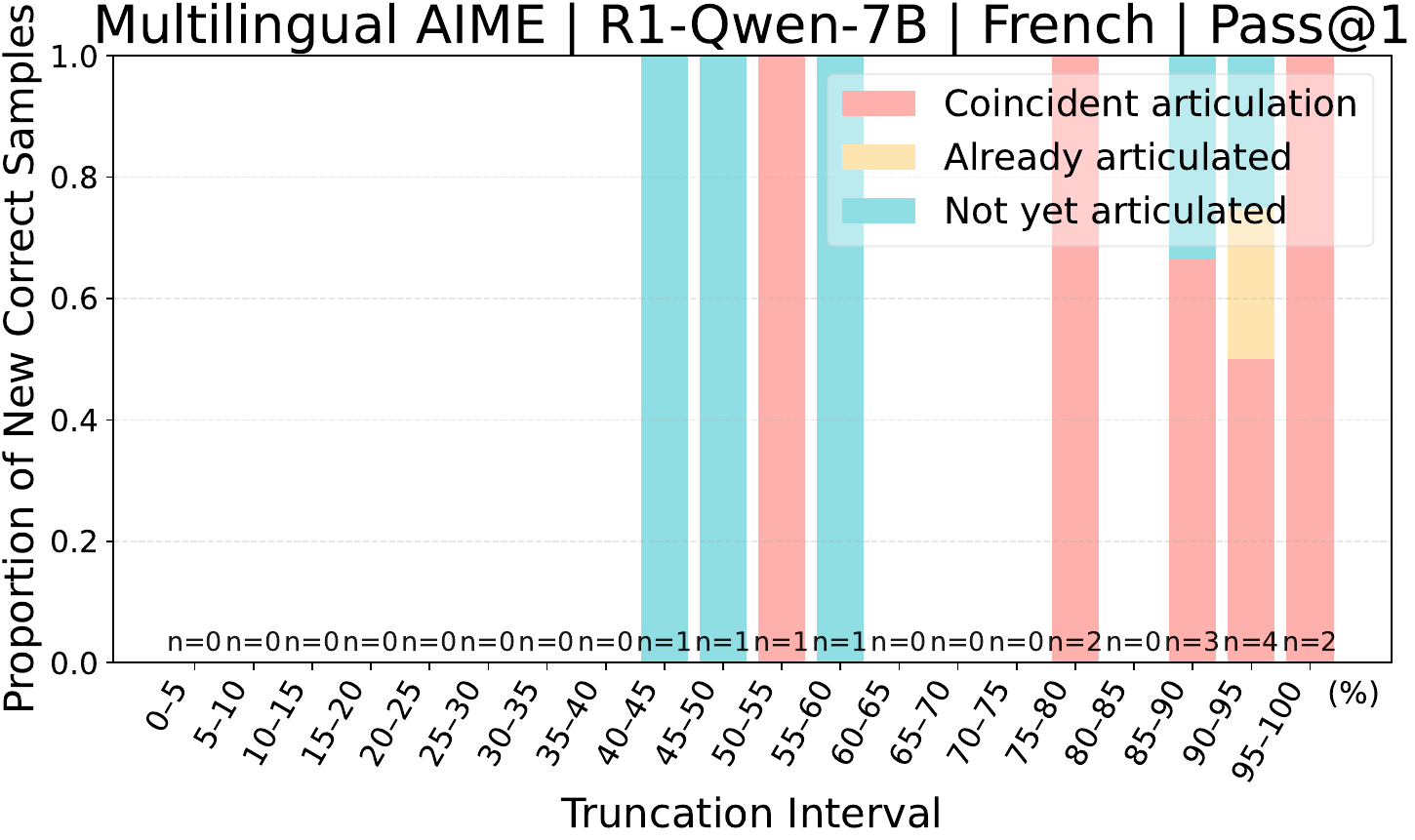}
    \includegraphics[width=0.24\textwidth]{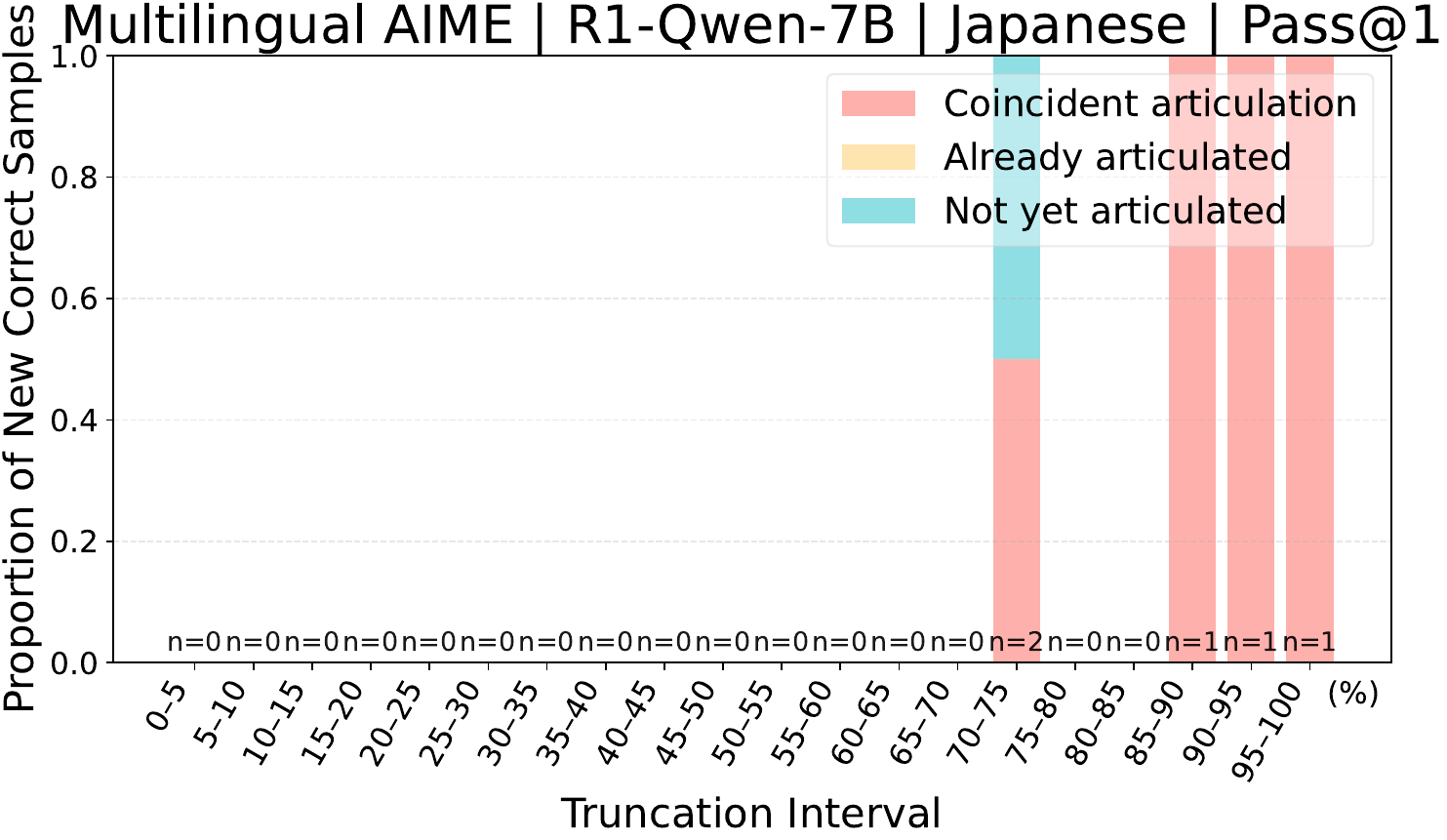}
    \includegraphics[width=0.24\textwidth]{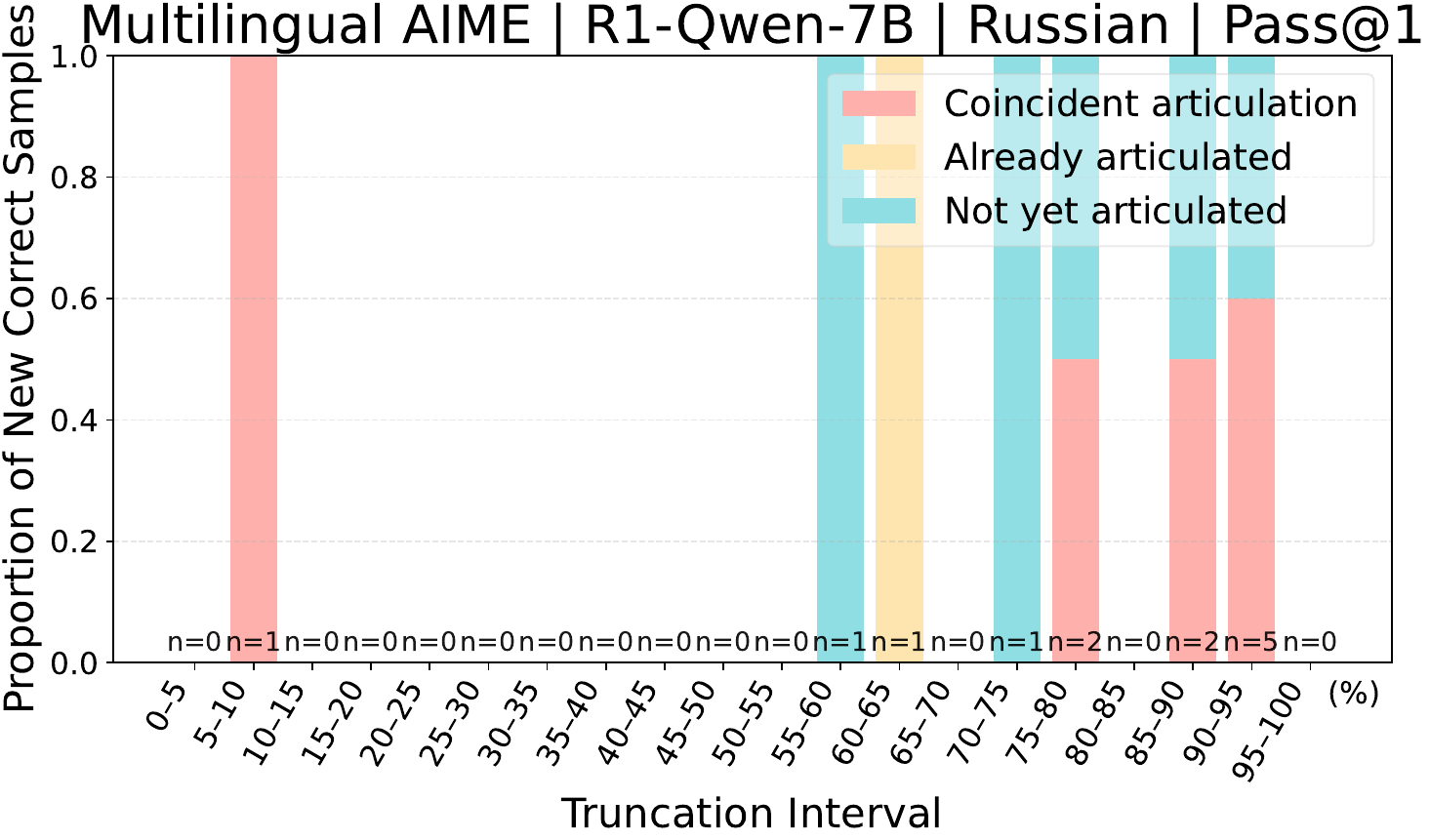}
    \includegraphics[width=0.24\textwidth]{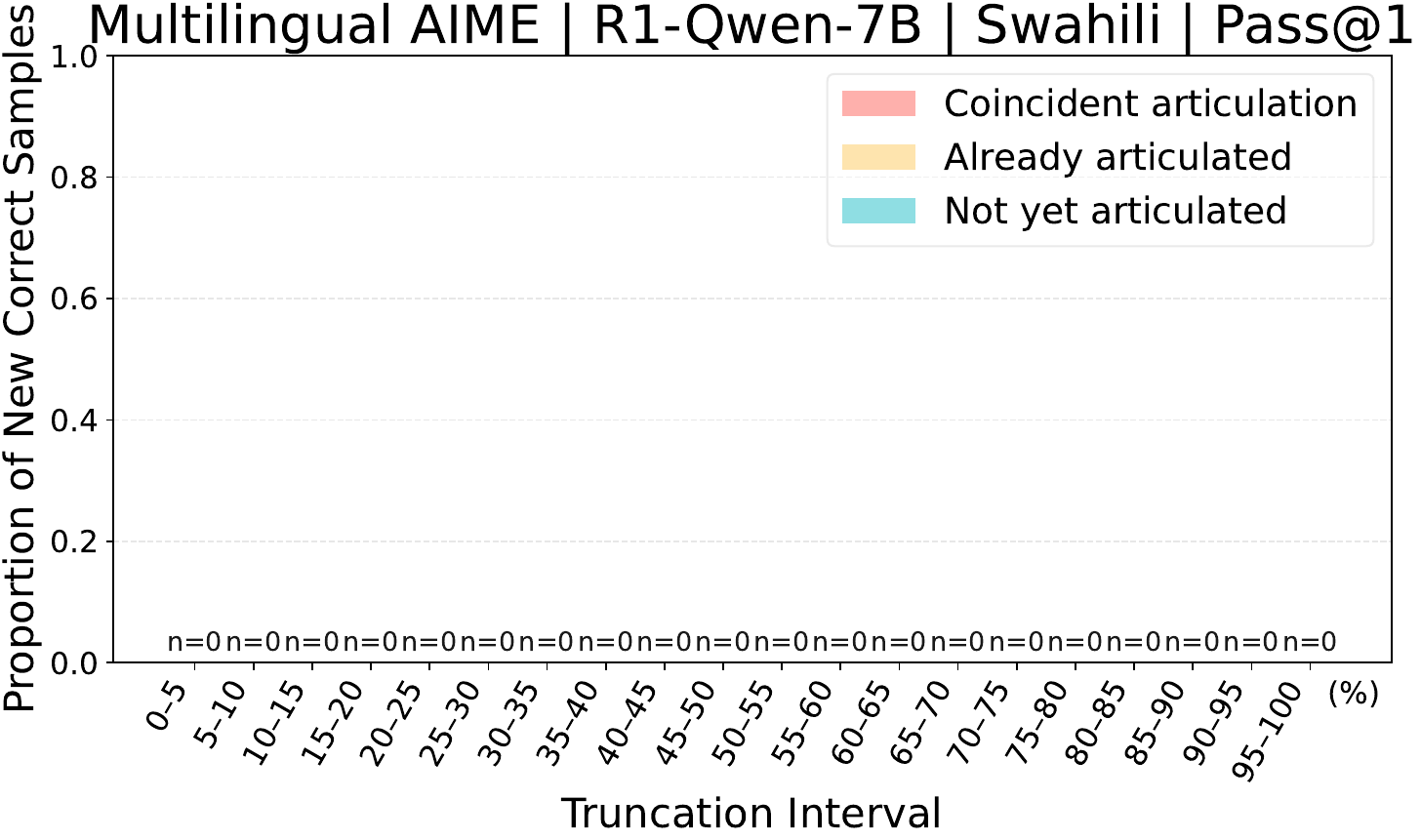}
    \includegraphics[width=0.24\textwidth]{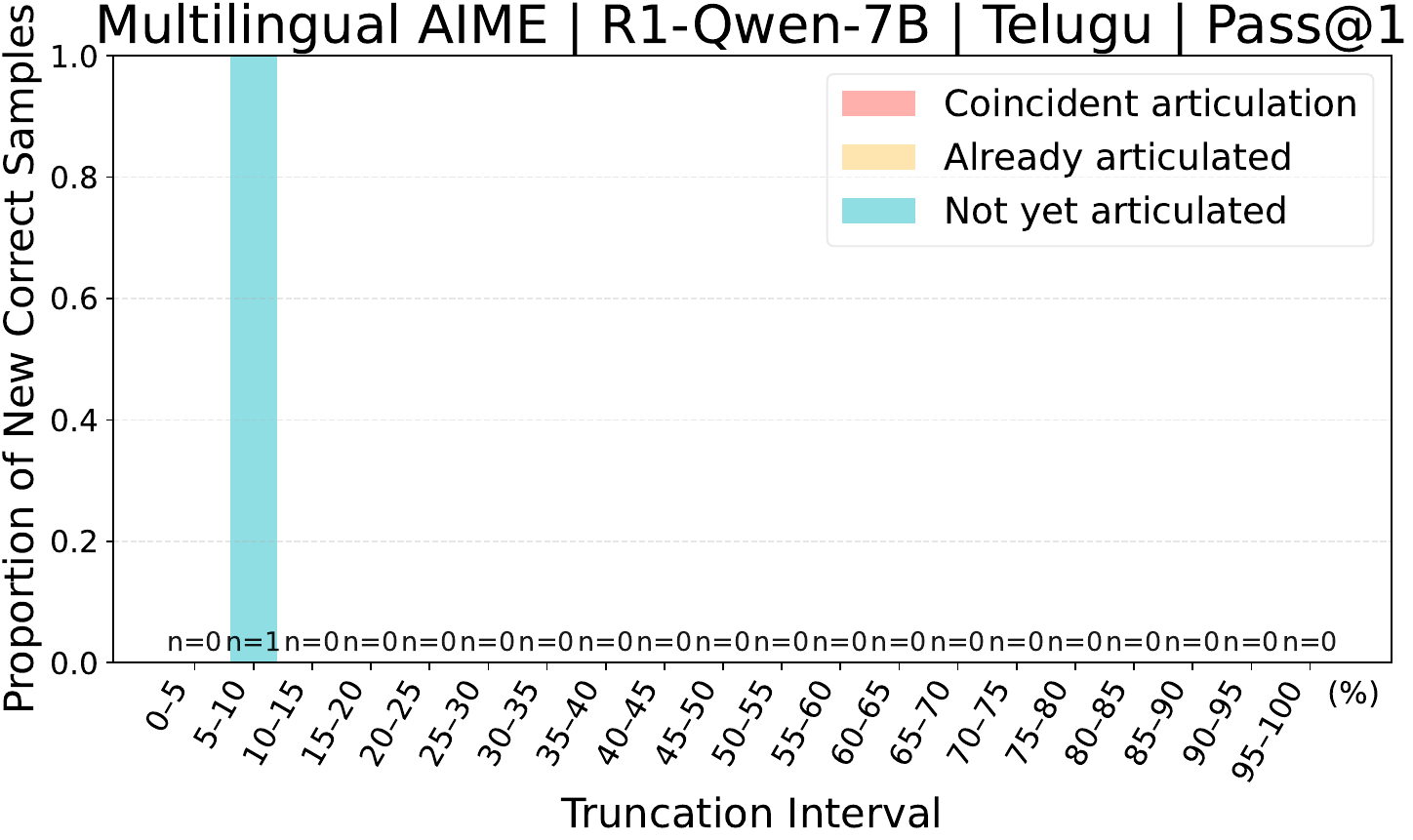}
    \includegraphics[width=0.24\textwidth]{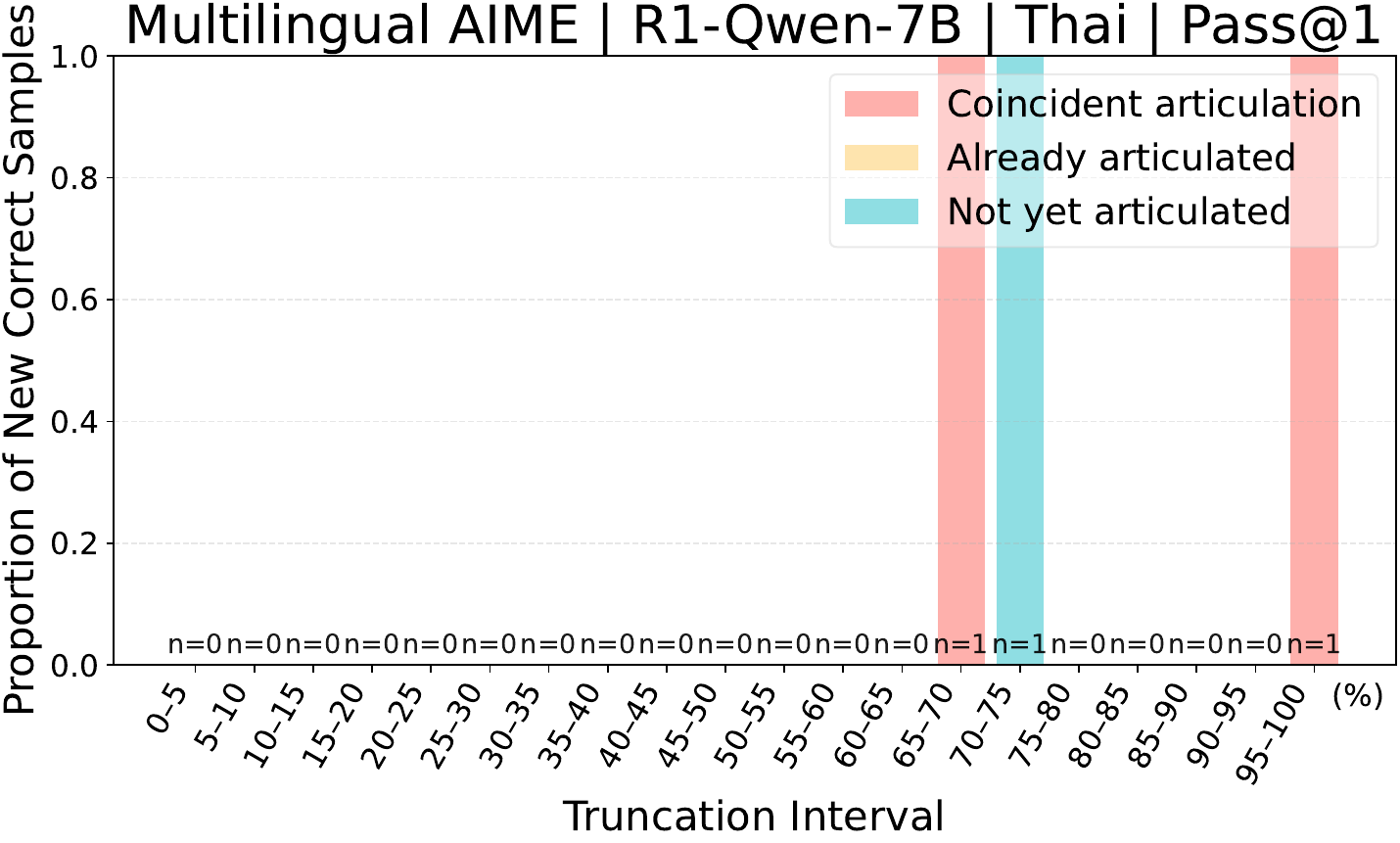}
    \includegraphics[width=0.24\textwidth]{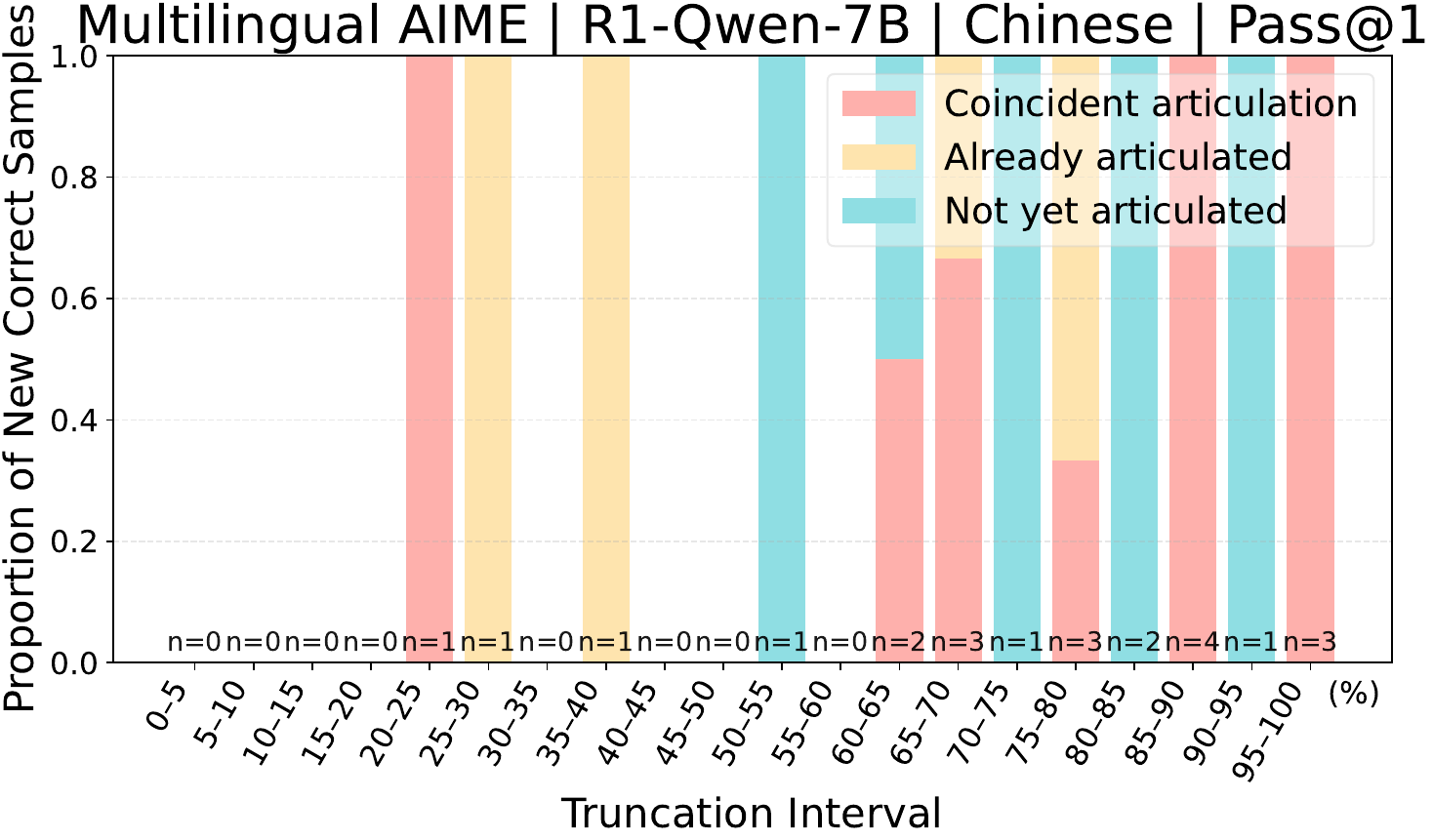}
    \caption{
    Causal decomposition of newly correct predictions across truncation intervals on \textbf{Multilingual AIME} with \textbf{R1-Qwen-7B}.
    Each bar partitions gains into three cases: (\textbf{i}) the gold answer is first articulated in the newly added reasoning steps,
    (\textbf{ii}) it was already articulated in earlier steps, or
    (\textbf{iii}) it has not yet appeared in the visible truncated trace.
    Compared to MGSM, gains are sparser and less dominated by latent reasoning.
    }
    \label{fig:interval_7b_aime}
\end{figure*}

\begin{figure*}
    \centering
    \includegraphics[width=0.24\textwidth]{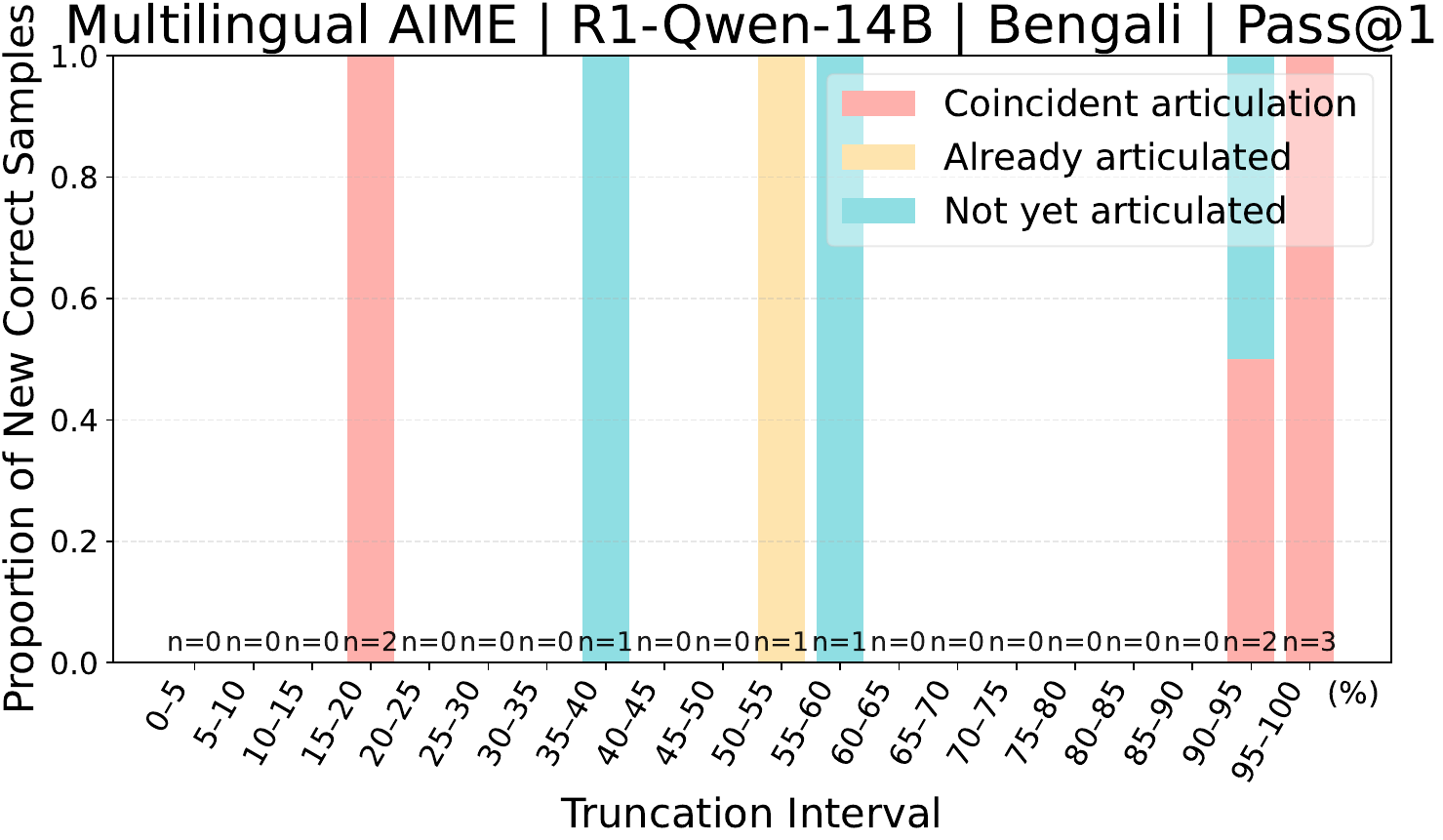}
    \includegraphics[width=0.24\textwidth]{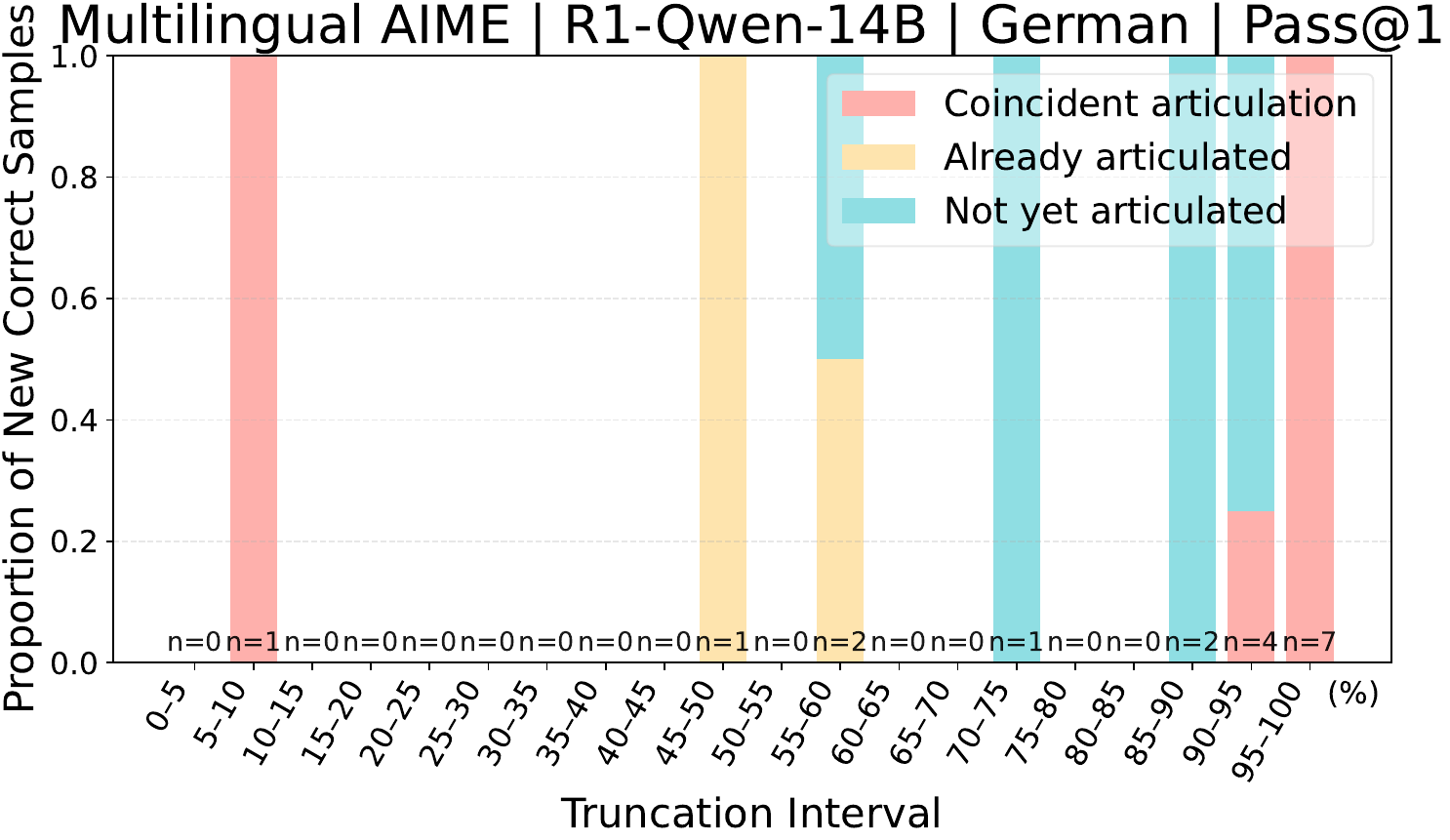}
    \includegraphics[width=0.24\textwidth]{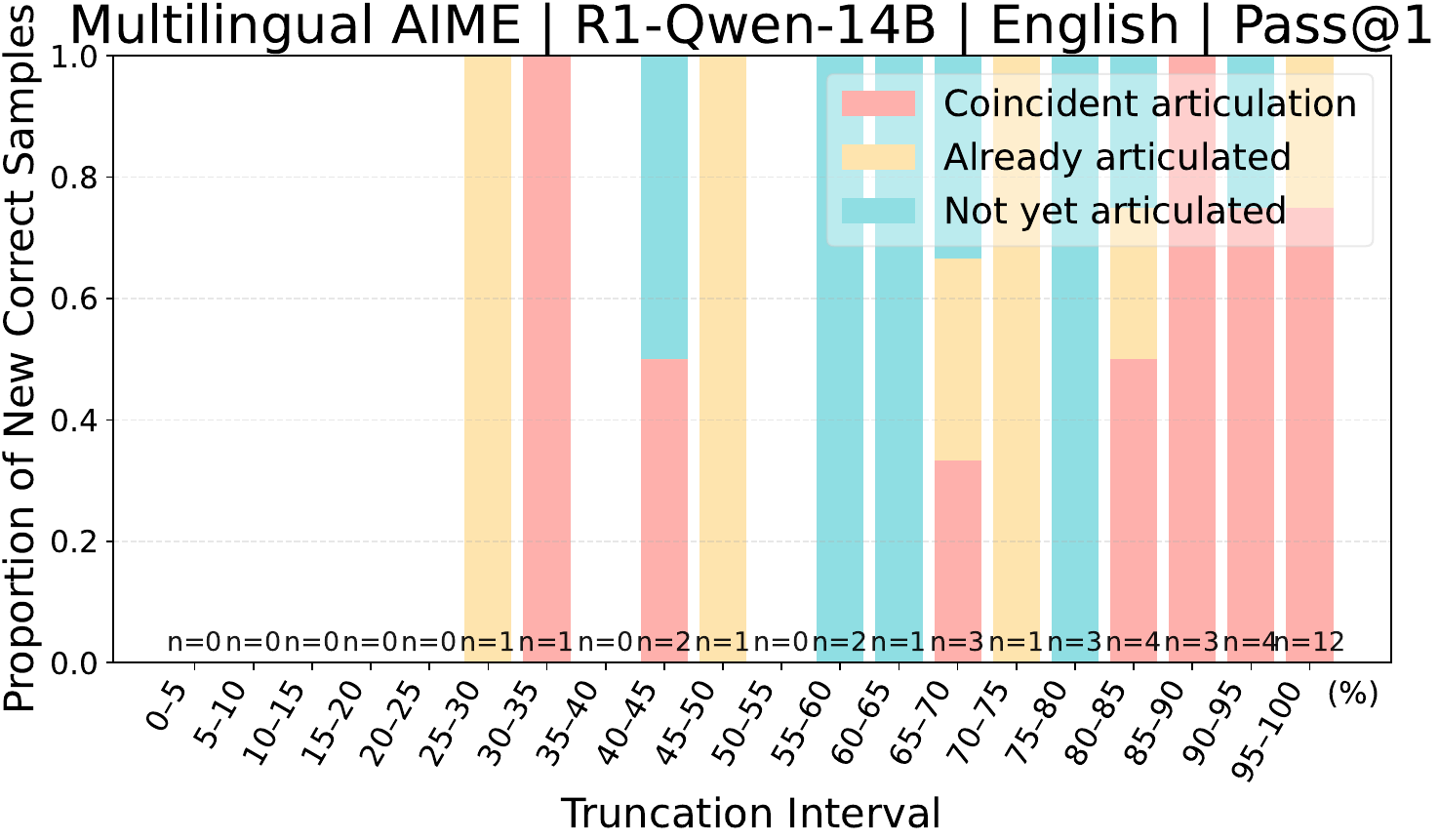}
    \includegraphics[width=0.24\textwidth]{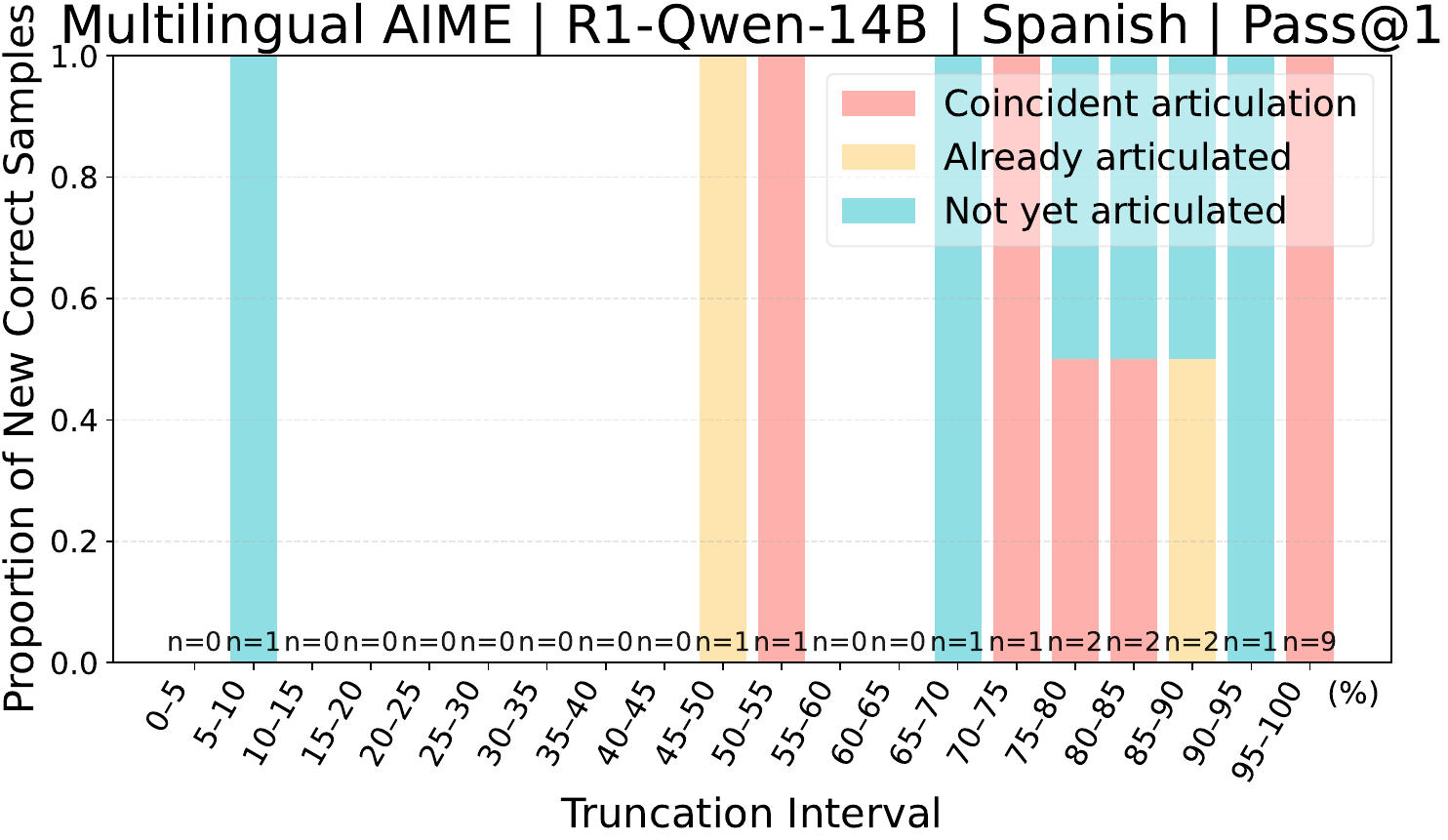}
    \includegraphics[width=0.24\textwidth]{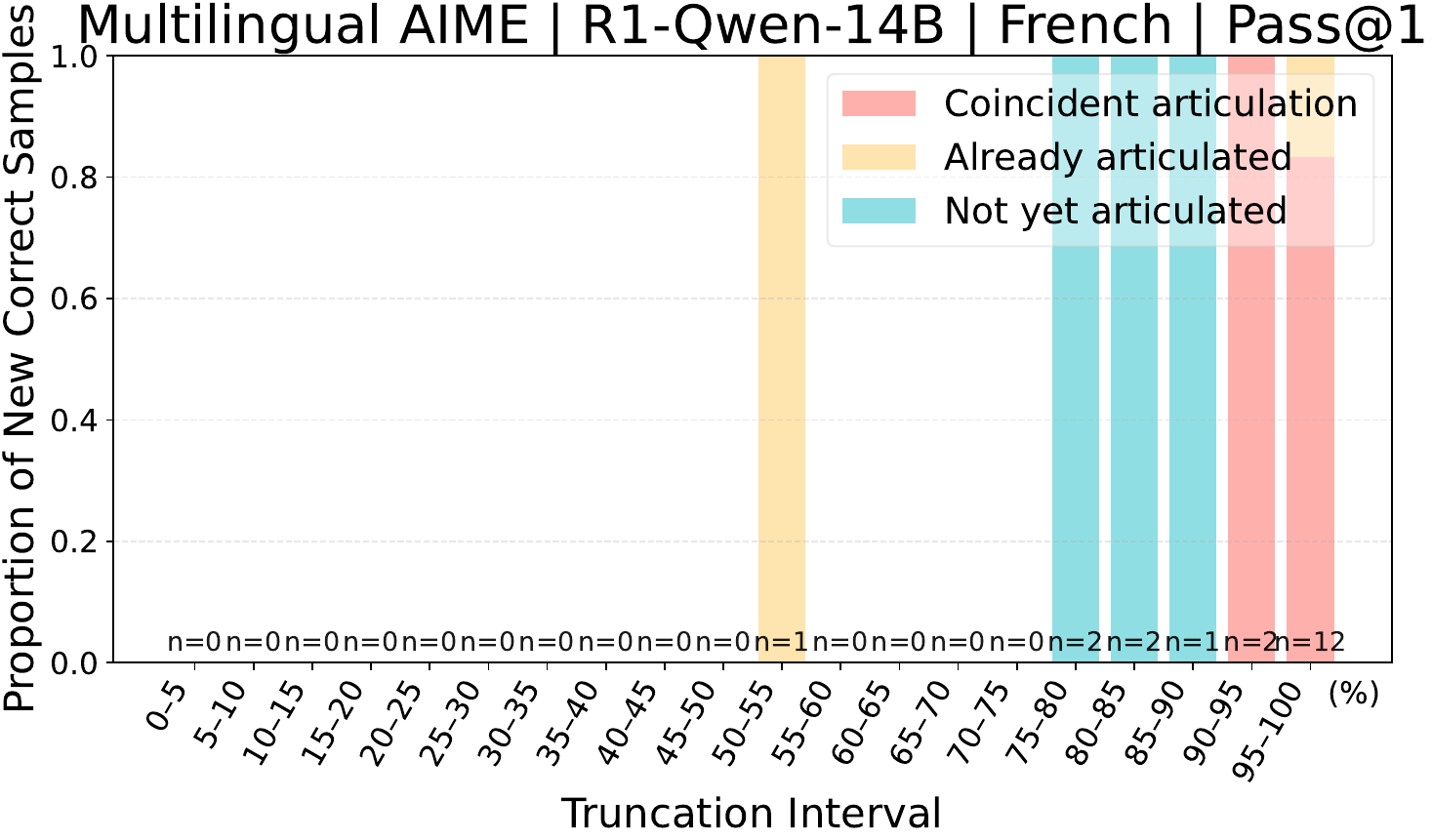}
    \includegraphics[width=0.24\textwidth]{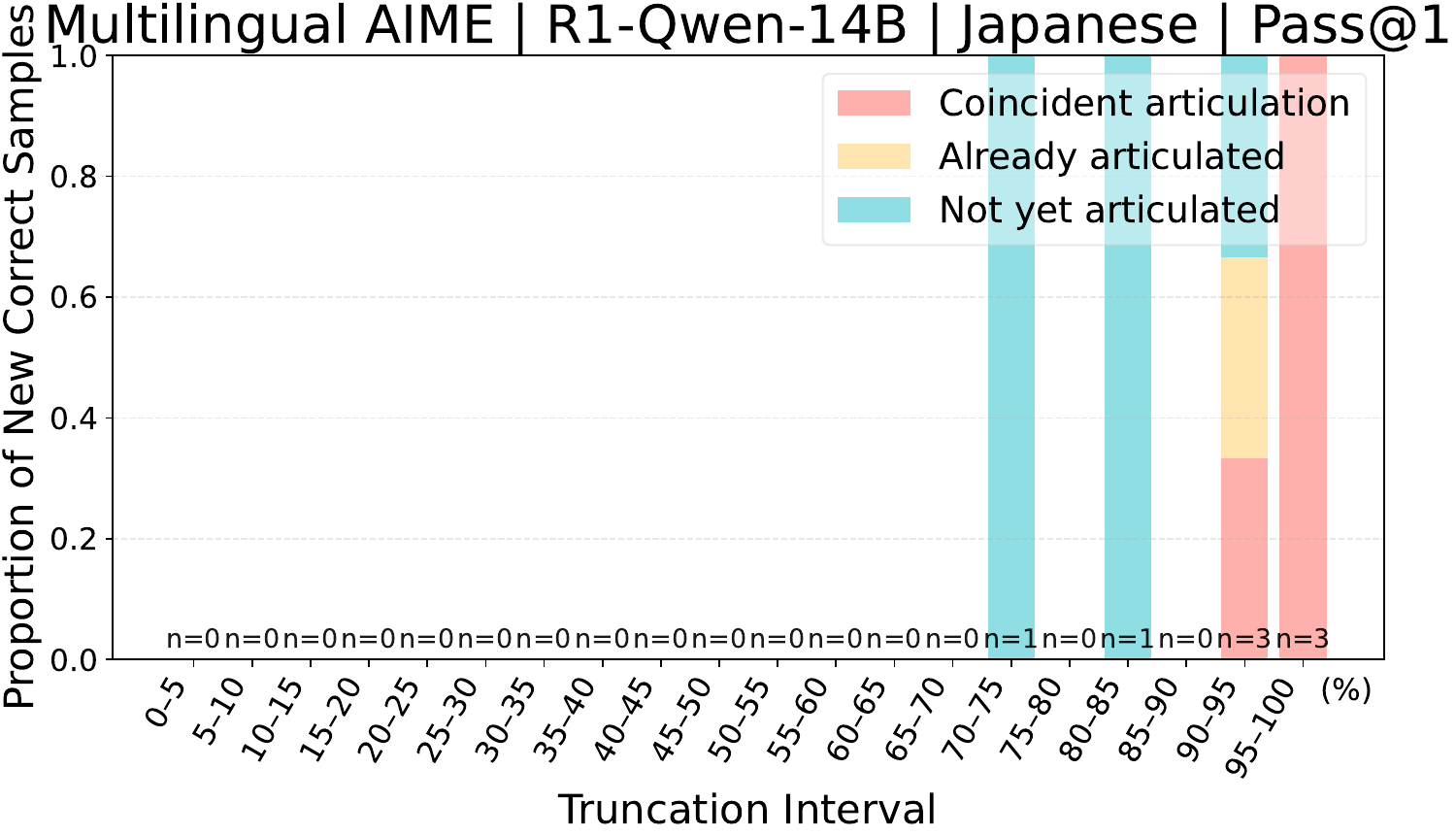}
    \includegraphics[width=0.24\textwidth]{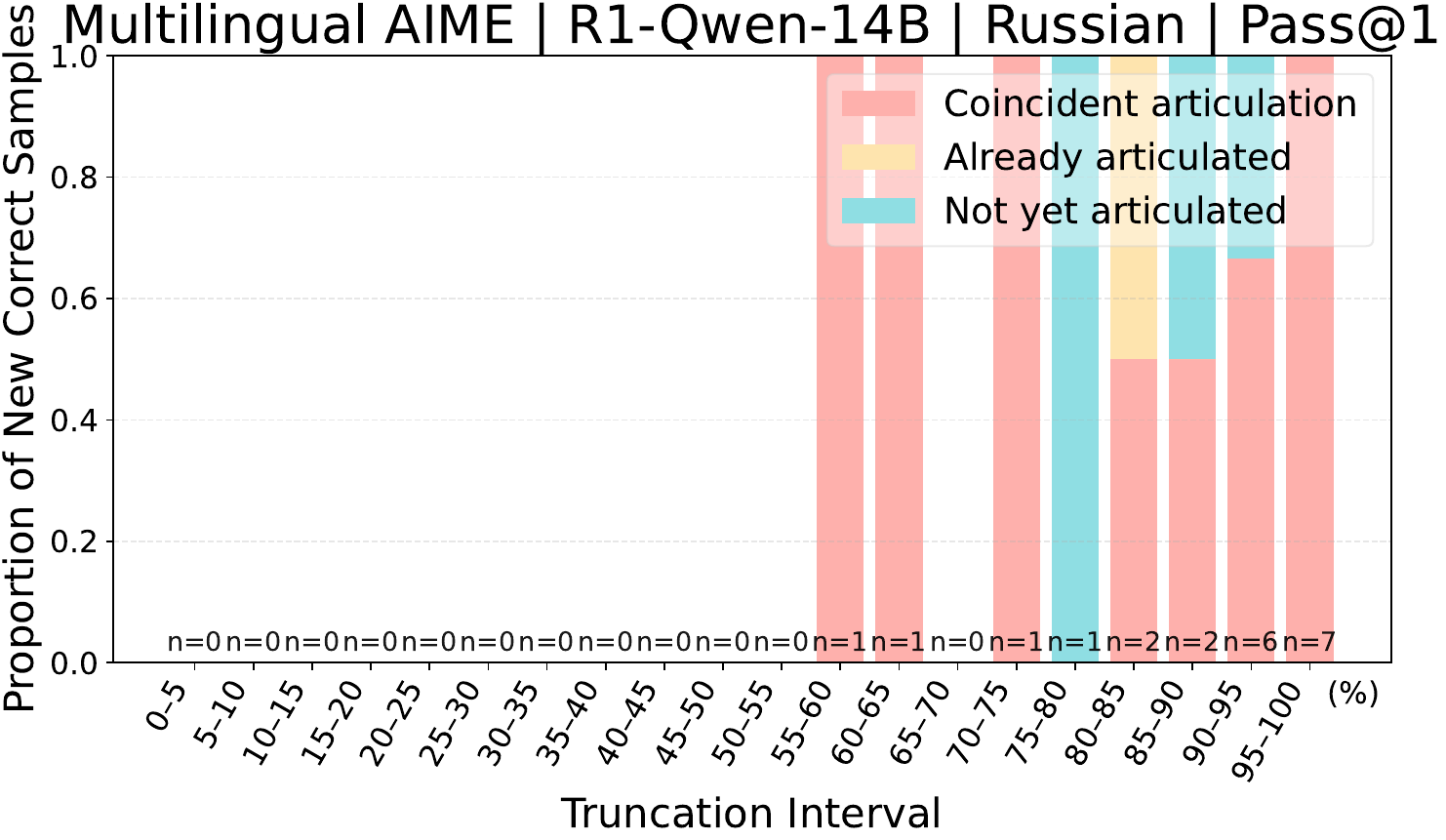}
    \includegraphics[width=0.24\textwidth]{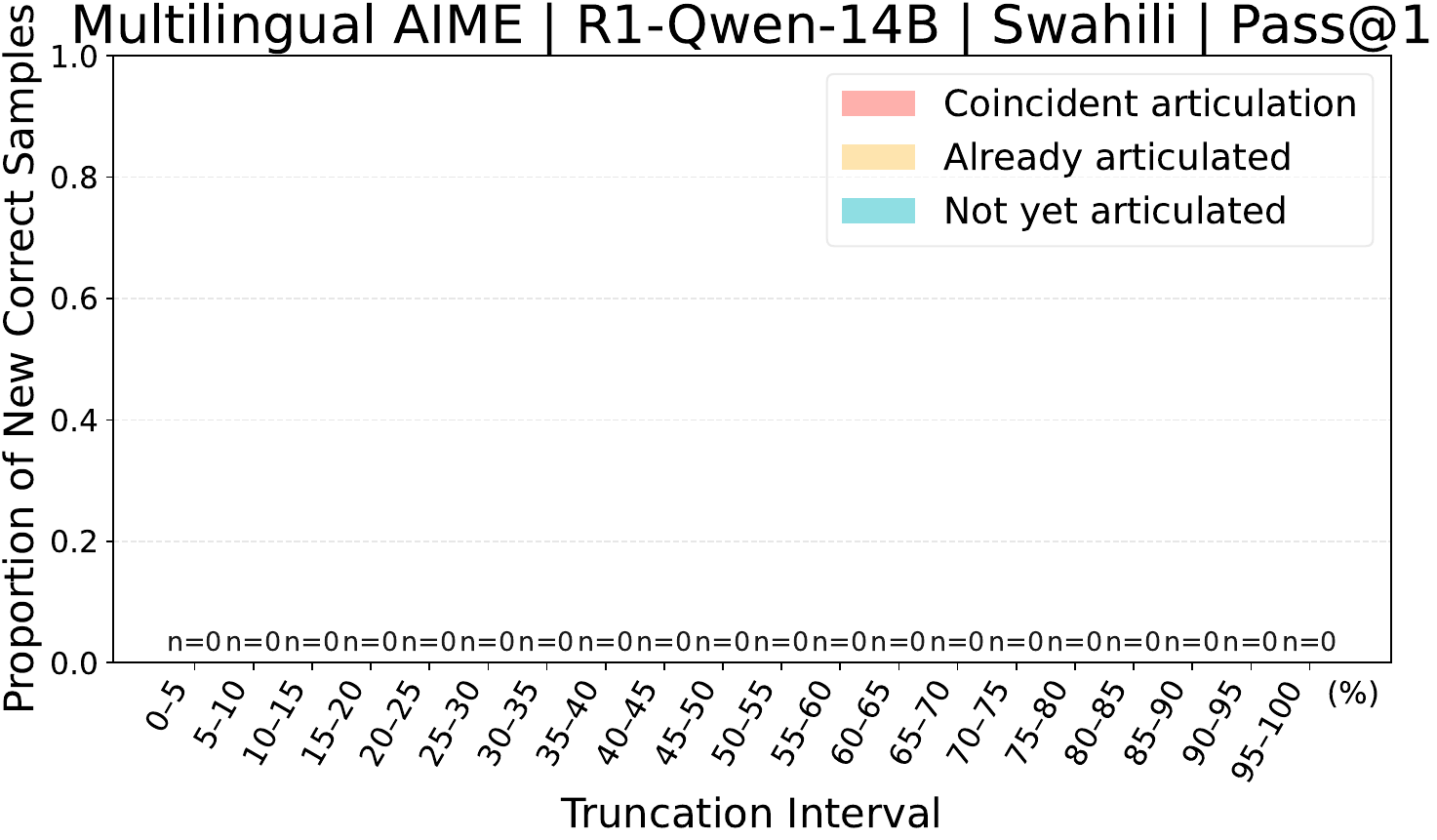}
    \includegraphics[width=0.24\textwidth]{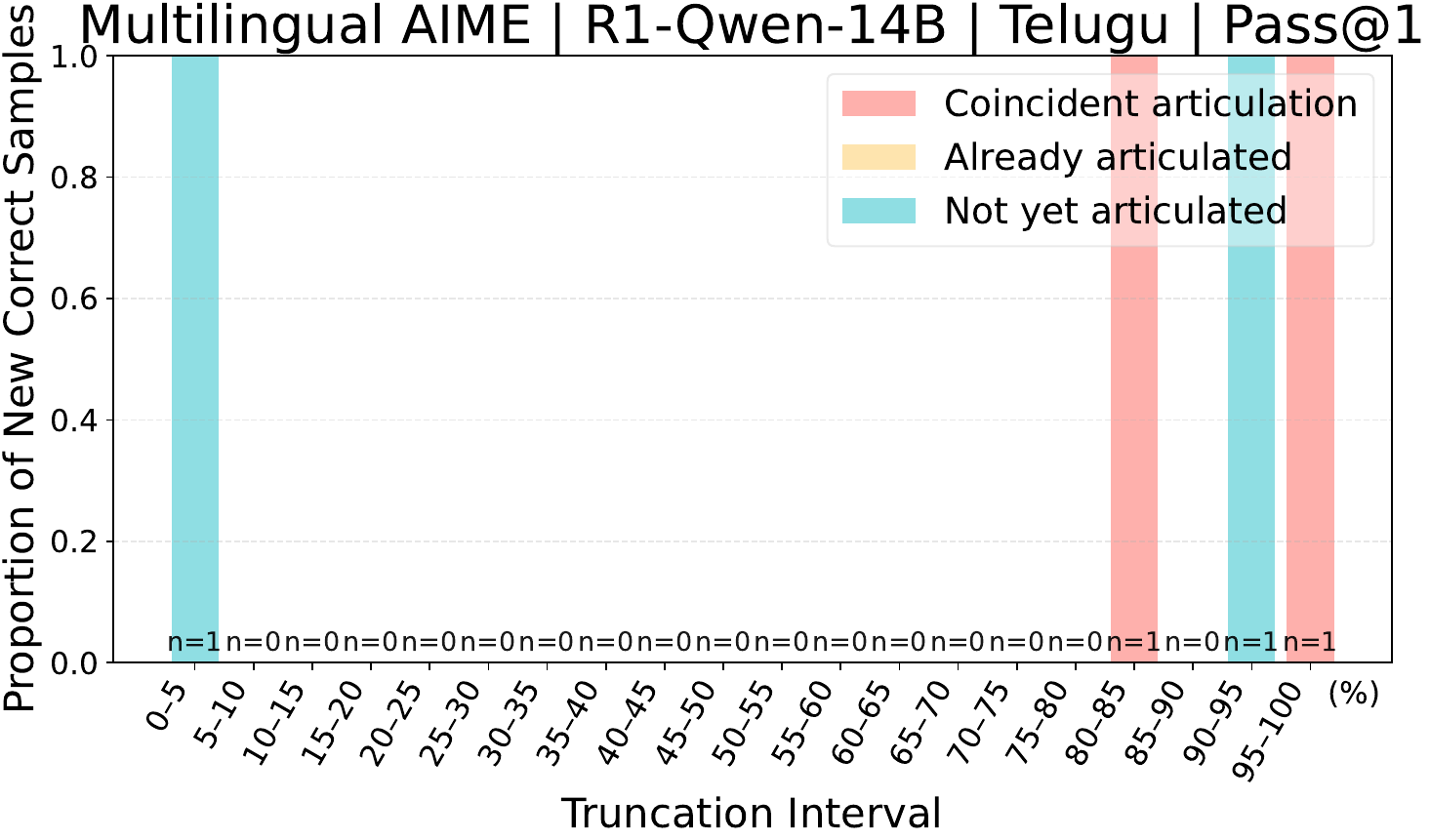}
    \includegraphics[width=0.24\textwidth]{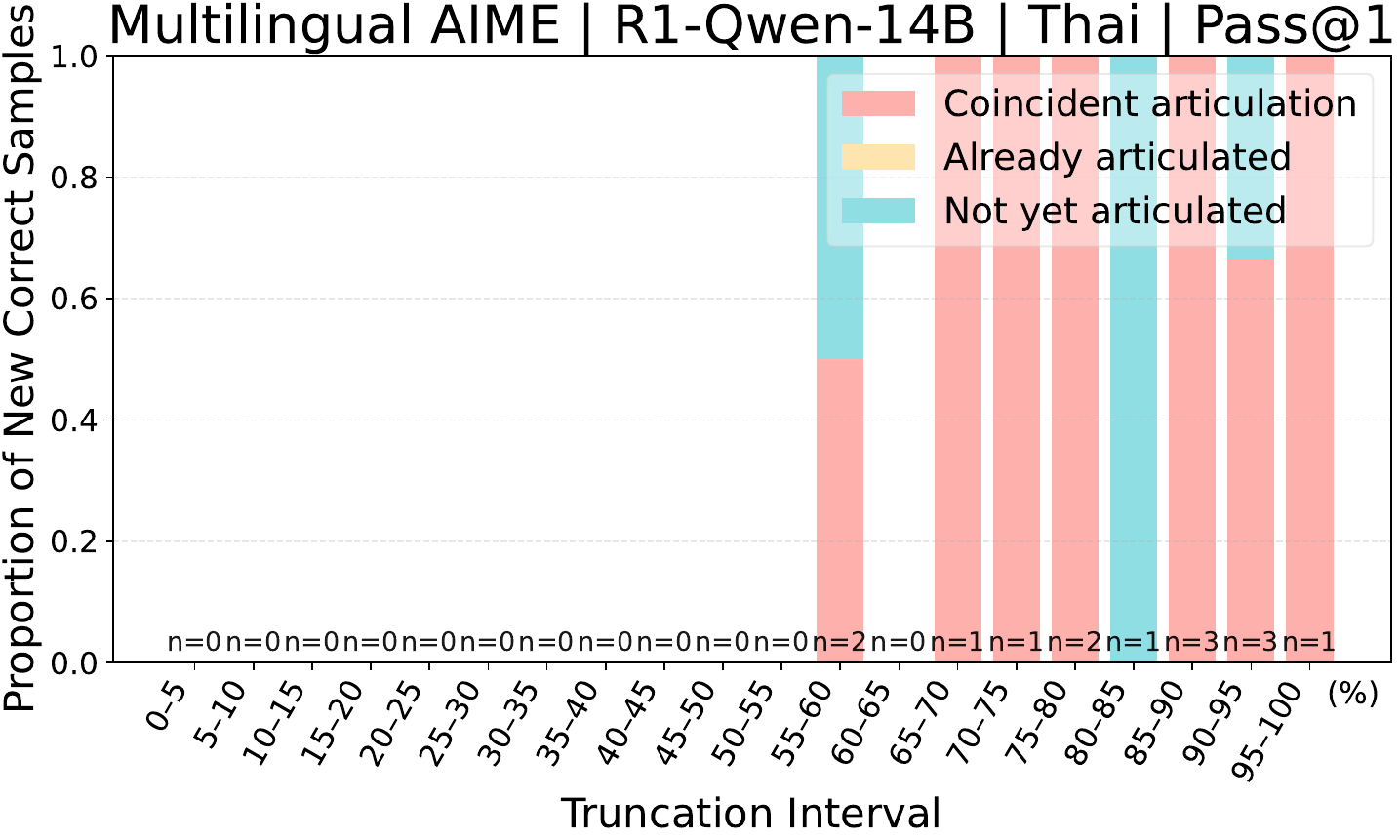}
    \includegraphics[width=0.24\textwidth]{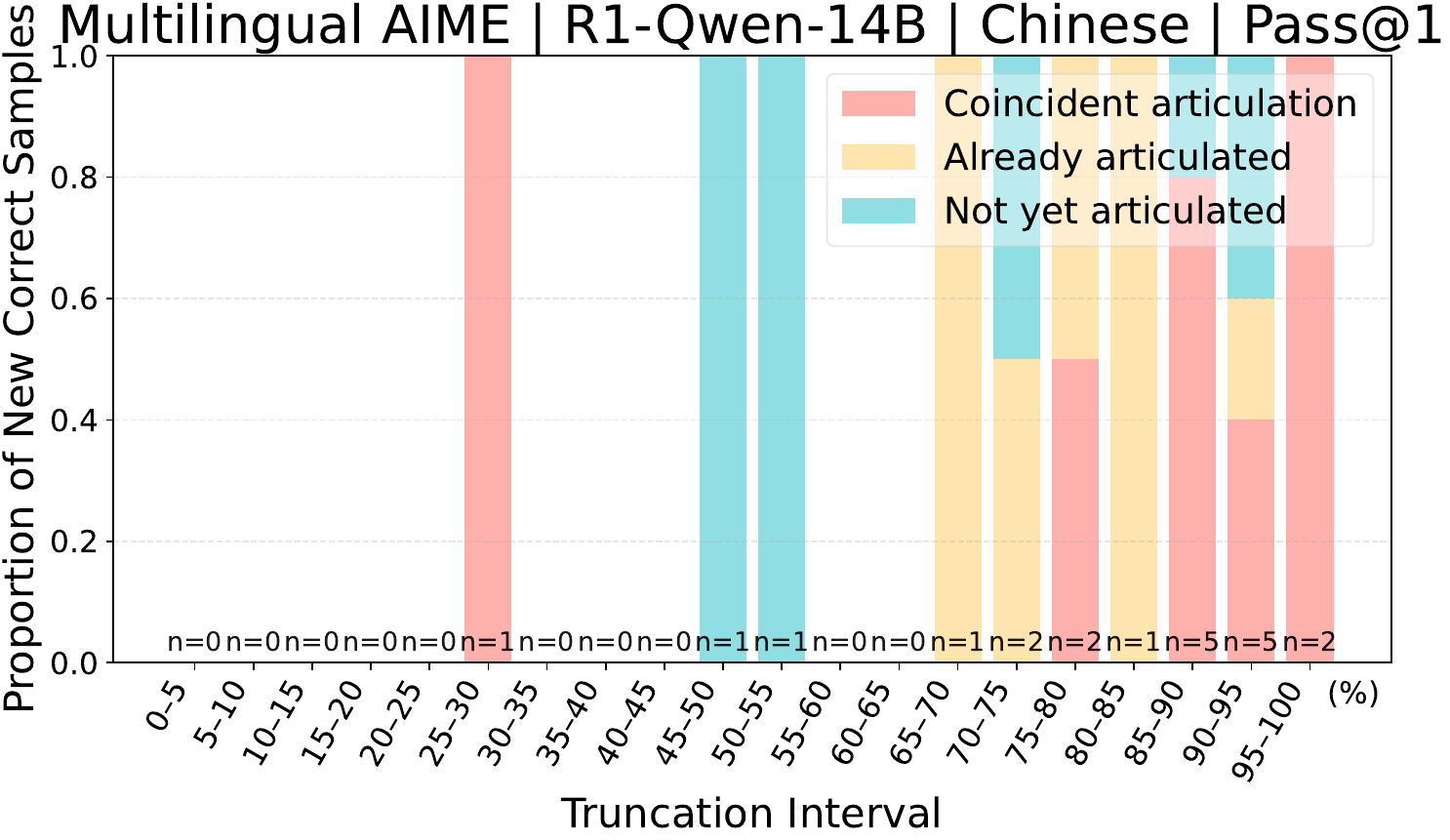}
    \caption{
    Causal decomposition of newly correct predictions across truncation intervals on \textbf{Multilingual AIME} with \textbf{R1-Qwen-14B}.
    Each bar partitions gains into three cases: (\textbf{i}) the gold answer is first articulated in the newly added reasoning steps,
    (\textbf{ii}) it was already articulated in earlier steps, or
    (\textbf{iii}) it has not yet appeared in the visible truncated trace.
    Compared to MGSM, gains are sparser and less dominated by latent reasoning.
    }
    \label{fig:interval_14b_aime}
\end{figure*}

\begin{figure*}
    \centering
    \includegraphics[width=0.24\textwidth]{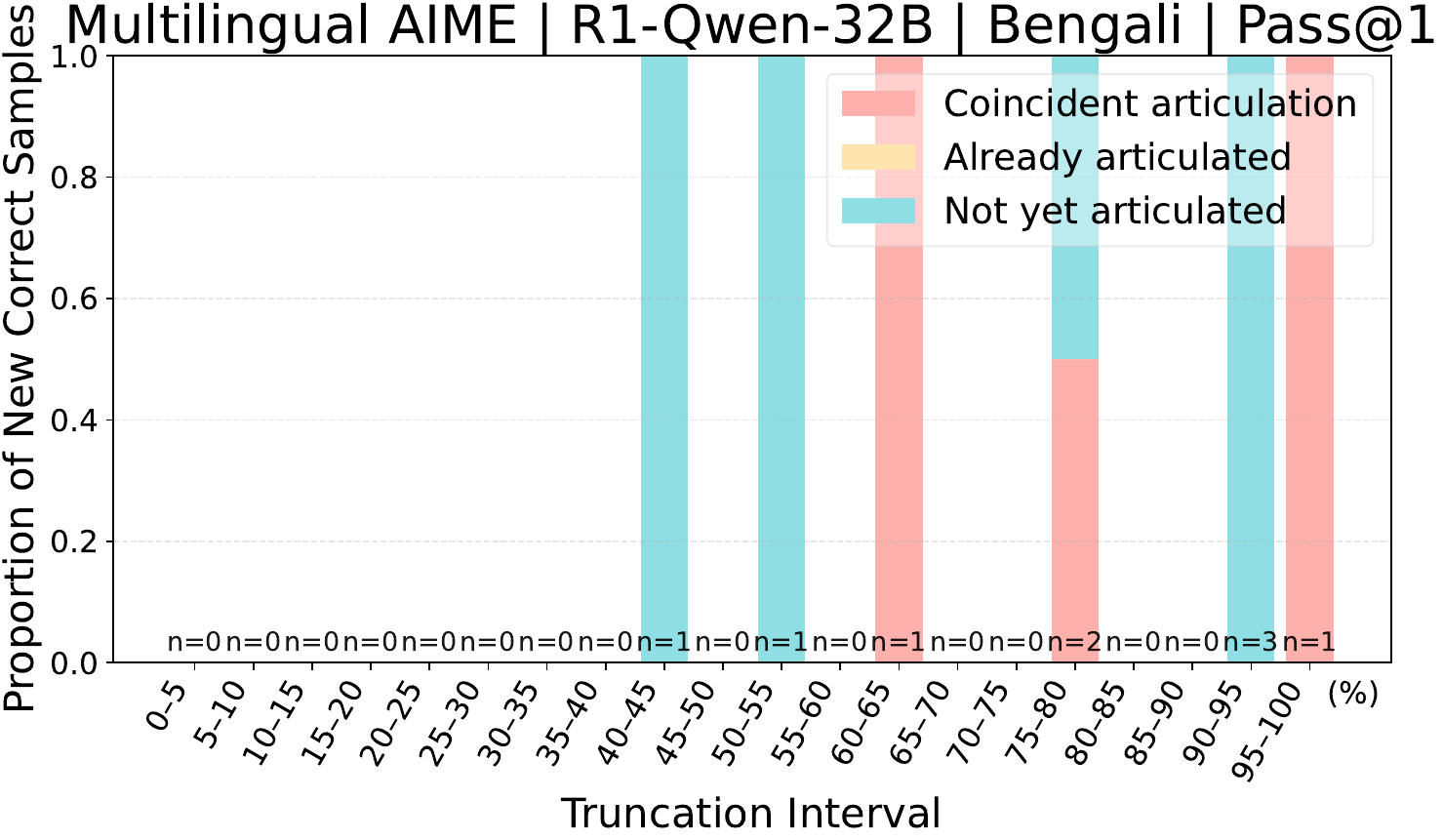}
    \includegraphics[width=0.24\textwidth]{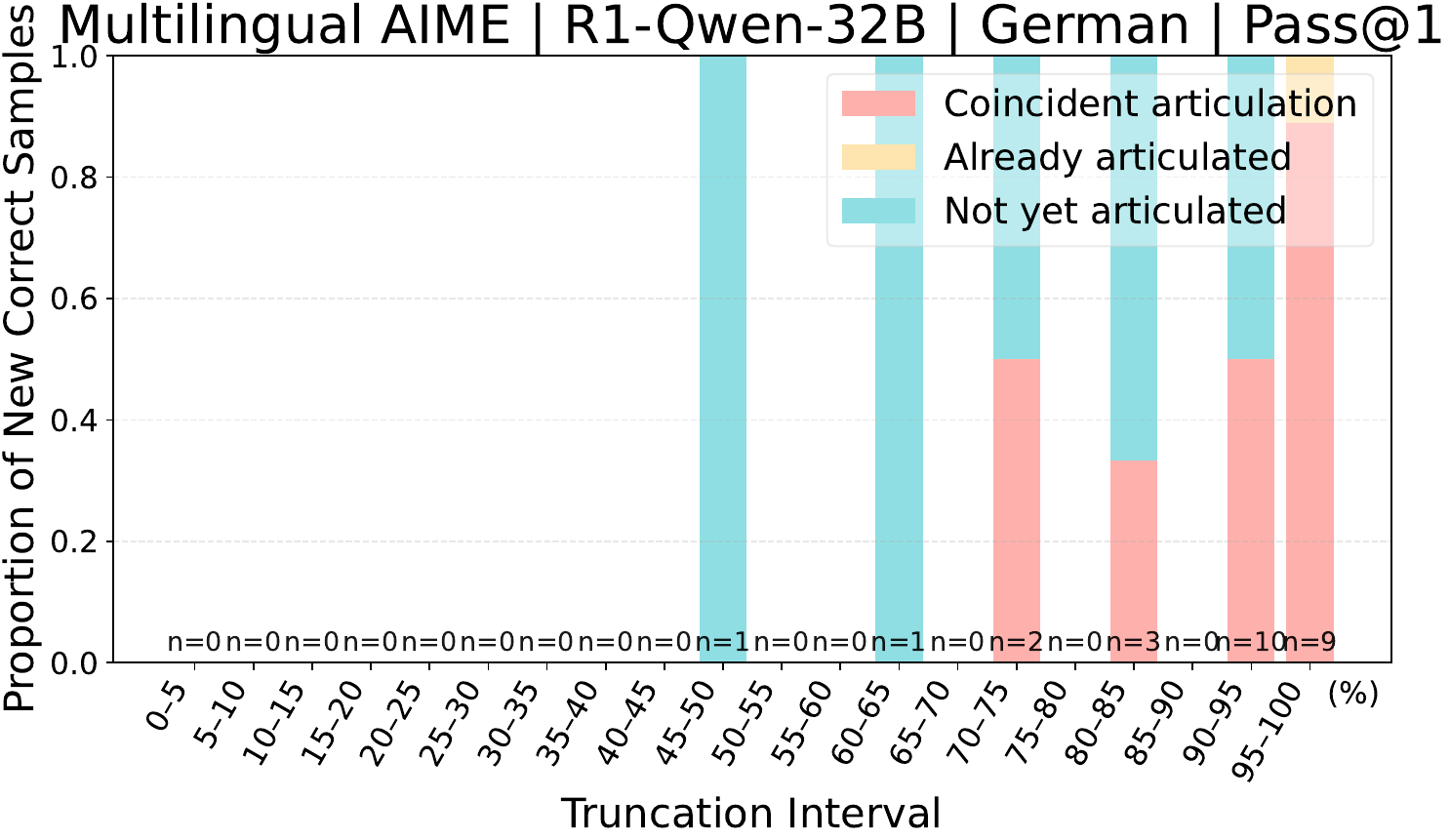}
    \includegraphics[width=0.24\textwidth]{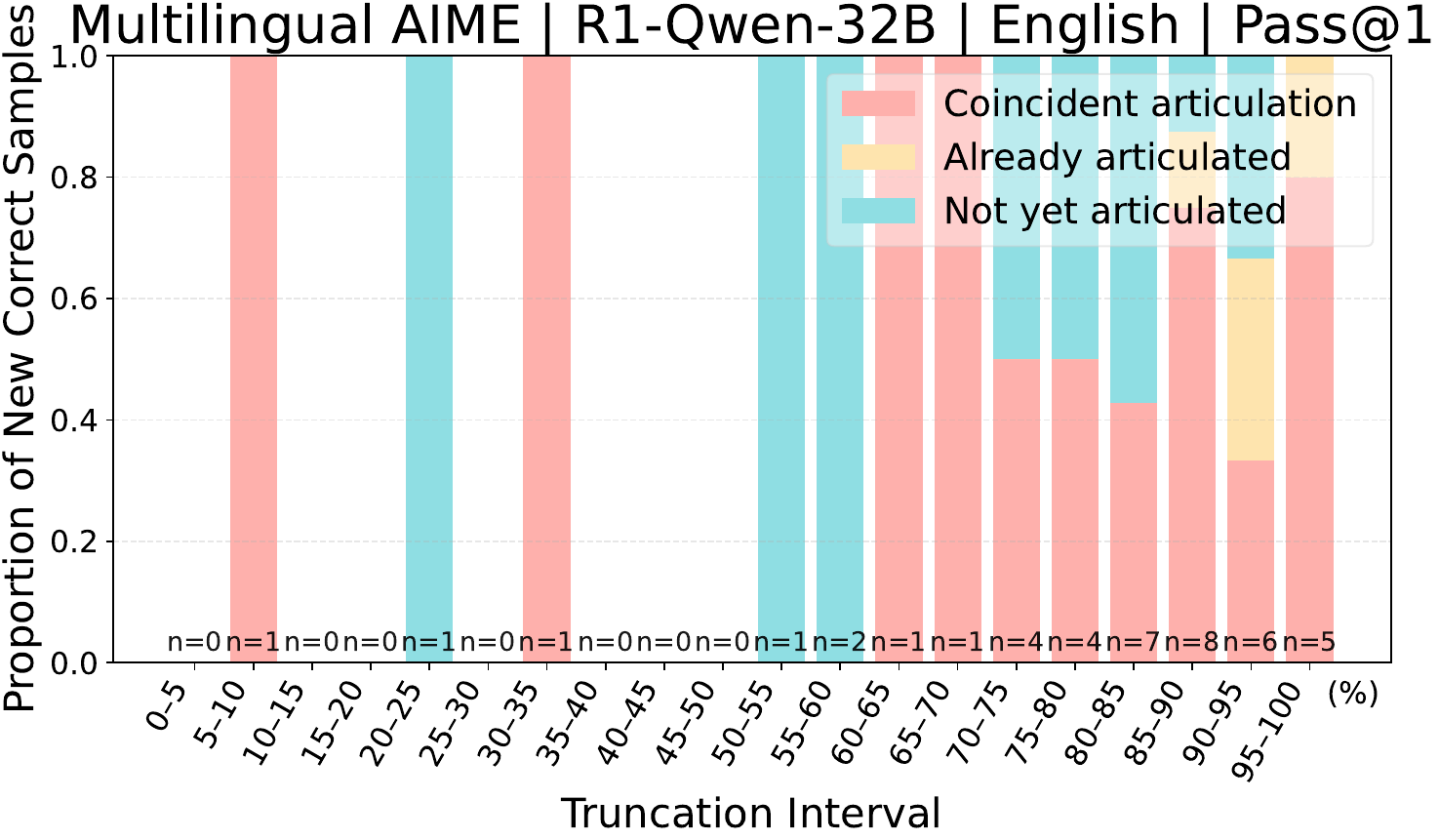}
    \includegraphics[width=0.24\textwidth]{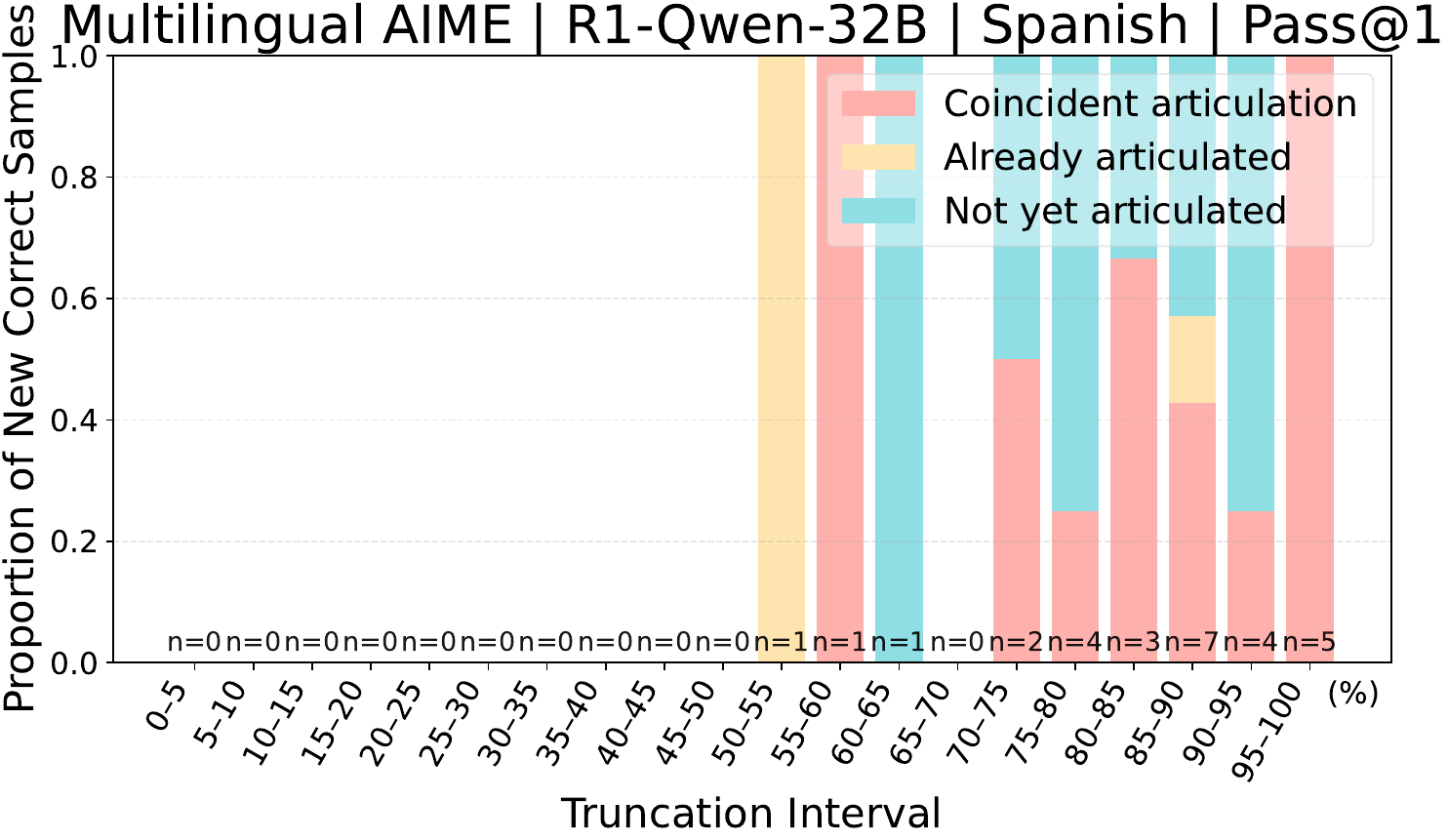}
    \includegraphics[width=0.24\textwidth]{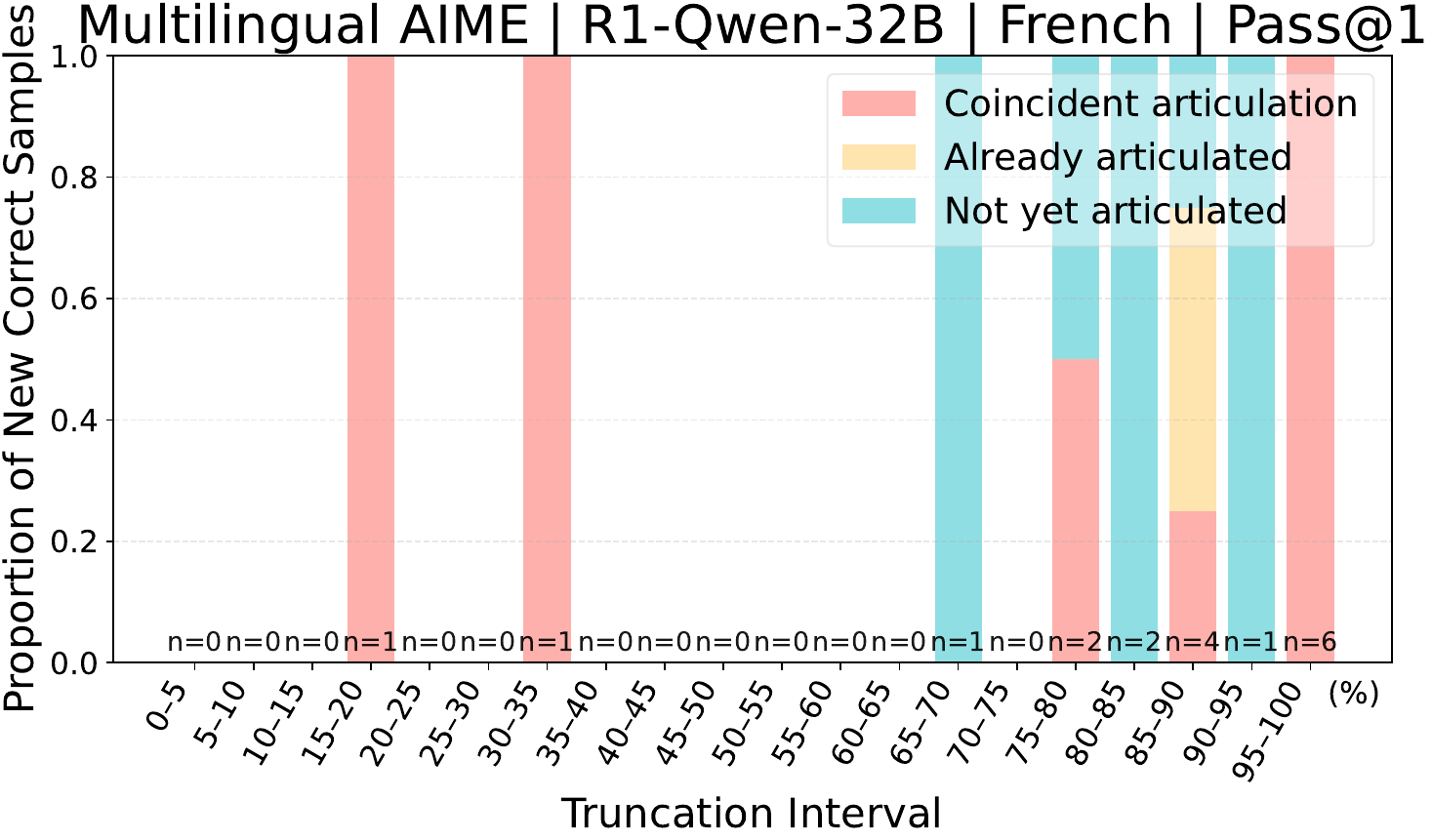}
    \includegraphics[width=0.24\textwidth]{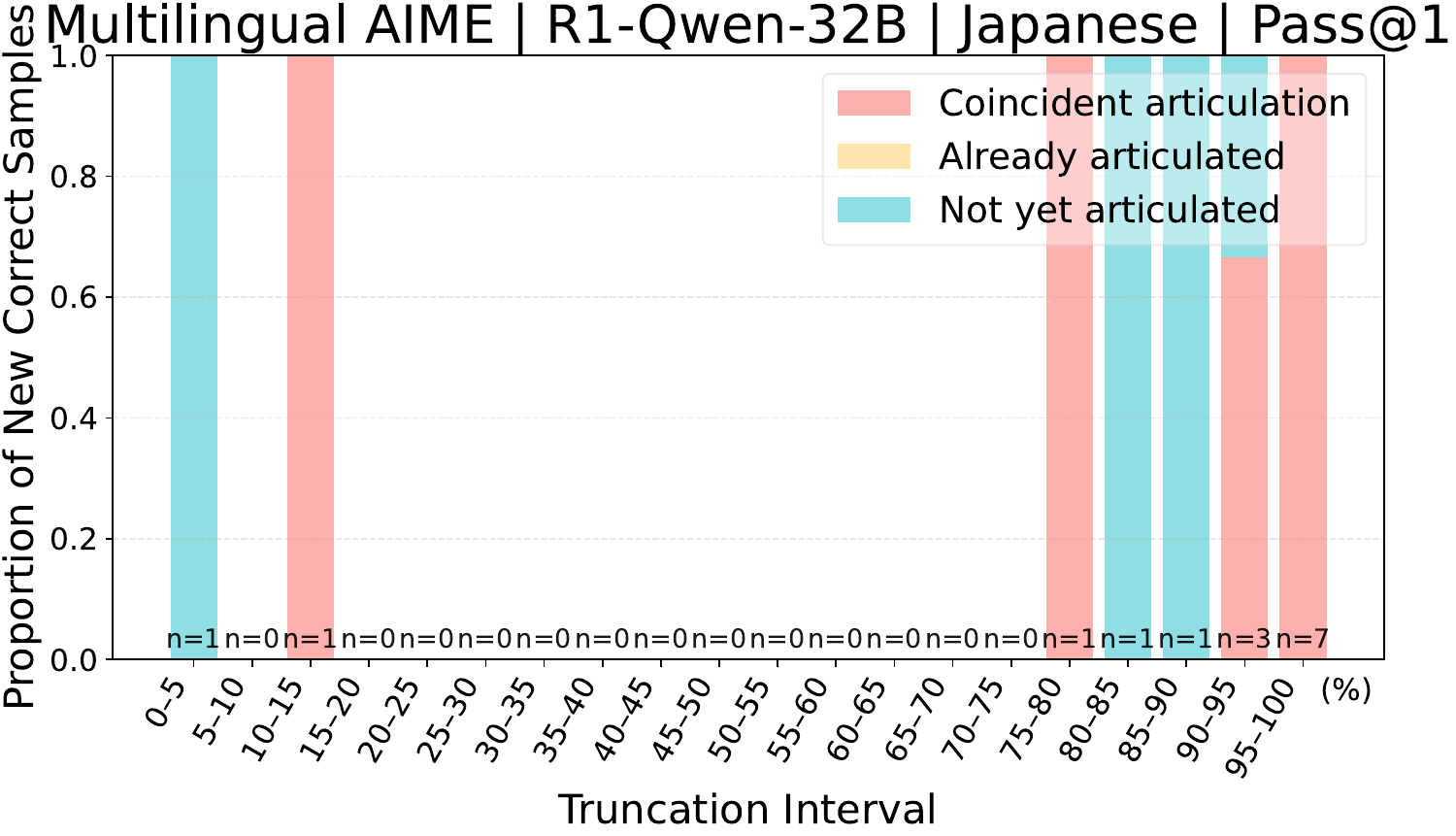}
    \includegraphics[width=0.24\textwidth]{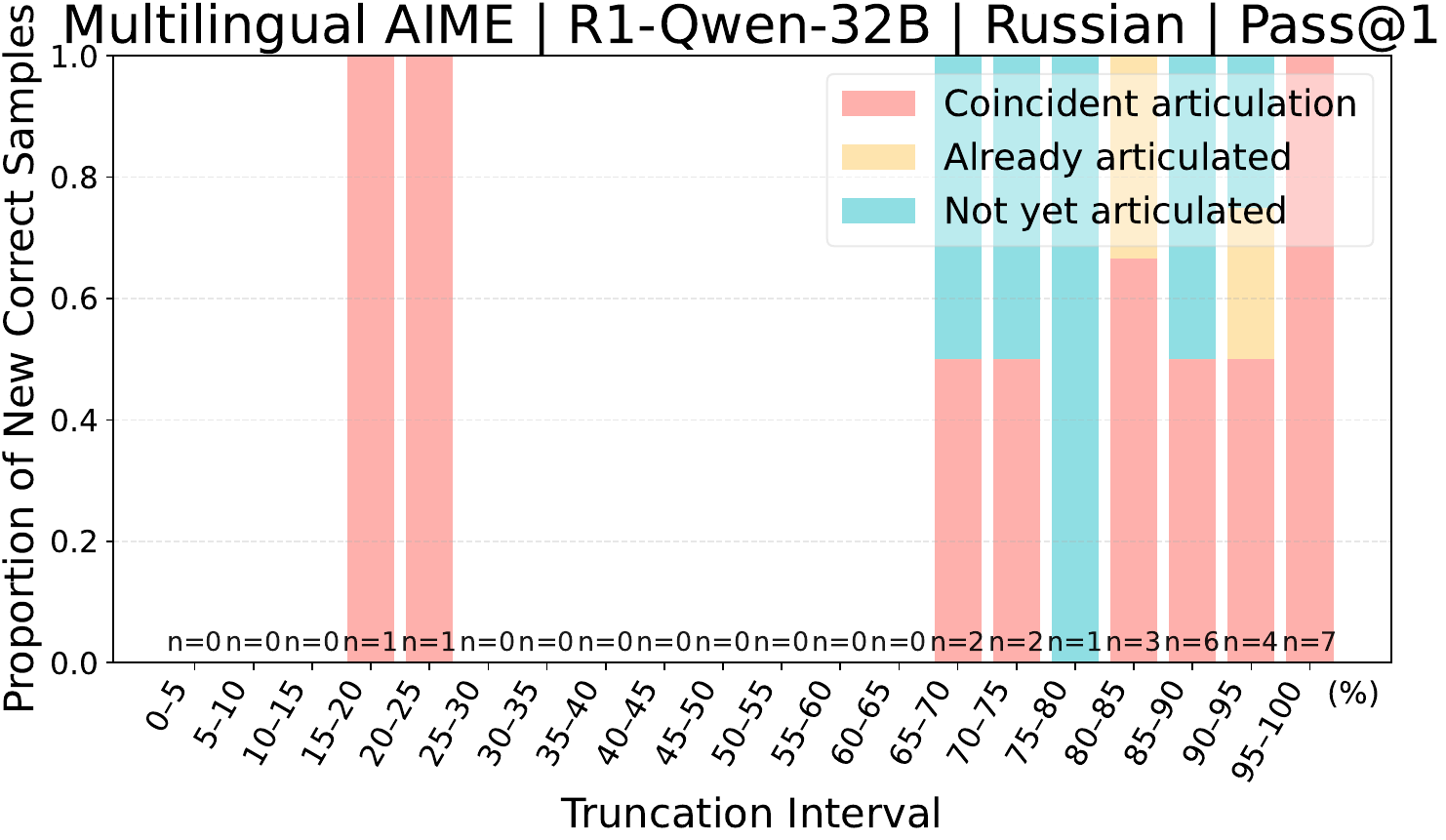}
    \includegraphics[width=0.24\textwidth]{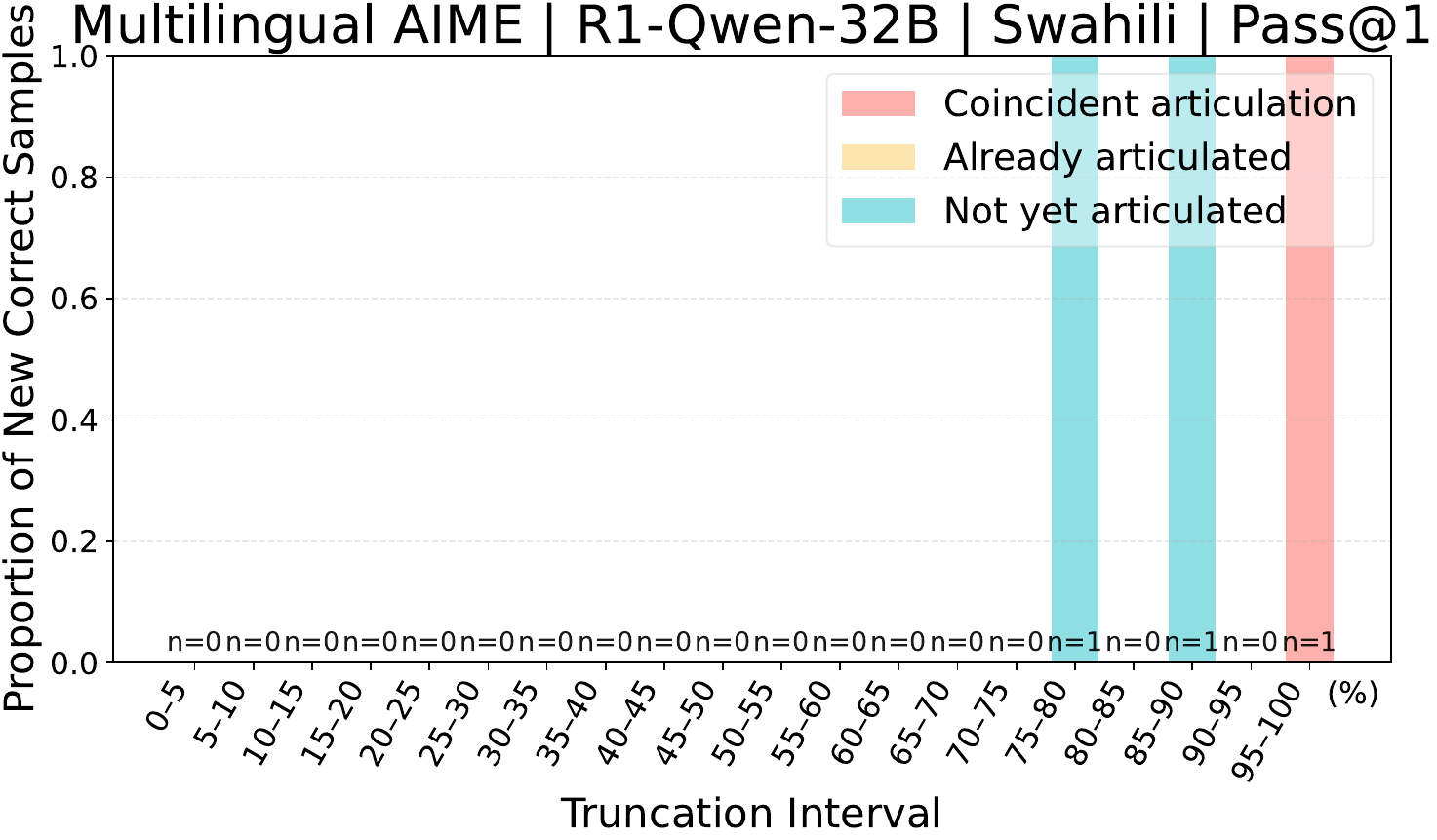}
    \includegraphics[width=0.24\textwidth]{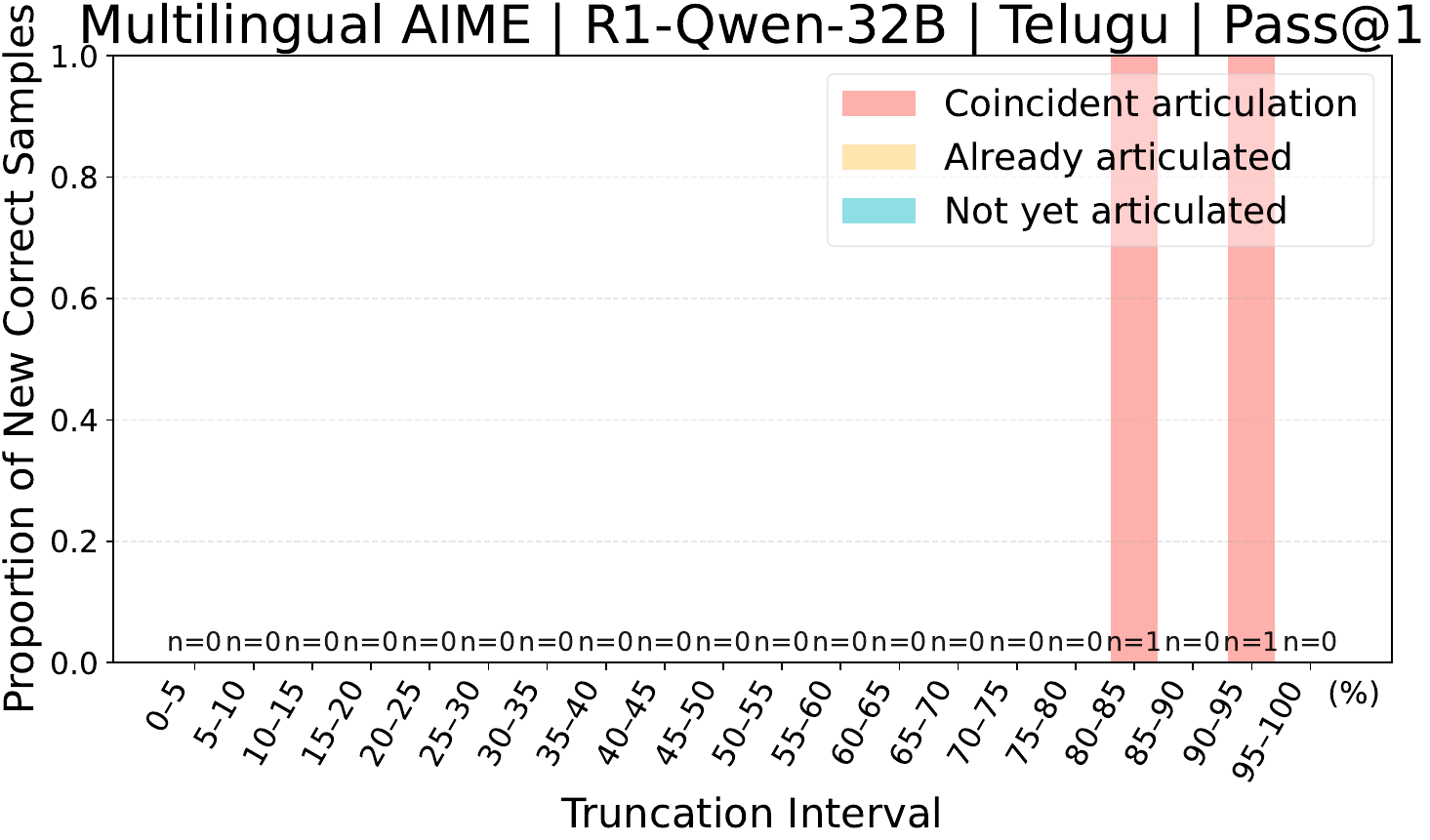}
    \includegraphics[width=0.24\textwidth]{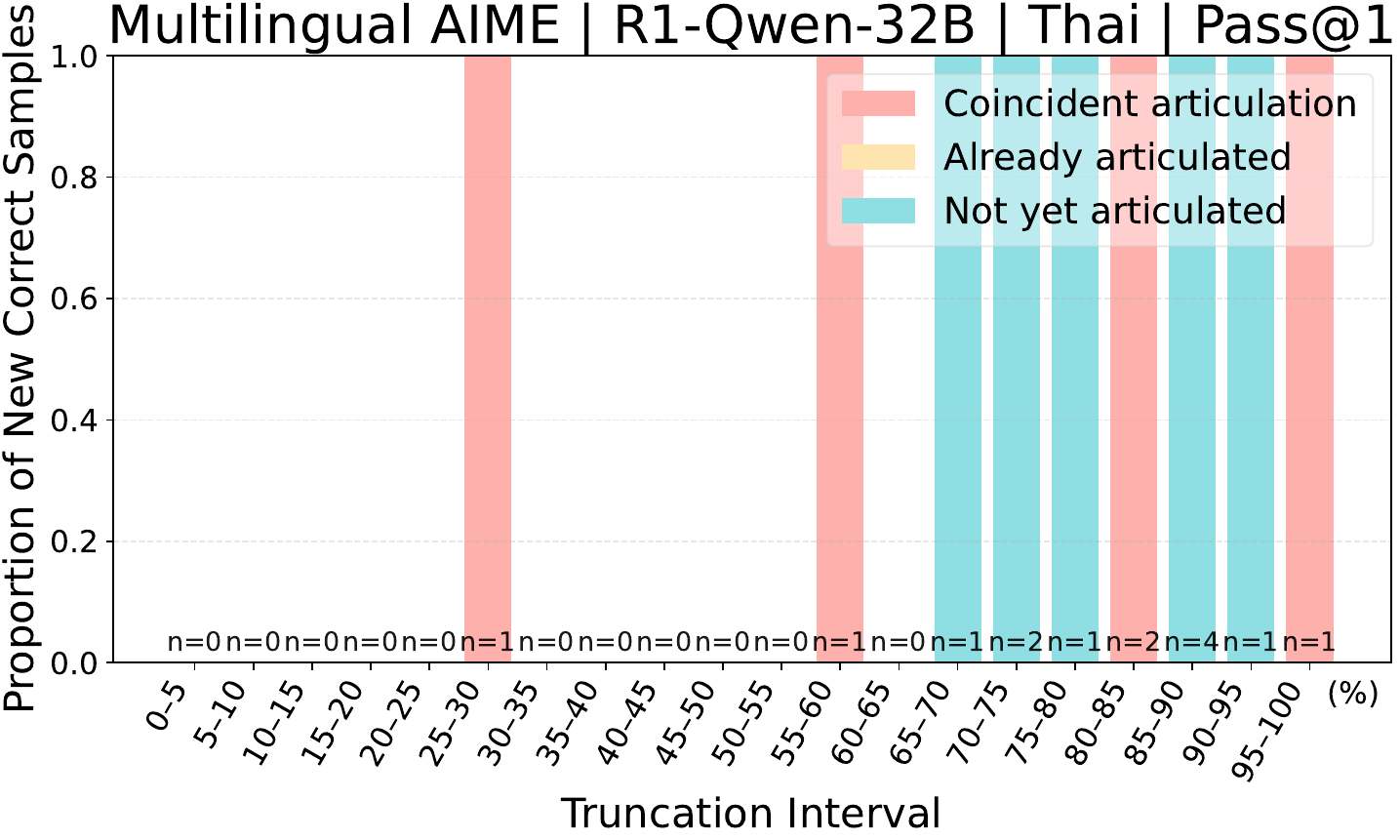}
    \includegraphics[width=0.24\textwidth]{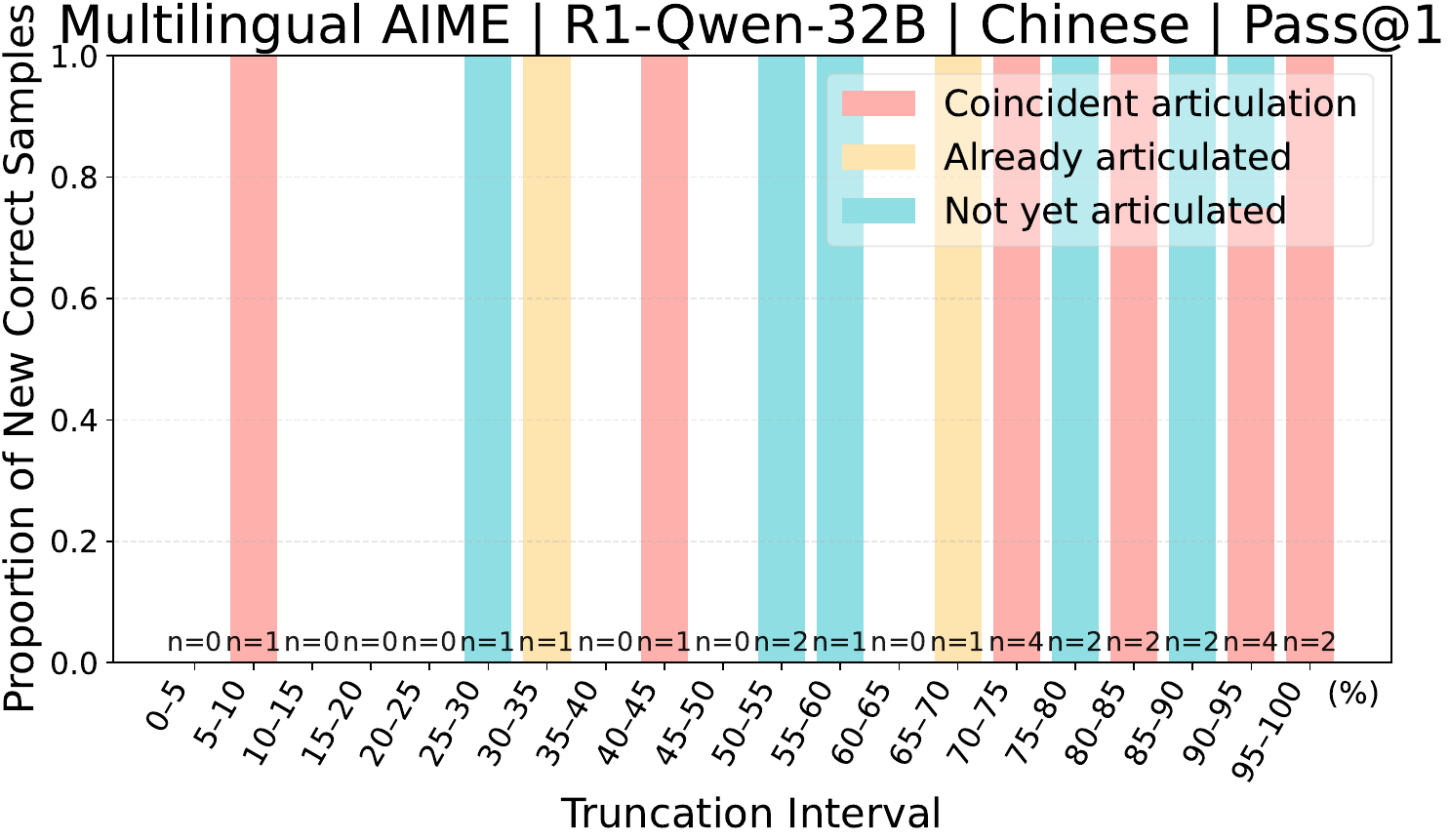}
    \caption{
    Causal decomposition of newly correct predictions across truncation intervals on \textbf{Multilingual AIME} with \textbf{R1-Qwen-32B}.
    Each bar partitions gains into three cases: (\textbf{i}) the gold answer is first articulated in the newly added reasoning steps,
    (\textbf{ii}) it was already articulated in earlier steps, or
    (\textbf{iii}) it has not yet appeared in the visible truncated trace.
    Compared to MGSM, gains are sparser and less dominated by latent reasoning.
    }
    \label{fig:interval_32b_aime}
\end{figure*}

Figures~\ref{fig:interval_7b_mgsm}, \ref{fig:interval_14b_mgsm}, \ref{fig:interval_32b_mgsm} (MGSM) and Figures~\ref{fig:interval_7b_aime}, \ref{fig:interval_14b_aime}, \ref{fig:interval_32b_aime} (Multilingual AIME) further decompose accuracy gains between consecutive truncation intervals into three cases:
(i) the gold answer is newly articulated in the added reasoning steps,
(ii) it was already present earlier in the visible trace, or
(iii) it has not yet appeared in the truncated trace.
This analysis disentangles improvements driven by explicit answer articulation from those arising prior to any verbalized solution.
On MGSM, gains at early and intermediate truncation ratios are largely attributed to case (iii), indicating that correct predictions often emerge before the answer is explicitly articulated.
In contrast, on Multilingual AIME, gains are sparser and increasingly dominated by cases (i) and (ii), reflecting a stronger dependence on explicit reasoning and a reduced role for early latent solution formation.

%% file: similarity_results.tex
\section{Complete Similarity results}\seclabel{similarity_results}

\begin{figure*}
    \centering
    \includegraphics[width=0.32\textwidth]{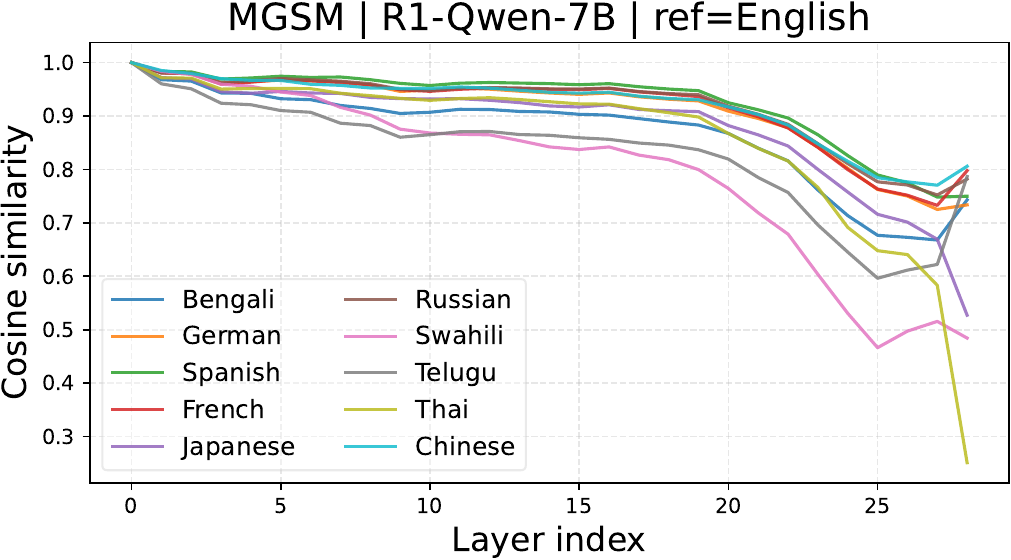}
    \includegraphics[width=0.32\textwidth]{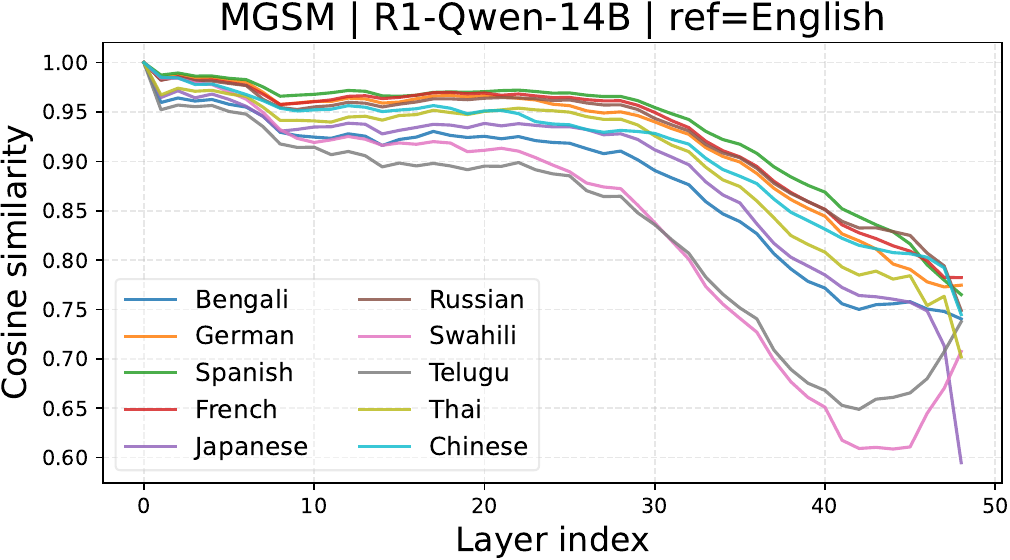}
    \includegraphics[width=0.32\textwidth]{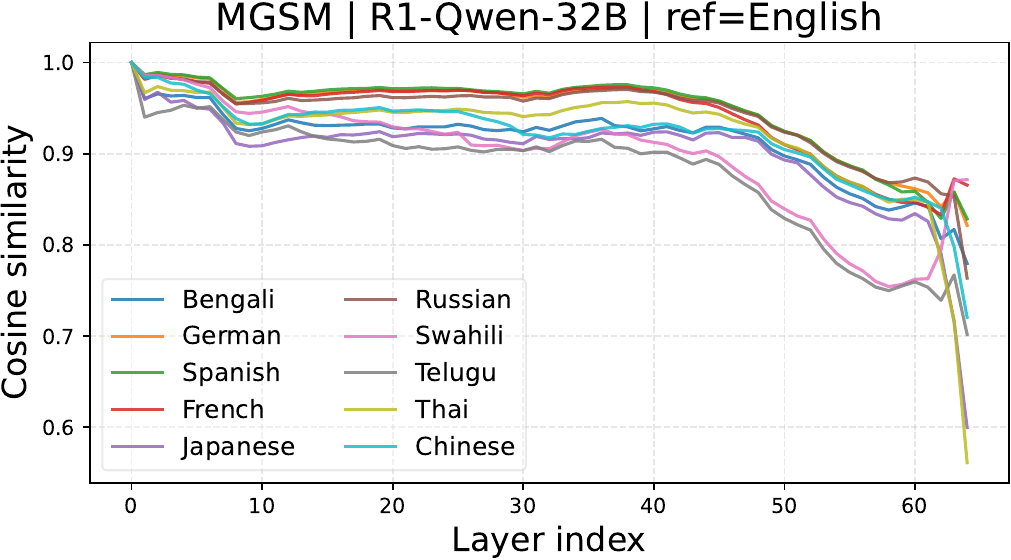}
    \includegraphics[width=0.32\textwidth]{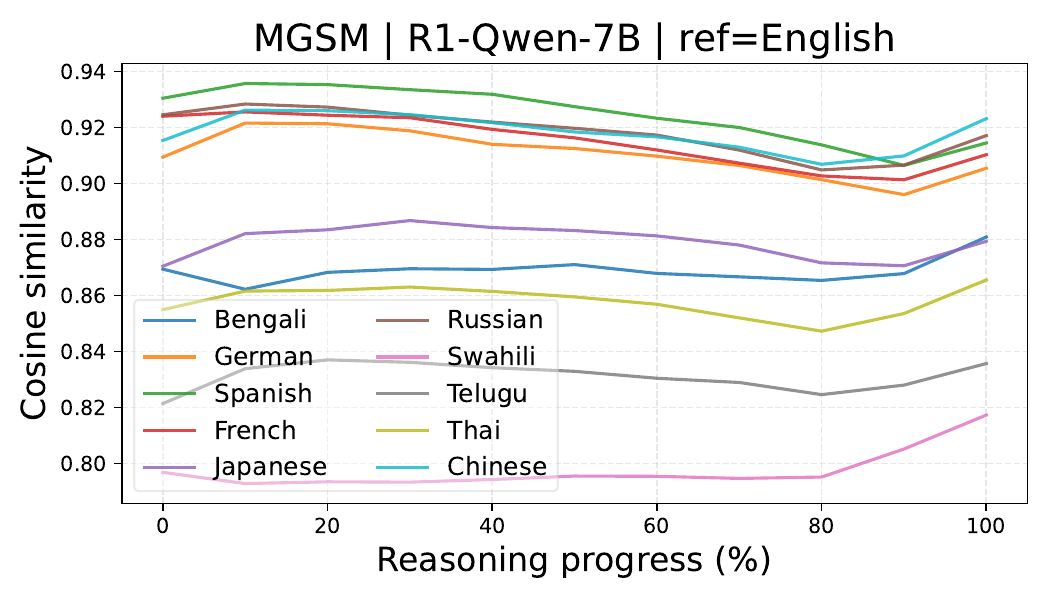}
    \includegraphics[width=0.32\textwidth]{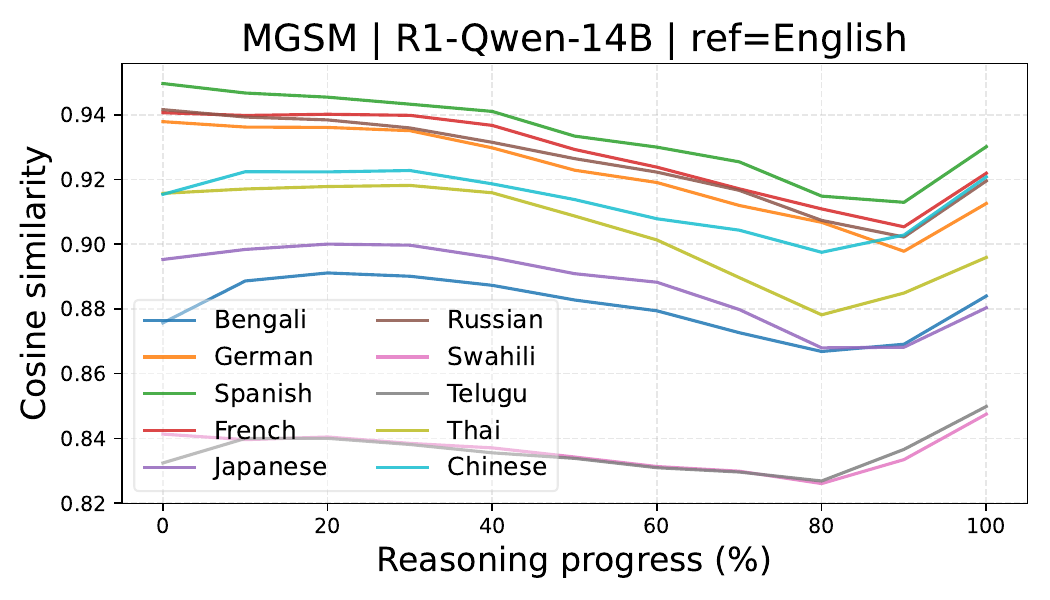}
    \includegraphics[width=0.32\textwidth]{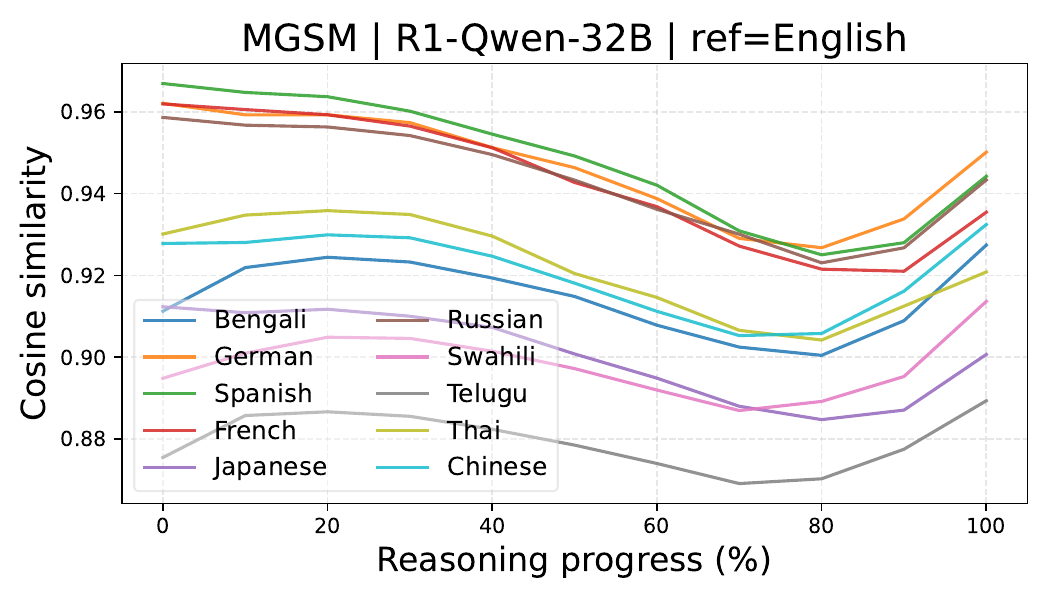}
    \caption{Aggregated cosine similarity on \textbf{MGSM} between hidden states in each language and English (reference), averaged over both reasoning steps and layers. 
    High-resource languages show consistently higher similarity to English, suggesting convergence toward an English-centered latent reasoning pathway.}
    \label{fig:cosine_sim_aggregated_mgsm}
\end{figure*}

\begin{figure*}
    \centering
    \includegraphics[width=0.32\textwidth]{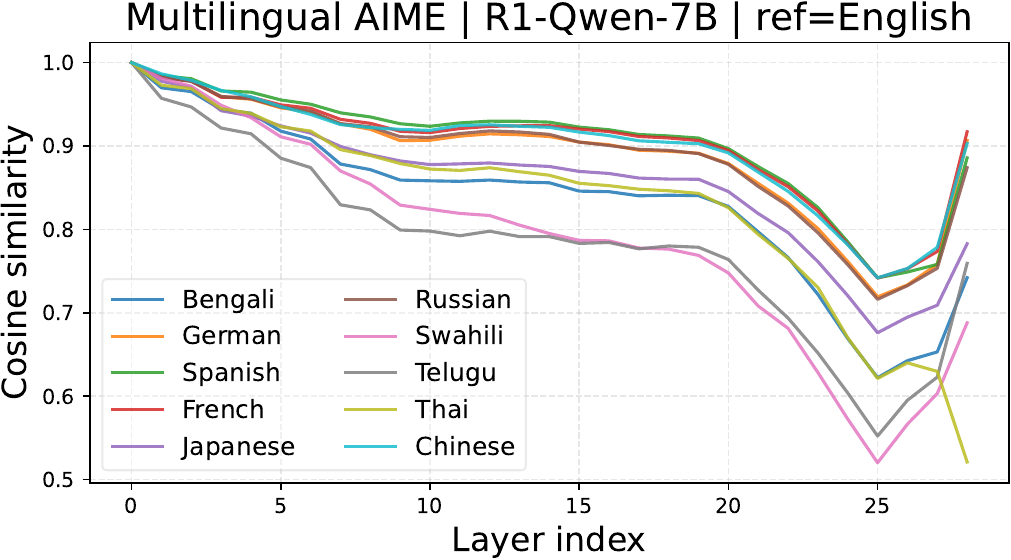}
    \includegraphics[width=0.32\textwidth]{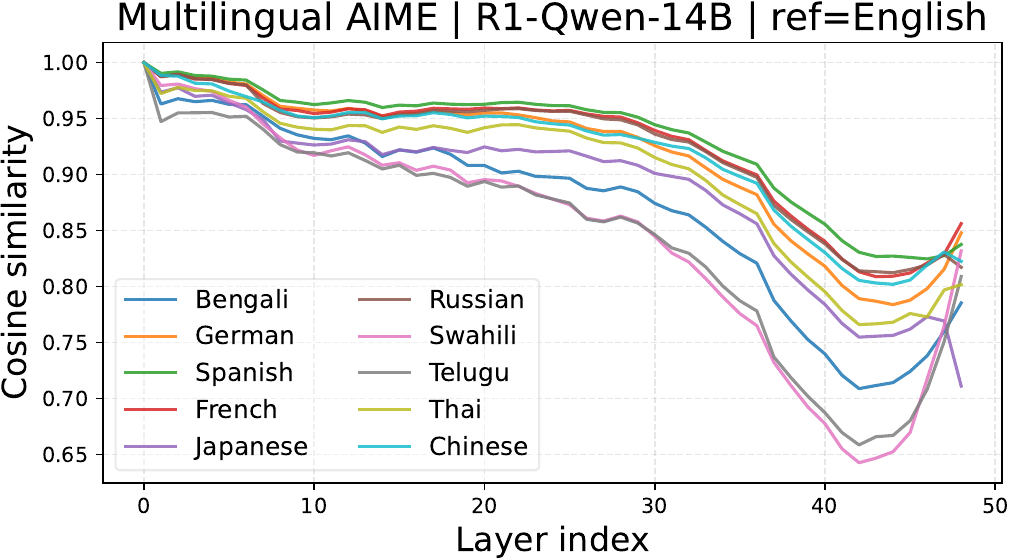}
    \includegraphics[width=0.32\textwidth]{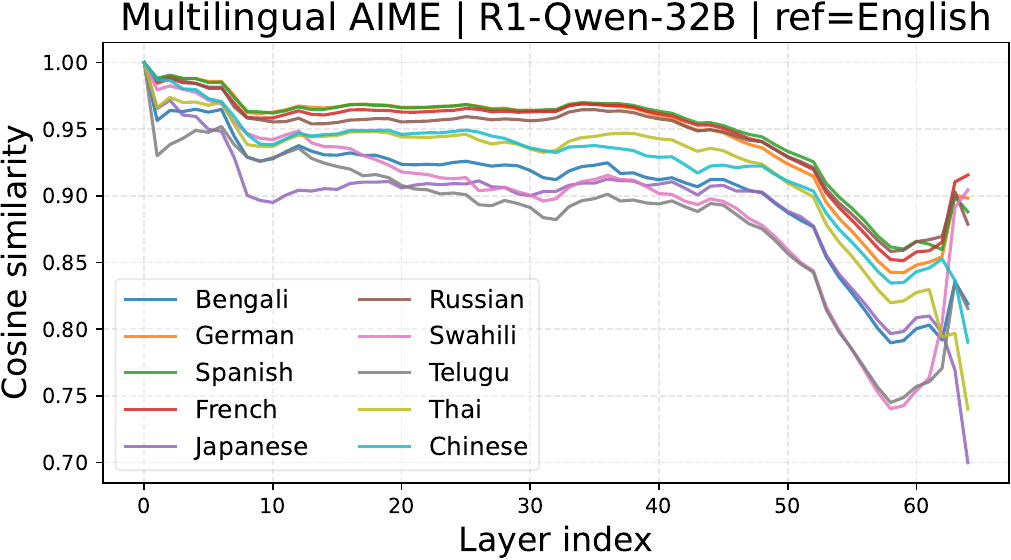}
    \includegraphics[width=0.32\textwidth]{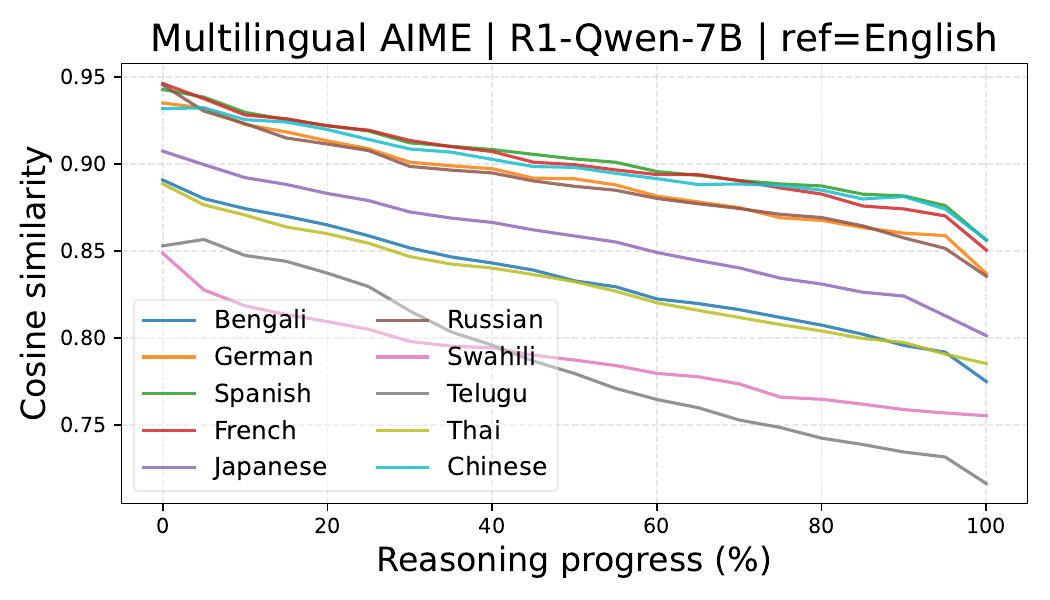}
    \includegraphics[width=0.32\textwidth]{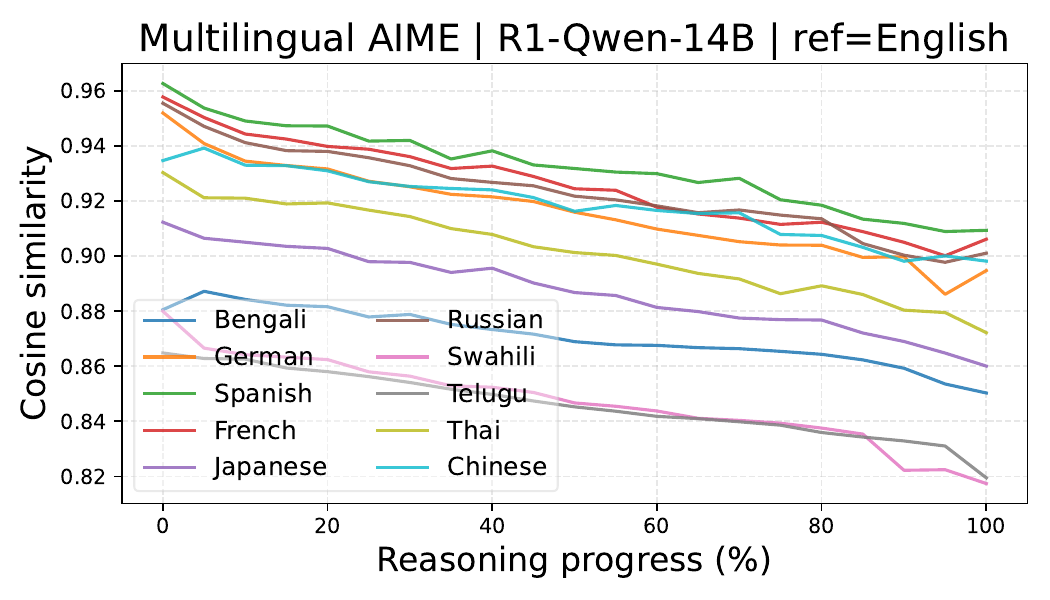}
    \includegraphics[width=0.32\textwidth]{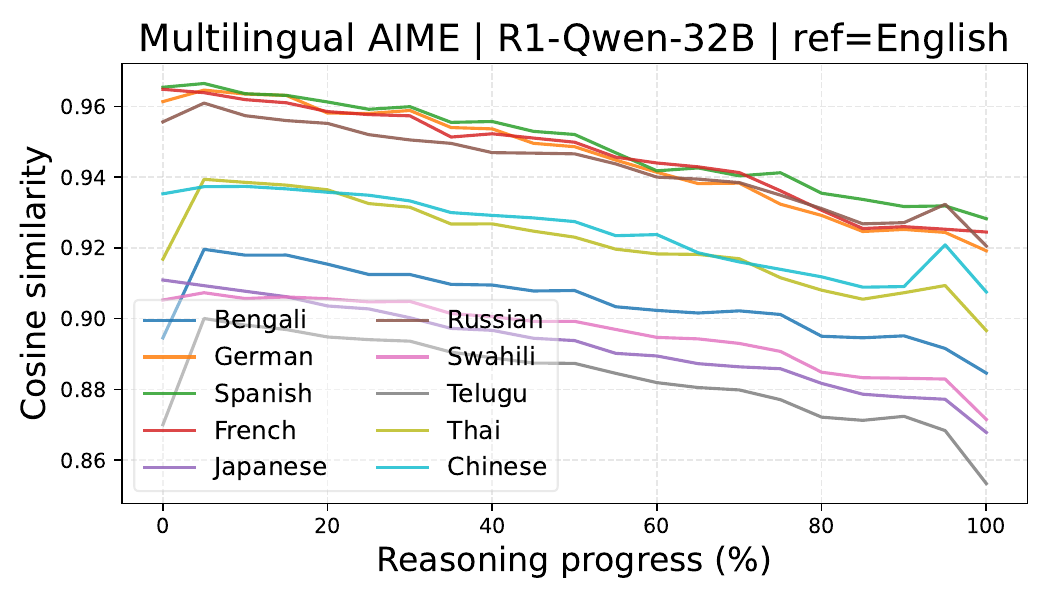}
    \caption{Aggregated cosine similarity on \textbf{Multilingual AIME} between hidden states in each language and English (reference), averaged over both reasoning steps and layers. 
    High-resource languages show consistently higher similarity to English, suggesting convergence toward an English-centered latent reasoning pathway.}
    \label{fig:cosine_sim_aggregated_aime}
\end{figure*}

\subsection{Similarity with English}\seclabel{sim_eng}

We conduct a focused analysis of crosslingual representational alignment by computing the cosine similarity between hidden states in each target language and those of English.
For each instance, and at each truncation ratio, we extract the hidden state corresponding to the final token of the partial reasoning trace and compare it with the hidden state obtained from the English version of the same problem.
To summarize these similarities, we aggregate them both across layers and across reasoning steps.
Figure~\ref{fig:cosine_sim_aggregated_mgsm} and Figure~\ref{fig:cosine_sim_aggregated_aime} present the resulting trends for MGSM and Multilingual AIME, respectively.

Across both benchmarks, we observe systematic differences in representational alignment with English that seem to correlate with language resource levels: high-resource languages exhibit consistently stronger alignment with English representations, whereas mid- and low-resource languages show reduced alignment.

\subsection{Similarity vs. Correctness}\seclabel{sim_correct}

\begin{figure*}
    \centering

    \includegraphics[width=0.23\textwidth]{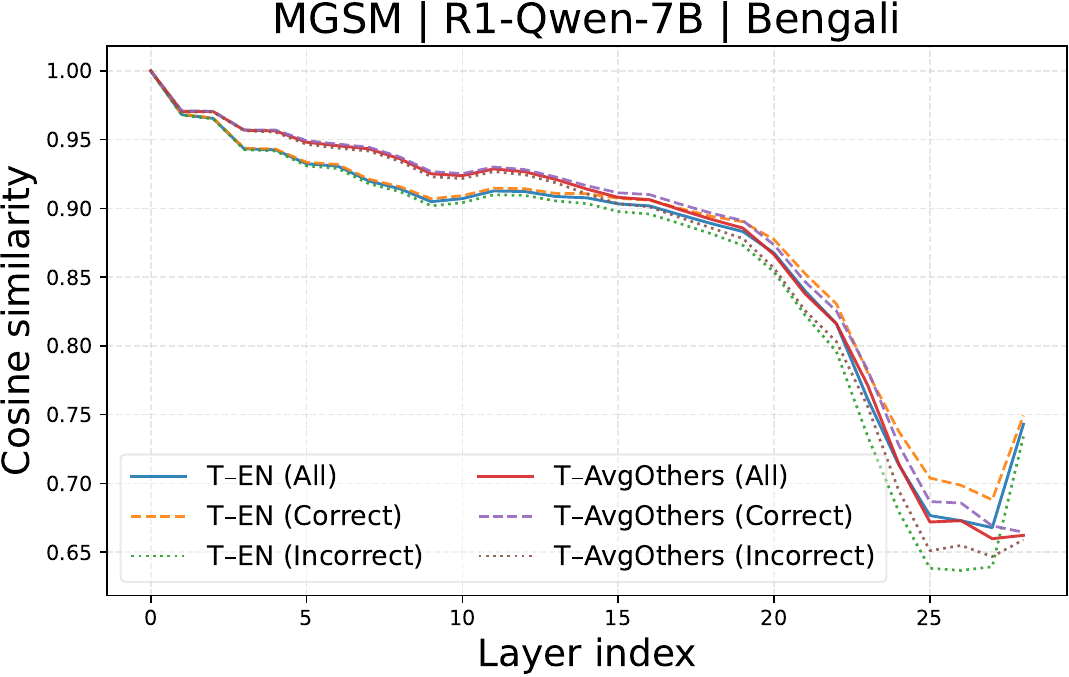}
    \includegraphics[width=0.23\textwidth]{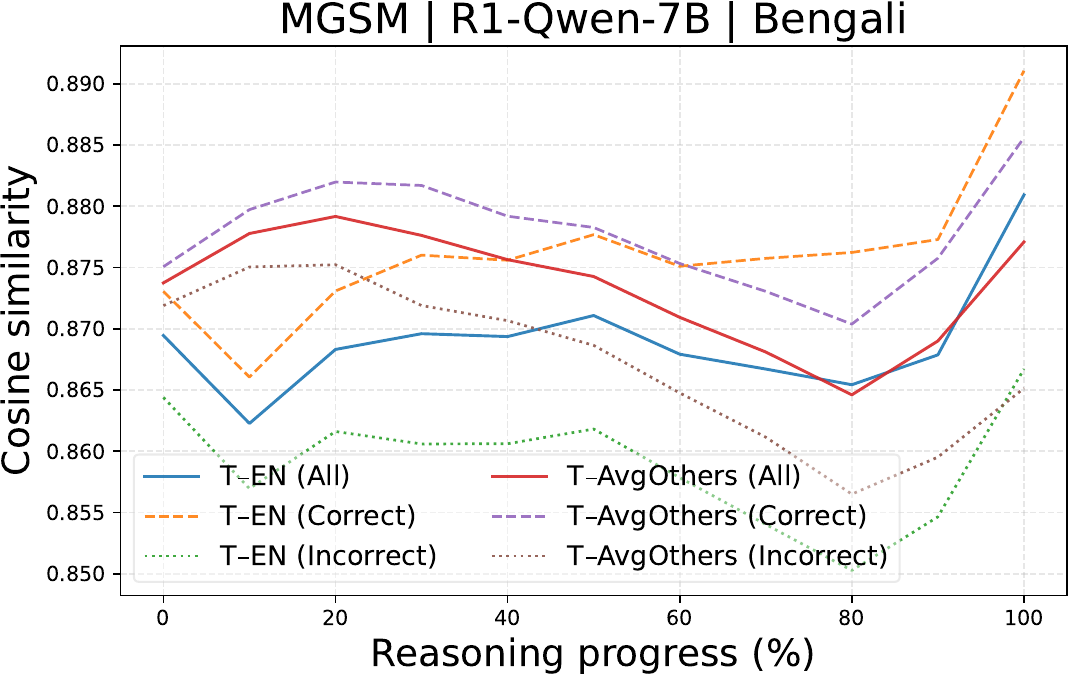}
    \includegraphics[width=0.23\textwidth]{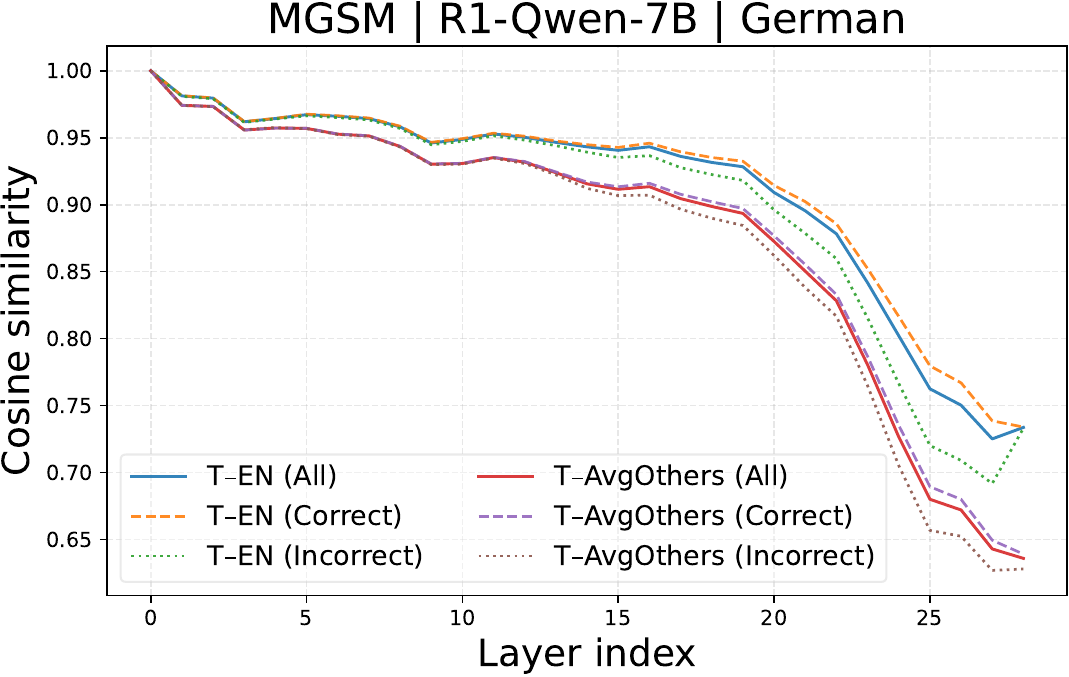}
    \includegraphics[width=0.23\textwidth]{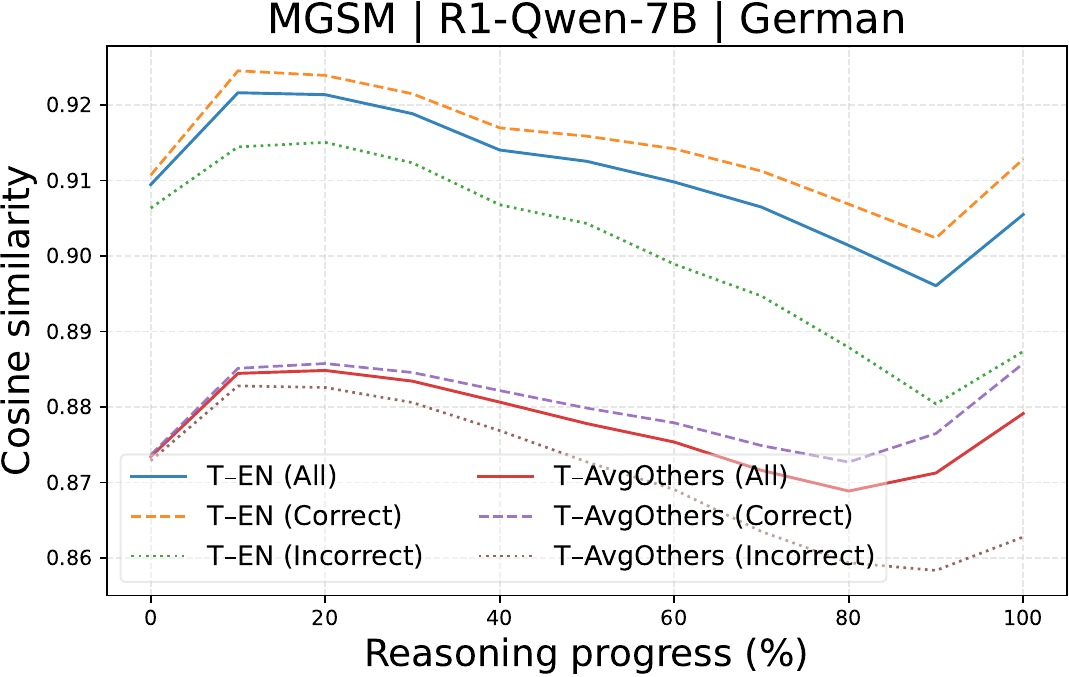}
    \includegraphics[width=0.23\textwidth]{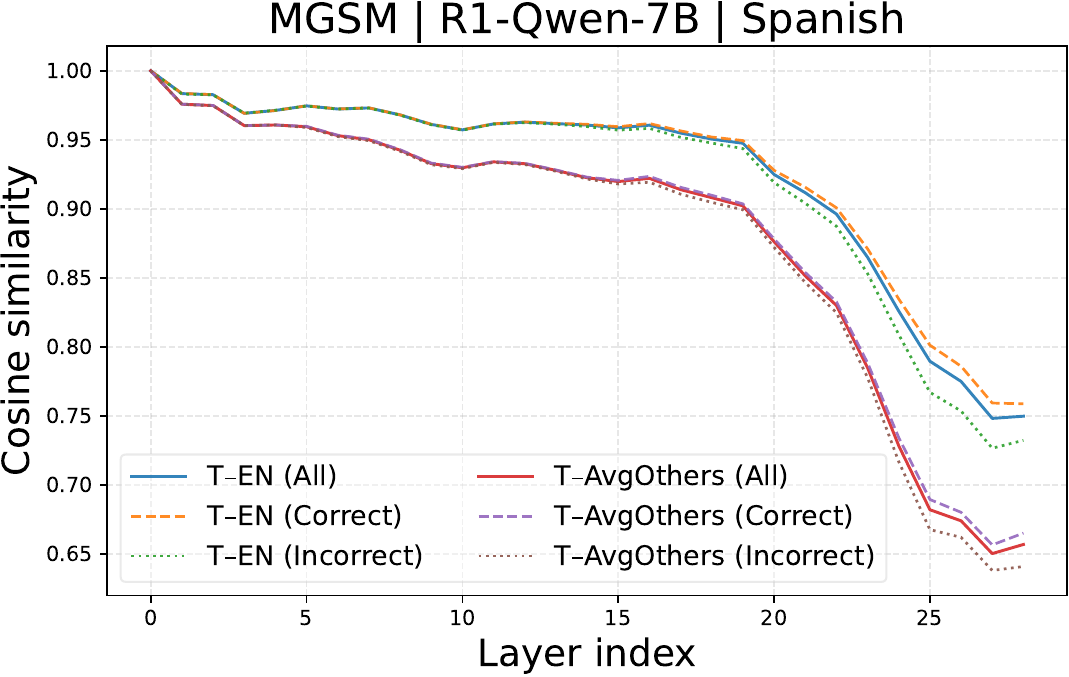}
    \includegraphics[width=0.23\textwidth]{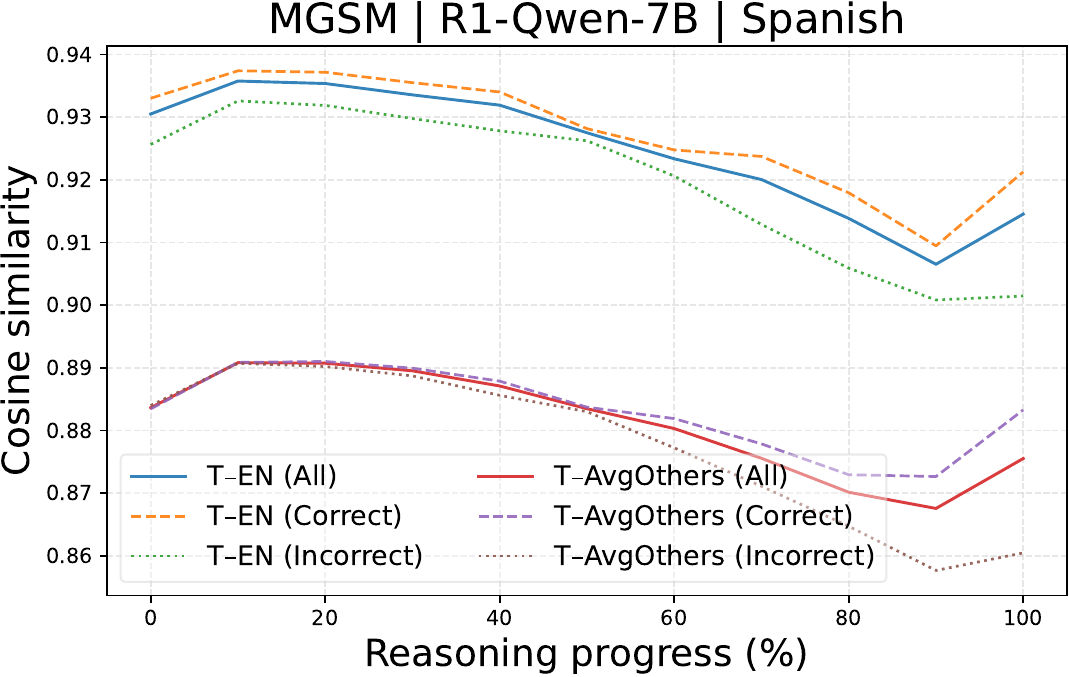}
    \includegraphics[width=0.23\textwidth]{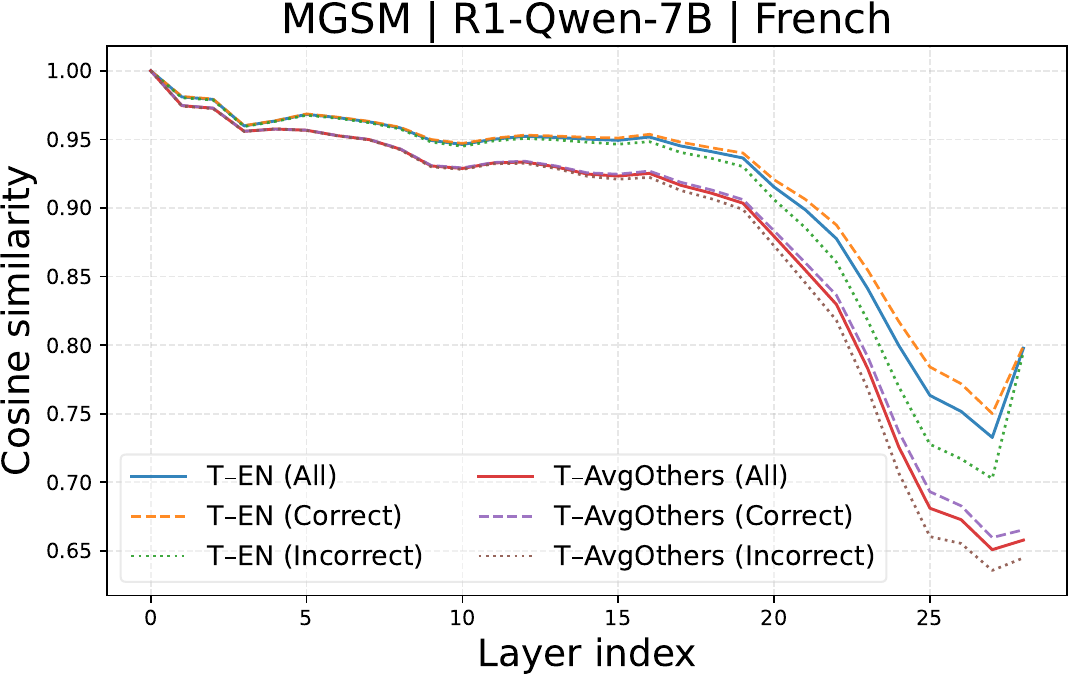}
    \includegraphics[width=0.23\textwidth]{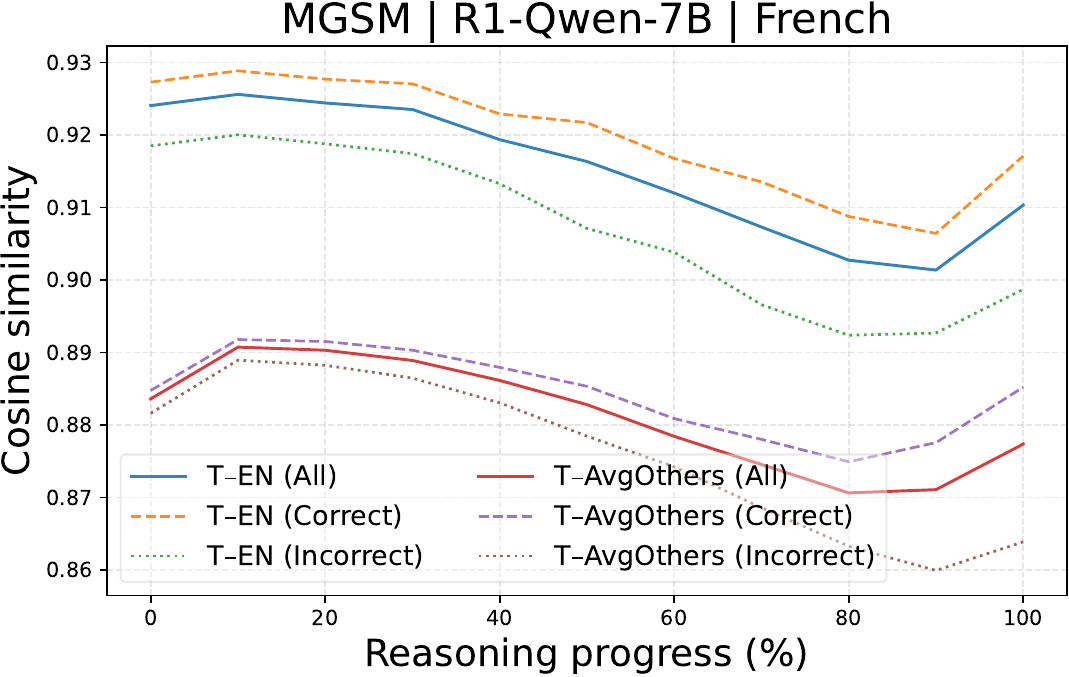}
    \includegraphics[width=0.23\textwidth]{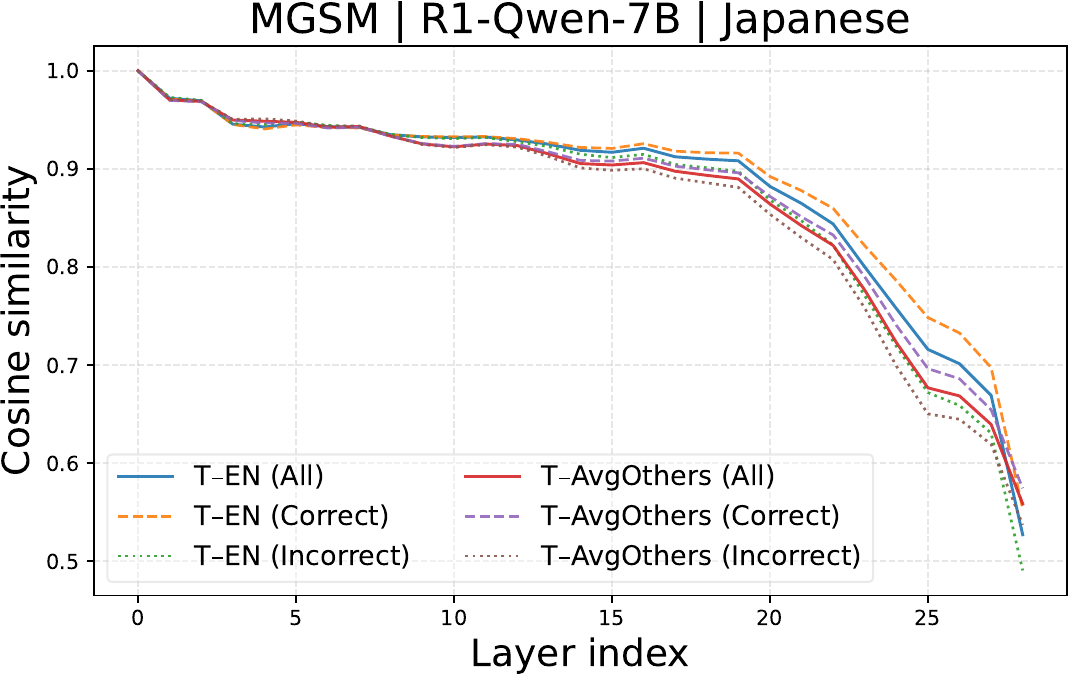}
    \includegraphics[width=0.23\textwidth]{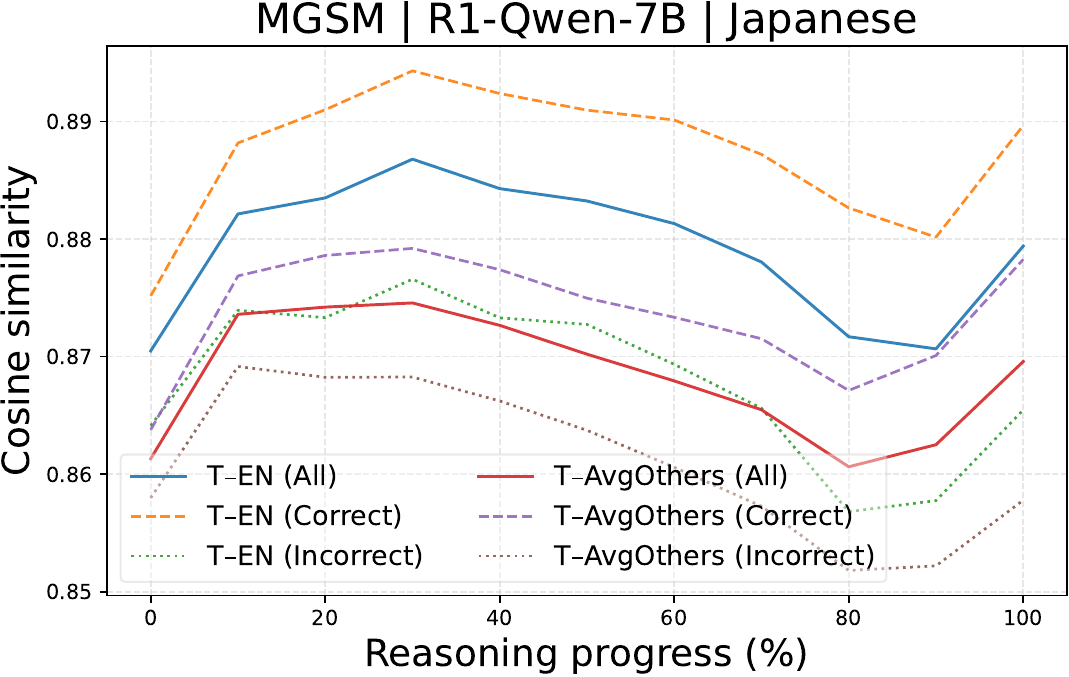}
    \includegraphics[width=0.23\textwidth]{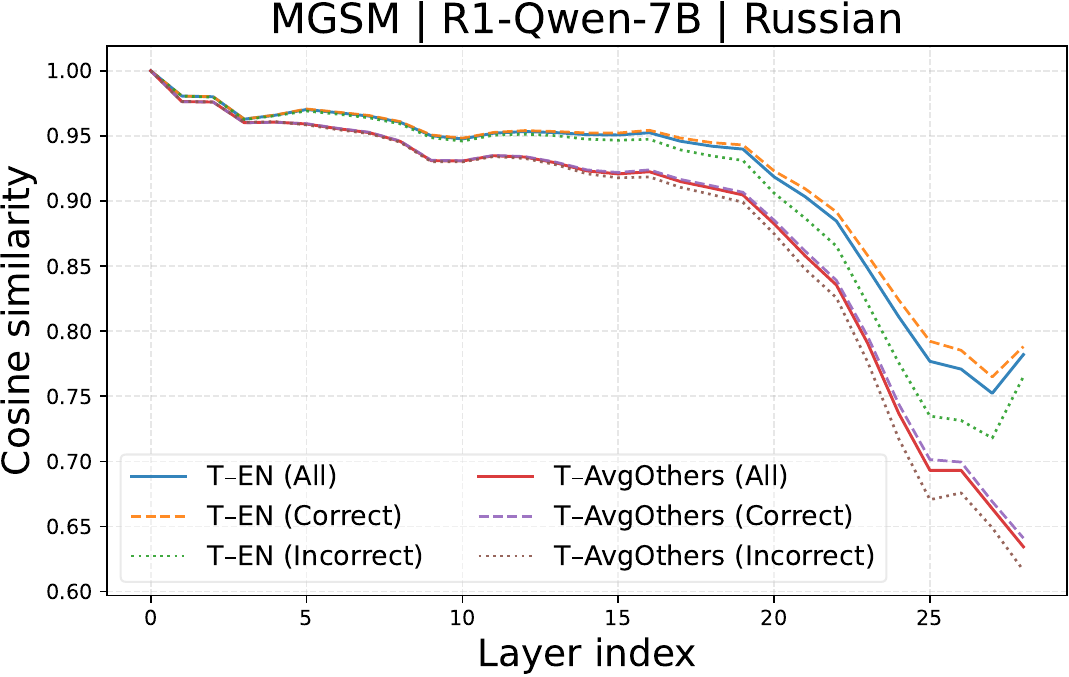}
    \includegraphics[width=0.23\textwidth]{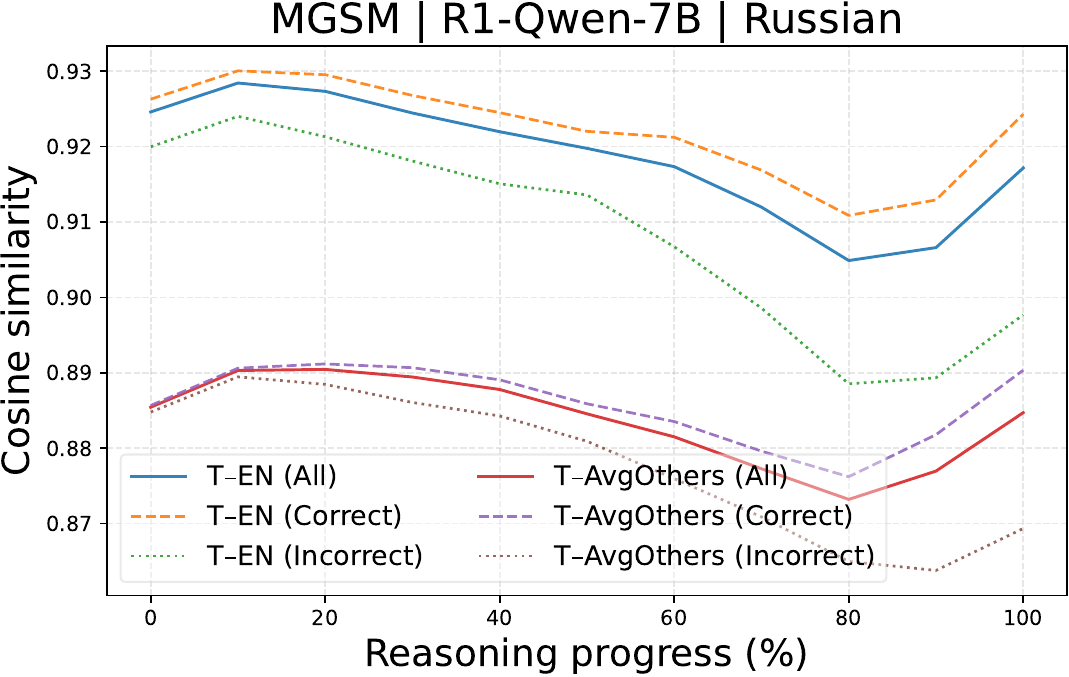}
    \includegraphics[width=0.23\textwidth]{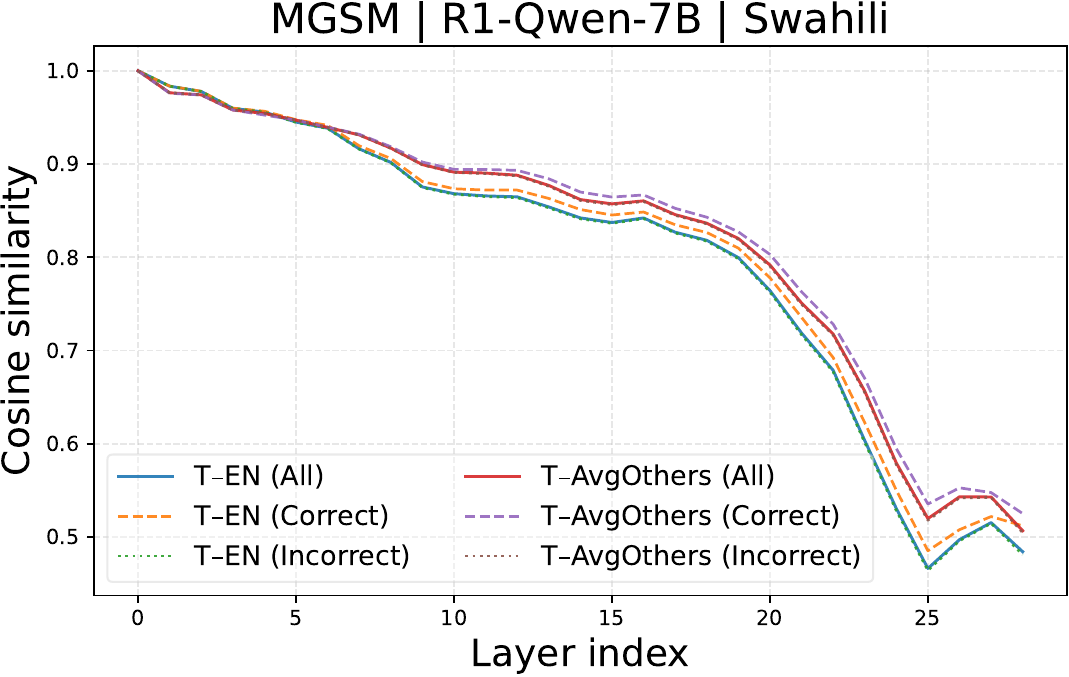}
    \includegraphics[width=0.23\textwidth]{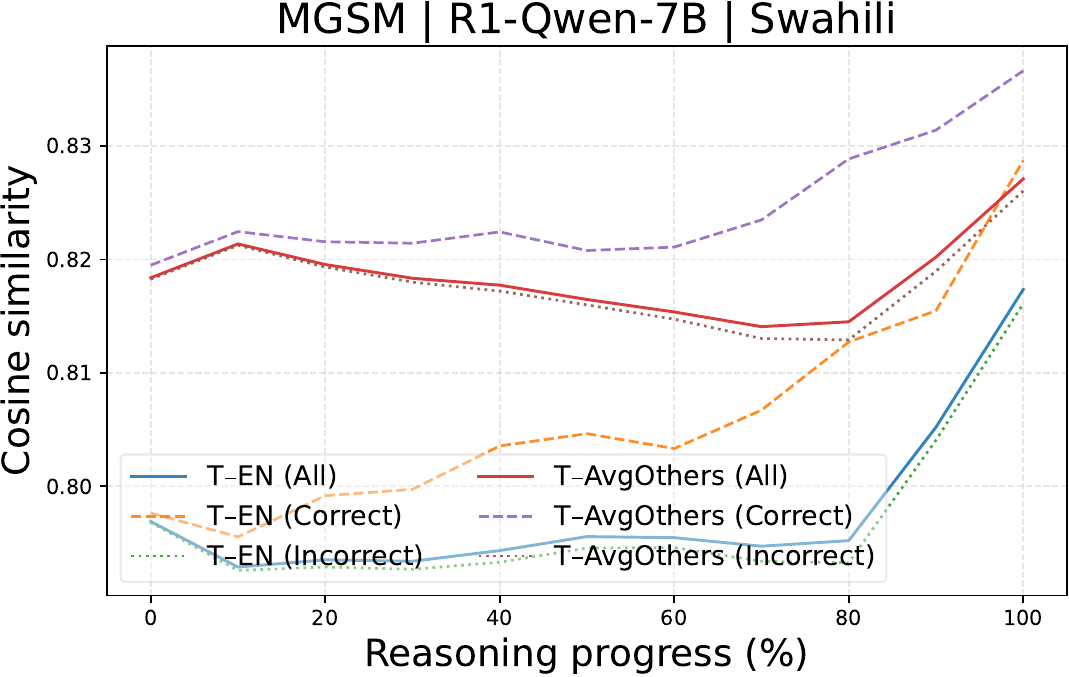}
    \includegraphics[width=0.23\textwidth]{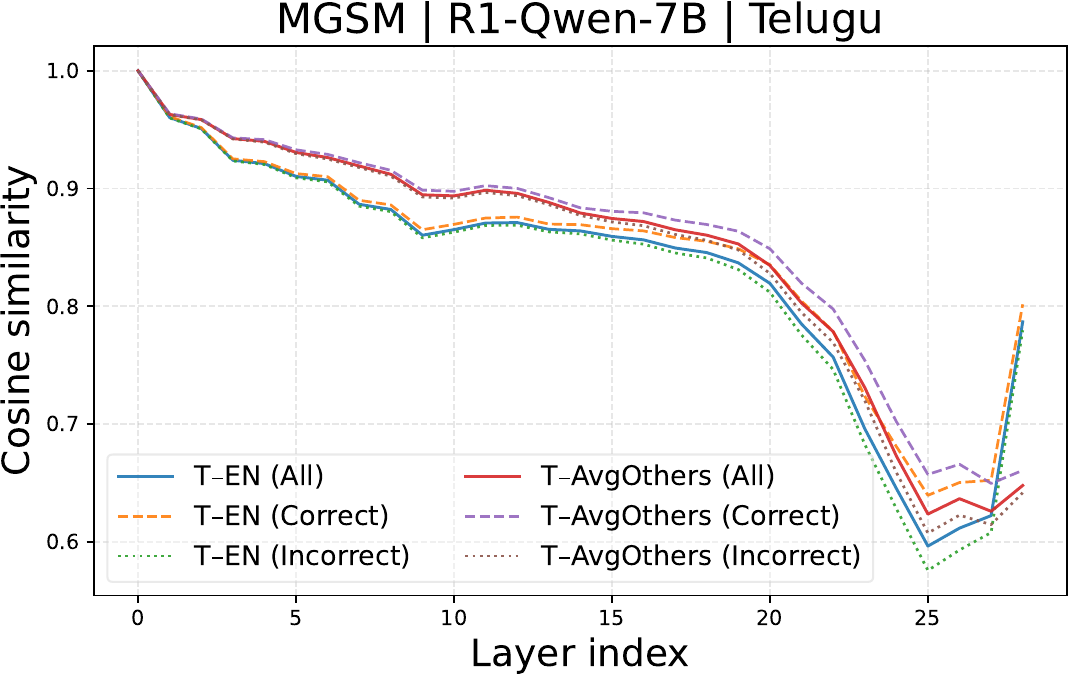}
    \includegraphics[width=0.23\textwidth]{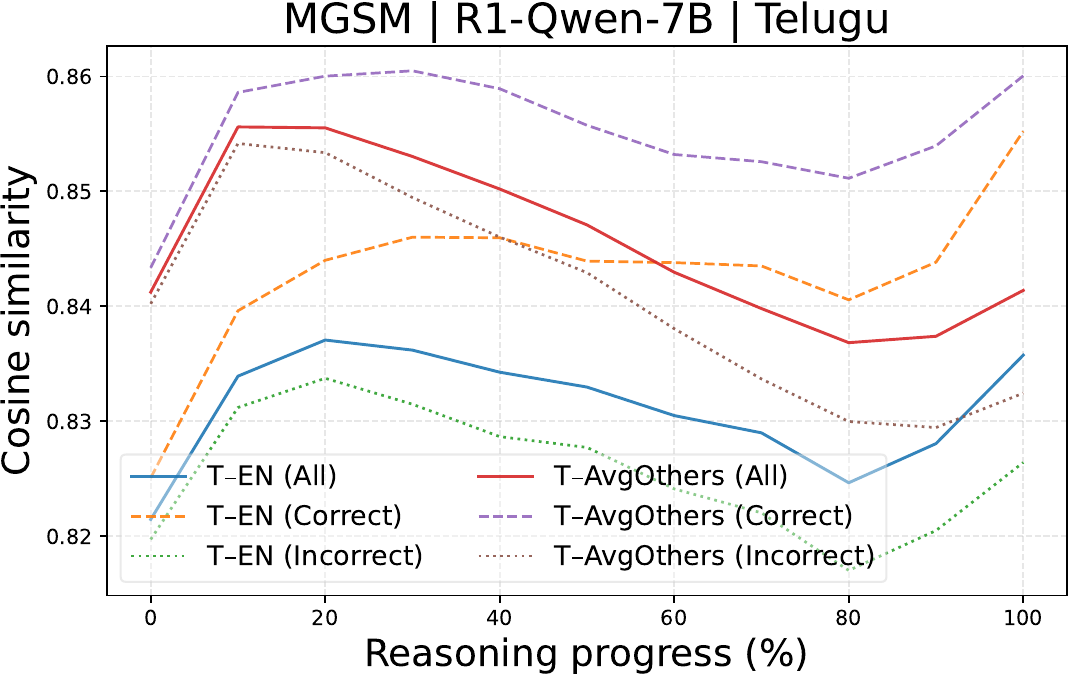}
    \includegraphics[width=0.23\textwidth]{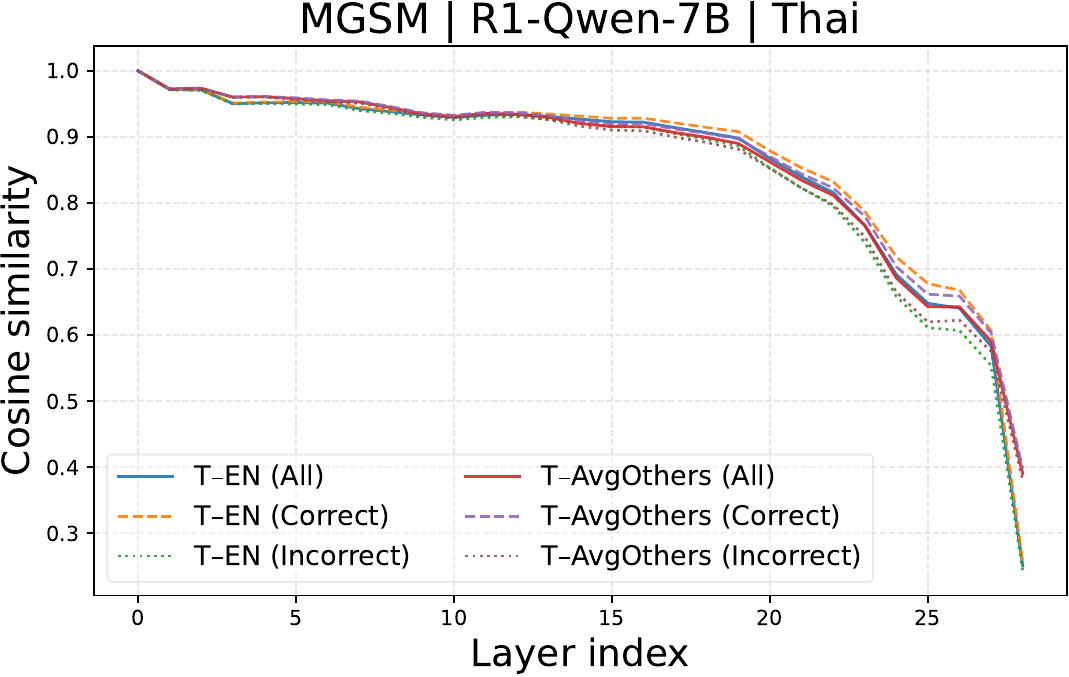}
    \includegraphics[width=0.23\textwidth]{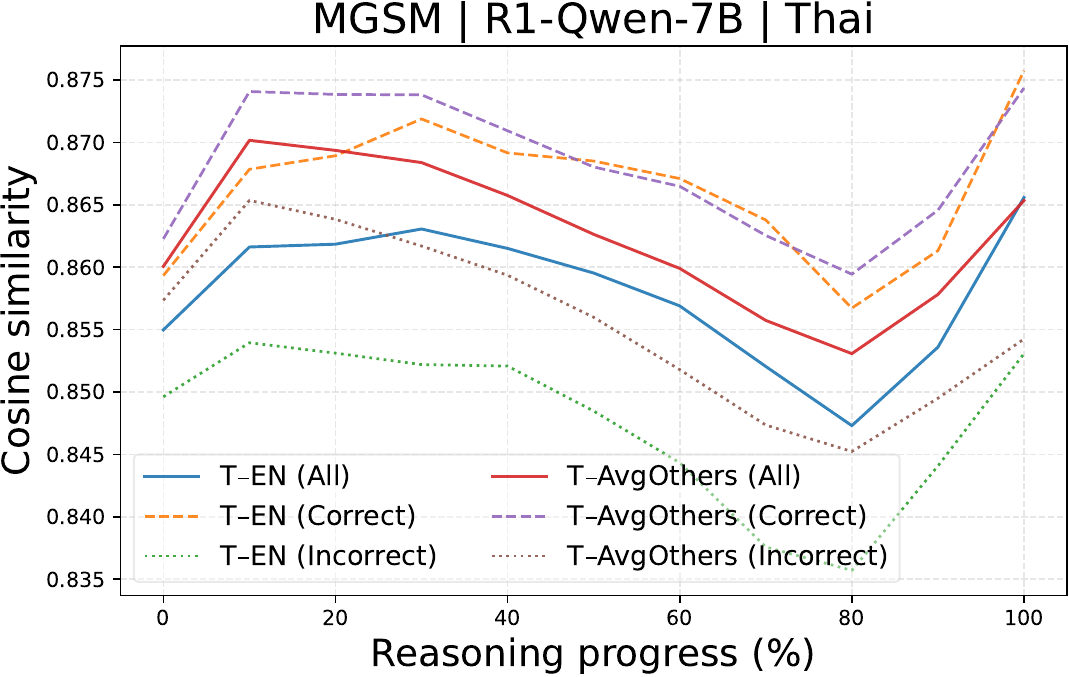}
    \includegraphics[width=0.23\textwidth]{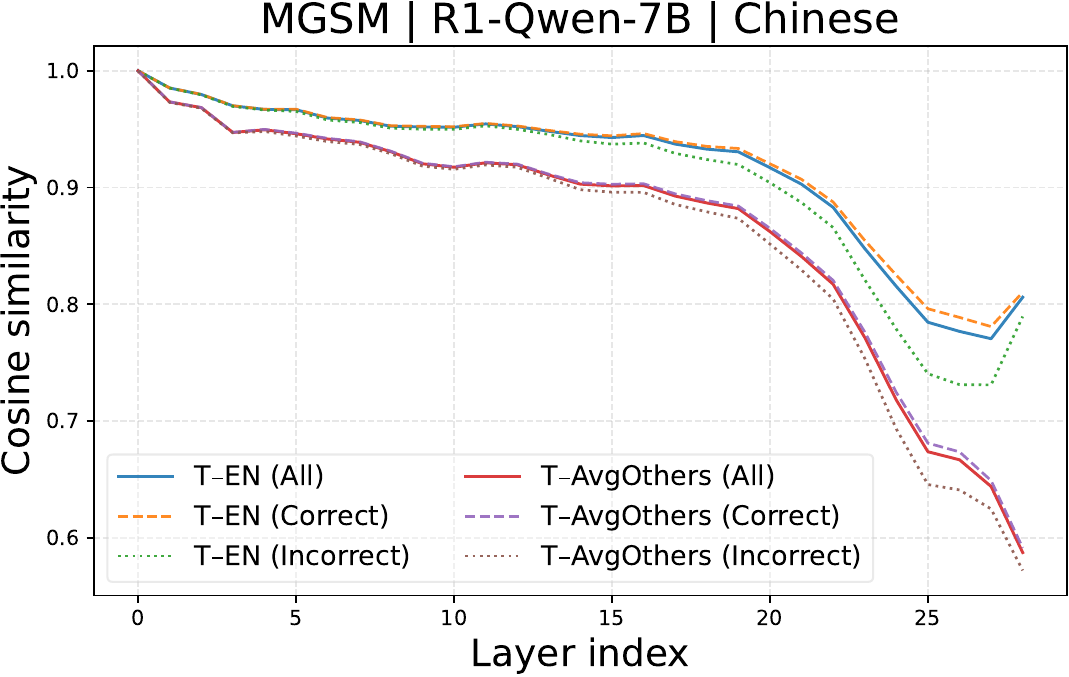}
    \includegraphics[width=0.23\textwidth]{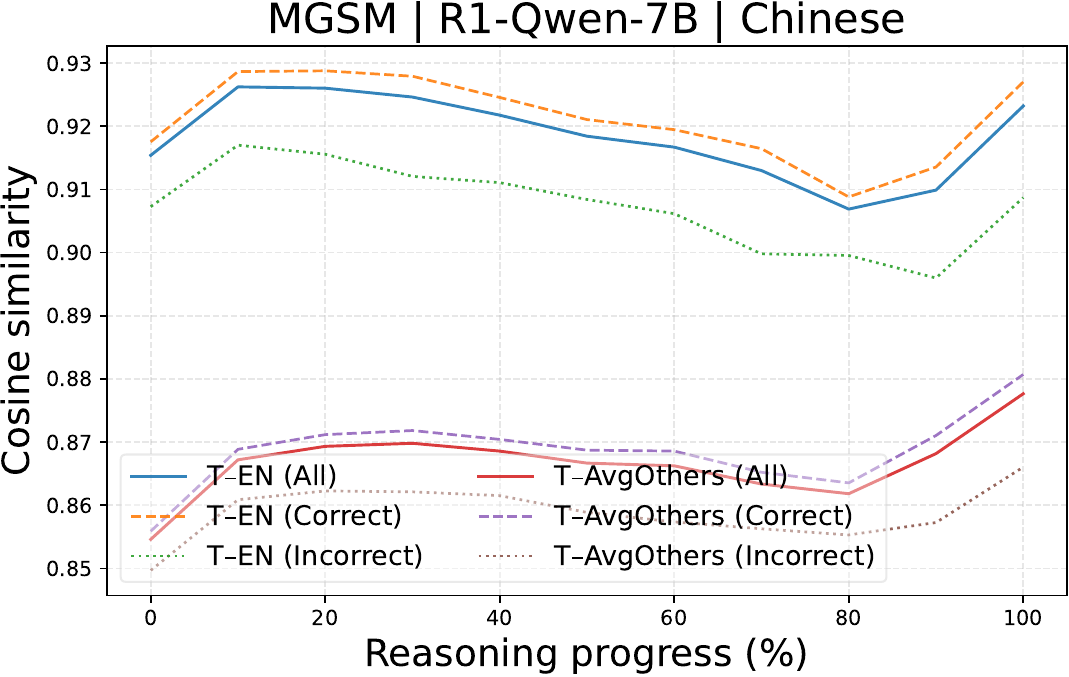}
    \caption{Comparison of cosine similarity with English versus average similarity with other languages, shown separately for correctly and incorrectly solved examples in \textbf{MGSM} with \textbf{R1-Qwen-7B}.}
    \label{fig:cosine_sim_vs_others_complete_7b}
\end{figure*}

\begin{figure*}
    \centering

    \includegraphics[width=0.23\textwidth]{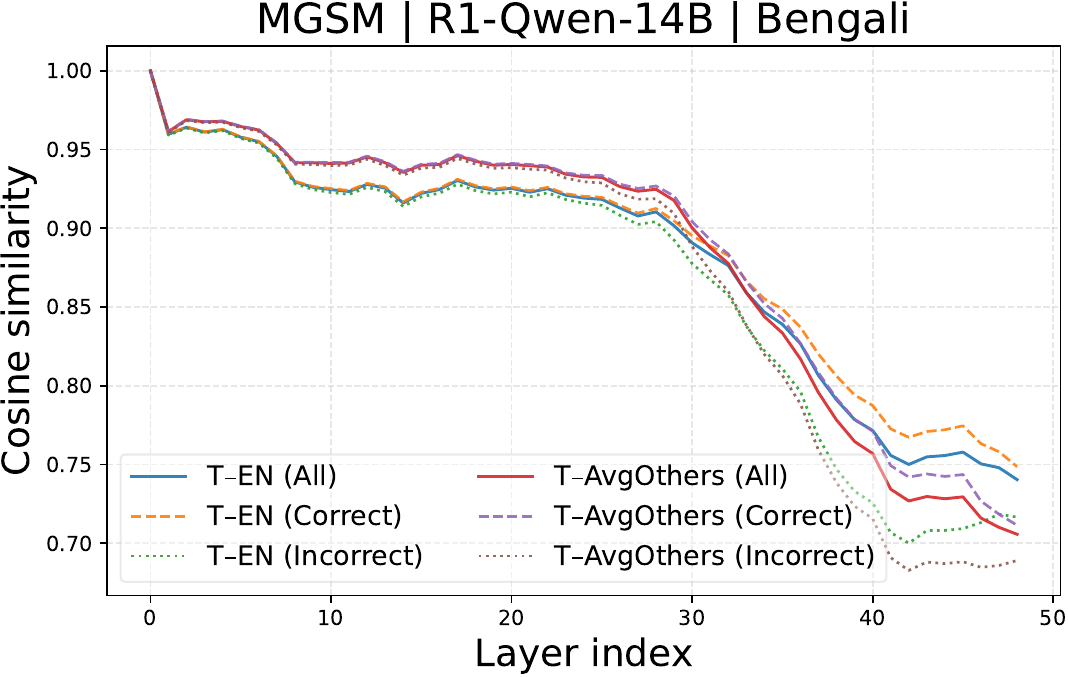}
    \includegraphics[width=0.23\textwidth]{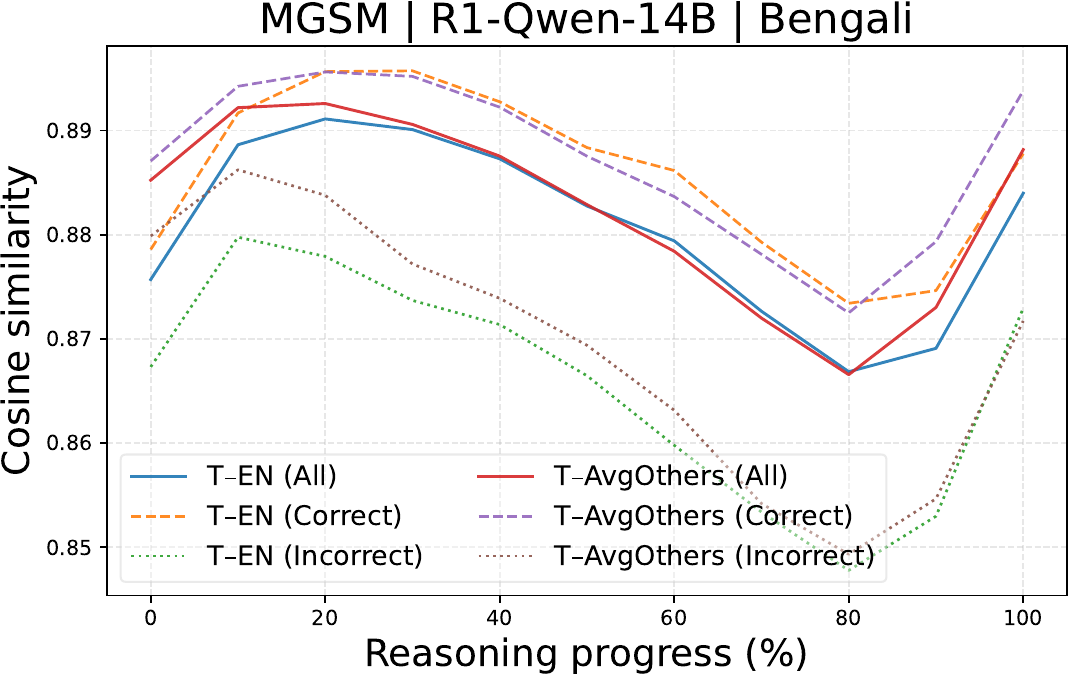}
    \includegraphics[width=0.23\textwidth]{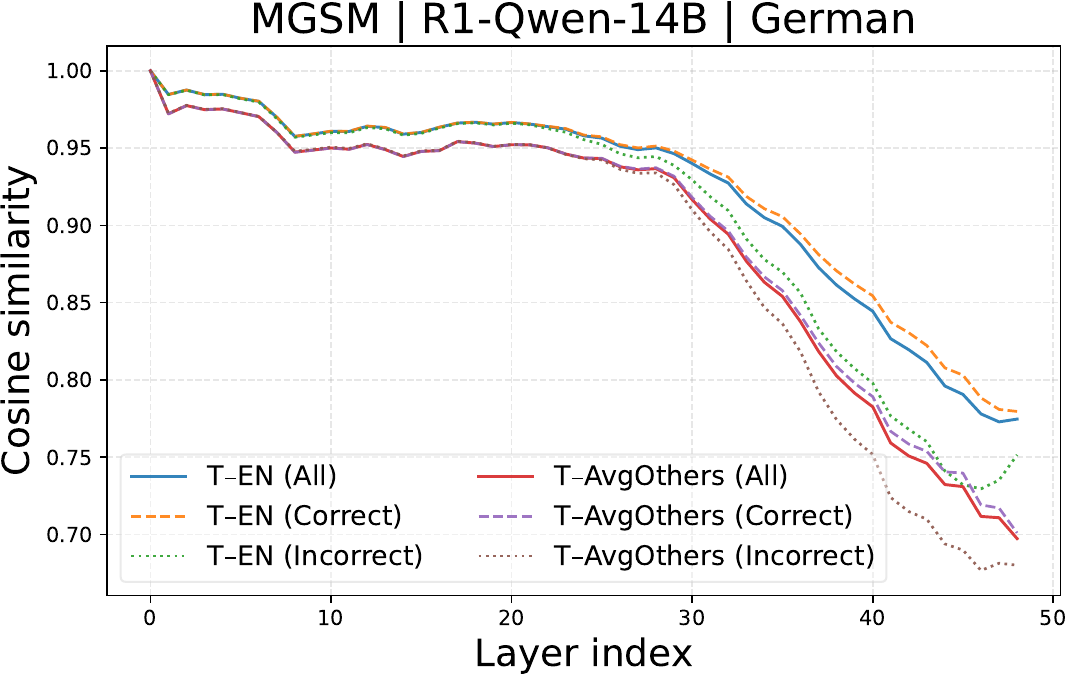}
    \includegraphics[width=0.23\textwidth]{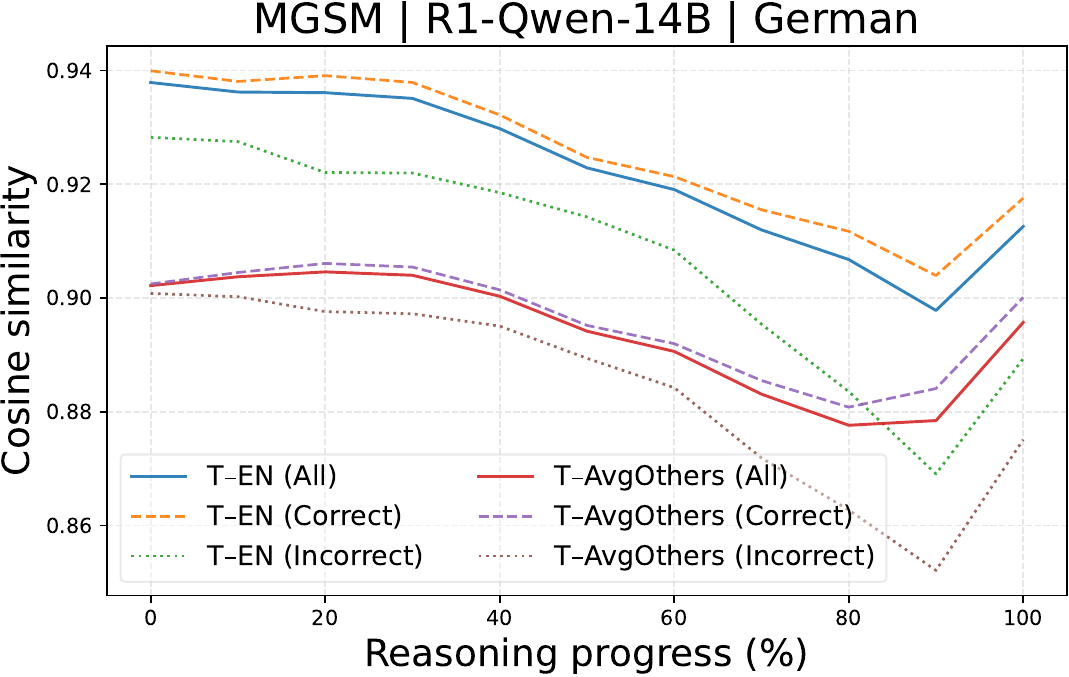}
    \includegraphics[width=0.23\textwidth]{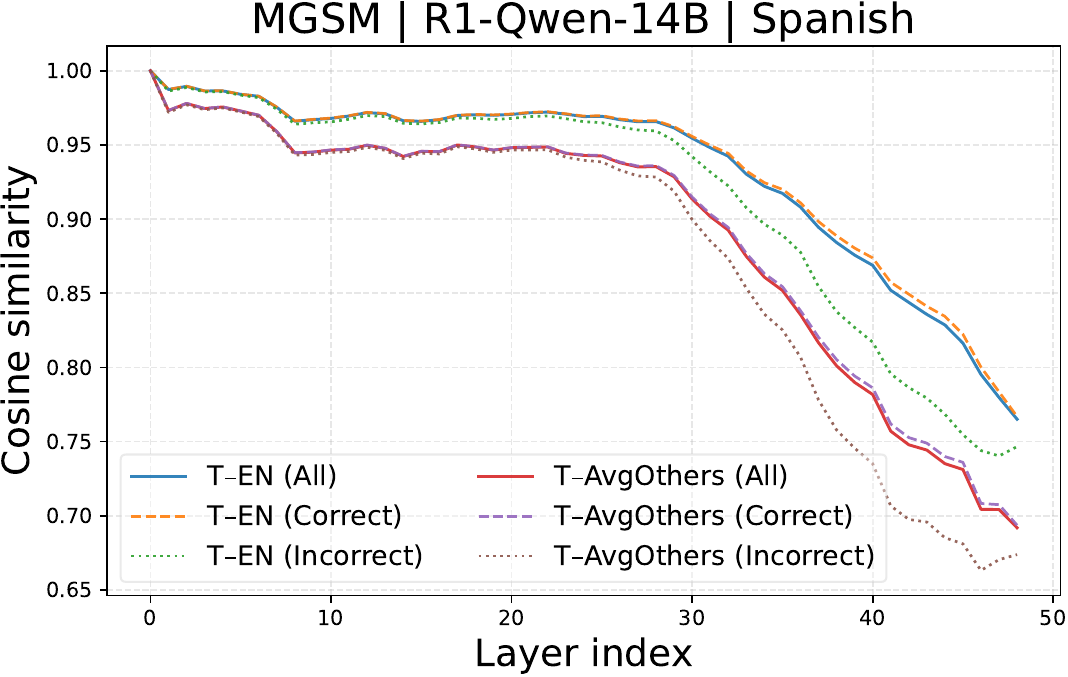}
    \includegraphics[width=0.23\textwidth]{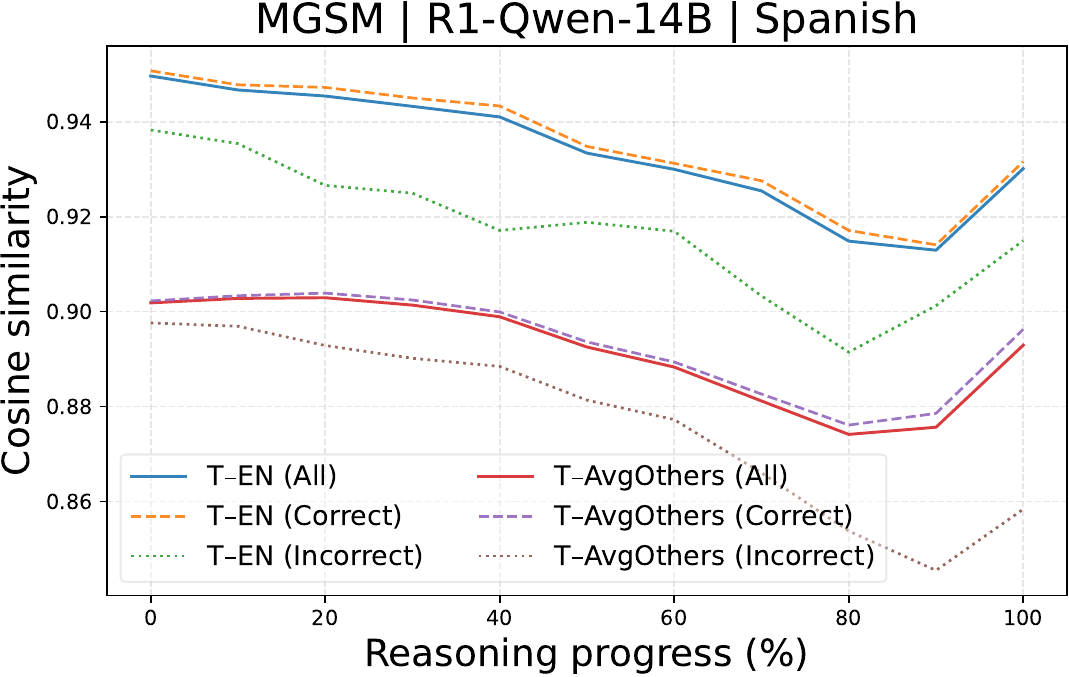}
    \includegraphics[width=0.23\textwidth]{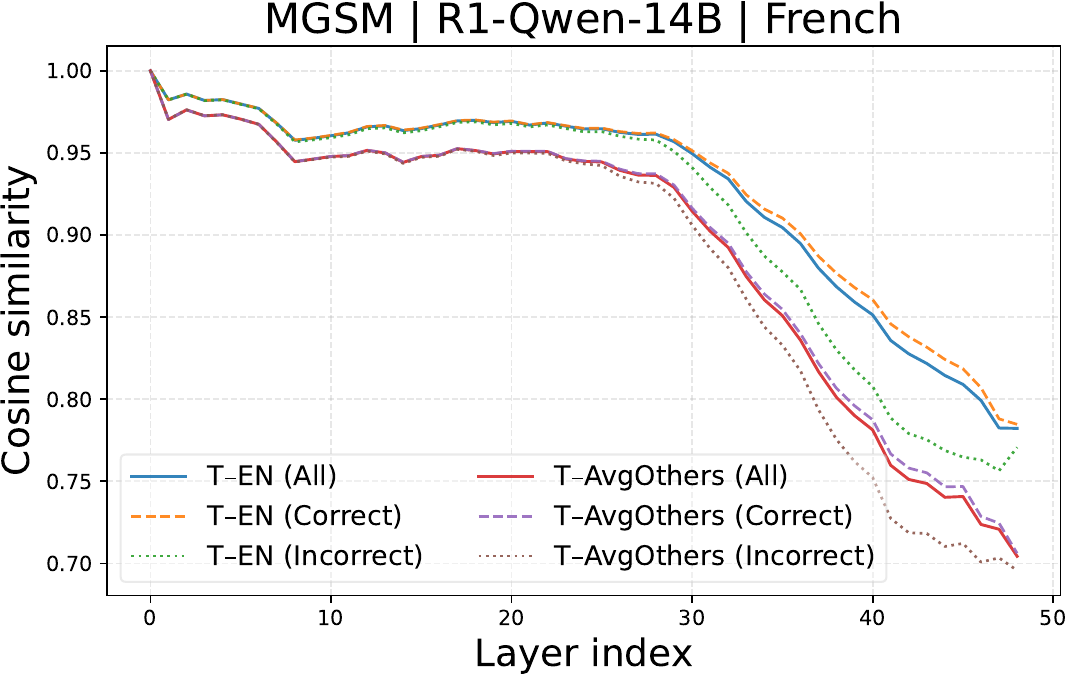}
    \includegraphics[width=0.23\textwidth]{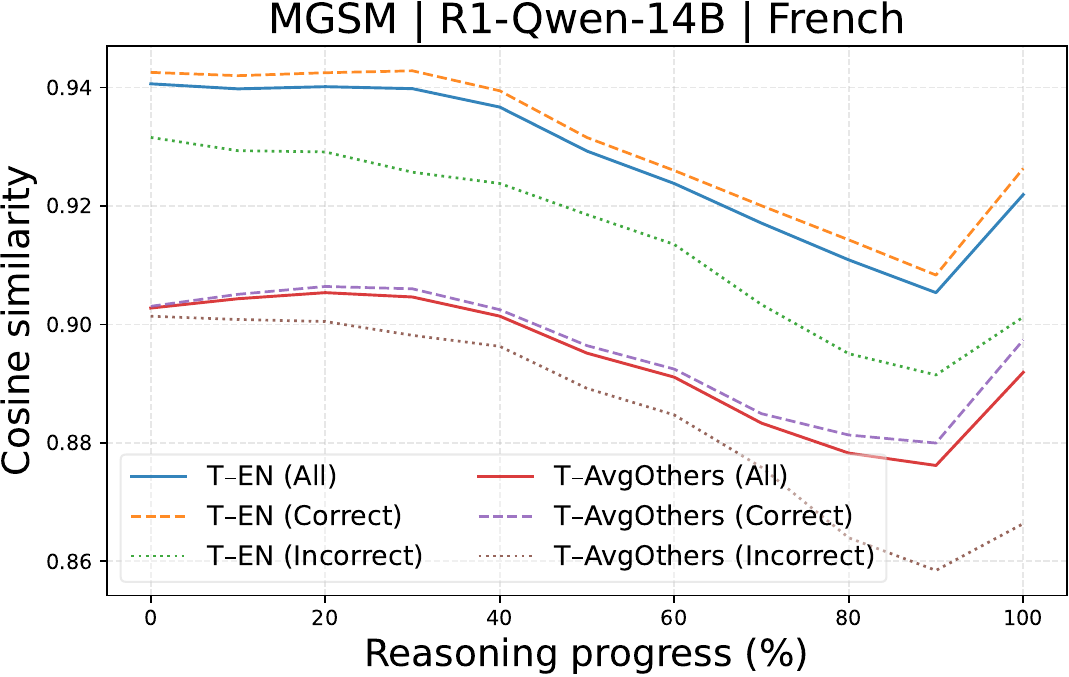}
    \includegraphics[width=0.23\textwidth]{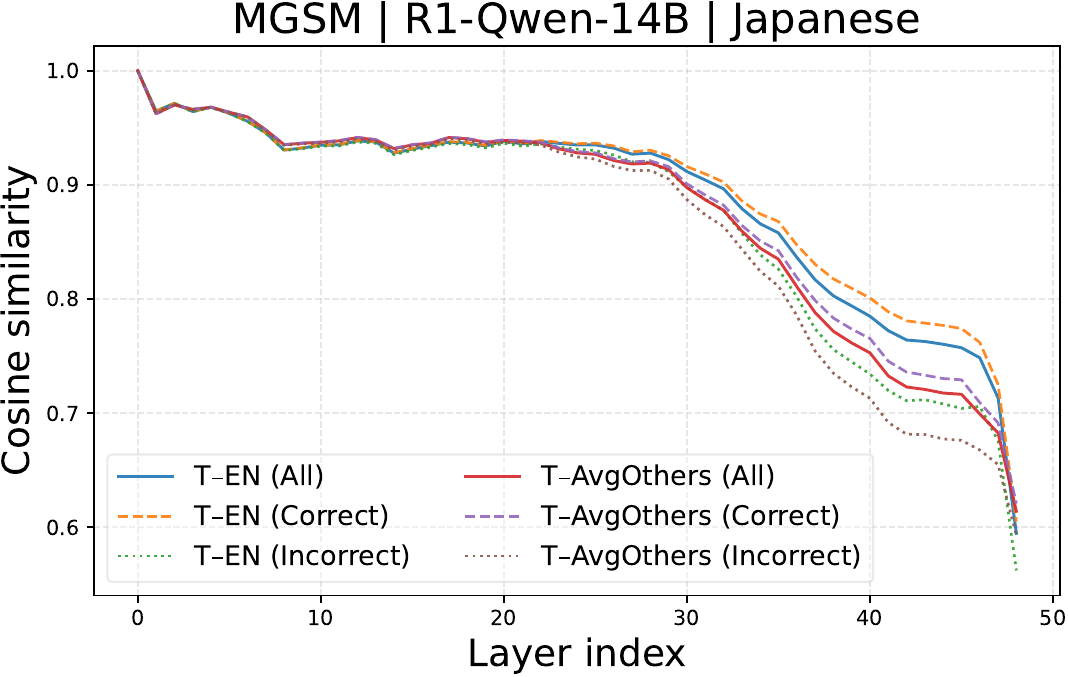}
    \includegraphics[width=0.23\textwidth]{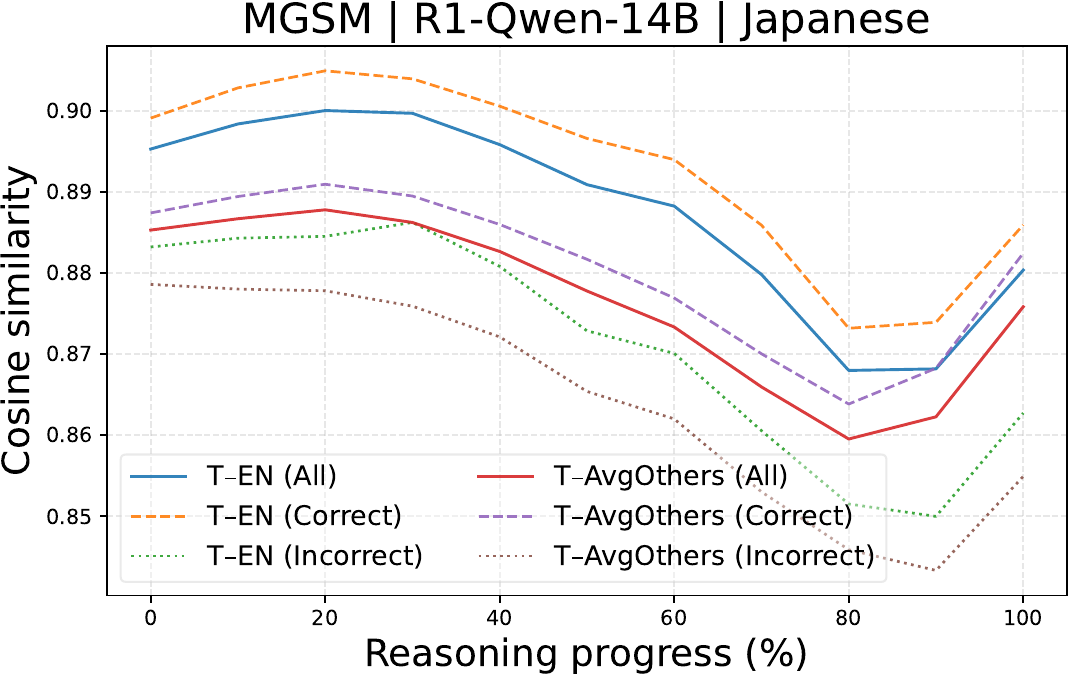}
    \includegraphics[width=0.23\textwidth]{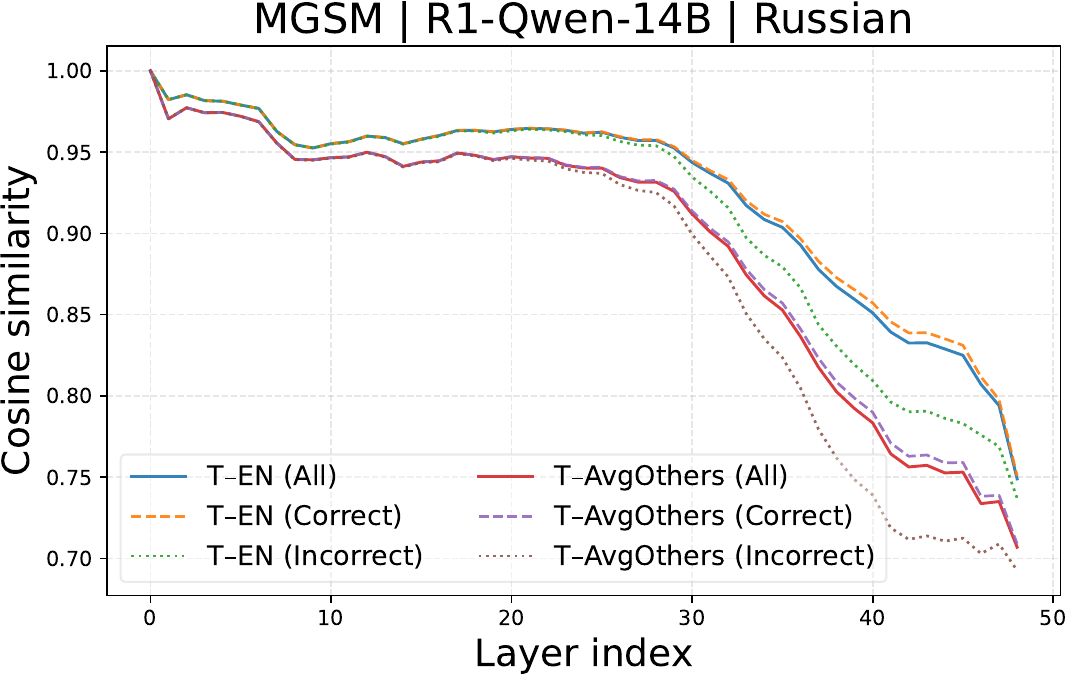}
    \includegraphics[width=0.23\textwidth]{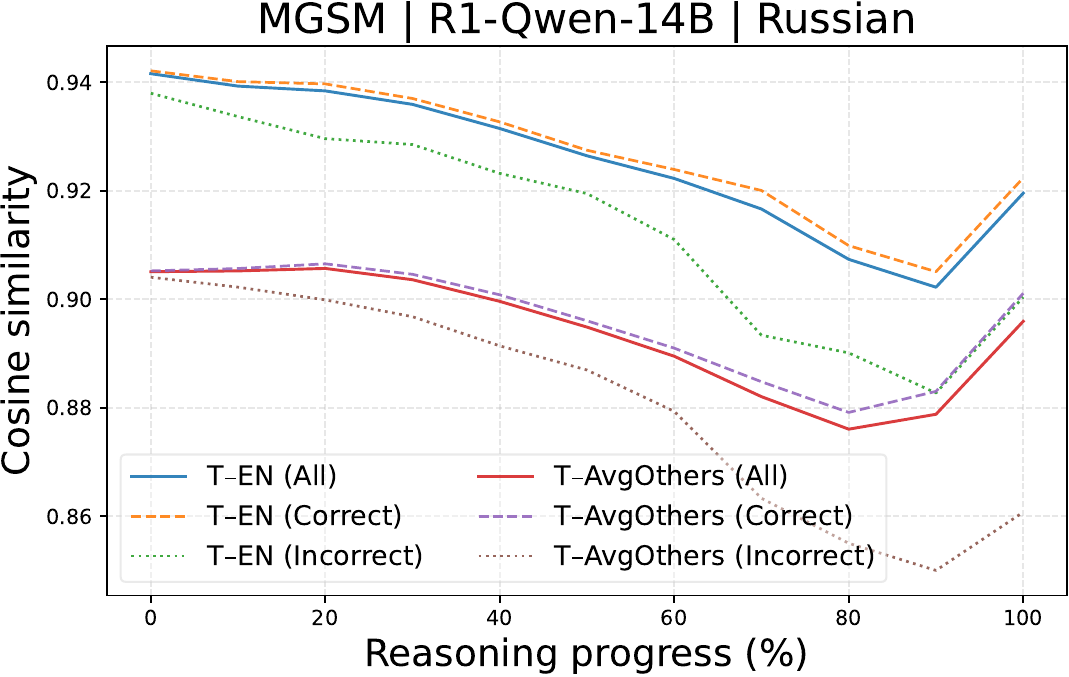}
    \includegraphics[width=0.23\textwidth]{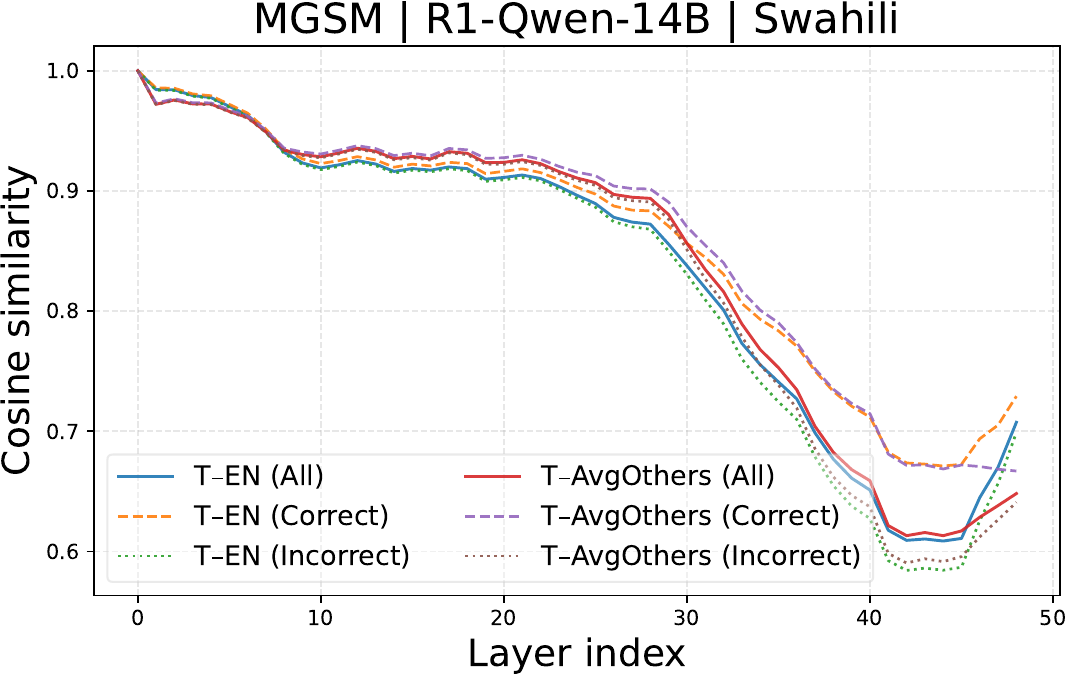}
    \includegraphics[width=0.23\textwidth]{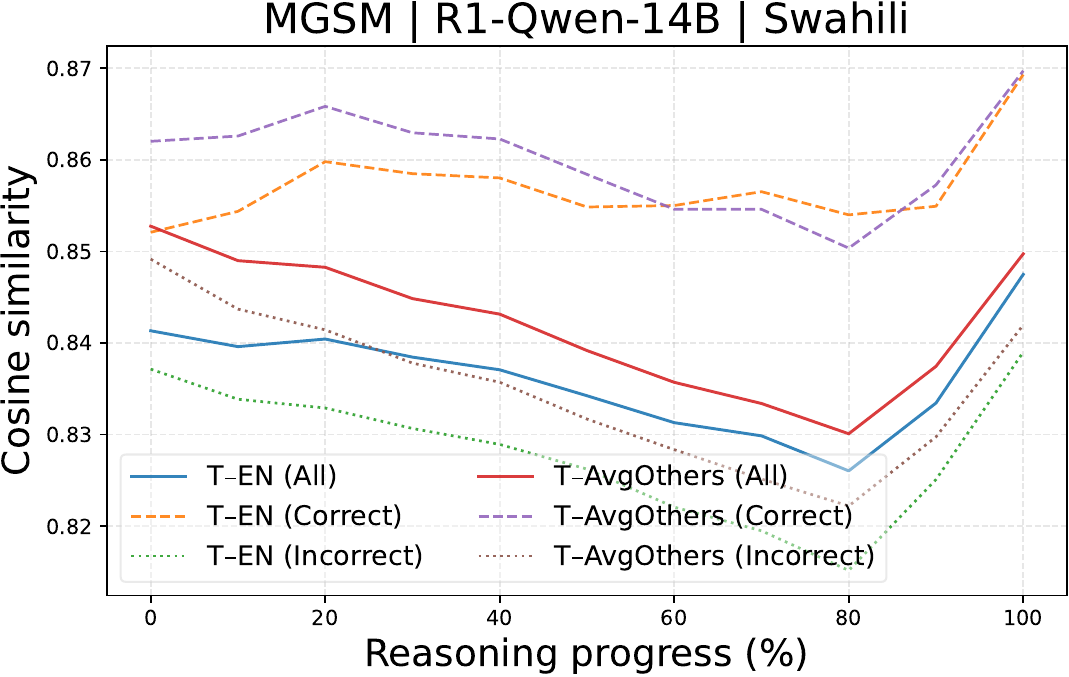}
    \includegraphics[width=0.23\textwidth]{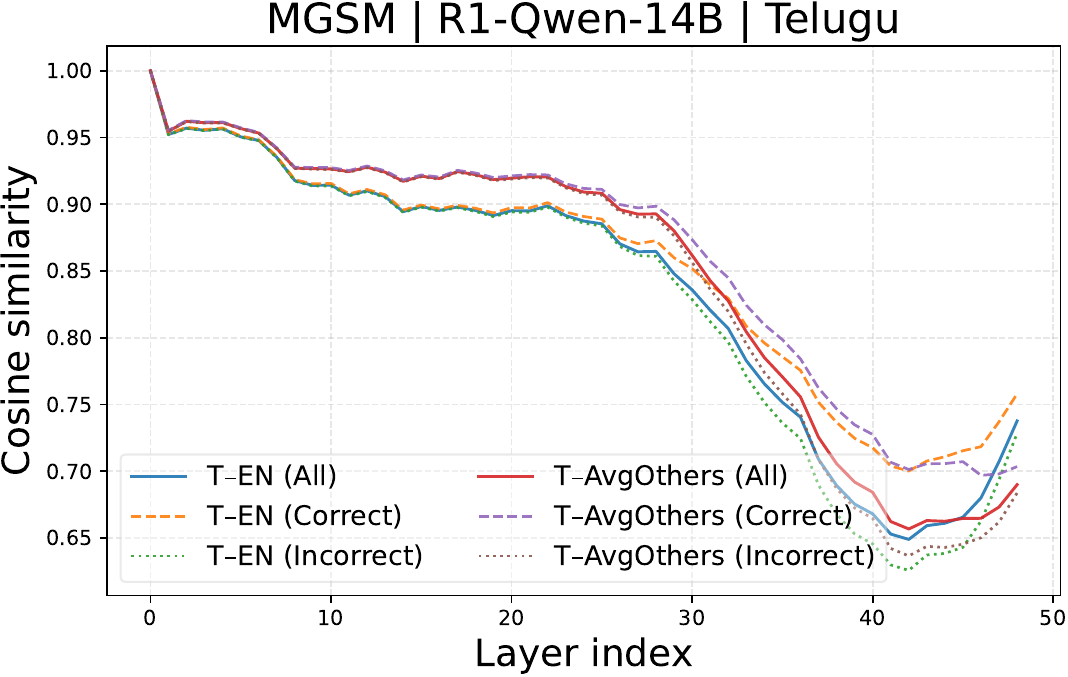}
    \includegraphics[width=0.23\textwidth]{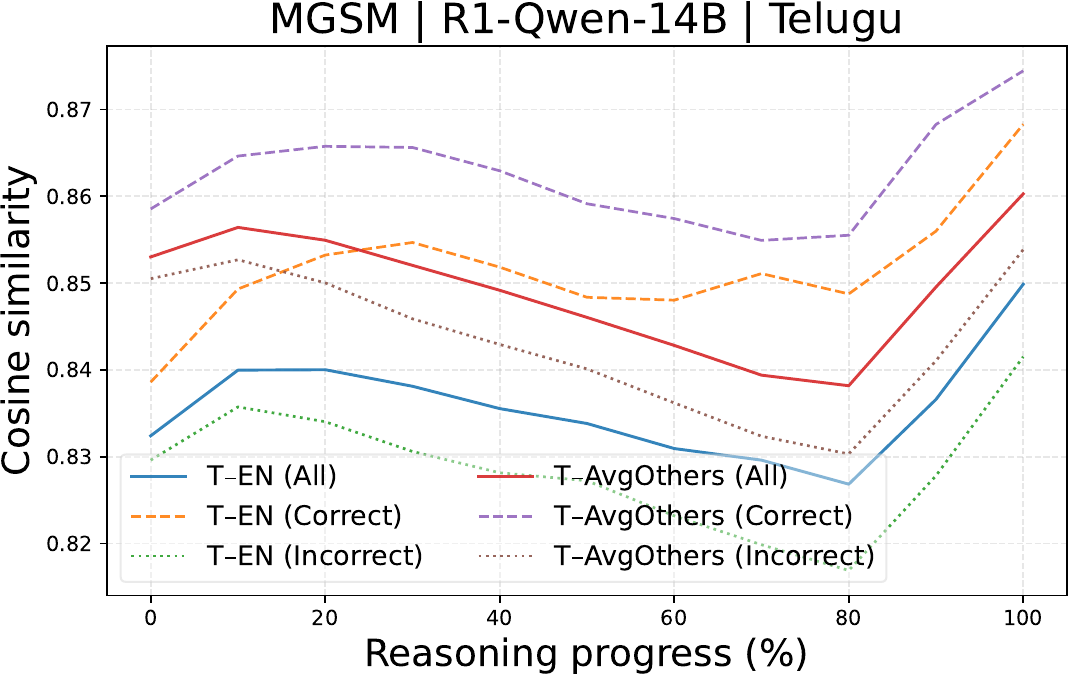}
    \includegraphics[width=0.23\textwidth]{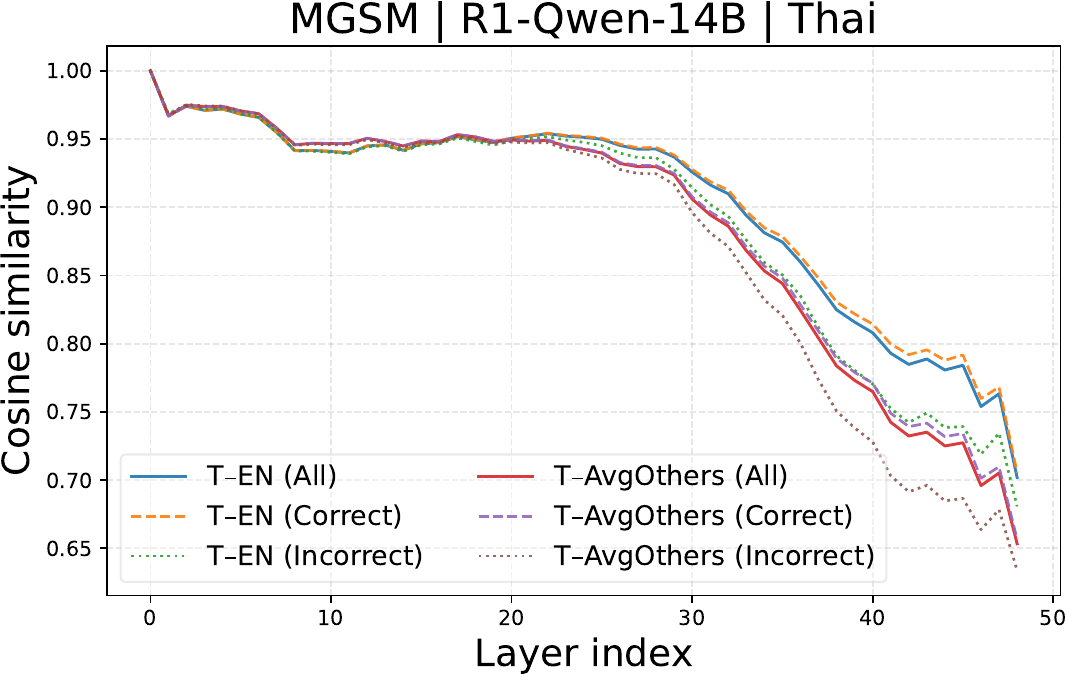}
    \includegraphics[width=0.23\textwidth]{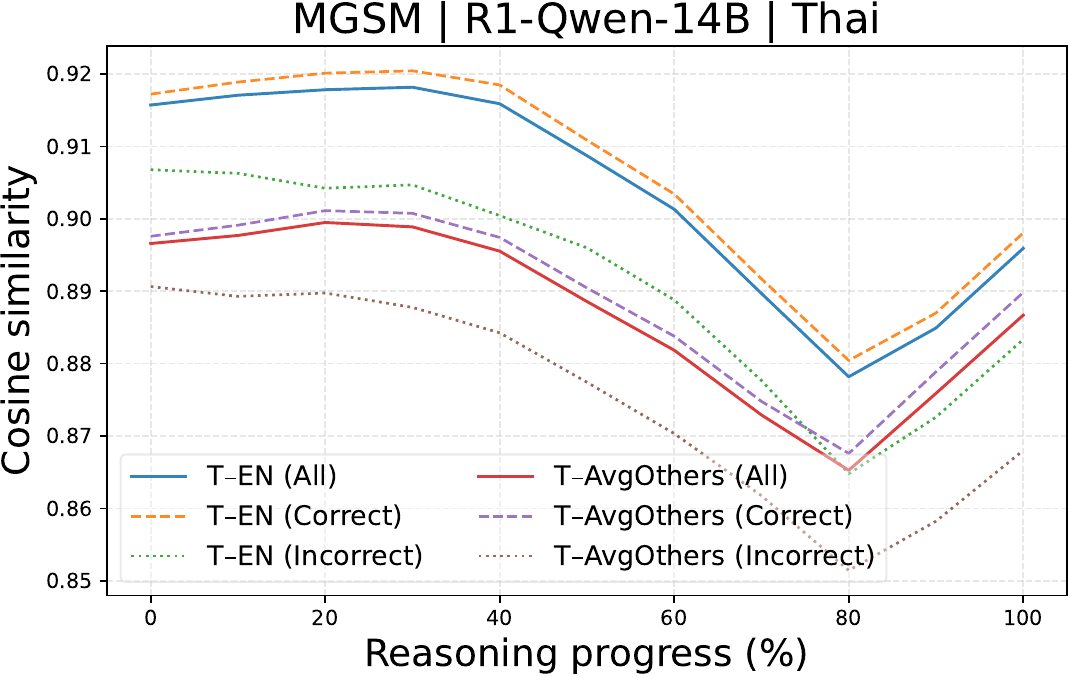}
    \includegraphics[width=0.23\textwidth]{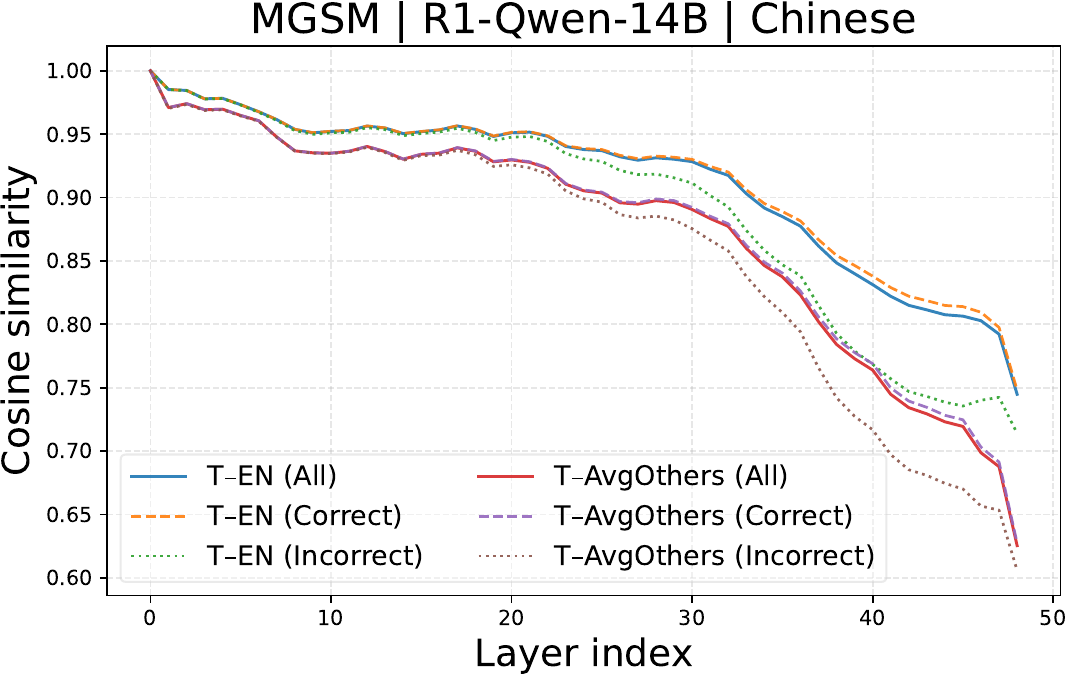}
    \includegraphics[width=0.23\textwidth]{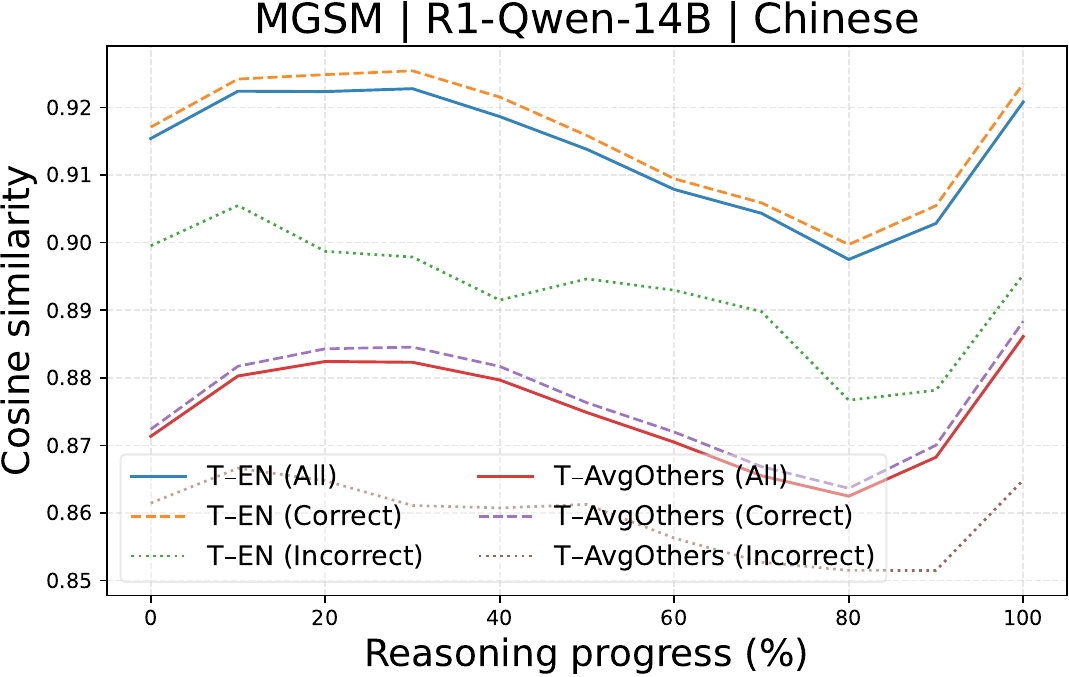}
    \caption{Comparison of cosine similarity with English versus average similarity with other languages, shown separately for correctly and incorrectly solved examples in \textbf{MGSM} with \textbf{R1-Qwen-14B}.}
    \label{fig:cosine_sim_vs_others_complete_14b}
\end{figure*}

\begin{figure*}
    \centering

    \includegraphics[width=0.23\textwidth]{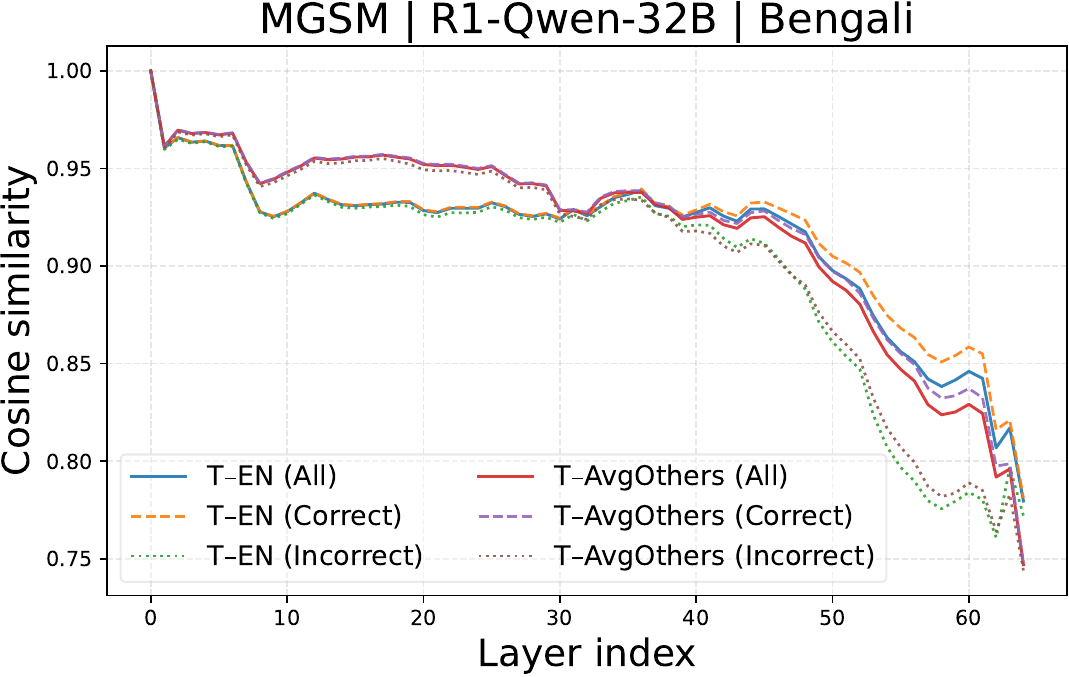}
    \includegraphics[width=0.23\textwidth]{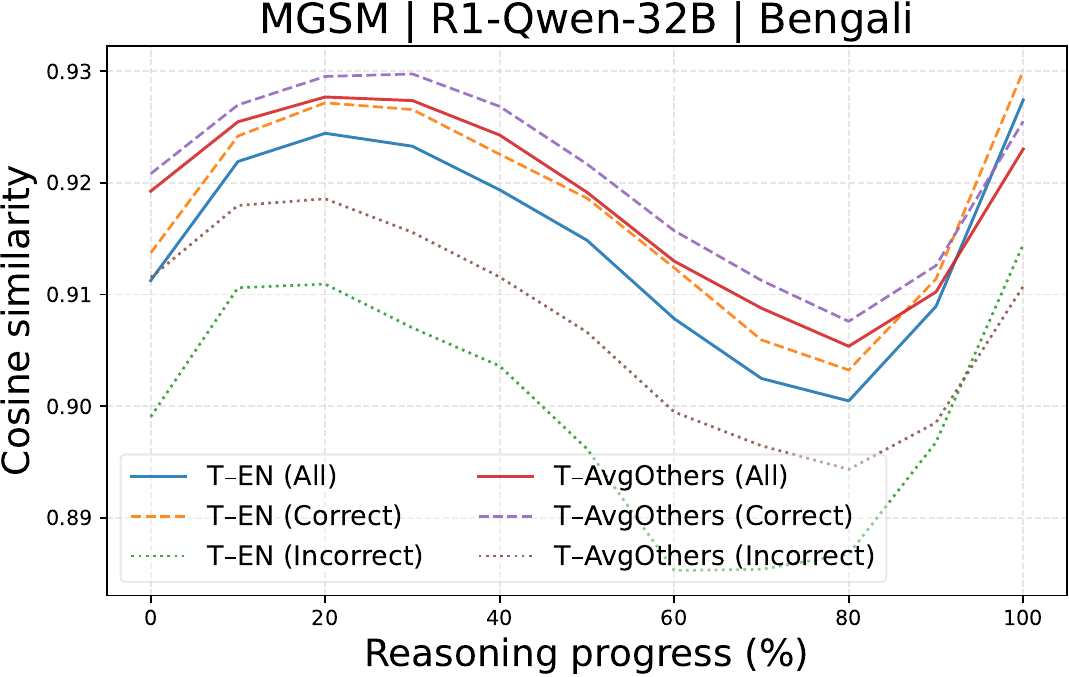}
    \includegraphics[width=0.23\textwidth]{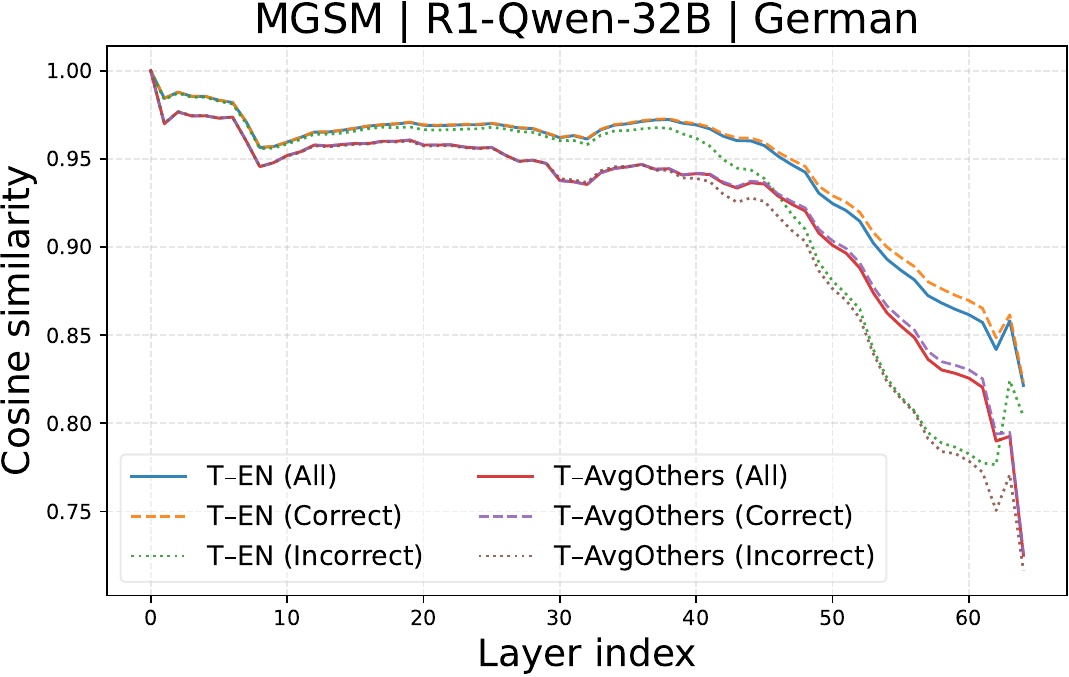}
    \includegraphics[width=0.23\textwidth]{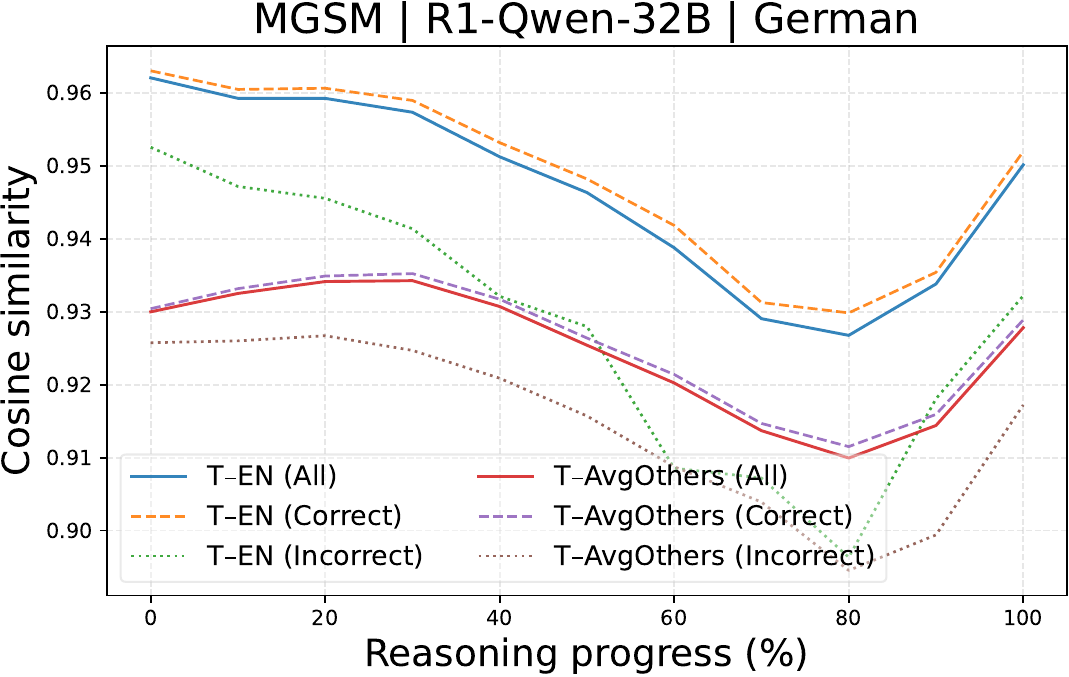}
    \includegraphics[width=0.23\textwidth]{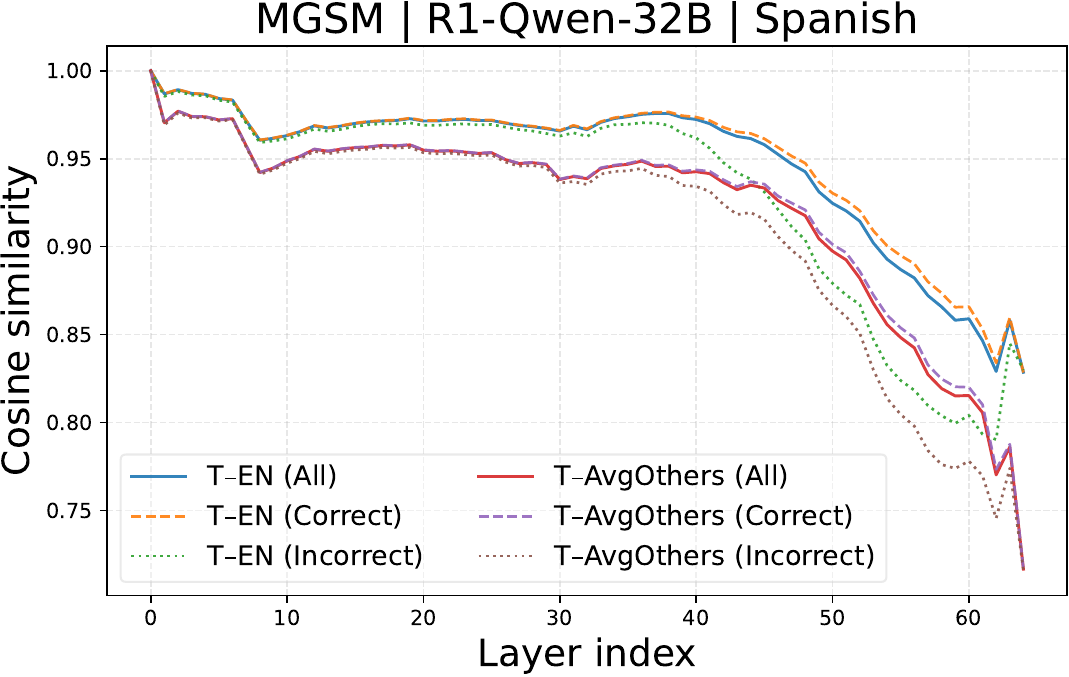}
    \includegraphics[width=0.23\textwidth]{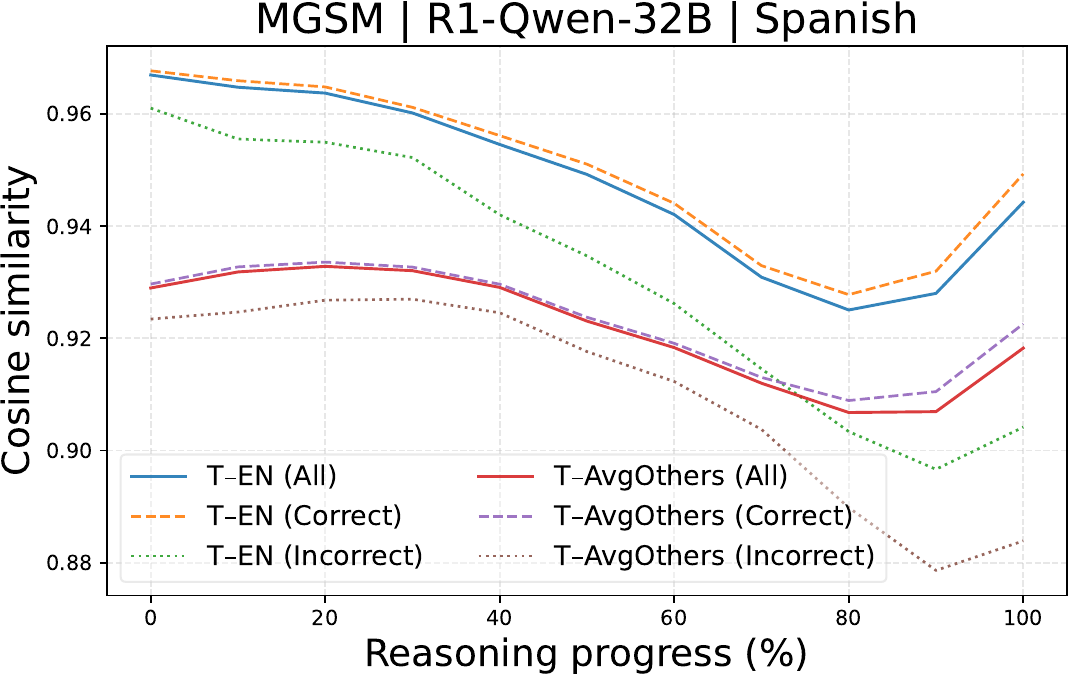}
    \includegraphics[width=0.23\textwidth]{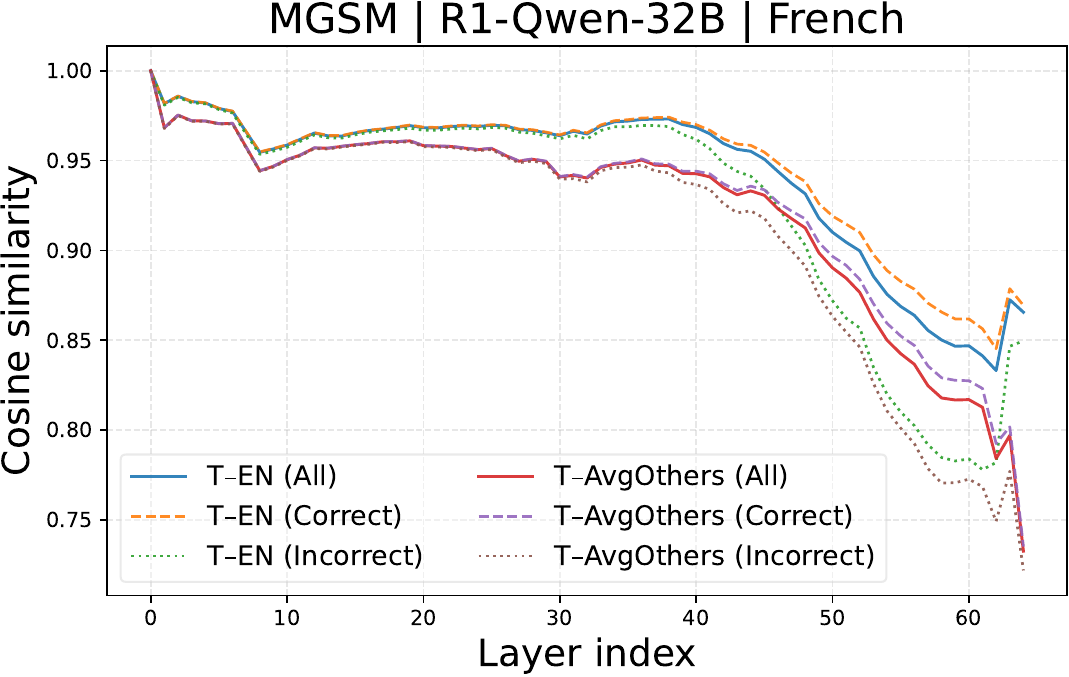}
    \includegraphics[width=0.23\textwidth]{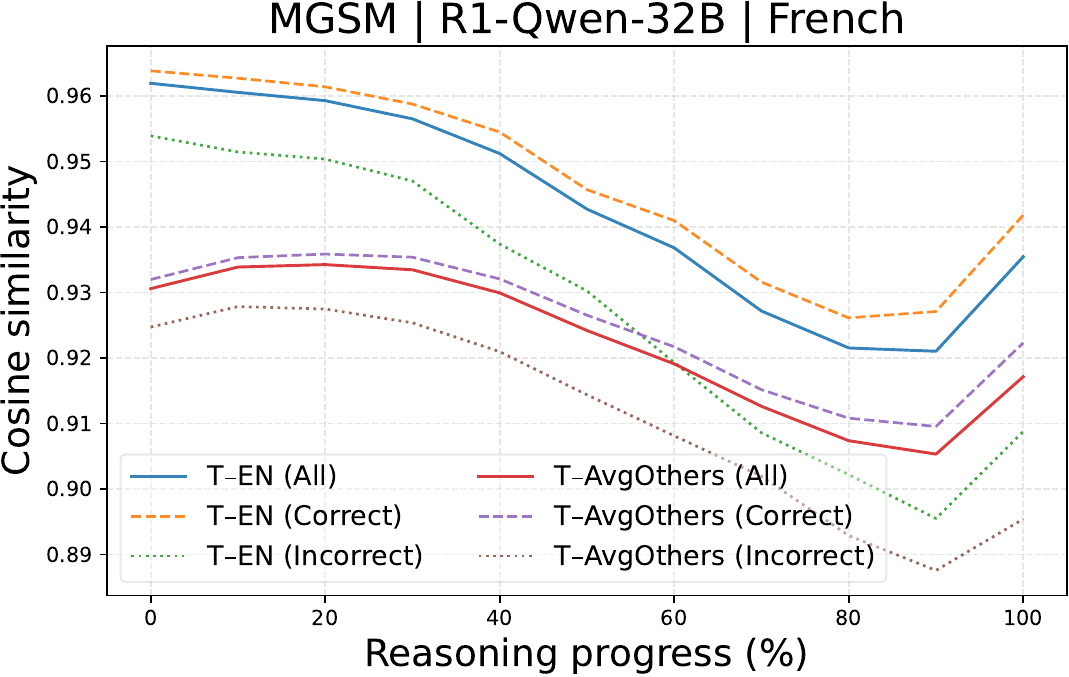}
    \includegraphics[width=0.23\textwidth]{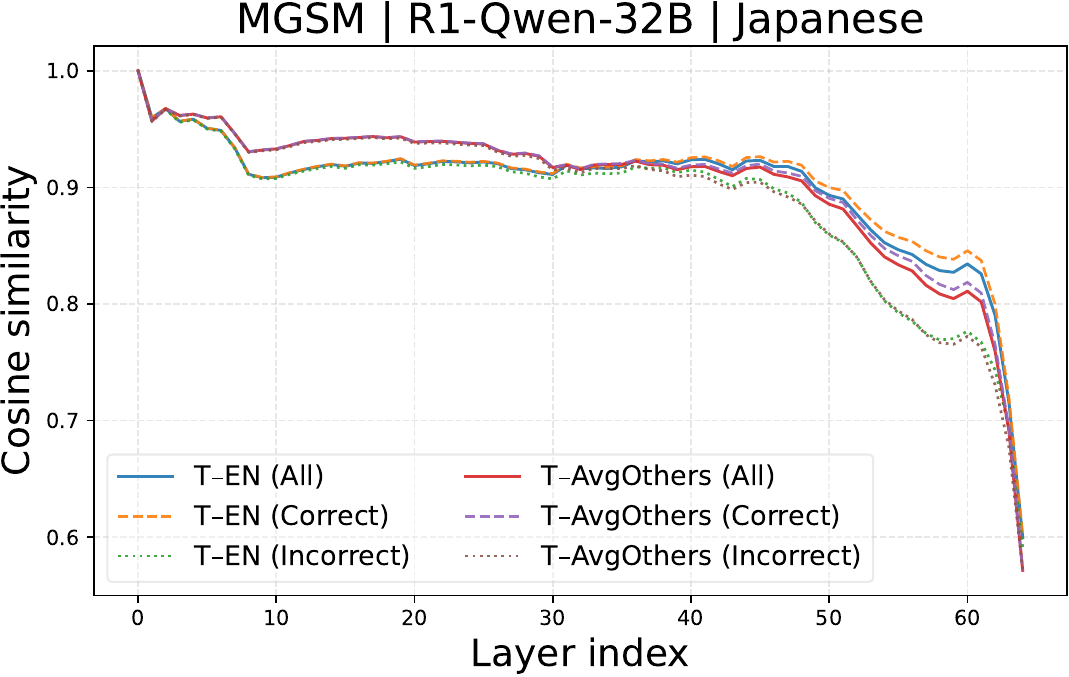}
    \includegraphics[width=0.23\textwidth]{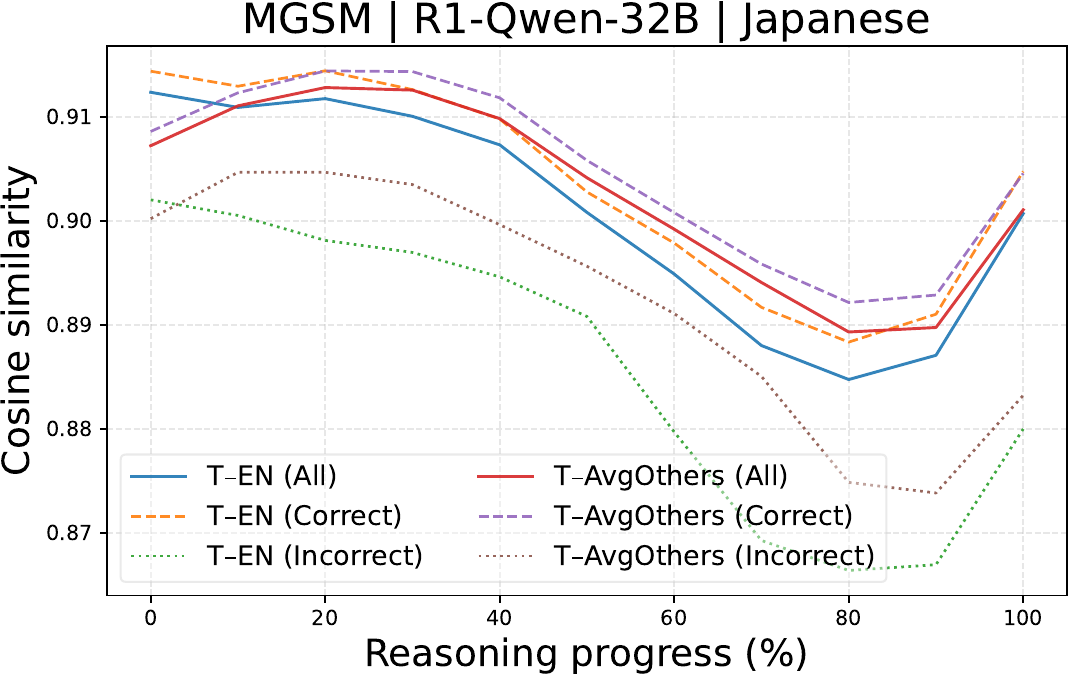}
    \includegraphics[width=0.23\textwidth]{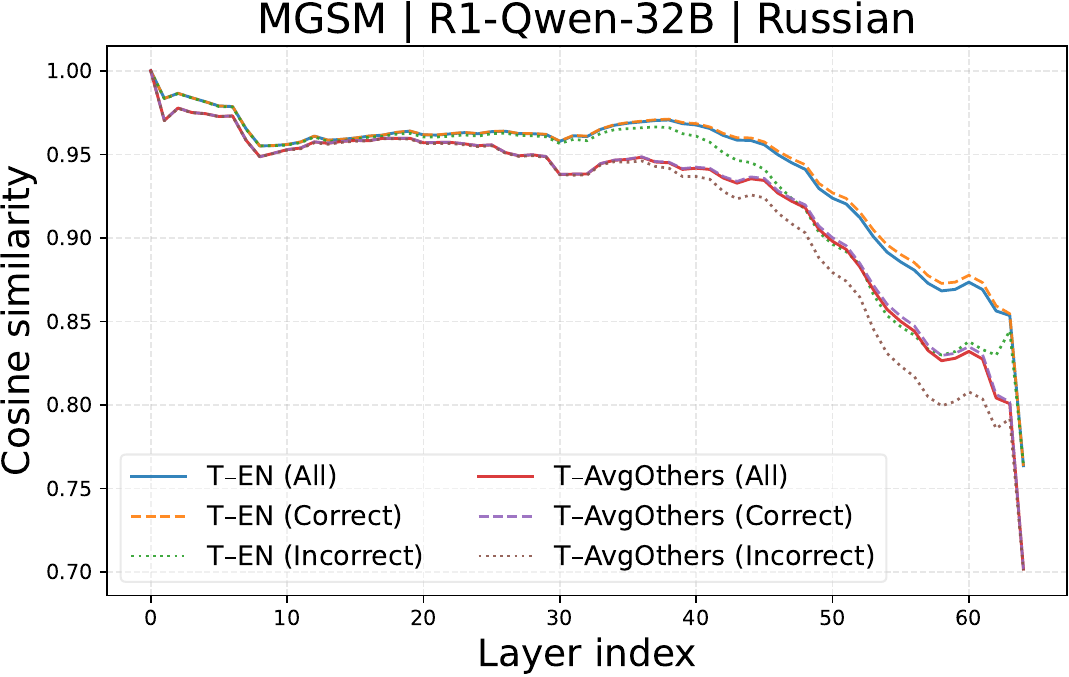}
    \includegraphics[width=0.23\textwidth]{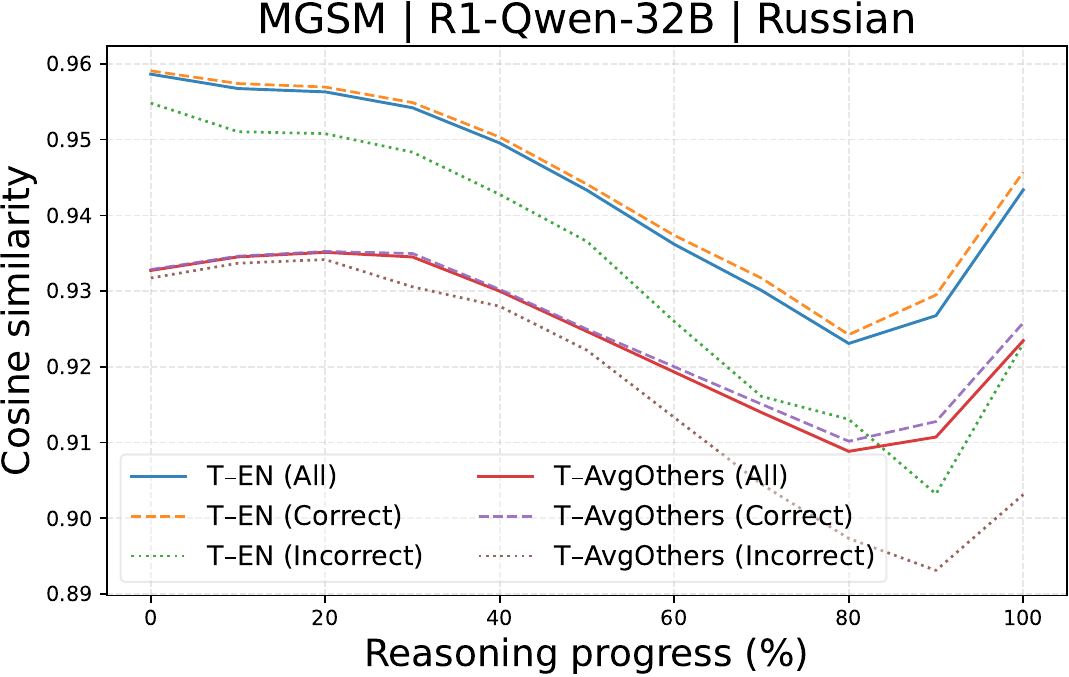}
    \includegraphics[width=0.23\textwidth]{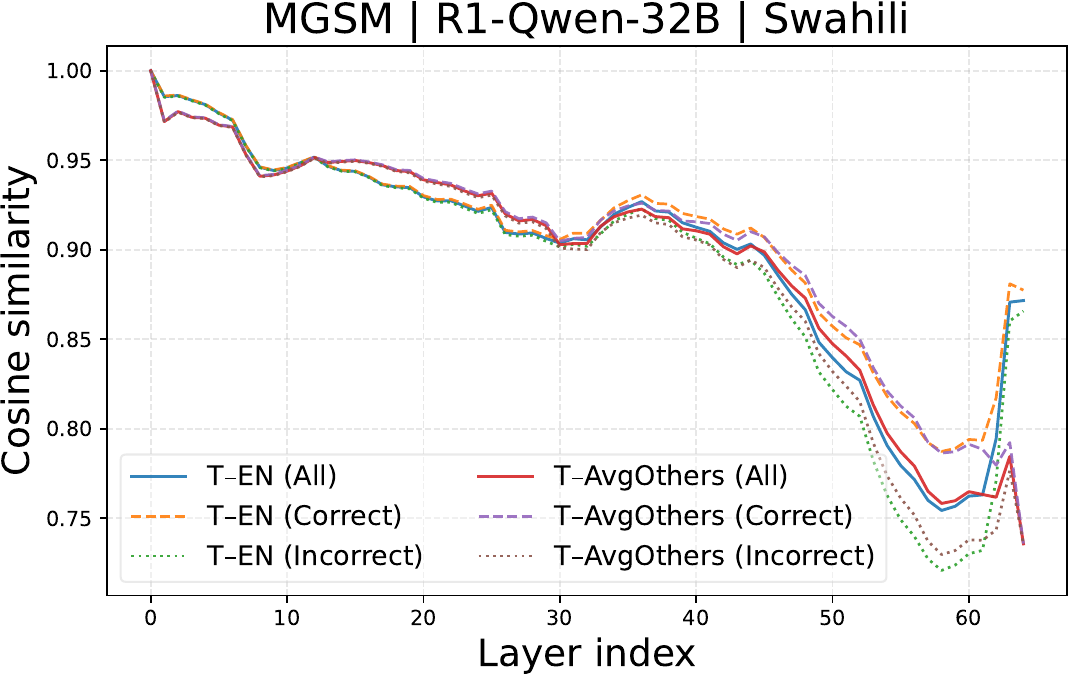}
    \includegraphics[width=0.23\textwidth]{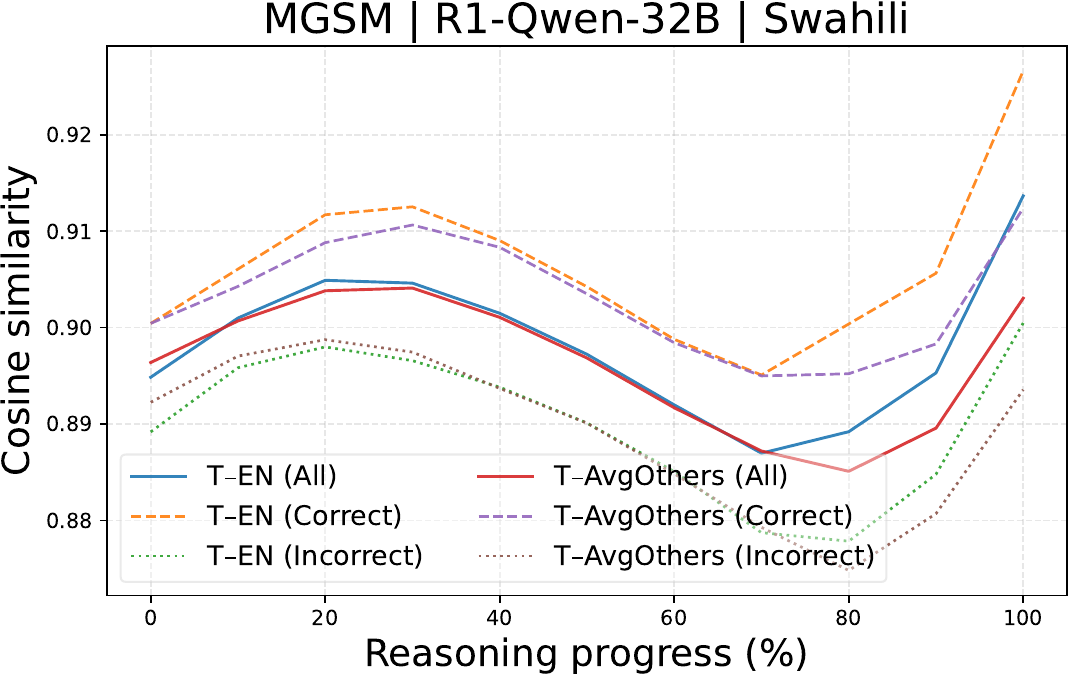}
    \includegraphics[width=0.23\textwidth]{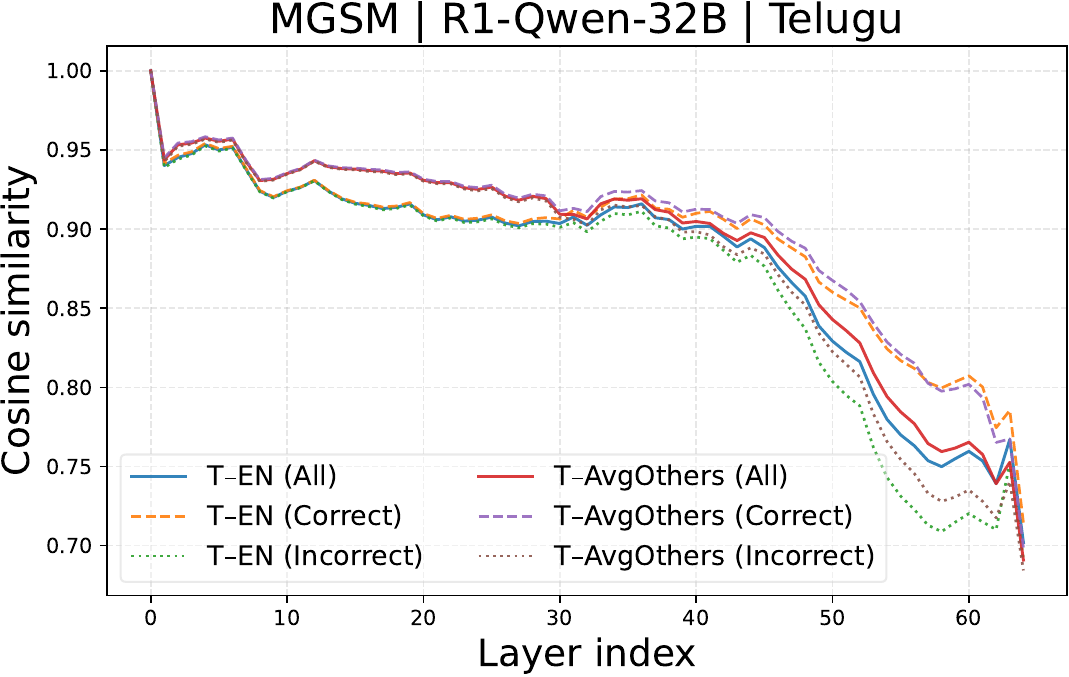}
    \includegraphics[width=0.23\textwidth]{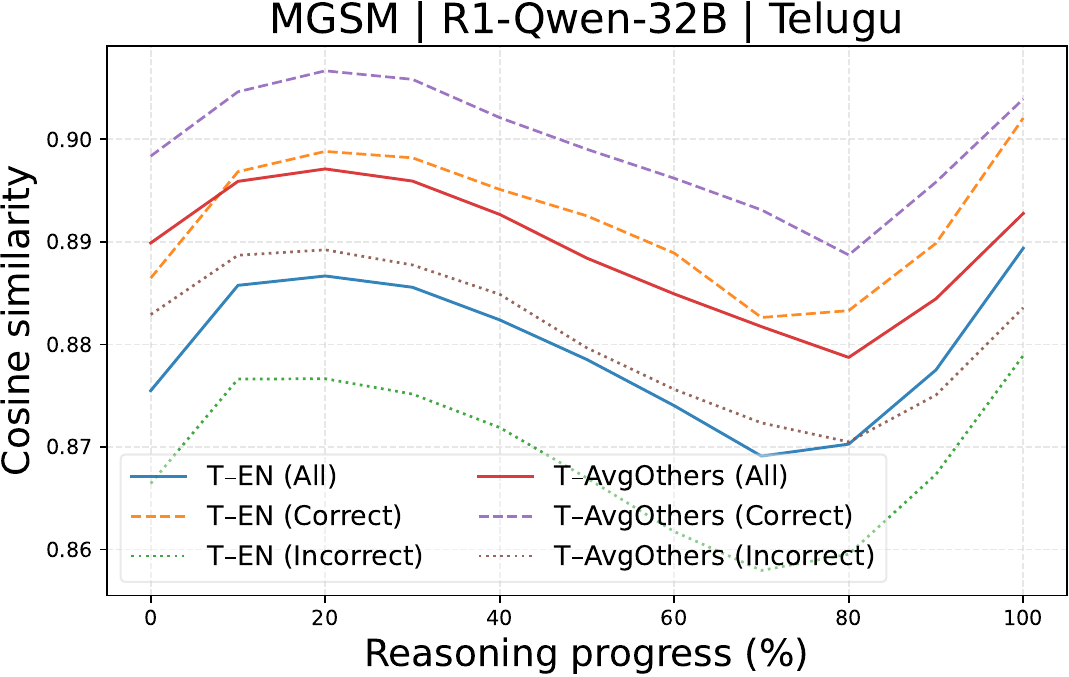}
    \includegraphics[width=0.23\textwidth]{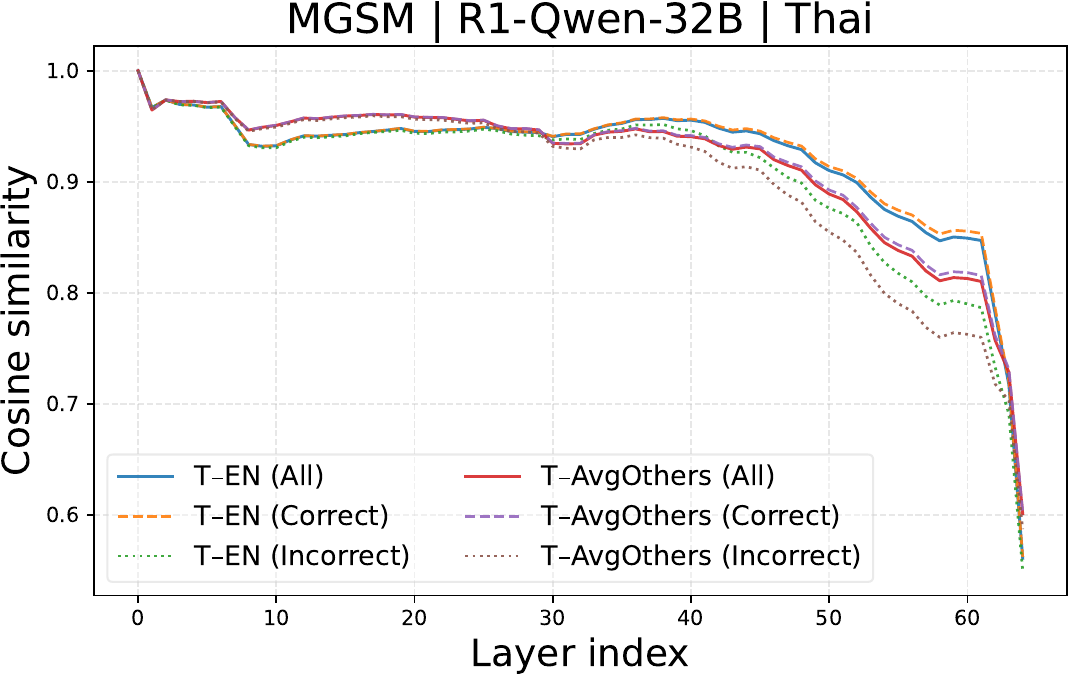}
    \includegraphics[width=0.23\textwidth]{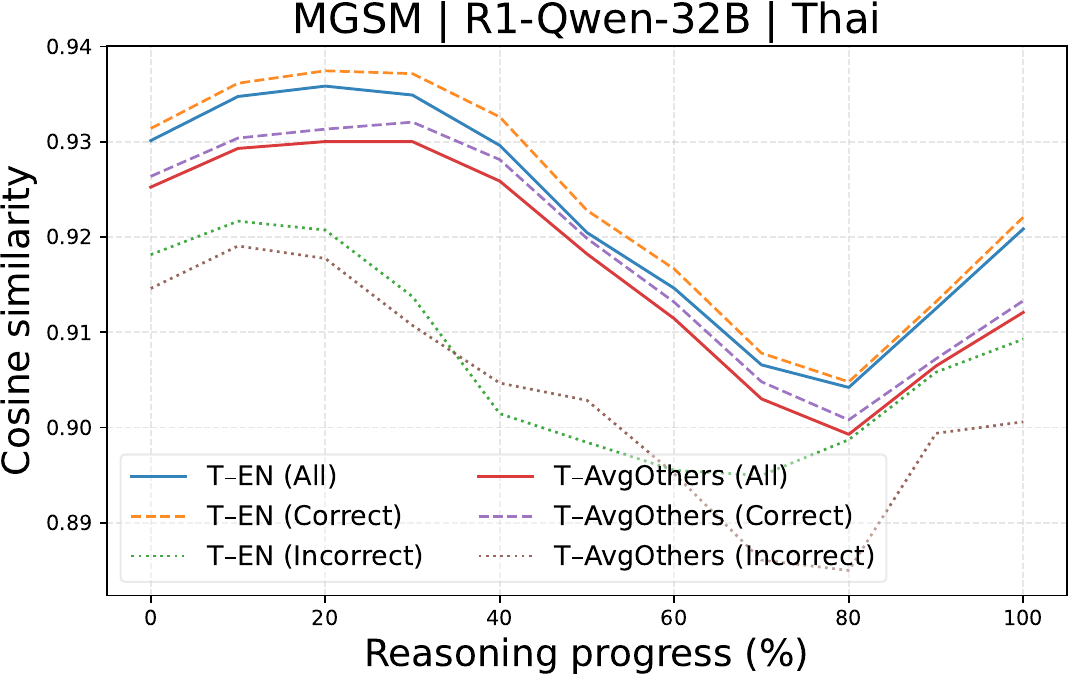}
    \includegraphics[width=0.23\textwidth]{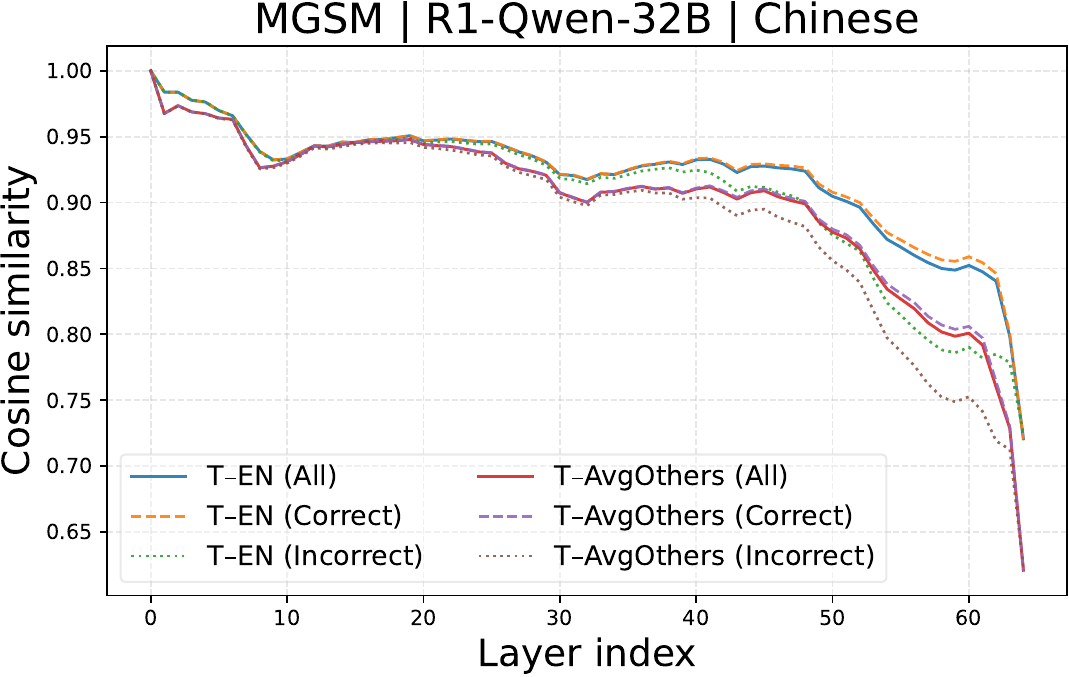}
    \includegraphics[width=0.23\textwidth]{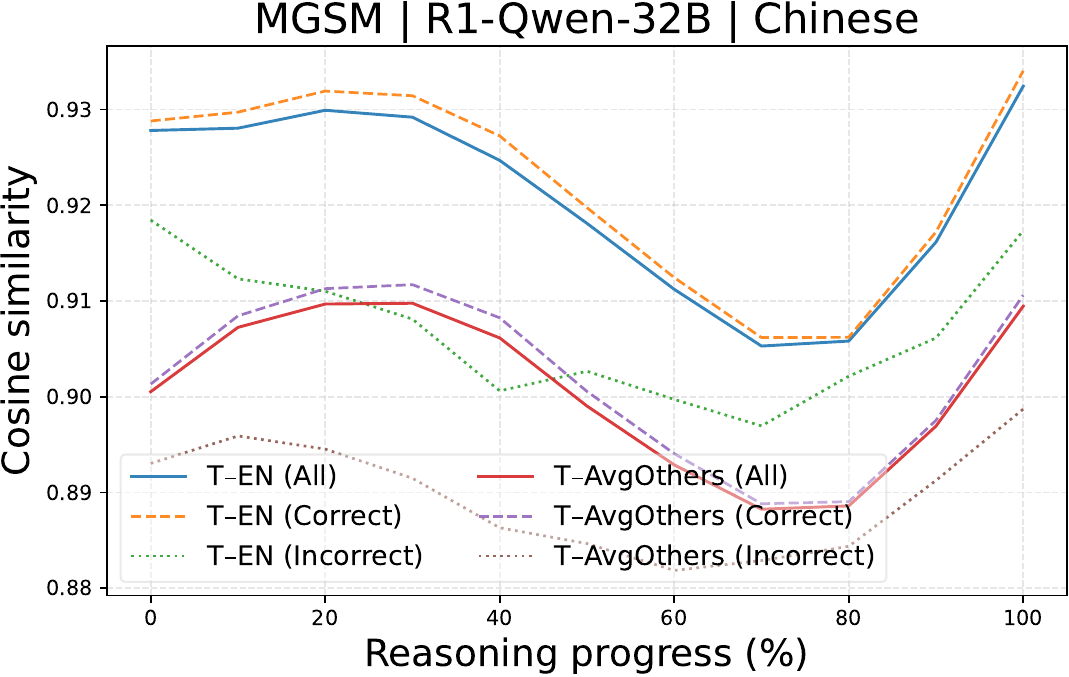}
    \caption{Comparison of cosine similarity with English versus average similarity with other languages, shown separately for correctly and incorrectly solved examples in \textbf{MGSM} with \textbf{R1-Qwen-32B}.}
    \label{fig:cosine_sim_vs_others_complete_32b}
\end{figure*}

We examine whether answer correctness affects crosslingual alignment by grouping MGSM examples into \emph{correct} and \emph{incorrect} sets and comparing their similarity to English and to other languages.
Results for R1-Qwen-\{7B,14B,32B\} are shown in Figures~\ref{fig:cosine_sim_vs_others_complete_7b}, \ref{fig:cosine_sim_vs_others_complete_14b}, \ref{fig:cosine_sim_vs_others_complete_32b}.
Across models, alignment with English seems to vary systematically with language resource level.
High-resource languages show consistently strong similarity to English regardless of correctness, whereas low-resource languages (e.g., Swahili) are more similar to other non-English languages, suggesting a more autonomous latent subspace.
Mid-resource languages exhibit an intermediate pattern, with stronger English alignment primarily for correctly solved instances.

%% file: perturbation.tex
\section{Perturbation Details}\seclabel{perturbation_details}

We used the two perturbation methods in \secref{edit_method} to probe memorization versus (latent) reasoning.
Both methods operate \emph{only} on the subset of MGSM instances that the model answers correctly under \textbf{pass@10} when the truncation ratio is \textbf{0\%} (i.e., with an empty \texttt{<think>\texttt{</think>}} block).
For each selected instance, we produce two edited variants: \textbf{NumEdit} (meaning-altering) and \textbf{Paraphrase} (meaning-preserving).

\subsection{NumEdit}\seclabel{numedit}

NumEdit creates a minimally changed but \emph{meaning-altering} counterfactual by perturbing exactly \emph{one} number in the question.
The expected behavior differs depending on whether the model relies on memorization or reasoning:
a memorization-driven model may continue to output the original answer, whereas a reasoning-driven model should adapt its answer to the changed quantity.

\paragraph{Numeric span detection.}
We identify candidate numbers using a conservative regular expression that matches standalone numeric tokens (including optional negative sign and decimal part), while avoiding matches that are embedded in words or common formats that are likely to break semantics:
\begin{itemize}
    \item \textbf{Exclude years:} numbers matching \texttt{19xx} or \texttt{20xx} are skipped.
    \item \textbf{Exclude ordinals:} tokens followed by \texttt{st/nd/rd/th} are skipped.
    \item \textbf{Exclude fractions:} occurrences adjacent to \texttt{/} (e.g., \texttt{1/2}) are skipped.
\end{itemize}

\paragraph{Perturbation rule.}
Among the remaining candidates, we perturb \emph{exactly one} numeric span with a small additive change:
\begin{itemize}
    \item If the token is an integer:
    \begin{itemize}
        \item For $\{0,1,2\}$, add $+1$.
        \item Otherwise add a small $\Delta\in\{1,2\}$ (seeded randomness).
    \end{itemize}
    \item If the token is a float, add a small fixed delta depending on magnitude:
    $+0.1$ for $|x|<1$, $+0.5$ for $|x|<10$, otherwise $+1.0$.
\end{itemize}

\subsection{Paraphrase}

Paraphrase produces a meaning-preserving rewrite intended to reduce lexical overlap with the original prompt while keeping the problem logically equivalent.
Unlike NumEdit, the gold answer remains the same.
Thus, high performance on paraphrased questions supports generalization beyond surface-form memorization.

\paragraph{LLM-based Question Paraphrasing.}
We leverage \texttt{Gemini-2.5-Flash} to paraphrase each question.
We have the following constraints in the prompt to ensure that the paraphrased question is equivalent to the original one:
\begin{enumerate}
    \item[-] Preserve all numbers exactly.
    \item[-] Preserve all LaTeX math segments (anything inside \texttt{\$...\$}) exactly as-is.
    \item[-] Maintain logical equivalence and ask for the same final quantity.
    \item[-] Do not add/remove constraints, entities, or units.
\end{enumerate}
To prevent crosslingual drift, we additionally specify that the paraphrase must be written in the same language as the input question.
The prompt template is shown in Figure~\ref{fig:paraphrase_prompt}.

\paragraph{Automatic validation.}
Each paraphrase is validated before acceptance: the multiset of numeric tokens in the paraphrase must match the original. If not, the same prompt will be applied again until the paraphrased question is valid.

\subsection{Perturbation Solvability}

To verify that the generated counterfactual questions remain \emph{solvable} and do not introduce unintended artifacts, we conduct an auxiliary evaluation using a strong commercial model again (\texttt{Gemini-2.5-Flash}).
This analysis serves as a sanity check that the perturbations preserve mathematical well-formedness while modifying surface form or numerical content as intended.

\paragraph{Setup.}
For each selected MGSM problem, we query the model on three inputs: the original question, its NumEdit variant, and its Paraphrase variant (for each input, we only query once).
The model is instructed to optionally generate intermediate steps but to output a single final answer in a strict, parseable format (see prompt in Figure~\ref{fig:gemini_solvability_prompt}).
We then compare Gemini's predictions across variants.
Specifically, we report:
(i) \textbf{Orig Acc}, Gemini's accuracy on the original question;
(ii) \textbf{NumEdit Match}, the proportion of NumEdit predictions identical to the original prediction (lower is better, as the gold answer is expected to be changed); and
(iii) \textbf{Paraphrase Match}, the proportion of Paraphrase predictions identical to the original prediction (higher is better, as the gold answer is expected to be preserved).

\paragraph{Results and Discussion.}
Table~\ref{tab:gemini_counterfactuals} shows that Gemini achieves consistently high accuracy on the original questions across models and languages, indicating that the selected problem subset is reliably solvable.
For NumEdit, the matching ratio remains low (typically below 10\%), confirming that the numerical perturbations effectively alter the solution and are not trivially ignored.
In contrast, Paraphrase variants exhibit very high matching ratios (often above 95\%), demonstrating that meaning-preserving rewrites retain solvability and solution consistency.
These trends are stable across model sizes and languages, including lower-resource settings.
Together, these results confirm that both perturbation methods produce mathematically valid and solvable questions.
NumEdit reliably changes the target answer, while Paraphrase preserves it, validating their use as controlled probes for disentangling memorization from latent reasoning.

\begin{table*}[t]
\centering
\small
\setlength{\tabcolsep}{4pt}
\resizebox{\linewidth}{!}{
\begin{tabular}{l l r r r r r r r r r r r}
\toprule
Original Model & Metric & EN & FR & DE & ZH & JA & RU & ES & SW & BN & TE & TH \\
\midrule
\multirow{3}{*}{R1-Qwen-7B} & \textbf{Orig Acc} $\uparrow$ & 0.98 & 0.93 & 0.95 & 0.94 & 0.88 & 1.00 & 0.96 & 0.94 & 0.93 & 0.91 & 0.96 \\
 & \textbf{NumEdit Match} $\downarrow$ & 0.07 & 0.05 & 0.08 & 0.06 & 0.16 & 0.07 & 0.07 & 0.16 & 0.05 & 0.08 & 0.09 \\
 & \textbf{Paraphrase Match} $\uparrow$ & 0.97 & 0.93 & 0.99 & 0.95 & 0.97 & 1.00 & 0.95 & 0.97 & 0.93 & 0.87 & 0.91 \\
 \hline
\addlinespace[3pt]
\multirow{3}{*}{R1-Qwen-14B} & \textbf{Orig Acc} $\uparrow$ & 0.97 & 0.96 & 0.94 & 0.94 & 0.88 & 0.95 & 0.94 & 0.86 & 0.99 & 0.89 & 0.96 \\
 & \textbf{NumEdit Match} $\downarrow$ & 0.07 & 0.06 & 0.07 & 0.09 & 0.16 & 0.08 & 0.12 & 0.06 & 0.07 & 0.04 & 0.08 \\
 & \textbf{Paraphrase Match} $\uparrow$ & 0.97 & 0.94 & 0.97 & 0.96 & 0.93 & 0.95 & 0.97 & 0.88 & 0.97 & 0.93 & 0.94 \\
 \hline
\addlinespace[3pt]
\multirow{3}{*}{R1-Qwen-32B} & \textbf{Orig Acc} $\uparrow$ & 0.98 & 0.97 & 0.93 & 0.93 & 0.94 & 0.98 & 0.97 & 0.95 & 0.96 & 0.89 & 0.96 \\
 & \textbf{NumEdit Match} $\downarrow$ & 0.10 & 0.06 & 0.09 & 0.11 & 0.15 & 0.11 & 0.08 & 0.09 & 0.11 & 0.13 & 0.09 \\
 & \textbf{Paraphrase Match} $\uparrow$ & 0.94 & 0.95 & 0.97 & 0.98 & 0.95 & 0.99 & 0.97 & 0.95 & 0.97 & 0.95 & 0.95 \\
\bottomrule
\end{tabular}
}
\caption{Gemini performance on original and counterfactual MGSM questions. \textbf{Orig Acc} compares Gemini's prediction on the original question to the gold answer. \textbf{NumEdit Match} measures the fraction of NumEdit predictions identical to the original prediction (lower is better). \textbf{Paraphrase Match} measures the fraction of Paraphrase predictions identical to the original prediction (higher is better).}
\label{tab:gemini_counterfactuals}
\end{table*}

\begin{figure*}[t]
\centering
\begin{tcolorbox}[
  enhanced,
  colback=white,
  colframe=black!50,
  boxrule=0.6pt,
  arc=2mm,
  left=3mm,right=3mm,top=2mm,bottom=2mm,
  width=\textwidth,
  title=\textbf{Paraphrase prompt template},
  fonttitle=\normalsize,
]
\small
\ttfamily
You are rewriting a math problem.

Language constraint (MUST follow):\\
- The paraphrase MUST be written in the SAME language as the original question.\\
- The original question language is: \{language\_name\}. Do NOT translate to any other language.
\\
Hard constraints:\\
1) Preserve ALL numbers exactly (character-for-character).\\
2) Preserve ALL LaTeX math exactly as-is (anything inside \$...\$ must appear unchanged).\\
3) Keep the question asking for the same final quantity; the problem must be logically equivalent.\\
4) Reduce lexical overlap by paraphrasing and reordering sentences outside math mode.\\
5) Do NOT include any solution steps, explanations, or the final answer.\\
6) Do NOT add or remove any facts, entities, units, or constraints.\\

Return ONLY valid JSON with exactly these keys:\\
\{``paraphrase'': ``...'', ``changes'': ``...''\}

Original problem:
\\
\{problem\}
\\
\end{tcolorbox}
\caption{Prompt used to generate meaning-preserving paraphrases using \texttt{Gemini-2.5-Flash}. Placeholders \{\texttt{language\_name}\} and \{\texttt{problem}\} are substituted per instance.}
\label{fig:paraphrase_prompt}
\end{figure*}

\begin{figure*}[t]
\centering
\begin{tcolorbox}[
  enhanced,
  colback=white,
  colframe=black!50,
  boxrule=0.6pt,
  arc=2mm,
  left=3mm,right=3mm,top=2mm,bottom=2mm,
  width=\textwidth,
  title=\textbf{Solvability evaluation prompt},
  fonttitle=\normalsize,
]
\small
\ttfamily
You are given a grade-school math word problem.\\

Language constraint:\\
- Write your solution in the SAME language as the problem.\\
- The problem language is: \{language\_name\}. Do not translate.\\

You may write intermediate steps.\\
Hard requirement:\\
- End your response with a SINGLE final line in the following exact format:\\
\quad FINAL\_ANSWER: <answer>\\

Rules for <answer>:\\
- Provide only the final numeric value (or a simplified number).\\
- Do not wrap it in LaTeX, do not add units, and do not add extra words.\\
- Do not output anything after the FINAL\_ANSWER line.\\

Problem:
\\
\{problem\}
\\
\end{tcolorbox}
\caption{Prompt used to evaluate the solvability of original and counterfactual MGSM questions using \texttt{Gemini-2.5-Flash}. Placeholders \{\texttt{language\_name}\} and \{\texttt{problem}\} are substituted per instance.}
\label{fig:gemini_solvability_prompt}
\end{figure*}